\documentclass[12pt,twoside]{UNIVRThesis}
\usepackage{lgrind}
\usepackage{longtable}
\usepackage{multirow}
\usepackage{multicol}
\usepackage{hyperref}
\usepackage{graphicx}
\usepackage{amsmath}
\usepackage{amssymb}
\usepackage{enumitem}
\usepackage{array}
\usepackage{float}
\usepackage{algorithm}
\usepackage{algorithmic}
\usepackage{times}
\usepackage{epsfig}
\usepackage{graphicx}
\usepackage{multirow}
\usepackage{color}
\usepackage{booktabs}
\usepackage{colortbl}
\usepackage{caption}
\usepackage{subcaption}
\definecolor{gre8}{RGB}{180,180,180}
\definecolor{gre5}{RGB}{224,224,224}
\usepackage[usenames,dvipsnames]{xcolor}

\floatname{algorithm}{Procedure}

\newcolumntype{C}[1]{>{\centering\arraybackslash}p{#1}}
\newcommand{\nullc}[0]{\multicolumn{1}{c|}{/}}
\newcommand{\red}[1]{\textcolor{black}{#1}}

\graphicspath{ {Figure/} }

\usepackage{amsfonts, amstext, amsthm, color}

\usepackage[normalem]{ulem}

\usepackage{booktabs} 

\usepackage[normalem]{ulem}
\usepackage{booktabs} 

\begin{document}

\title{Ranking to Learn and Learning to Rank:\\On the Role of Ranking in Pattern Recognition Applications} 

\author{Giorgio Roffo}

\department{Department of Computer Science}

\degree{European Doctor of Philosophy\\ S.S.D. ING-INF05\\Cycle XXIX/2014}

\degreemonth{May}
\degreeyear{2017}
\thesisdate{May 25, 2017}
 

\supervisor{Prof. Marco Cristani}{Associate Professor}

\chairman{Prof. Massimo Merro}{Chairman of the PhD School Council}

\pagenumbering{roman}

\maketitle


\newenvironment{bottompar}{\par\vspace*{\fill}}{\clearpage}

\begin{bottompar}
\begin{center}
\footnotesize \textit{This work is licensed under a Creative Commons Attribution-NonCommercial-NoDerivs 3.0 Unported License, Italy. To read a copy of the licence, visit the web page:}\\
\footnotesize \textit{http://creativecommons.org/licenses/by-nc-nd/3.0/}
\end{center}
\noindent
\includegraphics[width=0.025\textwidth]{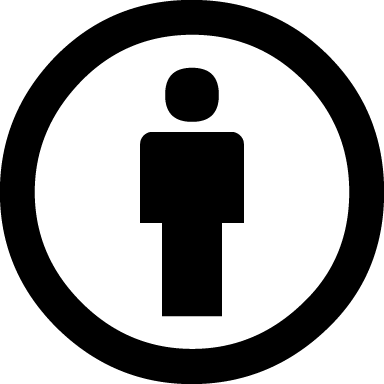} \footnotesize \textbf{Attribution --} You must give appropriate credit, provide a link to the license, and indicate if changes
were made. You may do so in any reasonable manner, but not in any way that suggests the licensor
endorses you or your use.

\noindent
\includegraphics[width=0.025\textwidth]{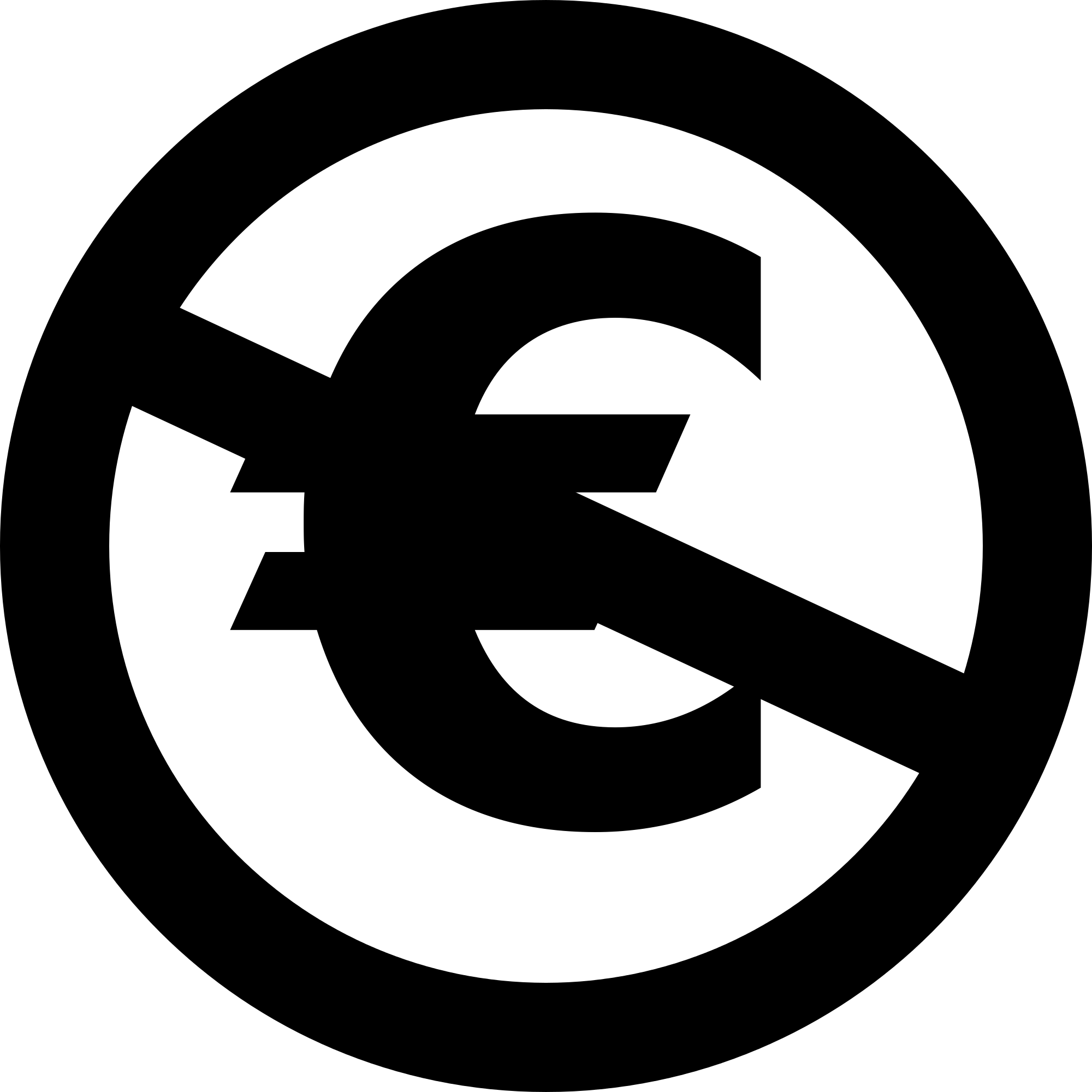} \footnotesize \textbf{NonCommercial --} You may not use the material for commercial purposes.\\
\noindent
\includegraphics[width=0.025\textwidth]{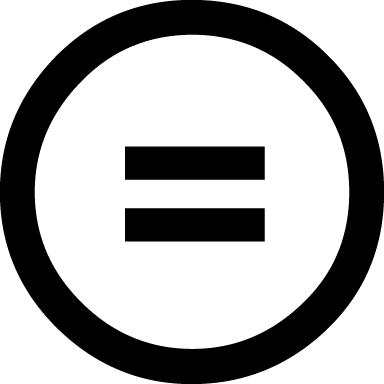} \footnotesize \textbf{NoDerivatives --} If you remix, transform, or build upon the material, you may not distribute the
modified material.\\

\begin{center}
\footnotesize \textit{Ranking to Learn and Learning to Rank: On the Role of Ranking in Pattern Recognition Applications} -- Giorgio Roffo\\ 
\footnotesize PhD Thesis, Verona, May 25, 2017 .\\
\footnotesize ISBN: XXXXXXXXXXXXX\\
\end{center}

\noindent
\footnotesize \textcopyright All rights reserved. This copy of the thesis has been supplied to ensure timely dissemination of scholarly
and technical work on condition that anyone who consults it is understood to recognize that its copyright
rests with its author and that no quotation from the thesis and no information derived from it may be
published without the author’s prior consent.
\end{bottompar}

\cleardoublepage


\begin{abstractpage}
\noindent
The last decade has seen a revolution in the theory and application of machine learning and pattern recognition. Through these advancements, variable ranking has emerged as an active and growing research area and it is now beginning to be applied to many new problems. The rationale behind this fact is that many pattern recognition problems are by nature ranking problems. The main objective of a ranking algorithm is to sort objects according to some criteria, so that, the most relevant items will appear early in the produced result list. Ranking methods can be analyzed from two different methodological perspectives: \textit{ranking to learn} and \textit{learning to rank}. The former aims at studying methods and techniques to sort objects for improving the accuracy of a machine learning model. Enhancing a model performance can be challenging at times. For example, in pattern classification tasks, different data representations can complicate and hide the different explanatory factors of variation behind the data. In particular, hand-crafted features contain many cues that are either redundant or irrelevant, which turn out to reduce the overall accuracy of the classifier. In such a case feature selection is used, that, by producing ranked lists of features, helps to filter out the unwanted information. Moreover, in real-time systems (e.g., visual trackers) ranking approaches are used as optimization procedures which improve the robustness of the system that deals with the high variability of the image streams that change over time. The other way around, learning to rank is necessary in the construction of ranking models for information retrieval, biometric authentication, re-identification, and recommender systems. In this context, the ranking model's purpose is to sort objects according to their degrees of relevance, importance, or preference as defined in the specific application.

This thesis addresses these issues and discusses different aspects of variable ranking in pattern recognition, biometrics, and computer vision. In particular, this work explores the merits of ranking to learn, by proposing novel solutions in feature selection that efficiently remove unwanted cues from the information stream. A novel graph-based ranking framework is proposed that exploits the convergence properties of power series of matrices thereby individuating candidate features, which turn out to be effective from a classification point of view. Moreover, it investigates the difficulties of ranking in real-time while presenting interesting solutions to better handle data variability in an important computer vision setting: \textit{Visual Object Tracking}. The second part of this thesis focuses on the problem of learning to rank. Firstly, an interesting scenario of automatic user re-identification and verification in text chats is considered. Here, we start from the challenging problem of feature handcrafting to automatic feature learning solutions. We explore different techniques which turn out to produce effective ranks, contributing to push forward the state of the art. Moreover, we focus on advert recommendation, where deep convolutional neural networks with shallow architectures are used to rank ads according to users' preferences. We demonstrate the quality of our solutions in extensive experimental evaluations. Finally, this thesis introduces representative datasets and code libraries in different research areas that facilitate large-scale performance evaluation.


\end{abstractpage}


\cleardoublepage

\section*{Acknowledgments}

This thesis could not have been completed without the help of many people. I would like to express my special appreciation and thanks to my advisor Professor Dr. Marco Cristani, you have been a tremendous mentor for me. I would like to thank you for encouraging my research and for allowing me to grow as a research scientist. I would also like to thank my internal committee members, Professor Dr. Gloria Menegaz and Dr. Manuele Bicego for serving as my committee members even at hardship. I also want to thank you for letting my doctoral thesis advancements be enjoyable moments, and for your brilliant comments and suggestions, thanks to you. I would especially like to thank Professor Dr. Alessandro Vinciarelli at University of Glasgow who provided me an opportunity to join his team as intern, and who gave access to the laboratory and research facilities. Without his precious support it would not be possible to conduct this research. Your advice on both research as well as on my career have been priceless. \\
I would like to thank Professor Dr. Vittorio Murino at Italian Institute of Technology for the great opportunity to collaborate with him and also for the insightful discussions about research which motivated me to widen my research from various perspectives. You inspired me to continue in the carrier of research and incited me to strive towards my goal. \\
A special thanks to my family. Words cannot express how grateful I am to my mother, father, and brothers for all of the sacrifices that you have made on my behalf. I would also like to thank all of my friends and colleagues (Matteo, Simone, Francesco, Pietro, Sami, Irtiza, Cristina, Davide) for their support and care. At the end I would like to thank the entire administrative staff of the Computer Science Department in Verona, especially Mrs Laura Marcazzan, the source from which all good things flow.


\noindent
\textit{Universit\`a degli Studi di Verona\\
Dipartimento di Informatica\\
Strada le Grazie 15, 37134 Verona\\
Italy}

\begin{bottompar}

\noindent
\footnotesize \textcopyright All rights reserved. This work is licensed under a Creative Commons Attribution-NonCommercial-NoDerivs 3.0 Unported License, Italy. This copy of the thesis has been supplied to ensure timely dissemination of scholarly and technical work on condition that anyone who consults it is understood to recognize that its copyright rests with its author and that no quotation from the thesis and no information derived from it may be published without the author’s prior consent.
\end{bottompar}

\pagestyle{plain}

\tableofcontents
\newpage

\pagenumbering{arabic}
\setcounter{savepage}{\thepage}

\chapter{Introduction}

Human beings have always looked at nature and searched for patterns. Eons ago we gazed at the stars and discovered patterns we call constellations, even coming to believe they might control our destiny. We have watched the days turn to night and back to day, and seasons, as they come and go, and called that pattern \textit{time}. We see symmetrical patterns in the human body and the tiger's stripes, and build those patterns into what we create, from art to our cities. 

In computer science, the problem of recognizing patterns and regularities in data has a long and successful history. The birth of the \textit{pattern} \textit{recognition} grew out of engineering schools, whereas \textit{machine} \textit{learning} has its origins in computer science. However, these areas of study can be viewed as two facets of the same medal, and together they have undergone substantial development over the past ten years. Indeed, the last decade has seen a revolution in the theory and application of machine learning and pattern recognition. Within these domains, variable \textit{ranking} is a key problem and it plays an increasingly significant role with respect to many pattern recognition applications. Recently, ranking methods based on machine learning approaches, called \textit{learning-to-rank}, become the focus for researchers and practitioners in the fields of information retrieval and recommendation systems. However, learning to rank exists beyond these domains as it remains a crucial operation within the pattern recognition community, since, many problems are by nature ranking problems. For example, areas of interest include, but are not limited to, feature ranking and selection, identity authentication, biometric verification, re-identification of anonymous people, and so forth.

Without loss of generality, learning to rank aims at designing and applying methods to automatically learn a model from training data, such that the model can be used to sort objects according to their degrees of relevance, preference, or importance as defined in a specific application \cite{hang2011short}. With few exceptions, all of these algorithms follow the same guiding principle: a sample's \textit{relevant} \textit{neighbors} should lie closer than its \textit{irrelevant} \textit{neighbors} \cite{mcfee2010metric}. The exact definitions of relevant and irrelevant vary across problem settings. For example, if the goal is classification, the aim is to find useful features that separate well two or more classes, therefore, \textit{relevant} is a feature which is highly representative of a class \cite{Guyon:2003:IVF:944919.944968}. A second, somewhat different, example is user re-identification (Re-ID). Re-ID is defined as a process of establishing correspondence between a probe user $q$ (unknown) and her true identity within a predefined database of users (i.e., gallery-set). In this context, relevancy is closely related to similarity between $q$ and her real identity. Therefore, a good ranking algorithm can sort the gallery-set by increasing distance from $q$ resulting in relevant (similar) neighbors at the front of the ranked list, and irrelevant neighbors at the end. In keeping with this principle, the ranking problem is a special case of information retrieval in the \textit{query}-\textit{by}-\textit{example} paradigm \cite{hang2011short,mcfee2010metric,phophalia2011survey}.

For most practical applications, before learning can take place, the original input variables are typically pre-processed to transform them into some new space of variables where, it is hoped, the pattern recognition problem will be easier to solve. This pre-processing stage is sometimes also called \textit{feature}  \textit{engineering} \cite{bengio2013representation,seide2011feature}. Clearly, the performance of machine learning methods is heavily dependent on the choice of features on which they are applied. Different features can entangle and hide the different explanatory factors of variation behind the data. As a result, a first challenging aspect for robust pattern recognition algorithms is related to feature extraction and engineering \cite{seide2011feature}. Designing features is often a laborious task and may require specific domain knowledge of the data that is sometimes difficult to obtain. Although feature engineering is important, it highlights difficulties to extract and organize the discriminative information from the data. Often, the designed features contain many cues that are either redundant or irrelevant with respect to the application. This unwanted stream of data results in a reduced learning rate and, then, a lower overall accuracy. 

In response to these issues, much efforts in deploying pattern recognition algorithms goes in improving the pre-processing stage by studying new solution for feature selection or different approaches to data representation that turn out to support effective machine learning. Recent work in the area of unsupervised feature learning and deep learning is expanding the scope of machine learning by making algorithms less dependent on feature engineering \cite{bengio2013representation,hamel2011temporal,seide2011conversational,yu2012conversational}, so that to make progress towards the automatic discovery of regularities in low-level sensory data. Among the various ways of learning representations, deep learning methods (e.g., convolutional neural networks) have gained a lot of interest in the recent years: ``\textit{those that are formed by the composition of multiple non-linear transformations, with the goal of yielding more abstract - and ultimately more useful - representations}" \cite{bengio2013representation}.

In other pattern recognition problems, the training data consists of a set of input vectors without any corresponding target values. The goal in such unsupervised learning problems may be to discover groups of similar examples within the data, where it is called clustering, or to determine the distribution of data within the input space, known as density estimation, or to project the data from a high-dimensional space down to two or three dimensions for the purpose of visualization. Within such situations, the curse of irrelevant or redundant cues is still present, and since this information is not important to the solution of the problem then the overall accuracy of the system can suffer. A crucial operation is of selecting a subset of the original features while retaining the optimal salient characteristics of the data.

Variable ranking might also be performed in order to help in handling data variability by improving performance in real-time settings. For example, in \textit{visual} \textit{tracking} (i.e., a highly popular research area of computer vision), the appearance of the target object may change drastically due to intrinsic or extrinsic factors such as sudden illumination variations, occlusions, motion blur, fast motion of the camera or target, and so on. Ranking and selecting the most suitable features for tracking, at the right time, allows the tracking system to be flexible enough to handle drastic changes in appearance in real-world scenarios.

This thesis explores the ranking process from feature engineering to representation learning and deep learning. We address motivations, advantages, and challenges of ranking under many different conditions while dealing with diverse scenarios. Moreover, we propose possible improved solutions in feature selection, biometric re-identification, identity authentication and verification that promise enhanced system accuracy, ranking quality and less of a need for manual parameter adaptation.

\section{Thesis Statement and Contributions}\label{ch1:goals}

This thesis is inserted in the above scheme, and is aimed at investigating the role of variable ranking in the wide field of pattern recognition considering many different conditions and dealing with diverse scenarios. The first contribution of this thesis is therefore the identification of some real-world scenarios which can be faced from a ranking perspective. For each scenario, motivations, advantages, and challenges of both feature engineering and ranking approaches are addressed, proposing possible solutions.

From a methodological point of view, this thesis contributes in different ways: i) the term ``learning-to-rank" has been exported from the information retrieval context and tailored to different pattern recognition scenarios; ii) the different problematics related to the curse of irrelevant and redundant cues have been discussed. This unwanted information affects negatively every pattern recognition algorithm, the merits of ranking features according to their degrees of relevance have been addressed contributing with novel ranking solutions that help in removing irrelevant cues from the information stream, enabling the learning algorithm to reach higher generalization levels; iii) supervised and unsupervised scenarios have been taken into account, contributing with a novel graph-based ranking method exploiting the convergence properties of power series of matrices and demonstrating the quality of being particularly good in extensive experimental evaluations; iv) a further contribution of this work is given with respect to real-time systems, in response to the need of handling data variability in visual object tracking, we investigated on the possibility to produce robust and stable ranked lists of features with the aim of selecting the most significant ones at each frame (i.e., features that better discriminate between foreground and background), under the hard constraint of speed required for such an application setting; v) learning to rank approaches have been explored in the context of biometric authentication from textual conversations, in particular text chats. Here, novel biometric features have been derived from recent literature on authorship attribution and keystrokes biometrics. From a more applicative perspective, we explored different techniques which turn out to produce effective ranks for the tasks of automatic user re-identification and verification, contributing to push forward the state of the art; vi) we provided evidence that deep learning approaches can be applied to the problem of ranking. In this work, we overview the history and training of modern convolutional neural networks (CNNs), then we proposed a possible solution to the ranking problem (i.e., user re-identification and ad recommendation) by deep CNNs with shallow architectures. vii) Finally, new representative datasets in different research fields have been introduced along with code libraries that integrate several algorithms and evaluation metrics with uniform input and output formats to facilitate large scale performance evaluation. For example, the Feature Selection Library  \href{https://uk.mathworks.com/matlabcentral/fileexchange/56937-feature-selection-library}{(FSLib)}, which was the most downloaded toolbox in 2016, received a \textbf{Matlab official recognition for the outstanding contribution}, the \href{http://www.votchallenge.net/vot2016/trackers.html}{DFST tracker} released to the international Visual Object Tracking (VOT) challenge 2016, and the \href{https://www.kaggle.com/groffo/ads16-dataset}{ADS-16} dataset for computational advertising released on kaggle repository, among others.  

More in detail, four applicative contexts have been analyzed:
\begin{itemize}
  \item \textbf{Feature Ranking and Selection:}
  
This thesis contributed in the feature selection context, by introducing a graph-based algorithm for feature ranking, called \textit{Infinite Feature Selection} (Inf-FS), that permits the investigation of the importance (relevance and redundancy) of a feature when injected into an arbitrary set of cues.
The Inf-FS is inspired to the Feynman's path integrals \cite{Feynman:1948aa}. The path integral formalism, is a tool for calculating quantum mechanical probabilities. The basic idea behind path integrals is that of measuring the quantum probability of a space-time event. In particular, if a particle is measured to be at a particular position at a particular time, to calculate the quantum amplitude (or its probability) to be in a different position at a later time, all the possible space-time paths the particle can take between those two space-time events must be considered. In other words, each path has its own amplitude, which is a complex number, in order to calculate the total amplitude between two space-time events the Feynman's receipt states that all the complex amplitudes have to be added up. The standard way to interpret these amplitudes is the probability to measure the particle at position $B$ at time $t_B$, knowing that it was at position $A$ at time $t_A < t_B$, which is given by the the square absolute value of the amplitude associated to those two events. Here is where the Inf-FS comes from. Thus, we derived a discrete form for path integral. Then, we mapped the space-time to a simplified discrete form without time, that is: \textit{a graph}. Finally, the framework used to estimate the most likely position where to find a particle has been switched to the novel problem of finding the most relevant feature (see Sec. \ref{sec5:Quantum} for details). 

From a computer science perspective, the idea is to map the feature selection problem to an affinity graph, and then to consider a subset of features as a path connecting them. The cost of the path is given by the combination of pairwise relationships between the features embedded in a cost matrix. By construction, the method allows to use convergence properties of the power series of matrices, and evaluate analytically the relevance of a feature with respect to all the other ones taken together. Indeed, considering a selection of features as a path among feature distributions and letting these paths tend to an infinite number permits to individuate candidate features, which turn out to be effective from a classification point of view, as proved by a thoroughly experimental section. The Inf-FS performs the ranking step in an unsupervised manner, followed by a simple cross-validation strategy for selecting the best $m$ features. 

The most appealing characteristics of this approach are 1) all possible subsets of features are considered in evaluating the rank of a given feature and 2) it is extremely efficient, as it converts the feature ranking problem to simply calculating the geometric series of an adjacency matrix.

  
The second original contribution is given by a second feature selection method, called EC-FS, that ranks features according to a graph centrality measure. The Eigenvector centrality is a way of measuring the total effects centrality of a node in a graph and then to rank nodes according to their importance in the network, also in this case nodes directly correspond to features. 
  
Given that, we investigated the interrelations of the two algorithms. These methods rank features based on path integrals (Inf-FS) and a centrality concept (EC-FS) on a feature adjacency graph. The purpose of this analysis was to obtain more insights into the Inf-FS formulation, with the aim of better understanding where these methods work well and where they fail. 

\item \textbf{Online Feature Ranking for Visual Object Tracking:}

Another important contribution of this thesis is related to real-time settings, in particular \textit{visual object tracking}. Object tracking is one of the most important tasks in many applications of computer vision. Many tracking methods use a fixed set of features ignoring that appearance of a target object may change drastically due to intrinsic and extrinsic factors. The ability to identify discriminative features at the right time would help in handling the appearance variability by improving tracking performance. We analysed different ranking methods in terms of accuracy, stability and speed, thereby indicating the significance of employing them for selecting candidate features for tracking systems, while maintaining high frame rates. In particular, the \textit{Infinite Feature Selection} embedded on the Adaptive Color Tracking \cite{Danelljan_2014_CVPR} system operates at over 110 FPS resulting in what is clearly a very impressive performance.

\item \textbf{Learning to Rank:}
  
In this context, we explored different application domains where learning is used to sort objects by relevance (e.g., users' identities or items) according to the specific application. We worked on two different tasks. Firstly, we dealt with biometric identification and verification in text chats. We start from the principle that the correct match of the probe template (unknown user) should be positioned in the top rank within the whole gallery set (database of known users). Secondly, we dealt with the task of recommending adverts to users. The goal is to produce ranked lists of ads according to the users' degrees of preference. 
  
\textbf{Biometric Verification and Identification}
  
The need to authenticate ourselves to machines is ever increasing in today's Internet society and it is necessary to fill the gap between human and computer to secure our transactions and networks. In this context, the first original contribution was in recognizing that re-identification and verification of people can be exported from the video-surveillance and monitoring contexts and tailored to online dyadic textual chats. Interacting via text chats can be considered as a hybrid type of communication, in which textual information delivery follows turn-taking dynamics, resembling spoken interactions. We presented interesting stylometric features that encode turn-taking conversational aspects, and the manner and rhythm in which users type characters to automatically recognize their identity using these distinguishing traits. These features measure the typing behaviour of people - meaning the way people type and not what they type - providing information about the most important characteristic of an individual: the identity.  
We investigated possible solutions to learning to rank. Firstly, we derived a plausible distance to match users descriptors which operates without learning (i.e., a similarity-based approach). Secondly, we used multi-view learning to compute the similarity between the descriptors by means of kernel matrices. Multi-view learning consists of estimating the parameters of the model given the training set. Given a probe signature, the testing phase consists of computing the similarity of each descriptor with the training samples and using the learned parameters to classify it. Finally, an effective algorithm is proposed to minimize the cost corresponding to the ranking disorders of the gallery. The ranking model is solved with a deep convolutional neural network (CNN) with a shallow architecture. The CNN builds the relation between each input template and its user's identity through representation learning from frequency distributions of a set of data. We provided a new CNN architecture with a reduced number of parameters which allows learning from few training examples. The designed network consists of 1D convolutional filters that results to be useful when learning from vectorized data. We also showed that it is possible to use heterogeneous data sources, combine them together and use this data to feed a CNN.
The main contribution of this pilot experiment is the investigation of possible protocols based on CNNs and limited amount of training samples. A shallow architecture has much less parameters and learning can be performed also without expensive hardware and GPUs. It results in shorter training times. By feeding the deep network with pre-processed features (represented as histograms and interpreted as an estimate of the probability distribution of a continuous variable) allows us, at least partially, to have a better understanding of the results. In other words, we know that features will be transformed into hierarchical representations in increasing levels of abstraction starting from specific cues designed a priori. As a result, when deep learning is not used on images or videos, but generic data, this fact makes results easier to interpret by researchers.  
\\
The proposed approaches have been tested for the task of re-identification and verification of users involved in online dyadic text chat conversations, contributing to push forward the state of the art in this domain.

\textbf{Ranking and Recommending}

Finally, this thesis addressed the \textit{recommendation} problem. We provided evidence that ranking approaches employed in re-identification can be tailored for \textit{computational advertising}. Firstly, the lack of a publicly available benchmark for computational advertising do not allow both the exploration of this intriguing research area and the evaluation of recent algorithms. This work tried to fill, at least partially, the gap above and proposed a novel dataset called ADS-16. It is a publicly available benchmark consisting of $300$ real advertisements (i.e., Rich Media Ads, Image Ads, Text Ads) rated by $120$ unacquainted individuals, enriched with Big-Five users' personality factors and $1,200$ personal users' pictures. Given that, we investigated the capabilities of different techniques in providing good ranked lists of ads (i.e., recommendations). We analysed two tasks (i) ad rating prediction, and (ii) ad click prediction, with the goal in mind to sort ads according to their degrees of preference as defined by the predicted values. Within these two scenarios we exported the deep network used for the Re-ID task obtaining a powerful deep ranking framework. Noteworthy, deep ranking represents a very recent research topic in the recommender systems community.
\end{itemize}

Summarizing, this work explores the different aspects of variable ranking in pattern recognition, computer vision, and biometrics. Firstly, ranking is performed to improve machine learning. In this scenario we contributed with a novel solution in feature selection called \textit{Infinite Feature Selection}, and a second algorithm based on the same concepts of the previous one but considering the ranking step from a graph theory perspective, therefore performing feature selection via \textit{Eigenvector Centrality}. Other contributions of this work fall under the area of computer vision, in particular, visual object tracking. In this context, we analysed the characteristics of a variety of feature selection methods to individuate some approaches suitable for real-time object tracking. 
The second part of this work focuses on \textit{learning to rank}, that is to say, using machine learning to provide high quality rankings. For example, ranking items for personalized recommendations or performing tasks such as re-identification. Indeed, re-identification is closely related (at least methodologically) to object retrieval, where the first step is usually ranking. In this scenario we considered textual chats, we firstly proposed a novel set of soft-biometric features able to measure the style and the typing behavior of the subjects involved in the conversations. Then, we explored different ways to re-identification and verification, among them, we moved a step toward deep ranking as a possible framework for future re-identification systems. This part of the thesis is also providing insights on what concerns social and psychological dimensions underlying the data, such as machine detectable traces of the user's gender within a text.
As a result of this work, we provided four public datasets and different code libraries and tools like the Feature Selection Library for Matlab.

\section{Thesis Structure}\label{ch1:structure}

This thesis is divided in two introductory chapters and three main parts. Chapter \ref{ch2:backgroung} presents the problem of ranking and summarizes the recent literature in diverse contexts. It also introduces the notation and formalism employed in the subsequent chapters. In Chapter \ref{ch3:DeepLearning} a compact and effective overview of modern convolutional neural networks is provided. \\
The first main part, Part \ref{part:one}, describes the data collection and benchmarking activities. In particular, Chapter \ref{ch4:DATASETS} lists four different interdisciplinary research projects and datasets collected in a number of fruitful collaborations with experts from different research backgrounds such as Psychology or Sociology.\\
Part \ref{part:two} focuses on ranking to learn and it reports two applications in which ranking is used to improve the accuracy of a system. Chapter \ref{ch5:FS} introduces and describes the importance of using feature selection approaches in classification, where ranking is analysed with the goal of improving machine learning (i.e., enhancing generalization by reducing overfitting, shortening training times, and simplifying models by making them easier to interpret). Chapter \ref{ch6:TRACKING} deals with online feature ranking for visual tracking, describing the need of handling data variability and especially of meeting the hard constraint of speed required for this application domain. \\
The last part of the thesis, Part \ref{part:trhee}, deals with learning to rank in two different settings. Chapter \ref{ch7:CHATS} describes the problem of recognize people from online dyadic textual chats, it discusses the novelty of extracting nonverbal features and ranking-based methods used to solve the re-identification problem. Then, Chapter \ref{ch8:ADS} presents an application of variable ranking for ad recommendation. It also describes how deep learning is applied to rank adverts according to tasks of click/rating prediction. Finally, Chapter \ref{ch9:conclusions} conclusions are drawn and future perspectives are envisaged.

\section{Publications List}\label{ch1:publications}

Some parts of this thesis have been published in international conference proceedings, including $A^*$ conferences such as IEEE International Conference on Computer Vision, ACM Multimedia, and The British Machine Vision Conference. The following publications were produced:

\begin{itemize}
\item{Discrete time Evolution Process Descriptor for shape analysis and matching. S. Melzi, M. Ovsjanikov, \textbf{G. Roffo}, M. Cristani, U. Castellani. ACM Transactions on Graphics (TOG) in \cite{melzi2017}.}
\item Ranking to Learn: Feature Ranking and Selection via Eigenvector Centrality. \textbf{G. Roffo}, S. Melzi. Springer Book Chapter: New Frontiers in Mining Complex Patterns, 2017 in \cite{roffomelzi2016}. 
\item Online Feature Selection for Visual Tracking. \textbf{G. Roffo}, S. Melzi. In Conf. The British Machine Vision Conference (BMVC 2016) in \cite{RoffoBMVC2016}.
\item The Visual Object Tracking VOT2016 Challenge Results. Joint Paper: \textbf{G. Roffo}, S. Melzi et Al. In Conf. IEEE European Conference on Computer Vision Workshops (ECCV 2016) in \cite{KristanLMFPCVHL16}. 
\item Feature Selection via Eigenvector Centrality. \textbf{G. Roffo}, S. Melzi. In Conf. ECML/PKDD - New Frontiers in Mining Complex Patterns, (NFMCP 2016) in \cite{RoffoECML16}. 
\item Personality in Computational Advertising: A Benchmark. \textbf{G. Roffo}, A. Vinciarelli. In International Workshop on Emotions and Personality in Personalized Systems at ACM RecSys (EMPIRE 2016) in \cite{RoffoEMPIRE16}. 
\item Infinite Feature Selection on SHORE based biomarkers reveals connectivity modulation after stroke. S. Obertino, \textbf{G. Roffo}, G. Menegaz. In Conf. International Workshop on Pattern Recognition in Neuroimaging (PRNI 2016) in \cite{obertino2016infinite}. 
\item Infinite Feature Selection. \textbf{G. Roffo}, S. Melzi and M. Cristani. In Conf. IEEE International Conference on Computer Vision (ICCV 2015) in \cite{Roffo_2015_ICCV}. 
\item Statistical Analysis of Personality and Identity in Chats Using a Keylogging Platform. \textbf{G. Roffo}, C. Giorgetta, R. Ferrario, W. Riviera and M. Cristani. In Conf. ACM International Conference on Multimodal Interaction (ICMI 2014) in \cite{Roffo:icmi2014}. 
\item Just The Way You Chat: Linking Personality, Style And Recognizability In Chats. \textbf{G. Roffo}, C. Giorgetta, R. Ferrario and M. Cristani. In Conf. European Conference on Computer Vision Workshops (ECCV 2014) in \cite{Roffo:HBU2014}.
\item Trusting Skype: Learning the Way People Chat for Fast User Recognition and Verification. \textbf{G. Roffo}, M. Cristani, L. Bazzani, H. Q. Minh, and V. Murino. IEEE International Conference on Computer Vision Workshops (ICCV 2013) in \cite{Roffo2013}. 
\item Statistical Analysis of Visual Attentional Patterns for Videosurveillance. \textbf{G. Roffo} M. Cristani, F. Pollick, C. Segalin and V. Murino. Iberoamerican Congress on Pattern Recognition (CIARP 2013) in \cite{Roffo2013pollick}. 
\item Reading Between the Turns: Statistical Modeling for Identity Recognition and Verification in Chats. \textbf{G. Roffo}, C. Segalin, V. Murino and M. Cristani. IEEE International Conference on Advanced Video and Signal-Based Surveillance (AVSS 2013) in \cite{Cristani:Skype:AVSS:2013}. 
\item Conversationally-inspired stylometric features for authorship attribution in instant messaging. M. Cristani, \textbf{G. Roffo}, C. Segalin, L. Bazzani, A. Vinciarelli, and V. Murino. ACM international conference on Multimedia, (ACMM 2012) in \cite{Cristani:2012}. 
\end{itemize}

 \chapter{Ranking in Pattern Recognition Scenarios}\label{ch2:backgroung}
 
Ranking plays a central role in many pattern recognition applications such as feature selection, where the goal is to produce ranked lists of features to defy the curse of dimensionality and improve machine learning, information retrieval (e.g., document/image retrieval, person identification) where ranking of query results is one of the fundamental problems, recommendation systems that produce ranked lists of items according to what the user would probably be interested in, and so on. This thesis examines the ranking problem from two different methodological perspectives: \textit{ranking to learn}, which aims at studying methods and techniques to sort objects with the aim to enhance models generalization by reducing overfitting, and \textit{learning to rank}, which makes use of machine learning to produce ranked lists of objects to solve problems in those domains that require objects to be sorted with some particular criteria.  

This chapter starts with ranking to learn and proposes, in Section \ref{ch2:classification}, the related literature of feature selection methods, since any set of hand-crafted features may contain many features that are either redundant or irrelevant, and can thus be removed without incurring loss of information. Indeed, in classification scenarios the aim is to find useful features that separate well two or more classes, therefore, ranking is used to individuate those features which are highly representative of a class \cite{Guyon:2003:IVF:944919.944968}. Section \ref{ch2:ReID} starts with a task of learning to rank in the context of behavioral biometric of keystroke dynamics. This section introduces the related literature of users' authentication, identification and verification in dyadic textual chat conversations. In this context ranking is used to validate the identity of a user who wishes to sign into a system by measuring some intrinsic characteristic of that user. 
In section \ref{ch2:Recommendation}, we briefly review the main recommender system frameworks, where their goal is to sort a set of items according to some criteria of preference, dictated by the user. \\
Beyond the application domain, it is crucial to employ well established metrics for performance evaluation and quality assessment of the produced rankings. At the base of the vast majority of the applications examined in this thesis lie a prediction engine. Given a sample input, this engine may rank objects according to their degrees of similarity with regard to the input sample, or it may predict a probability distribution over a set of classes and estimate the likelihood of such a sample to belong to one of these classes. A basic assumption in statistical pattern recognition is that a system that provides more accurate predictions will be preferred, therefore, where several algorithms are compared it is important to measure the prediction accuracy by using the right metric. Section \ref{ch2:PerfMetrics} overviews some standard performance metrics used for evaluating algorithms, independently from their application, such evaluation metrics are the Precision-Recall, Receiver Operating Characteristic (ROC) Curve, Cumulative Match Characteristic (CMC) Curve, among others.

\section{Scenario 1: Feature Selection as Ranking to Learn}\label{ch2:classification}

Since the mid-1990s, few domains explored used more than 50 features. The situation has changed considerably in the past few years and most papers explore domains with hundreds to tens of thousands of variables or features. New approaches are proposed to address these challenging tasks involving many irrelevant and redundant variables and often comparably few training examples. In this work, the term \textit{variable} is used for the \textit{raw} input variables, and \textit{features} for variables constructed from input variables. There is no distinction between the terms variable and feature if there is no impact on the ranking and selection algorithm. The distinction is necessary in the case of ``feature learning" and ``deep learning" methods for which features are not explicitly computed (hand-crafted), but learnt from input variables.

Two examples are typical of the feature selection domains and serve us as illustration throughout this section on \textit{ranking to learn}. One is gene selection from microarray data and the other is image classification. In the gene selection problem, the variables are gene expression coefficients corresponding to the abundance of mRNA in a sample (e.g. tissue biopsy), for a number of patients. A typical classification task is to separate healthy patients from cancer patients, based on their gene expression \textit{profile}. Usually fewer than 100 examples (patients) are available altogether for training and testing. But, the number of variables in the raw data ranges from 6,000 to 60,000. Some initial filtering usually brings the number of variables to a few thousand. Because the abundance of mRNA varies by several orders of magnitude depending on the gene, the variables are usually standardized. In the image classification problem, the images may be represented by a bag-of-words, that is a vector of dimension the size of the vocabulary containing word frequency counts. Vocabularies of hundreds of thousands of words are common, but an initial pruning of the most and least frequent words may reduce the effective number of words to 15,000. With the advent of deep learning, the number of variables has grown to hundreds of thousands. Large image collections of 150,000 images with the presence or absence of 1000 object categories, are available for research \cite{imagenet_cvpr09, KrizhevskyNIPS2012,ILSVRC15}. Typical tasks include the automatic sorting of images into a web directory, the object localization, object detection, object detection from video, scene classification, and scene parsing.

There are several benefits of selecting a subset of relevant features for use in model construction. The central premise when using a feature selection technique is that the data contains many features that are either redundant or irrelevant, and can thus be removed without incurring much loss of information. This section surveys the works proposed in last few years, that focus mainly on ranking and selecting subsets of features that are useful to build a good predictor. 

\begin{figure}[!]
\centering
\includegraphics[width=1.0\linewidth]{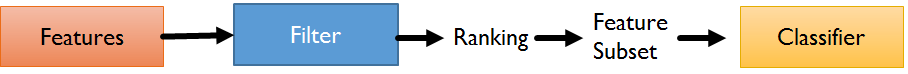}
\caption{Filter Methods: the selection of features is independent of the classifier used. They rely on the general characteristics of the training data to select features with independence of any predictor.}\label{fig2:families1}
\end{figure}

\begin{figure}[!]
\centering
\includegraphics[width=1.0\linewidth]{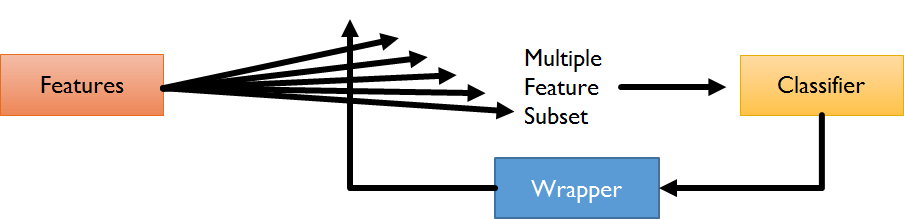}
\caption{Wrapper models involve optimizing a predictor as part of the selection process. They tend to give better results but filter methods are usually computationally less expensive than wrappers.}\label{fig2:families2}
\end{figure}

\begin{figure}[!]
\centering
\includegraphics[width=1.0\linewidth]{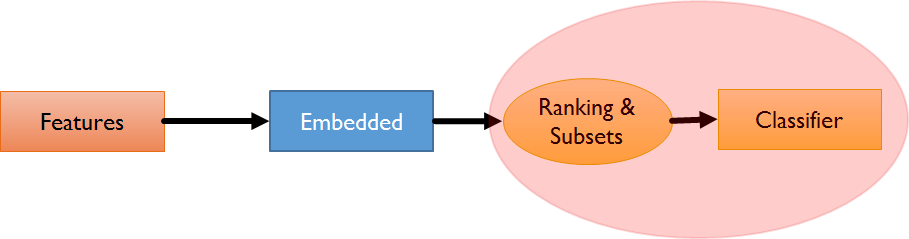}
\caption{In embedded methods the learning part and the feature selection part can not be separated.}\label{fig2:families3}
\end{figure}

Generally, FS techniques can be partitioned into three classes~\cite{guyon2006feature}: \emph{wrappers} (see Fig.~\ref{fig2:families2}), which use classifiers to score a given subset of features; \emph{embedded} methods  (see Fig.~\ref{fig2:families3}), which inject the selection process into the learning of the classifier; \emph{filter} methods (see Fig.~\ref{fig2:families1}), which analyze intrinsic properties of data, ignoring the classifier. Most of these methods can perform two operations, \emph{ranking} and \emph{subset selection}: in the former, the importance of each individual feature is evaluated, usually by neglecting potential interactions among the elements of the joint set~\cite{duch2004comparison}; in the latter, the final subset of features to be selected is provided. In some cases, these two operations are performed sequentially (ranking and selection) \cite{Guyon:2002,Bradley98featureselection,Grinblat:2010,Hutter:02feature,liu2008,LeiYi10.1109};  in other cases, only the selection is carried out \cite{Quanquanjournals}. Generally, the subset selection is always supervised, while in the ranking case, methods can be supervised or not.
While wrapper models involve optimizing a predictor as part of the selection process, filter models rely on the general characteristics of the training data to select features with independence of any predictor. Wrapper models tend to give better results but filter methods are usually computationally less expensive than wrappers. So, in those cases in which the number of features is very large, filter methods are indispensable to obtain a reduced set of features that then can be treated by other more expensive feature selection methods. 
Embedded methods, Fig.~\ref{fig2:families3}, differ from other feature selection methods in the way feature selection and learning interact. In contrast to filter and wrapper approaches, in embedded methods the learning part and the feature selection part can not be separated - the structure of the class of functions under consideration plays a crucial role. For example, Weston et al.\cite{Weston} measure the importance of a feature using a bound that is valid for Support Vector Machines only thus it is not possible to use this method with, for example, decision trees.

Feature selection is \emph{NP-hard}~\cite{guyon2006feature}; if there are $n$ features in total, the goal is to select the optimal subset of $m\!\!\ll \!\!n$, to evaluate $\binom{n}{m}$ combinations; therefore, suboptimal search strategies are considered.
Usually, with the filters, features are first considered individually, ranked, and then a subset is extracted, some examples are 
MutInf~\cite{Hutter:02feature} and Relief-F~\cite{liu2008}. Conversely, with wrapper and embedded methods, subsets of features are sampled, evaluated, and finally kept as the final output, for instance, FSV ~\cite{Bradley98featureselection,Grinblat:2010}, and SVM-RFE \cite{Guyon:2002}.

\begin{figure}[!]
\centering
\includegraphics[width=0.49\linewidth]{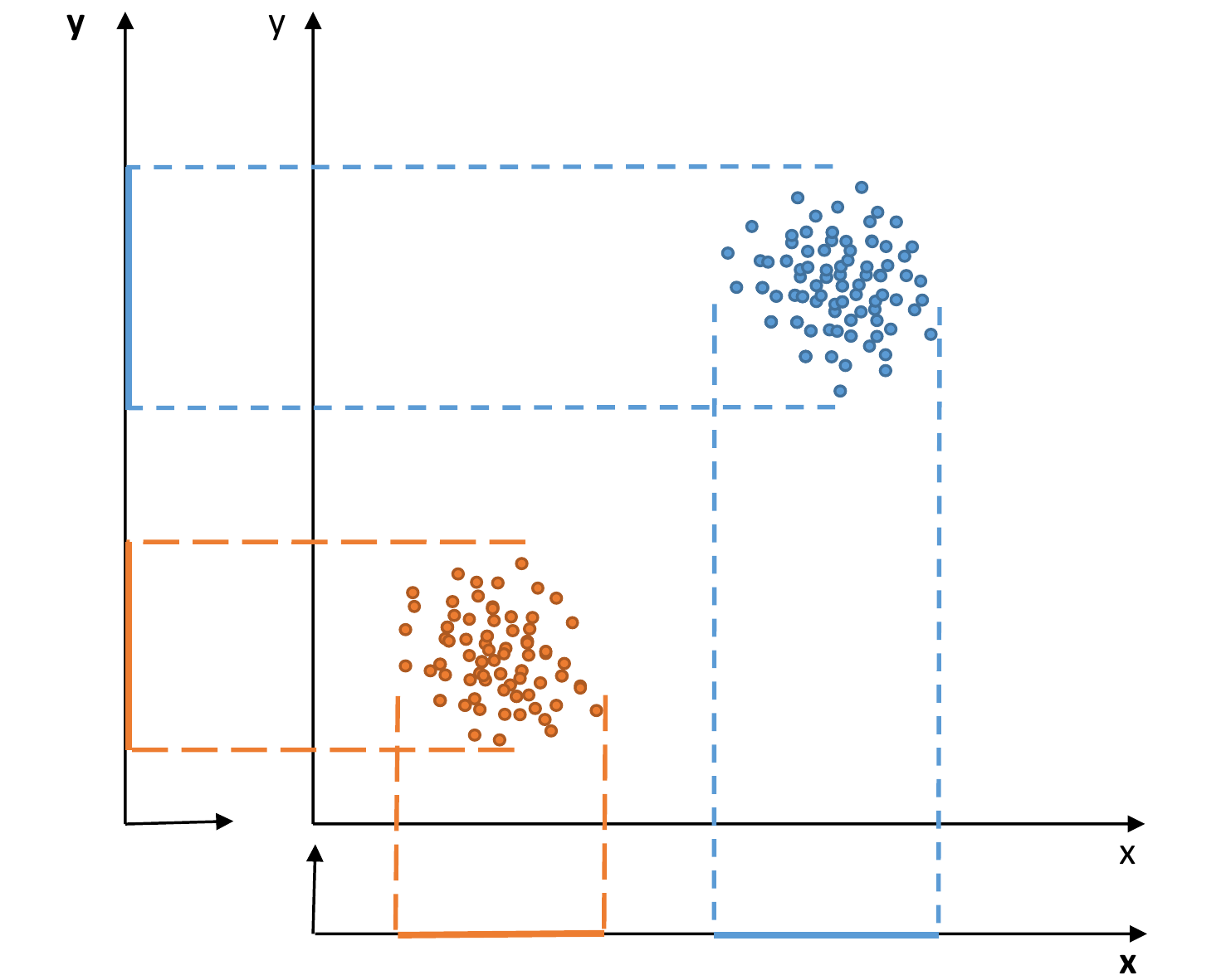}
\includegraphics[width=0.49\linewidth]{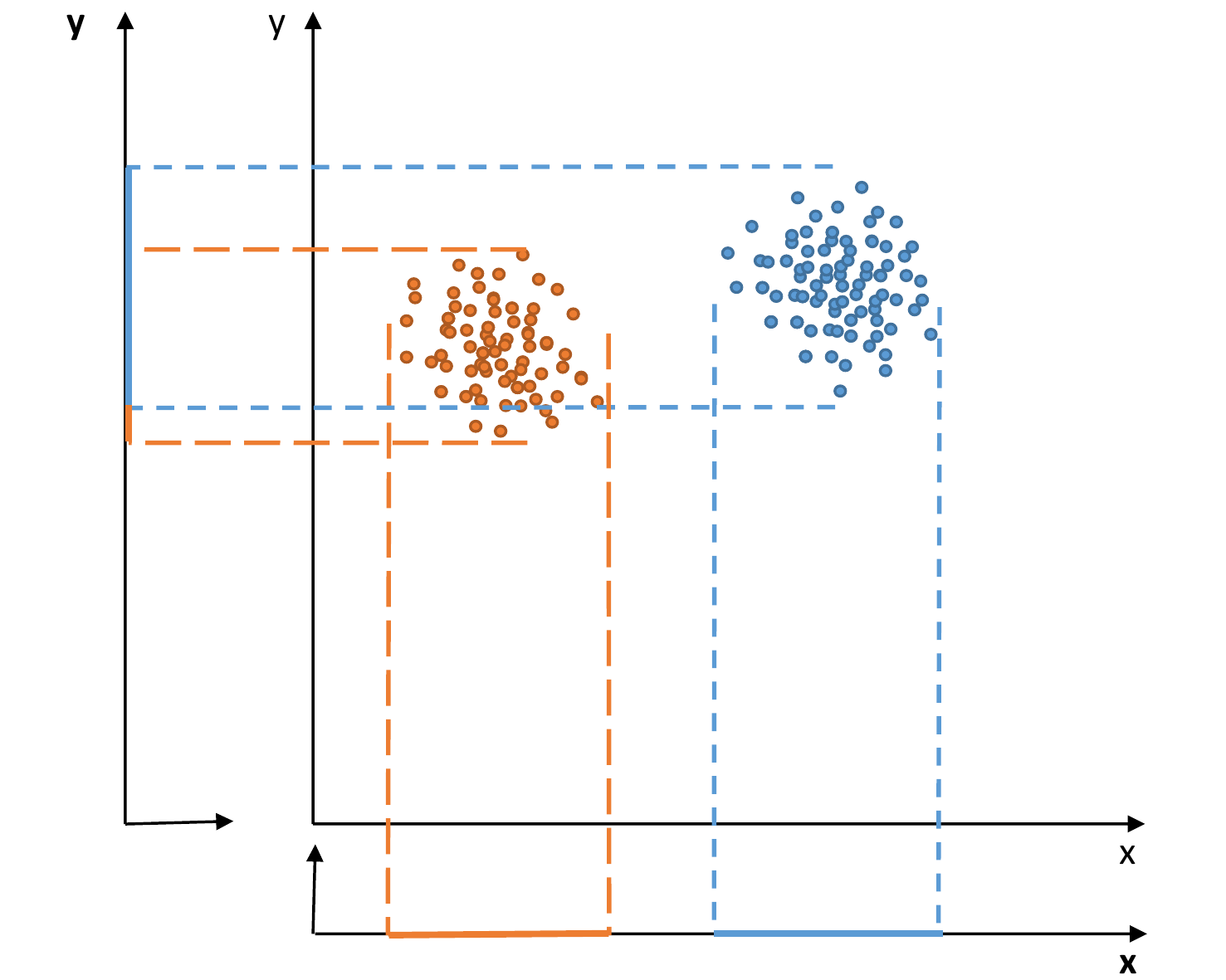}
\caption{(Left) In this example, features x and y are redundant, because feature x provides the same
information as feature y with regard to discriminating the two clusters. (Right) In this example, we consider feature y to be irrelevant, because if we omit x, we have only one class, which is uninteresting.}\label{fig2:redundant}
\end{figure}

Each feature selection method can be also classified as \textit{Supervised} or \textit{Unsupervised}. For supervised learning, feature selection algorithms maximize some function of predictive accuracy. Because class labels are given, it is natural to keep only the features that are related to or lead to these classes. Generally, feature selection for supervised machine learning tasks can be accomplished on the basis of the following underlying hypothesis: ``\emph{a good feature subset is one that contains features highly correlated with (predictive of) the class, yet uncorrelated with (not predictive of) each other} \cite{Gennari:1989}". A feature which is highly correlated with the class may be defined as ``relevant", whereas a feature which is uncorrelated with the others as not ``redundant". Redundant features are those which provide no more information than the currently selected features, and irrelevant features provide no useful information in any context. Figure~\ref{fig2:redundant}(Left) shows an example of feature redundancy. Note that the data can be grouped in the same way using only either feature x or feature y. Therefore, we consider features x and y to be redundant. Figure~\ref{fig2:redundant}(Right) shows an example of an irrelevant feature. Observe that feature y does not contribute to class discrimination. Used by itself, feature y leads to a single class structure which is uninteresting. Note that irrelevant features can misguide classification results, especially when there are more irrelevant features than relevant ones.
 
In unsupervised learning, we are not given class labels. Unsupervised learning is a difficult problem. It is more difficult when we have to simultaneously find the relevant features as well. 

Among the most used FS strategies, \emph{Relief-F}~\cite{liu2008} is an iterative, randomized, and supervised approach that estimates the quality of the features according to how well their values differentiate data samples that are near to each other; it does not discriminate among redundant features (i.e., may fail to select the most useful features), and performance decreases with few data.
Similar problems affect SVM-RFE (RFE)~\cite{Guyon:2002}, which is a wrapper method that
selects features in a sequential, backward elimination manner, ranking high a feature if it strongly separates the samples by means of a linear SVM.

Batti \cite{Battiti1994} has developed the Mutual Information-Based Feature Selection (MIFS) criterion, where the features are selected in a greedy manner. Given a set of existing selected features, at each step it locates the feature $x_i$ that maximizes the relevance to the class. The selection is regulated by a proportional term $\beta$ that measures the overlap information between the candidate feature and existing features. In \cite{Zhang2011} the authors proposed a graph-based filter approach to feature selection, that constructs a graph in which each node corresponds to each feature, and each edge has a weight corresponding to mutual information (MI) between features connected by that edge. This method performs dominant set clustering to select a highly coherent set of features and then it selects features based on the multidimensional interaction information (MII). Another effective yet fast filter method is the \textit{Fisher} method~\cite{Quanquanjournals}, it computes a score for a feature as the ratio of interclass separation and intraclass variance, where features are evaluated independently, and the final feature selection occurs by aggregating the $m$ top ranked ones.  Other widely used filters are based on mutual information, dubbed \emph{MI} here \cite{Hutter:02feature}, which considers as a selection criterion the mutual information between the distribution of the values of a given feature and the membership to a particular class. Mutual information provides a principled way of measuring the mutual dependence of two variables, and has been used by a number of researchers to develop information theoretic feature selection criteria. Even in the last case, features are evaluated independently, and the final feature selection occurs by aggregating the $m$ top ranked ones. Maximum-Relevance Minimum-Redundancy criterion (MRMR) \cite{Peng05featureselection} is an efficient incremental search algorithm. Relevance scores are assigned by maximizing the joint mutual information between the class variables and the subset of selected features. The computation of the information between high-dimensional vectors is impractical, as the time required becomes prohibitive. To face this problem the mRMR propose to estimate the mutual information for continuous variables using Parzen Gaussian windows. This estimate is based on a heuristic framework to minimize redundancy and uses a series of intuitive measures of relevance and redundancy to select features. Note, it is equivalent to MIFS with $\beta = \frac{1}{n - 1}$, where $n$ is the number of features.

Selecting features in unsupervised learning scenarios is a much harder problem, due to the absence of class labels that would guide the search for relevant information. In this setting, an unsupervised method for feature selection is the Laplacian Score (LS)~\cite{HCN05a}, where the importance of a feature is evaluated by its power of locality preserving. In order to model the local geometric structure, this method constructs a nearest neighbor graph. LS algorithm seeks those features that respect this graph structure. Finally, for the embedded methods, the \emph{feature selection via concave minimization} (\emph{FSV})~\cite{Bradley98featureselection} is a quite used FS strategy, where the selection process is injected \emph{into} the training of an SVM by a linear programming technique.

\section{Scenario 2: Keystroke Dynamics}\label{ch2:ReID}

On May 24, 1844 Samuel Morse sent the first telegraph message ``\textit{What hath God wrought}" from the U.S. Capitol in Washington, D.C. to the Baltimore and Ohio Railroad outer depot in Baltimore, Maryland, a new era in long-distance communications had begun. Some years later, the telegraph revolution was in full swing and telegraph operators were a valuable resource. With experience, each operator developed their unique signature and was able to be identified simply by their tapping rhythm. As late as World War II the military transmitted messages through Morse Code, using a methodology called \textit{The Fist of the Sender}, Military Intelligence identified that an individual had a unique way of keying in a message's \textit{dots} and \textit{dashes}, creating a rhythm that could help distinguish ally from enemy.

Nowadays, biometric technologies are used to secure several electronic communications, including access control and attendance, computer and enterprise network control, financial and health services, government, law enforcement, and telecommunications. These systems can be logically divided into two, namely, enrollment phase and authentication/verification phase. During the enrollment phase user biometric data is acquired, processed and stored as reference file in a database. This is treated as a template for future use by the system in subsequent authentication operations. Authentication is the process of determining whether a user is, in fact, who they are declared to be. For example, in the information security world, user authentication is analogous to entering a \textit{password} for a given \textit{username}, which is an authentication form based on something that the user have. Instead, Biometric Authentication (BA) is performed via something the user \textit{is}. In fact, BA uniquely identifies the user through one or more distinguishing biological traits, such as fingerprints, hand geometry, earlobe geometry, retina and iris patterns, voice waves, keystroke dynamics, DNA and signatures. Among the many types of biometric authentication technologies this thesis focuses on \textit{keystroke biometrics}. Comprehension of the statistical and structural mechanisms governing these dynamics in online interaction (e.g., text chats) plays a central role in many contexts such as user identification, verification, profile development, and recommender systems. 

The behavioral biometric of keystroke dynamics uses the manner and rhythm in which a user types characters on a keyboard \cite{Deng2013,A2004,Shepherd1995}. The keystroke rhythms of a user are measured to develop a unique biometric template of the user's typing pattern for future authentication. Data needed to analyze keystroke dynamics is obtained by keystroke logging. Normally, all that is retained when logging a text chat session is the sequence of characters corresponding to the order in which keys were pressed and timing information is discarded. However, research is interested in using this keystroke dynamic information to verify or even try to determine the identity of the person who is producing those keystrokes. This is often possible because some characteristics of keystroke production are as individual as handwriting or a signature. In other words, keystroke dynamics is a behavioral measurement and it aims to identify users based on the typing of the individuals or attributes such as duration of a keystroke or key hold time, latency of keystrokes (inter-keystroke times), typing error, force of keystrokes etc.

Very simple rules can be used to recognize users of a given text sample, based on the analysis of \emph{stylometric} cues introduce in literature in the Authorship Attribution (AA) domain. Indeed, AA attempts date back to the 15th century\cite{ShalCSIS:2010}: since then, many stylometric cues have been designed, usually partitioned into five major groups: \emph{lexical, syntactic, structural, content-specific and idiosyncratic}~\cite{AbbW:2008}. Table~\ref{table2:FeatTaxonomy} is a synopsis of the features applied so far in the literature.
\begin{table*}[!t]
\centering
  \resizebox{1\textwidth}{!}{%
\begin{tabular}{|m{2.9cm}|m{3.7cm}|m{9.1cm}|m{3.3cm}|}
\hline
\textbf{Group}  &\textbf{Category} & \textbf{Description} & \textbf{References}\\\hline
\multirow{6}{*}{\textbf{Lexical}} & Word level & Total number of words (M), \# short words (/M), \# char in words (/C),  \# different words, percent
characters per word , frequency of stop words& \cite{ AbaS:2008, Iqbal:2011,   Oreb:2009,ShalCSIS:2010, Stamatatos_2009,Zheng:2006}
\\\cline{2-4}

&Character level& Total number of characters (C), \# uppercase characters (/C), \# lowercase characters(/C), \# digit characters(/C),
frequency of letters, frequency of special characters, percent character per message & \cite{ AbaS:2008,  Oreb:2009, Stamatatos_2009, Zheng:2006}\\  \cline{2-4}

&Character/Digit n-grams& Count of letter /digit n-gram (a, at, ath, 1 , 12 , 123) &\cite{ AbaS:2008, Stamatatos_2009, Zheng:2006}\\ \cline{2-4}

&Word-length distribution & Frequency distribution, average word length & \cite{ AbaS:2008, Iqbal:2011, Oreb:2009, ShalCSIS:2010, Stamatatos_2009, Zheng:2006} \\ \cline{2-4}										

&Vocabulary richness & Hapax legomena/dislegomena &\cite{ AbaS:2008,  Iqbal:2011,   Stamatatos_2009, Zheng:2006}\\\hline

\multirow{2}{*}{\textbf{Syntactic}} & Function words & Frequency of function words (of,  for,  to ) &\cite{  AbaS:2008,  Iqbal:2011, Oreb:2009, Stamatatos_2009, Zheng:2006}\\\cline{2-4}

 & Punctuation & Occurrence of punctuation marks (!, ?, : ), multi !!!, multi ??? &\cite{ AbaS:2008,   Iqbal:2011,    Oreb:2009, Stamatatos_2009, Zheng:2006}\\\cline{2-4}

 &Emoticons/Acronym& :-), L8R, Msg, :( , LOL &\cite{ Oreb:2009, ShalCSIS:2010, Stamatatos_2009}\\\hline

\textbf{Structural} & Message level & Has greetings, farewell, signature &\cite{ AbaS:2008, Iqbal:2011,  Oreb:2009,  Stamatatos_2009, Zheng:2006} \\
\hline

\textbf{Content-specific}  &Word n-grams & Bag of word n-gram, agreement  (ok, yeah, wow),  discourse markers/onomatopee  (uhm, but, oh, aaarrr, eh), \# stop words, \# abbreviations , word based gender/age, slang/ out of dictionary words &\cite{ AbaS:2008,  Iqbal:2011, Oreb:2009,  Stamatatos_2009, Zheng:2006} \\\hline

\textbf{Idiosyncratic} &Misspelled word& belveier instead of believer &\cite{ AbaS:2008, Iqbal:2011,   Oreb:2009, Stamatatos_2009}\\
\hline	

\hline
\end{tabular}}
  \caption{Authorship Attribution in Instant Messaging: features employed in the different approaches.}
  \label{table2:FeatTaxonomy}
\end{table*}
AA on chats is a recent application, and very few are the approaches. A nice review, that considers also emails and blog texts, is \cite{Stamatatos_2009}. In the typical AA approaches, stylometric features do not take into account timing information that is crucial to measure human dynamics like writing behaviors. For example, if user \textbf{A} types at $30$ words per minute, and the user at the keyboard, \textbf{B}, is typing at $75$ words per minute, the identification system will not re-identify and authenticate the user \textbf{B}. This form of test is based simply on raw speed, motivated by the fact that it is always possible for people to go slower than normal, but it is unusual for them to go twice their normal speed. The time to get to and depress a key (seek-time), and the time the key is held-down (hold-time) may be very characteristic for a person, regardless of how fast they are going overall. Most people have specific letters that take them longer to find or get to than their average seek-time over all letters, but which letters those are may vary dramatically but consistently for different people. Right-handed people may be statistically faster in getting to keys they hit with their right hand fingers than they are with their left hand fingers. Index fingers may be characteristically faster than other fingers to a degree that is consistent for a person day-to-day regardless of their overall speed that day \cite{journals/asc/KarnanAK11}.
In addition, sequences of letters may have characteristic properties for a person. In English, the word ``the" is very common, and those three letters may be known as a rapid-fire sequence and not as just three meaningless letters hit in that order. Common endings, such as ``ing", may be entered far faster than, say, the same letters in reverse order (``gni") to a degree that varies consistently by person. This consistency may hold and may reveal the person's native language's common sequences even when they are writing entirely in a different language, just as revealing as an accent might in spoken English \cite{Shen2013}.

Common typos may also be quite characteristic of a person, and there is an entire taxonomy of errors, such as the user's most common \textit{substitutions, reversals, drop-outs, double-strikes, adjacent letter hits, homonyms}, hold-length-errors \cite{ AbaS:2008, Iqbal:2011,   Oreb:2009, Stamatatos_2009}. Even without knowing what language a person is working in, by looking at the rest of the text and what letters the person goes back and replaces, these errors might be detected. Again, the patterns of errors might be sufficiently different to distinguish two people. Based on these and other cues it is possible to recognize a user from their typing patterns, this operation is usually performed by learning a model that allows to rank user's templates according to their similarity with respect to a query unknown user template. 

\subsection{Re-Identification \& Verification}

Biometric authentication \cite{Nanavati:2002:BIV:560404} is an automatic method that identifies a user or verifies the identity based upon the measurement of their unique physiological traits (face \cite{Voth2003}, palm \cite{Shu1.601756}, iris \cite{MaLi}, etc.) or behavioral characteristics (voice \cite{Kinnunen:2010:OTS:1640546.1640908}, handwriting, signature, keystroke dynamics \cite{Lin97}, etc.). During the authentication/verification phase user biometric data is acquired, and processed. The authentication decision shall be based on the outcome of a matching process of the newly presented biometric template (belonging to the \textit{Probe} set) to the pre-stored reference templates (the \textit{Gallery} set). Ranking plays a crucial role in authentication and verification. The matching process that leads to the identification decision is usually performed by ranking the users in the gallery set according to the similarity with the probe template \cite{Farenzena:CVPR10,Bazzani:CVIU13,Roffo2013,Roffo:icmi2014}. There are two basic approaches of recognition errors based on the Receiver Operating Characteristic curve and the Cumulative Match Characteristic curve discussed in section \ref{ch2:PerfMetrics}.

The advantages of keystroke dynamics are obvious in the information security world as it provides a simple natural method for increased computer security. Static keystroke analysis is performed on typing samples produced using predetermined text for all the individuals under observation. Dynamic analysis implies a continuous or periodic monitoring of issued keystrokes. It is performed during the log-in session and continues after the session. Over the years, researchers have evaluated different features/attributes, feature extraction, feature subset selection and classification methods in an effort to improve the recognition capabilities of keystroke biometrics. The feature extraction is used to characterize attributes common to all patterns belonging to a class. A complete set of discriminatory features for each pattern class can be found using feature extraction. Some widely used features are reported in Table~\ref{table2:FeatTaxonomy}. 

Many other feature extraction methods and features/attributes which make use of keystrokes have been proposed in the last ten years. Gaines et al. \cite{gates:1980} use statistical significance tests between 87 lowercase letter inter-key latencies to check if the means of the keystroke interval times are the same. Young and Hammon \cite{young1989method} experimented with the time periods between keystrokes, total time to type a predetermined number of characters, or the pressure applied to the various keys and used it to form the feature template.
Joyce and Gupta \cite{joyce1990identity} propose two additional login sequences: the user's first name and the last name as the feature subset. This improved the performance of the method considerably. Obaidat and Sadoun \cite{obaidat1997verification} suggest inter-key and key hold times to be recorded using Terminate
and Stay resident (TSR) program in MS-DOS based environment. The standard keyboard interrupt handler was replaced by a special scan codes to record the time stamp. Lin \cite{lin1997computer} suggests a modified latency measurement to overcome the limitation of negative time measure, i.e. when the second key is pressed before the release of the first key. William and Jan \cite{de1997enhanced} propose typing difficulty feature in the feature subset to increase categorization. Robinson and Liang \cite{robinson1998computer} use user's login string to provide a characteristic pattern that is used for identity verification. In \cite{monrose2002password}, the authors examine the use of keystroke duration and latency between keystrokes and combine it with the user's password. Monrose and Rubin \cite{monrose2000keystroke} propose that users can be clustered into groups comprising disjoint feature sets in which the features in each set are pair wise correlated. In \cite{wong2001enhanced}, box plot algorithm was used as a graphical display with many features and the features extracted was normalized using normal bell curve algorithm. Bergadano et al. \cite{bergadano2002user} use timing information to obtain the relative order of trigraphs. It is used to compare two different sets of sorted trigraphs and to measure the difference in the ordering between them. In \cite{magalhaes2005improved}, the users were asked to type the usual password, or passphrase twelve times to get the digraph for further processing.
Chen et al. \cite{loy2005development} extracted features from the frequency domain signal which include mean, root mean square, peak value, signal in noise and distortion, total harmonic distortion, fundamental frequency, energy, kurtosis, and skew ness. Nick and Bojan \cite{bartlow2006evaluating} incorporated shift-key pattern along with keystroke inter-key and hold times. Kenneth \cite{revett2007bioinformatics} proposes motif signature which is used for classification. In order to improve the quality of data Pilsung et al. \cite{kang2008improvement} and Sungzoon Cho and Seongseob Hwang \cite{cho2006artificial} proposed artificial rhythms which increase uniqueness and cues to increase consistency.
Hu et al. \cite{hu2008k} use trigraphs (three consecutively typed keys) also known as Degree of Disorder as features and normalize the distance to find the feature subset. Mariusz et al. \cite{rybnik2008keystroke} propose an approach to select the most interesting features and combine them to obtain viable indicator of user's identity. Christopher et al. \cite{leberknight2008investigation} used three stage software design process along with data capture device hardware together with pattern recognition techniques like Bayesian and Discrimination function for classification. They used keystroke pressure and duration as features. Woojin \cite{chang2006reliable} applies discrete wavelet transformation (DWT) to a user's keystroke timing
vector (KTV) sample in the time domain, and then produces the corresponding keystroke wavelet coefficient vector (KWV) in the wavelet domain. In \cite{giot2009keystroke}, a comparative study of many techniques
considering the operational constraints of use for collaborative systems was made. Majority of the studies have identified three fundamental attributes: duration, latency and digraph. Many statistical properties of the attributes such as mean, standard deviation and Euclidean distance are measured and are used to construct the user reference profile. Each method has its own pros and cons as the number of test subjects differs.

\section{Scenario 3: On Recommender Systems}\label{ch2:Recommendation}

No other technology penetrated our everyday life as quickly and ubiquitously as social media. Billions of users are using these platforms every day, exploiting novel communication means. Among the computer-mediated technologies, \textit{Facebook}, \textit{Twitter}, \textit{YouTube}, \textit{Google+}, \textit{Skype}, and \textit{WhatsApp} are the most pervasive. In social media recommender systems play an increasingly important role. The main reason is that social media allow effective user profiling. For example, Facebook, MySpace, LinkedIn, and other social networks use collaborative filtering to recommend new friends, groups, and other social connections (by examining the network of connections between a user and their friends). Twitter uses many signals and in-memory computations for recommending who to follow to its users. The goal of recommender systems is to estimate a user’s preference and deliver an ordered list of items that might be preferred by the given user. 

In other words, the recommendation problem can be defined as estimating the response of a user for new items, based on historical information stored in the system, and suggesting to this user novel and original items for which the predicted response is high. The type of user-item responses varies from one application to the next, and falls in one of three categories: scalar, binary and unary. Scalar responses, also known as ratings, are numerical (e.g., 1-5 stars) or ordinal (e.g., strongly agree, agree, neutral, disagree, strongly disagree) values representing the possible levels of appreciation of users for items. Binary responses, on the other hand, only have two possible values encoding opposite levels of appreciation (e.g., like/dislike or interested/not interested). Finally, unary responses capture the interaction of a user with an item (e.g., purchase, online access, etc.) without giving explicit information on the appreciation of the user for this item. Since most users tend to interact with items that they find interesting, unary responses still provide useful information on the preferences of users. Ranking items, according to predicted responses, and selecting a subset from the top of the ranked list produces a \textit{recommendation}. 

One approach to the design of recommender systems that has wide use is user-based collaborative filtering \cite{Breese:1998:EAP:2074094.2074100}. These methods are based on collecting and analyzing a large amount of information on users' behaviors, activities or preferences and predicting what users will like based on their similarity to other users. Many algorithms have been used in measuring user similarity or item similarity in recommender systems. For example, the k-nearest neighbor (k-NN) approach and the Pearson Correlation as first implemented by Allen \cite{reference/rsh/DesrosiersK11}. 
\begin{figure*}[t]
\centering
\includegraphics[width=1.0\textwidth]{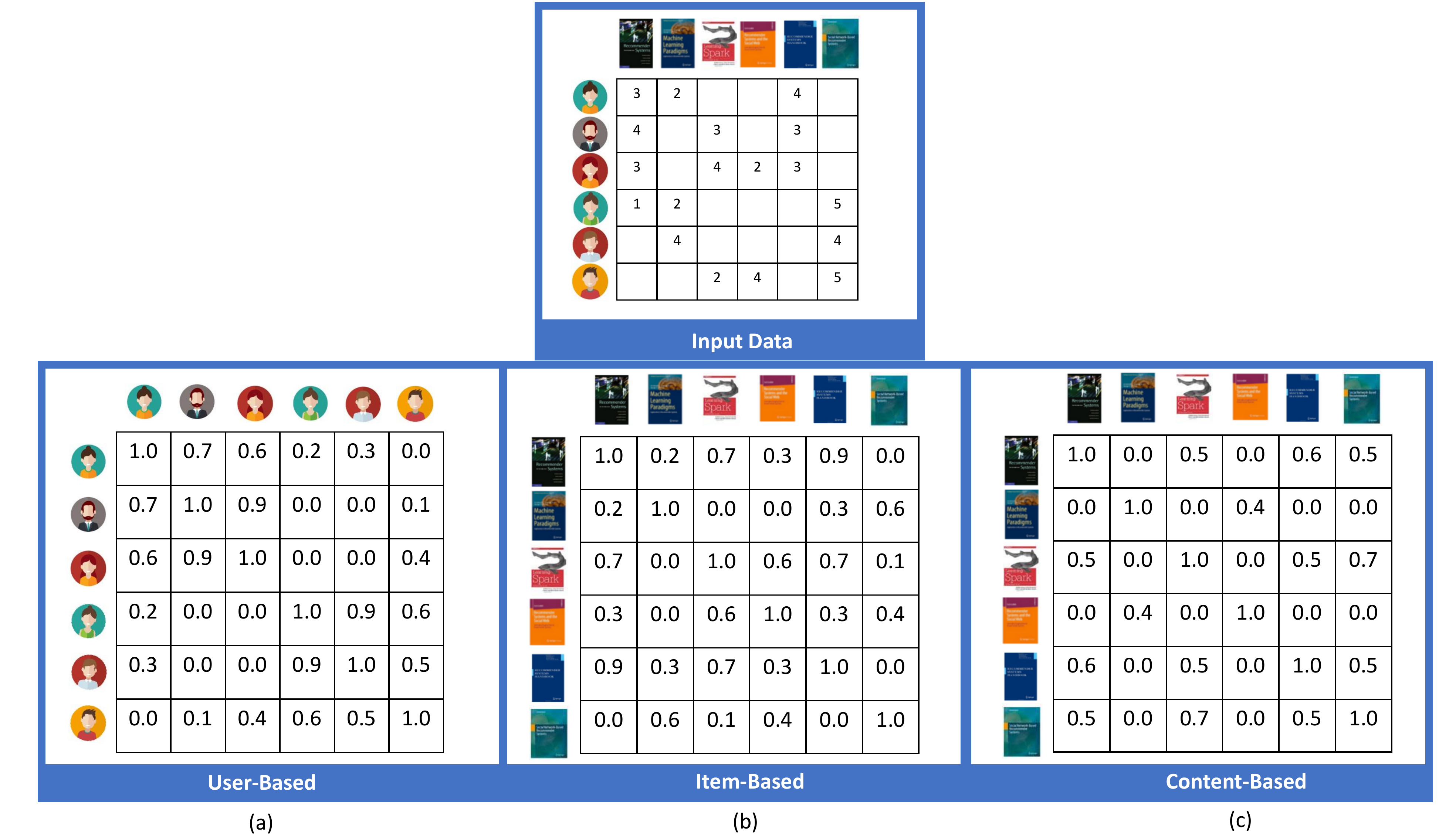}
\caption{An overview of collaborative filtering (CF) approaches. Given a sparse input matrix, in (a) the similarity matrix computed by a user-based CF. In (b) the matrix produced by an item-based CF, and in (c) the output of a content-based system.}\label{fig2:RecSys}
\end{figure*}
Other common approaches when designing recommender systems are item-based and content-based systems. The general principle of content-based (or cognitive) approaches \cite{Billsus:1998:LCI:645527.657311,Billsus:2000:UMA:598285.598352,Pazzani:1997:LRU:261092.261098} is to identify the common characteristics of items that have received a favorable rating from a user, and then recommend to the user new items that share these characteristics. User-based collaborative approaches overcome some of the limitations of content-based ones. For instance, items for which the content is not available or difficult to obtain can still be recommended to users through the feedback of other users. Furthermore, collaborative recommendations are based on the quality of items as evaluated by peers, instead of relying on content that may be a bad indicator of quality. Finally, unlike content-based systems, collaborative filtering (user-based) ones can recommend items with very different content, as long as other users have already shown interest for these different items. 

In typical commercial recommender systems, where the number of users far exceeds the number of available items, item-based approaches are typically preferred since they provide more accurate recommendations, while being more computationally efficient and requiring less frequent updates. On the other hand, user-based methods usually provide more original recommendations, which may lead users to a more satisfying experience. However, a common characteristic of all these approaches is the construction of a similarity matrix which encodes all the information needed to perform the final ranking. A user-based collaborative filtering will build a $users \times users$ similarity matrix (Figure \ref{fig2:RecSys}.\textbf{a}), while item-based and content-based will produce $items \times items$ and $contents \times contents$ matrices respectively (see Figure \ref{fig2:RecSys}.\textbf{b}-\textbf{c}). Based on these similarity matrices, the basic assumption is that people who agreed in the past will agree in the future, and that they will like similar kinds of items (or contents) as they (or similar users) liked in the past. 

Among collaborative recommendation approaches, methods based on nearest-neighbors still enjoy a huge amount of popularity, due to their simplicity, their efficiency, and their ability to produce accurate and personalized recommendations. An advantage of such an approach is that it can obtain meaningful relations between pairs of users or items, even though these users have rated different items, or these items were rated by different users. 

In chapter \ref{ch8:ADS}, we proposed a neighborhood-based approach, where deep convolutional neural networks are used to estimate the similarity among pairs of users. Then, very accurate rankings are produced in a task of advert recommendation.

\section{Performance Evaluation Measures}\label{ch2:PerfMetrics}

Evaluation is important in assessing the effectiveness of methodologies and algorithms. In this section we provide an overview of the most commonly used performance evaluation measures.

\subsection{Precision and Recall}\label{ch2:precRec}

We start with two of the simplest performance measures, \textit{precision} and \textit{recall} \cite{Sokolova2006,davis2006}. Precision and recall have been used to measure the accuracy of a system in different application domains such as classification, information retrieval, and recommender systems. \newline Suppose that we deal with an information retrieval task. Let us consider the task of retrieving the right user from a database of known users. $U_q \in P$ (Probe set) is a query user-template, and $U_i$ the $i^{th}$ user in some user collection $G$ (i.e., Gallery set). Let $rel_{U_q}(U_i) \in \{0, 1\}$ denote the relevance. Further, let $R_{U_q} = (U_{r_1}, ... , U_{r_n})$ denote the ranking returned for this query by the retrieval system being evaluated. The ranking is performed on the user collection $G$ where, for each user, more than one example is given. In other words, given a user $U_q$ from $P$, there are multiple relevant examples in $G$ that satisfies the query $q =$ ``\textit{who is $U_q$?}". In this scenario, the relevant users are simply those that belong to the relevant category (i.e., have the same identity). Recall is defined as the number of relevant users retrieved by a search divided by the total number of existing relevant users, while precision is defined as the number of relevant users retrieved by a search divided by the total number of users retrieved by that search. 
\begin{table}[!t]
\small
\centering
\begin{tabular}{ l | c | c  }
&   Retrieved & Not Retrieved  \\\hline
 Relevant & True-Positive (tp) & False-Negative (fn) \\
 Not Relevant & False-Positive (fp) & True-Negative (tn)\\
\end{tabular} 
\caption{Classification of the possible results of a retrieval task }
\label{tab:table1x22}
\end{table} 
Given the rank $R_{U_q}$, Table ~\ref{tab:table1x22} reports the set of retrieved users that can be found at the top of the ranked list $R_{U_q} = (U_{r_1}, ... , U_{r_k})$, while the not retrieved at the tail $R_{U_q} = (U_{r_{k+1}}, ... , U_{r_n})$. We can count the number of examples that fall into each cell in the table and compute the following quantities: 
\[
\textbf{Precision}  \qquad \qquad  \frac{tp}{tp + fp}, 
\]
\[
\textbf{Recall (True Positive Rate)} \qquad  \frac{tp}{tp + fn}.
\]
Recall is also referred to as the True Positive Rate (TPR) or \textit{Sensitivity}, especially when it is evaluated on different amounts of retrieved items, and precision is also referred to as positive predictive value (PPV). Other related measures used include true negative rate and accuracy:
\[
\textbf{False Positive Rate (1 - Specificity)} \qquad  \frac{fp}{fp + tn},
\]
\[
\textbf{Accuracy} \qquad  \frac{tp + tn}{tp + tn + fp + fn},
\]
where true negative rate is also called \textit{Specificity}. We can expect a trade-off between these quantities; while allowing longer retrieved lists typically improves recall, it is also likely to reduce the precision. 

Precision and Recall are important in assessing the effectiveness of recommendation systems and many classification algorithms as well. Let us consider a binary classification problem, in which the outcomes are labeled either as positive (p) or negative (n). Again, there are four possible outcomes from a binary classifier. If the outcome from a prediction is $p$ and the actual value is also $p$, then it is called a true positive (tp). However if the actual value is $n$ then it is said to be a false positive (fp). Conversely, a true negative (tn) has occurred when both the prediction outcome and the actual value are $n$, and false negative (fn) is when the prediction outcome is $n$ while the actual value is $p$.

It is always possible to compute curves comparing precision to recall and true positive rate to false positive rate. Curves of the former type are known simply as precision-recall curves, while those of the latter type are known as a Receiver Operating Characteristic or ROC curves (see section \ref{ch2:ROC}). The precision-recall plot is a model-wide evaluation measure. The precision-recall plot uses recall on the x-axis and precision on the y-axis (see Figure \ref{fig2:example}). Recall is identical with sensitivity, and precision is identical with positive predictive value. A precision-recall point is a point with a pair of x and y values in the precision-recall space where x is recall and y is precision. A precision-recall curve is created by connecting all precision-recall points of a classifier. Two adjacent precision-recall points can be connected by a straight line.

Measures that summarize the precision and recall are useful for comparing algorithms independently of application. For example, the area under the precision-recall curve (AUC-pr). AUC-pr is probably the second most popular metric, after accuracy. The reason is that accuracy deals with binary classification outputs, meaning it compares the binary outputs of a classifier (ones or zeros) with the ground truth. But many classifiers are able to quantify their uncertainty about the answer by outputting a probability value. To compute accuracy from probabilities we need a threshold to decide when zero turns into one. The most natural threshold is of course 0.5. Let us suppose to have a quirky classifier. It is able to get all the answers right, but it outputs 0.7 for negative examples and 0.9 for positive examples. Clearly, a threshold of 0.8 would be perfect.
That is the whole point of using AUC-pr, it considers all possible thresholds. Various thresholds result in different true positive/false positive rates. As you decrease the threshold, you get more true positives, but also more false positives. 

A second widely used score that considers both the precision and the recall of the test is the F-Score:  
\[
\textbf{F-Score} = \frac{2 \cdot Precision \cdot Recall}{Precision + Recall},
\]
which is the harmonic mean of precision and recall \cite{fawcett2006introduction}.

\subsubsection{Interpretation of precision-recall curves}\label{sec2:precrec}
\begin{figure}[!t]
    \centering
    \begin{subfigure}[b]{0.44\textwidth}
        \includegraphics[width=\textwidth]{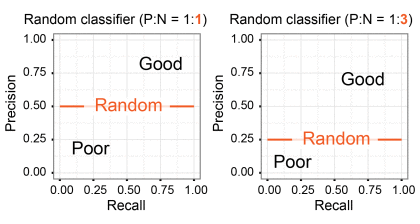}
        \caption{Random Classifier}
        \label{fig2:rand}
    \end{subfigure}
    ~ 
    \begin{subfigure}[b]{0.44\textwidth}
        \includegraphics[width=\textwidth]{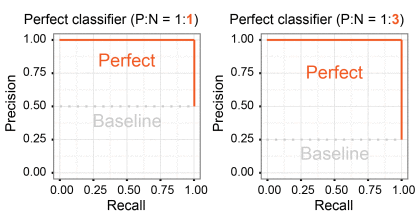}
        \caption{Perfect Classifier}
        \label{fig2:perfect}
    \end{subfigure}
    ~ 
    \begin{subfigure}[b]{0.4\textwidth}
        \includegraphics[width=\textwidth]{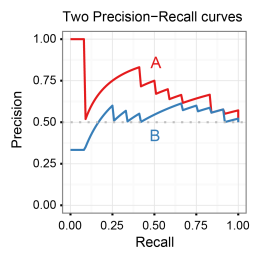}
        \caption{Comparison}
        \label{fig2:example}
    \end{subfigure}
    \caption{Interpretation of Precision-Recall curves. (a) A random classifier shows a straight line as $\frac{p}{p+n}$. (b) A perfect classifier shows a combination of two straight lines. The end point depends on the ratio of positives and negatives. (c) Two precision-recall curves represent the performance levels of two classifiers A and B. Classifier A clearly outperforms classifier B in this example.}\label{fig2:PRCurves}
\end{figure}
We propose several examples to explain how precision-recall curves can be interpreted. A classifier with the random performance level shows a horizontal line as $\frac{p}{p+n}$. This line separates the precision-recall space into two areas (see Figure \ref{fig2:rand}). The separated area above the line is the area of good performance levels. The other area below the line is the area of poor performance. For instance, in Figure \ref{fig2:rand} the line is $y = 0.5$ when the ratio of positives and negatives is $1:1$, whereas $0.25$ when the ratio is $1:3$. 

In Figure \ref{fig2:perfect}, a classifier with the perfect performance level that shows a combination of two straight lines – from the top left corner (0.0, 1.0) to the top right corner (1.0, 1.0) and further down to the end point (1.0, $\frac{p}{p+n}$). For example, in Figure \ref{fig2:perfect}, the end point is (1.0, 0.5) when the ratio of positives and negatives is 1:1, whereas (1.0, 0.25) when the ratio is 1:3. It is easy to compare several classifiers in the precision-recall plot (see Figure \ref{fig2:example}). Curves close to the perfect precision-recall curve have a better performance level than the ones closes to the baseline. In other words, a curve above the other curve has a better performance level. 

\subsection{Receiver Operating Characteristic Curve}\label{ch2:ROC}

\begin{figure*}[t]
\centering
\includegraphics[width=0.8\textwidth]{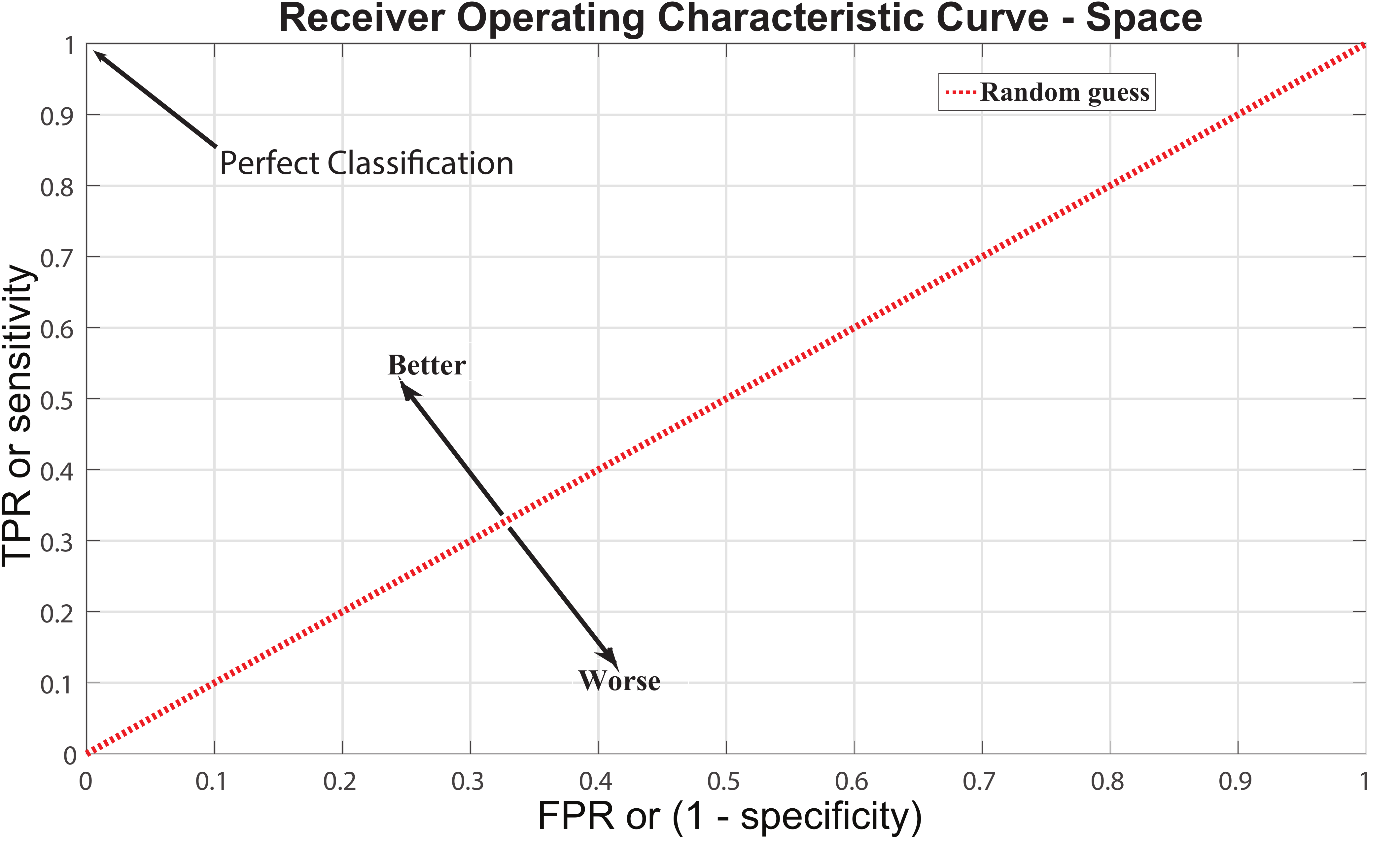}
\caption{An illustration of a ROC space that depicts relative trade-offs between true positive and false positive. The best possible prediction method would yield a point in the upper left corner or coordinate of the ROC space.}\label{fig2:rocspace}
\end{figure*}

A Receiver Operating Characteristic (ROC) Curve \cite{fawcett2006introduction,BAMBER1975}, is a graphical plot that illustrates the performance of a binary classifier or retrieval system as its discrimination threshold is varied. The curve is created by plotting the true positive rate (TPR) against the false positive rate (FPR) at various threshold settings (see Figure \ref{fig2:rocspace}). A widely used measurement that summarizes the ROC curve is the \textbf{Area Under the ROC Curve (AUC-roc or simply AUC)}~\cite{fawcett2006introduction,BAMBER1975} which is useful for comparing algorithms independently of application. 
In information retrieval, When evaluating ROC curves for multiple queries, a number of strategies can be employed in aggregating the results, depending on the application at hand. The usual manner in which ROC curves are computed in the community~\cite{Harman02overviewof,Roffo2013,Sarwar:2000,Schein:2002} is to average the resulting curves over queries. Such a curve can be used to understand the global trade-off between false positives and false negatives of the system. The same strategy is used for precision-recall curves as well.

One of the most important problems associated with recommender systems is the top-N recommendation problems. In this scenario, Precision-Recall and ROC curves measure the proportion of preferred items that are actually recommended. In particular, precision-recall curves emphasize the proportion of recommended items that are preferred, while ROC curves emphasize the proportion of items that are not preferred that end up being recommended. 
Generally, given two algorithms, we can compute a pair of such curves, one for each algorithm. If one curve completely dominates the other curve, the decision about the superior algorithm is easy. However, when the curves intersect, the decision is less obvious, and will depend on the application in question. Knowledge of the application will dictate which region of the curve the decision will be based on.

\subsubsection{Mean Average Precision}

Precision and recall are single-value metrics based on the whole list of objects returned by the system. For systems that return a ranked sequence of objects, it is desirable to also consider the order in which the returned objects are presented. By computing a precision and recall at every position in the ranked sequence of objects, one can plot a precision-recall curve, plotting precision $p(r)$ as a function of recall $r$. Average precision computes the average value of $p(r)$ over the interval from $r=0$ to $r=1$, that is the area under the precision-recall curve. This integral is in practice replaced with a finite sum over every position in the ranked sequence of objects:
\[
\textbf{AP} = \sum_{k=1}^n \Big( p(k) \Delta r(k)  \Big),
\]
where $k$ is the rank in the sequence of retrieved objects, $n$ is the number of retrieved objects, $P(k)$ is the precision at cut-off $k$ in the list, and $\Delta r(k)$ is the change in recall from items $k-1$ to $k$. This finite sum is equivalent to:
\[
\textbf{AP} = \frac{\sum_{k=1}^n \Big( p(k) rel(k)  \Big)}{tp},
\]
where $\operatorname{rel}(k)$ is an indicator function equaling 1 if the item at rank $k$ is a relevant object, zero otherwise. Note that the average is over all relevant objects and the relevant objects not retrieved get a precision score of zero.

Mean average precision is used for a set of queries. It is the mean of the average precision scores for each query.
\[
\textbf{mAP} = \frac{\sum_{q=1}^Q AP(q)}{|Q|},
\]
where $Q$ is the number of queries. 

\subsection{Cumulative Match Characteristic Curve}\label{ch2:CMC}

In the context of biometric verification systems, performance evaluation is achieved through estimating the FPR and the TPR and using these estimates to construct a ROC curve, described above, that expresses the trade-off between the FPR and TPR. This is used for the so called \textit{One-to-One } $(1:1)$ identification systems, whereas the task is verification. Verification performance is defined as the ability of the system in verifying if the biometric template that the probe user claims to be is truly themselves. The ROC is a well-accepted measure to express the performance of $1:1$ matchers. On the other hand, a Cumulative Match Characteristic curve (CMC) judges the ranking capabilities of biometric identification systems and it is used as a measure of \textit{One-to-Many} $(1:N)$ identification system performance \cite{Bolle2005}. Identification tasks involve a comparison against the entire biometric database. The CMC is an effective performance measure for identification systems. 

Specifically, when a system always returns the identities associated with the $K$ highest-scoring biometric samples from an enrollment database (gallery). To estimate the CMC, the match scores between a query sample and the $N$ biometric samples in the database are sorted. The lower the rank of the genuine matching biometric in the gallery, the better the $1:N$ identification system.

There is a relationship between the ROC and the CMC associated with a $1:1$ matcher. That is, given the ROC or given the FPR/TPR of a $1:1$ matcher, the CMC can be computed and expresses how good this particular $1:1$ matcher is at sorting galleries with respect to input query samples.  

The gallery is an integral part of a $1:N$ search engine because without an enrollment database there is no identification system. Therefore, before discussing the CMC curve estimation, we introduce the main concepts beyond this evaluation metric. Given a large database of biometric templates (samples) $S_i$ with associated ground truth $GT(S_i)$. Key to measuring a CMC curve associated with a 1:1 matcher is the assembly of two subsets of samples: a \textit{gallery set} $G$ which contains $M$ biometrics samples of different subjects, and a \textit{probe set} $P$ which is a set of $N$ unknown samples associated with the $M$ subjects.

The probe (or from an information retrieval perspective: query set) can be formed of any set of individuals, but usually probe identities are presumed to be in the gallery set $G$. Given a query biometric $S_q \in P$ and a biometric sample $S_i$ from $G$, the output of a biometric matcher is a similarity score $s(S_q,S_i)$. In order to estimate the Cumulative Match Curve, each probe biometric is matched to every gallery biometric obtaining a similarity matrix $N \times M$. The scores for each probe biometric are ordered, so as to obtain the most similar samples at the top of the ranked list.
\begin{figure*}[t]
\centering
\includegraphics[width=0.8\textwidth]{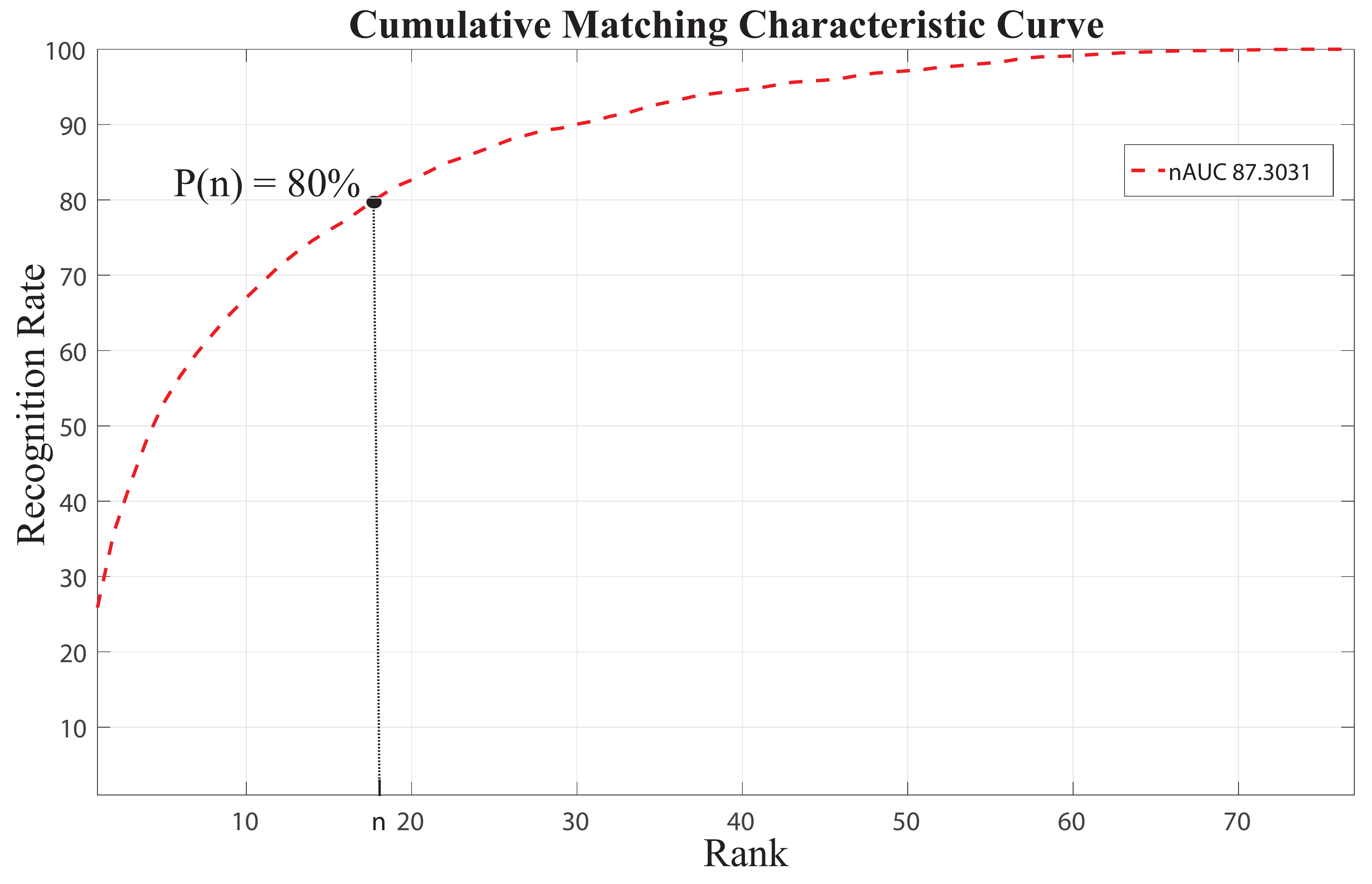}
\caption{A Cumulative Match Characteristic Curve. The value of the curve at position $n$ is the probability of finding the correct match in the first n positions of the output ranked list.}\label{fig2:CMCsExample}
\end{figure*}

\subsection{Estimating the CMC}

Given a set of $n$ ranked estimates $R = (Up_{n_1}, ... , Up_{n_M})$, where each $Up_{n_i}$ is an estimate of a probe biometric $S$, each rank is defined only if the correct identity is in the ordered list of gallery biometrics $G$. The Cumulative Match Curve estimates the distribution of the ranks $Up_{n_i}$  of probes $P$ and the curve $CMC(n)$ is the fraction of probe biometrics $S$ that have $n_i < n$. 
\[
    CMC(n) = \frac{1}{M} \sum_{i=1}^{M} n_i < n
\]
A CMC curve represent the discrete Rank Probability Mass (RPM) function (see Figure \ref{fig2:CMCsExample}). The discrete rank probabilities $P(n), n=1,...,M$ of a biometric search engine are probabilities (summing to $1$) that the identity associated with a probe has rank $n$. Therefore, the CMC estimate is the distribution of the estimated ranks $n_i$ and estimates the probability $P( n_i < n)$, where $n_i$ takes on discrete values $1,2,...,M$. 

\subsection{RMSE, MSE, and MAE}

In most online advertising platforms the allocation of adverts is dynamic, tailored to user interests based on their observed feedback. In this scenario, recommender systems want to predict the feedback a user would give to an advert (e.g. 1-star through 5-stars). In such a case, the goal is to measure the accuracy of the system's predicted ratings. \textbf{Root Mean Squared Error (RMSE)} is perhaps the most popular metric used in evaluating the accuracy of predicted ratings. The system generates predicted ratings $\hat r_{u,a}$ for a test set $T$ of user-advert pairs (u,a) for which the true ratings $r_{u,a}$ are known. The RMSE between the predicted and actual ratings is given by:

\begin{equation}\label{eq:RMSE}
	RMSE = \sqrt{\frac{1}{|T|}  \sum_{(u,a) \in T} {(\hat r_{u,a} - r_{u,a})}^2  }.
\end{equation}

\textbf{Mean square error (MSE)} is an alternative version of RMSE, the main difference between these two estimators is that RMSE penalizes more large errors, and MSE has the same units of measurement as the square of the quantity being estimated, while RMSE has the same units as the quantity being estimated. 
Therefore, MSE is given by

\begin{equation}\label{eq:MSE}
	MSE = \frac{1}{|T|}  \sum_{(u,a) \in T} {(\hat r_{u,a} - r_{u,a})}^2.
\end{equation}

\textbf{Mean Absolute Error (MAE)} is a popular alternative, given by
\begin{equation}\label{eq:MAE}
	MAE = \sqrt{\frac{1}{|T|}  \sum_{(u,a) \in T} |{\hat r_{u,a} - r_{u,a}}|  }.
\end{equation}
As the name suggests, the MAE is an average of the absolute errors $err_{u,a}=|{\hat r_{u,a} - r_{u,a}}|$, where $\hat r_{u,a}$ is the prediction and $r_{u,a}$ the true value. The MAE is on same scale of data being measured.

\section{Conclusions}

Ranking is a pervasive operation used in many different scenarios and settings. In this chapter we surveyed some of the possible domains where ranking is a fundamental task. Among these, we presented scenarios where ranking is used to improve the learners accuracy and enhance their generalization properties, and other settings in which learning is used to rank objects with the aim to retrieve them from a database, or to perform recommendations. We also provided a brief and compact introduction to standard performance metrics used for evaluating algorithms in those scenarios.


\chapter{Deep Convolutional Neural Nets: An Overview}\label{ch3:DeepLearning}

Over last decade, successful results obtained in solving pattern recognition problems (i.e., from vision to language) have been heavily based on hand-crafted features. Obviously, performance of algorithms relied crucially on the features used. As a result, progress in pattern recognition was based on hand-engineering better sets of features. Over time, these features started becoming more and more complex - resulting in difficulty with coming up with better, more complex features. From a methodological perspective, there were principally two steps to be followed: \textit{feature design} and \textit{learning algorithm design}, both of which were largely independent. Meanwhile, research in the machine learning community has been focusing on learning models which incorporated learning of features from raw data. These models typically consisted of multiple layers of non-linearity. This property was considered to be very important and lead to the development of the first deep learning models (e.g., Restricted Boltzmann Machines \cite{Hinton:2002}, Deep Belief Networks \cite{Hinton:2006}). Nowadays, deep learning is expanding the scope of machine learning by making algorithms less dependent on feature engineering. These techniques offers a compelling alternative: \textit{the automatic learning of problem specific features}. 

In 2012 at the Large Scale Visual Recognition Challenge (ILSVRC2012), for the first time, a deep learning model called ``Convolutional Neural Network" (CNN) \cite{KrizhevskyNIPS2012} decreased the error rate on the image classification task by half for the first time, outperforming traditional hand-engineered approaches. The network designed by Alex Krizhevsky in \cite{KrizhevskyNIPS2012}, popularly called ``\textit{AlexNet}" will be used and modified in the coming years for various vision problems.  

This chapter introduces Convolutional Neural Networks (CNN) from a computer vision perspective. There are two principal reasons supporting the choice of this domain. Firstly, CNNs are biologically inspired models, designed to emulate the behaviour of the human visual system. Secondly, CNNs have emerged as an active and growing area of research, predominantly in computer vision \cite{Srinivas2015}, and every problem is now being re-examined from a deep learning perspective. In the next chapter we will discuss the proposed architecture suit to solve variable ranking for the task of Re-Identification. 

\section{A Biologically-Inspired Model}

\begin{figure*}[ht]
\centering
\includegraphics[width=0.8\textwidth]{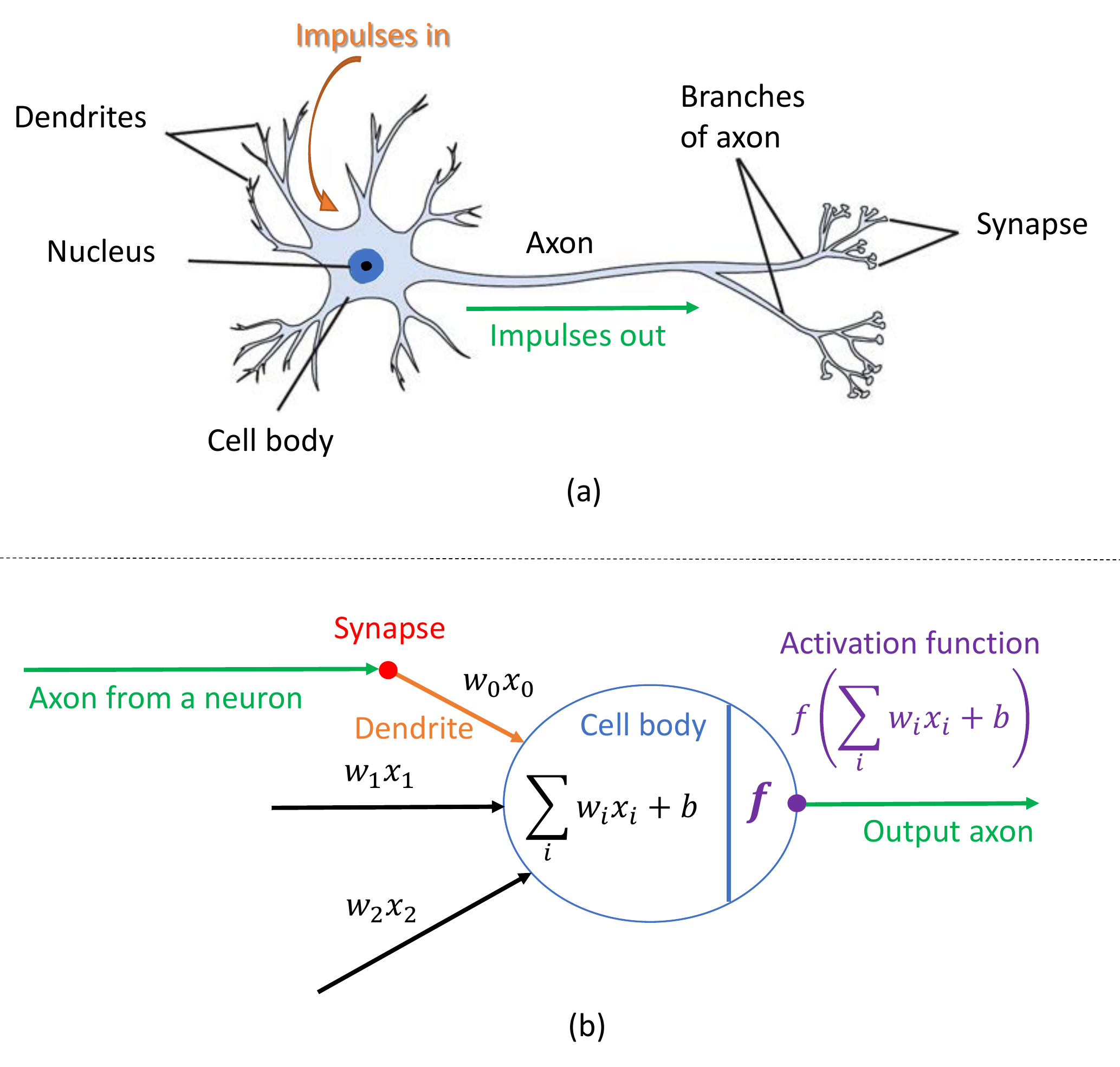}
\caption{A comparison between a human neuron and an ANN neuron. (a) Shows an illustration of a human neuron; (b) reports the associated artificial neuron, where synapses are modeled by the set of inputs $x_{1:n}$. The cell body is modeled by the biological counterpart functionality, that is in collecting together weighted inputs and filter them throughout an activation function.}\label{fig3:neurons}
\end{figure*}

The history of biologically inspired algorithms stretches far back into the history of computing. McCulloch and Pitts in \cite{McCulloch1943} formalized the notion of an ``integrate and fire" neuron in 1943, and Hebb \cite{hebb49} first proposed the idea of associative learning in neurons, ``what fires together, wires together", in the late 1940s \cite{McCulloch1943,McCulloch1988,hebb49}. One of the earliest instantiations of a neural network which could learn was the \textit{perceptron} of Rosenblatt \cite{Rosenblatt58theperceptron,Rosenblatt62}. It corresponds to a two-class model in which the input vector $x$ is first transformed using a fixed nonlinear transformation to give a feature vector $\phi(x)$. This is then used to construct a generalized linear model of the form $y(x) = f(w^T\phi(x)))$, where the vector $\phi(x)$ will typically include a bias component $\phi_0(x) = 1$. In the context of neural networks, a perceptron is an \textit{artificial neuron} using the Heaviside step function as an activation function (see Figure \ref{fig3:neurons}.\textbf{b}). An example of this function $f(\cdot)$ is given by
\[ 
f(z) = \left\{ \begin{array}{ll}
         +1 & \mbox{if $z \geq 0$};\\
          0 & \mbox{if $z < 0$}.
         \end{array} \right. 
\] 
Figure \ref{fig3:neurons} reports an interesting comparison between a human neuron and the artificial model used in ANNs. 
The perceptron algorithm is also termed the single-layer perceptron, to distinguish it from a Multi-Layer Perceptron (MLP), which is a misnomer for a more complicated neural network. In fact, MLP is really a misnomer, because the model comprises multiple layers of logistic regression models (with continuous non-linearities) rather than multiple perceptrons (with discontinuous non-linearities). Therefore, an MLP can be viewed as a logistic regression classifier where the input is first transformed using a learnt non-linear transformation $\phi$. This transformation projects the input data into a space where it becomes linearly separable. This intermediate layer is referred to as a \textit{hidden layer}. A single hidden layer is sufficient to make MLPs a universal approximator \cite{Hornik:1991}. If an MLP has a linear activation function in all neurons, that is, a linear function that maps the weighted inputs to the output of each neuron, then it is easily verified with linear algebra that any number of layers can be reduced to the standard single-layer perceptron model. What makes an MLP different is that some neurons (also called hidden units) use a \textit{nonlinear activation function} which was developed to model the frequency of action potentials, or firing, of biological neurons in the brain. This function is modelled in several ways. The main activation functions (or rectifiers) used in the context of ANNs are
\begin{equation}\label{eq:nonlinearity}
 f(z) = \left\{  
        \begin{array}{ll}
            max(0,z) & \mbox{ReLU \cite{KrizhevskyNIPS2012}} \\
            tanh(z)  & \mbox{Hyperbolic tangent} \\
            \frac{1}{1 + e^{-z} } & \mbox{Sigmoid} \\
            ln(1 + e^{z}) & \mbox{Softplus \cite{DugasBBNG00}} \\
            \frac{e^{z_j}}{\sum_{k=1}^K e^{z_k}} & \mbox{Softmax \cite{Bishop2006}} \\
            max(w^T_1x+b_1,w^T_2x+b_2) & \mbox{Maxout \cite{goodfellowBengio2013maxout}} \\
        \end{array}  \right.
\end{equation}
In particular, the sigmoid function has seen frequent use historically since it has a favourable interpretation as the firing rate of a neuron: from not firing at all (0) to fully-saturated firing at an assumed maximum frequency (1). However, each activation function has its pros and cons, we will discuss these issues in section \ref{sec3:relu}. Each hidden unit is constituted of activation functions that control the propagation of neuron signal to the next layer (e.g. positive weights simulate the excitatory stimulus whereas negative weights simulate the inhibitory ones as in its biological counterpart). A hidden unit is composed of a regression equation that processes the input information into a non-linear output data. Therefore, if more than one neuron is used to compose an ANN, non-linear correlations can be treated.


\begin{figure*}[!h]
\centering
\includegraphics[width=0.95\textwidth]{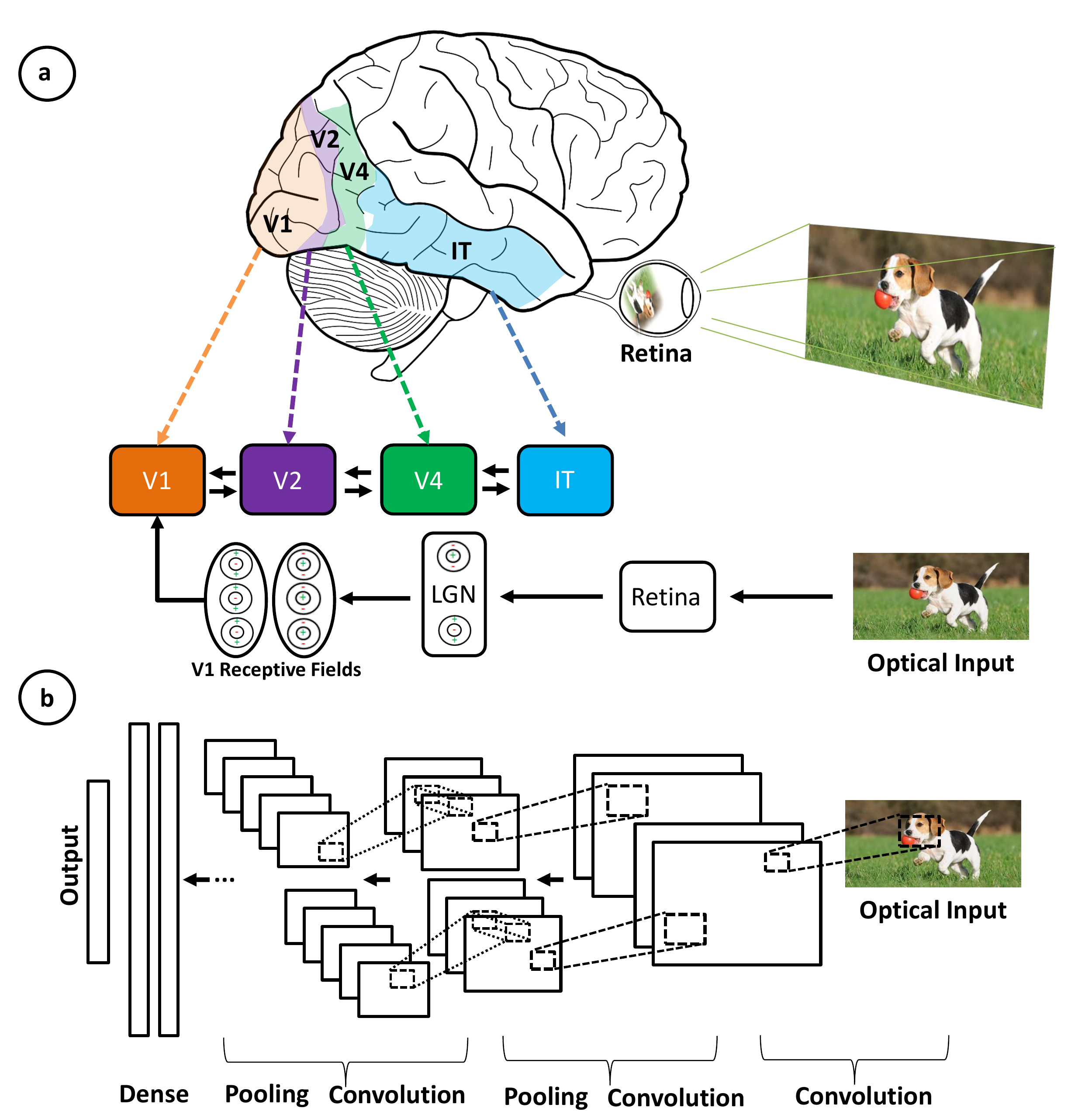}
\caption{Illustration of the corrispondence between the areas associated with the primary visual cortex and the layers in a convolutional neural network. (a) Four Brodmann areas associated with the ventral visual stream. The figure reports also a block diagram showing just a few of the many forward and backward projections between these areas. (b) The sketch of the \textit{AlexNet} convolutional neural network (adapted from \cite{KrizhevskyNIPS2012}) in which pairs of convolution operator followed by a max pooling layer are roughly analogous to the hierarchy of the biological visual system.}\label{fig3:ReceptiveFields}
\end{figure*}

\section{CNNs Architecture}

\begin{figure*}[!]
\centering
\includegraphics[width=1.0\textwidth]{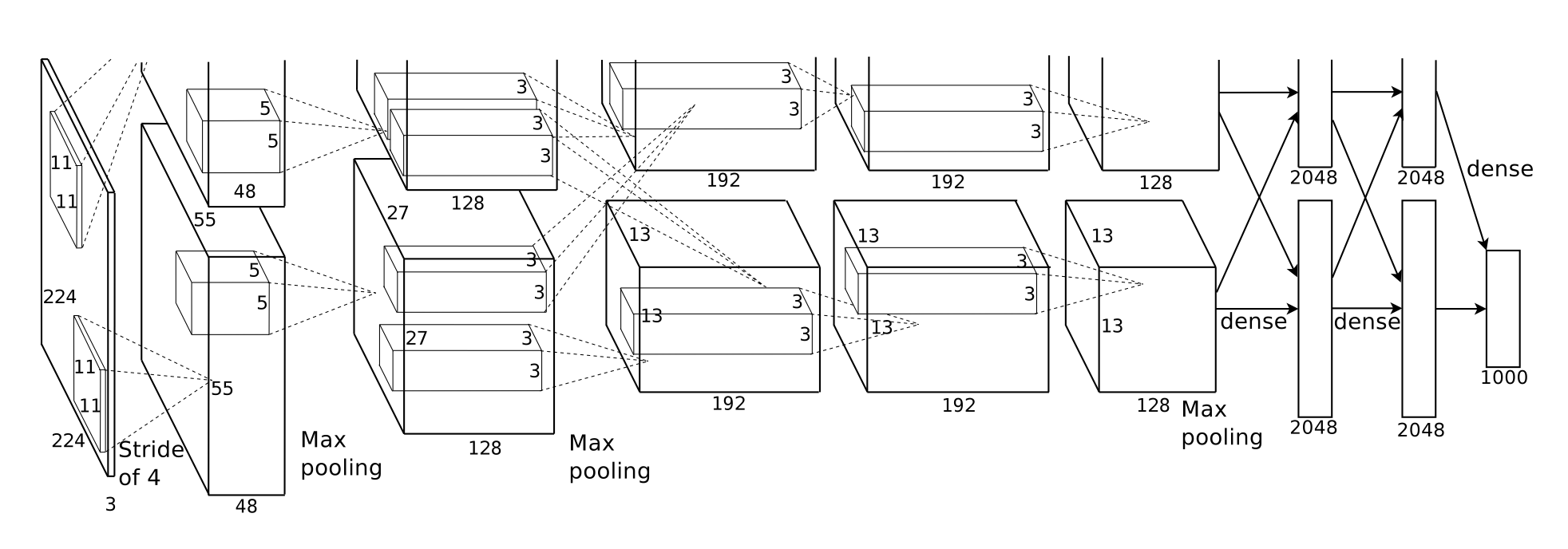}
\caption{This figure from \cite{KrizhevskyNIPS2012} illustrates the AlexNet model. Number of filters and dimensions are mentioned in the picture. Note that after every layer, there is an implicit rectified linear unit (ReLU) non-linearity. }\label{fig3:alexnet}
\end{figure*}

A Convolutional Neural Network (CNN) does not represent a new idea. This model had been proven to work well for hand-written digit recognition by \cite{Lecun98gradient} in 1998. However, due to the inability of these networks to scale to larger images, they slowly lost the community's interest. The for most reason was largely related to memory, hardware constraints and the unavailability of sufficiently large amounts of training data. With the increase of computational power, thanks to a wide availability of GPUs, and the introduction of large scale datasets such as the ImageNet $2015$ \cite{ILSVRC15} and the MIT Places dataset (see \cite{ZhouNIPS2014}), it was possible to train larger, more complex models. This was first demonstrated by the popular \textit{AlexNet} model in \cite{KrizhevskyNIPS2012}. This largely stimulated the usage of deep networks in computer vision.
This section is started by discussing the AlexNet architecture. 
Figure \ref{fig3:alexnet} reports the scheme of the $8$ layered AlexNet architecture. According to Figure \ref{fig3:alexnet}, the network consists of 96 filters (dim. 11 $\times$ 11 $\times$ 3) at the first layer, 256 filters (dim. 5 $\times$ 5 $\times$ 96) at layer two, 384 filters (dim. 3 $\times$ 3 $\times$ 256) at layer 3, and so on. While the first five layers are convolutional, the final three are traditional neural networks hereinafter referred to as ``fully connected layers". This network was trained on the ILSVRC 2012 training data, which contained 1.2 million training images belonging to 1000 classes. This was trained on 2 GPUs over the course of one month. The same network can be trained today in little under a week using more powerful GPUs. 
The hyper-parameters of the learning algorithms like learning rate, momentum, dropout and weight decay were tuned by hand. The initial layers tend to learn gabor-like oriented edges and blob-like features, followed by layers that seem to learn higher order features like shapes. The final layers seem to learn semantic attributes equivalent to eyes or wheels, which are crucial parts in several categories. A method to visualize these was provided by \cite{DBLPZeilerF13} in 2014.

\subsection{The Role of Convolutional Layers}

The AlexNet introduces 2D convolutions instead of matrix multiplications used in traditional neural networks. One of the reasons to employ convolutional filters is that learning such a kind of weight is much more tractable than learning a large matrix (50000 $\times$ 30000). 2D convolutions also naturally account for the 2D structure of images. According to \cite{Bishop2006}, convolutions can also be thought of as regular neural networks with two constraints: \textit{Local connectivity} and \textit{Weight sharing}. The former comes from the fact that a convolutional filter has much smaller dimensions than the image on which it operates. This contrasts with the global connectivity paradigm, typically relevant to vectorised images. The latter comes from the fact that in such a framework the same filter is applied across the image. This means we use the same local filters on many locations in the image. In other words, the weights between all these filters are shared. There is also evidence from visual neuroscience for similar computations within the human brain \cite{hubel1968,Fukushima1980}. 

Note, in practical CNNs, the convolution operations are not applied in the traditional sense wherein the filter shifts one position to the right after each multiplication. Instead, it is common to use larger shifts (commonly referred to as stride, see Figure \ref{fig3:alexnet}). This is equivalent to performing image down-sampling after regular convolution.

\subsection{Pooling Layers}

\begin{figure*}[!]
\centering
\includegraphics[width=0.8\textwidth]{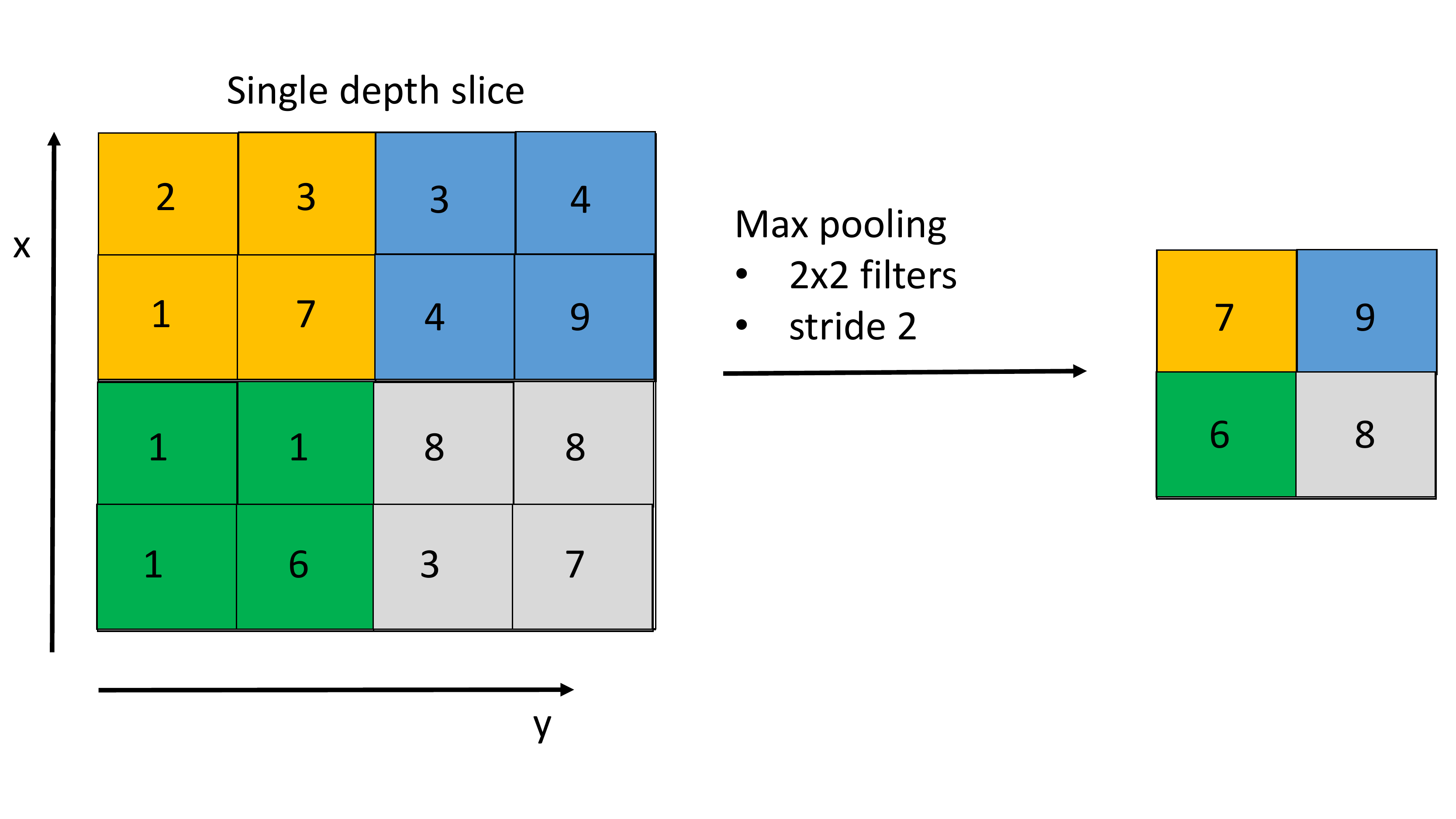}
\caption{In figure an example of max-pooling on a 4 $\times$ 4 depth slice. Max pooling is done by applying a max filter to non-overlapping sub-regions of the initial representation.}\label{fig3:maxpooling}
\end{figure*}

In addition to the convolutional layers just described, convolutional neural networks also contain pooling layers. Pooling layers are usually used immediately after convolutional layers. The pooling layers' function is to simplify the output information from the convolutional layer. Note, \textit{max-pooling} models the behaviour of the receptive fields of cells in the lateral geniculate nucleus (LGN) (see Figure \ref{fig3:ReceptiveFields}.\textbf{a} for further details). In detail, the max-pooling layer partitions the input image into a set of non-overlapping rectangles and, for each such sub-region, outputs the maximum value (see Figure \ref{fig3:maxpooling}). 

After obtaining the convolved features as described earlier and predefining the size of the region, say $n \times n$, to transfer the convolved features over. The convolved features are divided into disjoint $n \times n$ regions, only the maximum (in max-pooling) is taken as a kind of feature activation over these regions to obtain the pooled convolved features. These pooled features can then be used for the next steps. Max-pooling is useful for two reasons. Firstly, by eliminating non-maximal values, it reduces computation for upper layers (by a factor of $n^2$). The smaller the activation size, the smaller the number of parameters to be learnt in the later layers. Secondly, it provides a small degree of spatial invariance  meaning that the same (pooled) feature will be active even when the image undergoes (small) translations.

\subsection{Commonly Used Activation Functions}\label{sec3:relu}

In deep networks, convolutional layers are followed by a non-linear operation that substitutes the neuron activation function. Non-linearities between layers ensure that the model is more expressive than a linear model \cite{goodfellowBengio2013maxout}. Every activation function or non-linearity (see Equation \ref{eq:nonlinearity} for some examples) takes a single number on which it performs a certain fixed mathematical operation. There are several activation functions, but the key idea behind all of them is to define a new type of output layer for our neural networks. A widely used activation function is the sigmoid $\sigma(z) = \frac{1}{1 + e^{-z} }$. The main idea behind this solution is that large negative numbers become 0 and large positive numbers become 1. In practice, the sigmoid non-linearity has recently fallen out of favor and it is rarely used. Before starting the discussion, we would like to point out to the reader that details on the learning algorithm (back-propagation) are provided with section \ref{sec3:learningdeep}. 

\begin{figure*}[!]
\centering
\includegraphics[width=0.98\textwidth]{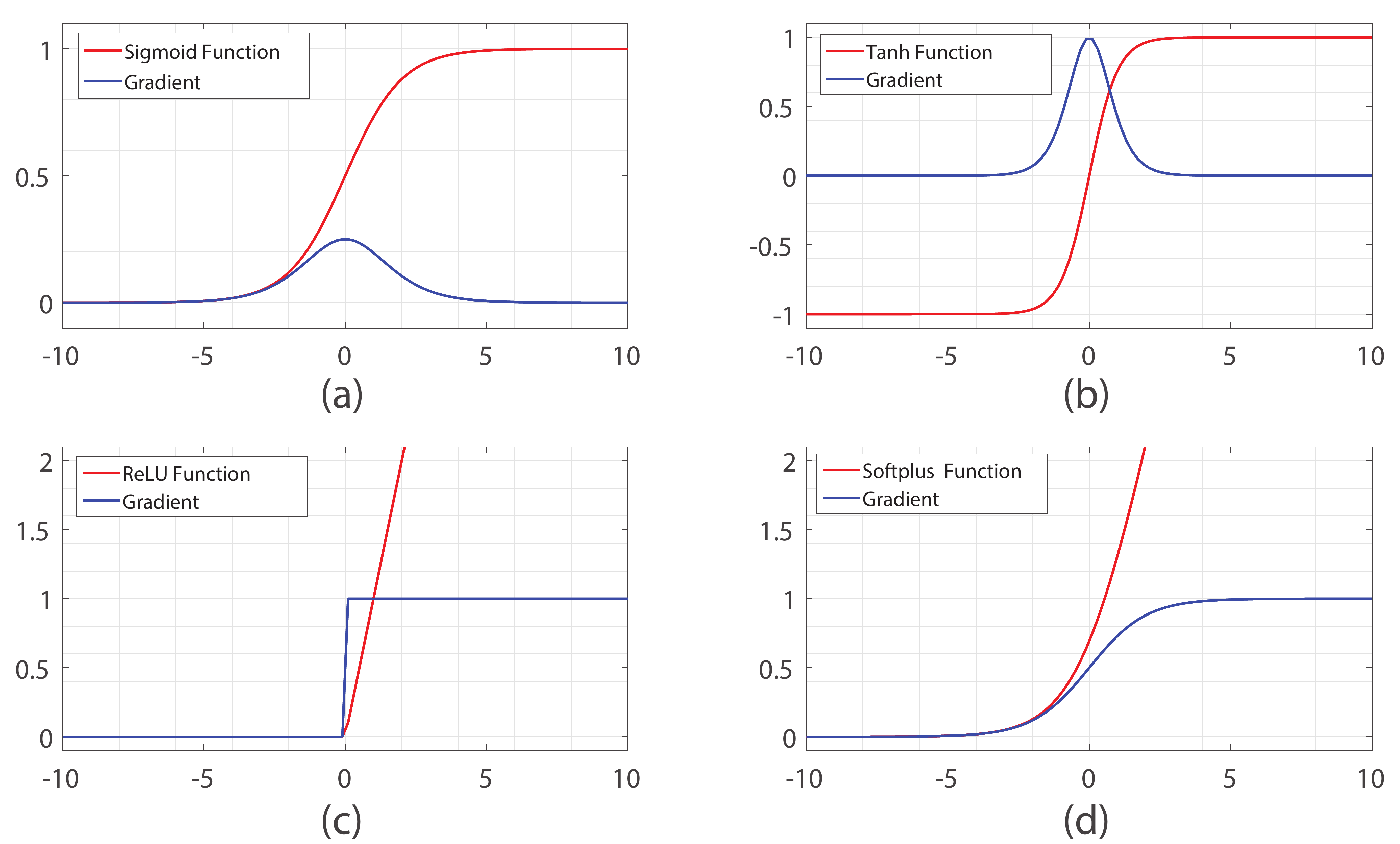}
\caption{Activation functions in comparison. Red curves stand for, respectively, sigmoid, hyperbolic tangent, ReLU, and Softplus functions. Their first derivative is plotted in blue. }\label{fig3:actfun}
\end{figure*}

There are two major drawbacks of sigmoid neurons. Firstly, sigmoid functions saturate and negate gradients. In Figure \ref{fig3:actfun}.\textbf{a} the plot of a sigmoid non-linearity and its gradient. An extremely undesirable property of the sigmoid neuron is that when the neuron's activation saturates at either tail of 0 or 1, the gradient at these regions is almost zero. If the local gradient is very small, during back-propagation it will effectively negate the gradient and almost no signal will flow through the neuron to its weights and recursively to its data. Furthurmore, if the initial weights are too large then most neurons would become saturated and the network’s ability to learn will all but come to a halt. 
Secondly, sigmoid outputs are not zero-centered. This has implications on the dynamics during gradient descent, because if the data coming into a neuron is always positive, then the gradient on the weights will become either all be positive, or all negative during back-propagation (depending on the gradient of the whole expression). Subsequently, this could introduce undesirable zig-zagging dynamics in the gradient updates for the weights. Therefore, this is an inconvenience but it has less severe consequences compared to the saturated activation problem above. 

A second widely used activation function is the hyperbolic tangent function $tanh(z) = \frac{e^{z} - e^{-z}}{e^{z} + e^{-z}} $ in Figure \ref{fig3:actfun}.\textbf{b}. It compresses a real-valued number to the range $[-1, 1]$. Like the sigmoid neuron, its activations saturate, but unlike the sigmoid neuron its output is zero-centered. Therefore, in practice the tanh non-linearity is always preferred to the sigmoid nonlinearity.  

Modern convolutional networks use the Rectified Linear Unit (ReLU) which is defined as $ReLU(z) = max(0, z)$ non-linearity. The ReLU non-linearity is shown in Figure \ref{fig3:actfun}.\textbf{c} along with its gradient. Roughly speaking, in the ReLU the activation is simply set to a threshold of zero. There are several pros and cons to using the ReLUs. Pros of using a ReLU can be summarized as follows:
\begin{itemize}
    \item ReLU is zero-centered. It avoids zig-zagging dynamics in the gradient descent optimization algorithms (e.g., stochastic gradient descent, momentum, adaptive gradient algorithm, etc.).
    \item ReLU greatly accelerate the convergence of stochastic gradient descent compared to the sigmoid/tanh functions \cite{KrizhevskyNIPS2012}. 
    \item ReLU is simple and it can be implemented by thresholding a matrix of activations at zero.
\end{itemize}
Unfortunately, a disadvantage becomes apparent when a large gradient flows through a ReLU neuron. This could cause the weights to update in such a way that the gradient flowing through the unit will constantly be zero from that point on. 

In order to solve this problem, different variants to ReLU have been proposed. One of the most interesting is called leaky-ReLU \cite{maas2013rectifier}. It was defined as $LReLU(z) = max(z, \alpha z )$, where $\alpha \leq 1$ (e.g., $\alpha = 0.01$). The leaky-ReLU sacrifices hard-zero sparsity for a gradient which is potentially more robust during optimization.
 
A smooth approximation to the ReLU is the analytic function Softplus$(z) = ln(1 + e^x )$ (see Figure \ref{fig3:actfun}.\textbf{d}), the derivative of softplus is the logistic function (i.e. sigmoid) \cite{DugasBBNG00}. Another interesting activation function is called \textit{SoftMax} which is a generalization of the logistic function (see Eq. \ref{eq:nonlinearity}). Interestingly, the output from the SoftMax function can be thought of as a probability distribution (i.e., it is a set of positive numbers which sum up to $1$). The fact that a SoftMax layer outputs a probability distribution is a pleasing result. In many problems, it is convenient to be able to interpret the output activation $a_j$ as the network's estimate of the probability that the correct output is $j$. Finally, one relatively popular solution is the \textit{Maxout} neuron \cite{goodfellowBengio2013maxout} that generalizes both the ReLU and its “leaky” version. The Maxout neuron computes the function $max(w^T_1x+b_1, w^T_2x+b_2)$, where ReLU (and leaky-ReLU) is a special case and by setting $w_1,b_1 = 0$, the Maxout reduces to ReLU. The Maxout neuron therefore takes advantage of all the benefits of a ReLU unit (linear regime of operation, no saturation) and does not have its drawbacks (dying ReLU). However, unlike the ReLU neurons it doubles the number of parameters for every single neuron, leading to a high total number of parameters.
 
\section{On the Need to be Deep}\label{sec3:needToDeep}

Hornik in 1991 introduced the popular \textit{Universal Approximation} theorem \cite{Hornik:1991}. This well-known theorem states that a neural network with a single, hidden layer is sufficient to model any continuous function. However, a few years later \cite{Bengio:2009} showed that such networks need an exponentially large number of neurons when compared to a neural network with many hidden layers. It also showed that poor generalization may be expected when using an insufficiently deep architecture for representing some functions. With the introduction of greedy layerwise pre-training by \cite{Hinton:2006}, researchers were able to train much deeper networks. This fact played a major role in bringing the so-called \textit{Deep Learning} systems into mainstream machine learning.

Modern deep networks such as AlexNet have 8 layers. A more recent network is the \textit{VGGnet} \cite{Simonyan14c} proposed in 2014 with 19 layers. The main difference between the AlexNet and the VGG architecture is that the VGGnet consists of filters with very small receptive field (3 $\times$ 3) and that not all the convolutional layers are followed by max-pooling. Another recent solution is the \textit{GoogleNet} \cite{Szegedy2015} architecture, proposed in 2015, which consists of 22 layers. In \cite{Szegedy2015}, they introduce a new level of organization in the form of the popular ``Inception module", which is repeated many times, leading to a deeper model. In 2016, \cite{he2016deep} proposed the \textit{ResNet} that reformulates the layers as learning residual functions with reference to the layer inputs. On the ImageNet dataset, the authors evaluated residual nets with a depth of up to $152$ layers (i.e., 8$\times$ deeper than VGG nets but still having lower complexity), where their result won the $1$st prize on the ILSVRC 2015 \cite{ILSVRC15} classification task. 

Very deep convolutional networks with hundreds of layers have led to significant reductions in errors on competitive benchmarks. Although, it is worth noting that research on the possibility of using shallow architectures has been conducted over the last few years. This is the case of \cite{BaCaruanaNIPS2014} where the authors provided evidence that shallow feed-forward nets could learn the complex functions, previously learned by deeper networks, and achieved accuracies previously only achievable with deeper models. They empirically showed that single-layer fully connected feedforward nets trained to mimic deep models can perform similarly to well-engineered complex deep convolutional architectures.  

While depth tends to improve network performances, it also makes gradient-based training more difficult since deeper networks tend to be more non-linear. To address these problems, \cite{Huang2016} proposes \textit{stochastic depth}. The principal idea is to start with very deep networks but during training, for each mini-batch, randomly drop a subset of layers and bypass them with the identity function. This simple approach complements the recent success of residual networks. With stochastic depth the number of layers increases beyond $1,200$ and the network still yields meaningful improvements in test error.

Finally, a further improvement of ResNets leads to Dense Convolutional Networks (DenseNets). They connect each layer to every other layer in a feed-forward fashion \cite{huang2016densely}. These networks can be substantially more accurate since they contain shorter connections between layers close to the input and those close to the output. Moreover, DenseNets alleviate the vanishing-gradient problem, strengthen feature propagation, encourage feature reuse, and substantially reduce the number of parameters.

\section{Learning Deep Architectures}\label{sec3:learningdeep}

Learning is generally performed by minimization of certain loss functions. Neural networks are generally trained using the back-propogation algorithm \cite{Rumelhart1988} which uses the chain rule to increase the computation speed of the gradient for the gradient descent (GD) algorithm. However, for training-set with tens of thousands of samples, using GD is impractical. In such cases, an approximation called the Stochastic Gradient Descent (SGD) is used. It has been found that training using SGD generalizes much better than training using GD. One disadvantage is that SGD is very slow to converge and so to counteract this, SGD is typically used with a \textit{mini-batches}, where a mini-batch typically contains a small number of data-points ($\approx$ 100). 

This section reviews the basics of supervised learning for deep architectures. Firstly, it discusses the choice of the cross-entropy cost function. Secondly, the back-propagation algorithm is presented and detailed. Thirdly, in section \ref{sec3:MinSGD} the minibatch stochastic gradient descent algorithm is presented along with one of its popular variants known as \textit{Momentum}. Finally, section \ref{sec3:Overfitting} presents the regularization methods (such as $L_2$ and $L_1$ regularization and dropout) which make the deep networks better at generalizing beyond the training data, and considerably reducing the effects of overfitting.

\subsection{The Cross-Entropy Cost Function}\label{sec3:crossEntropyCost}

To introduce the cross-entropy cost function we use the simple neuron model illustrated in Figure \ref{fig3:neurons}.\textbf{b}. The notation used is reported in Table \ref{tab:notation11}.
\begin{table}[!]
\begin{center}
  \begin{tabular}{ | l | l |}
    \hline
        $x$ & \mbox{denotes a training input}\\
        $y$ & \mbox{stands for a desired output}\\
        $f(\cdot)$ & \mbox{denotes an activation function}\\
        $a$ & \mbox{denotes an intermediate neuron's output}\\
        $z$ & \mbox{stands for the intermediate weighted input to a neuron}\\ 
        $w$ & \mbox{denotes the collection of all weights in the network }\\
        $b$ & \mbox{all the biases }\\
        $n$ & \mbox{is the total number of training inputs}\\
    \hline
  \end{tabular}
  \caption{The notation used in section \ref{sec3:learningdeep}.}\label{tab:notation11}
\end{center}
\end{table}
When training neural networks cross entropy is a better choice of cost function than a quadratic cost function. Quadratic cost is defined as 
\begin{equation}\label{eq:derErrC}
 C = \frac{(y - a)^2}{2},
\end{equation}
where $a = f(z)$, and $z = w^Tx + b$. Using the chain rule to differentiate with respect to the weight and bias we obtain
\begin{equation}\label{eq:derErrw}
 \frac{\partial C}{\partial w} = (a - y)f'(z)x,
\end{equation}
\begin{equation}\label{eq:derErrb}
 \frac{\partial C}{\partial b} = (a - y)f'(z)x.
\end{equation}
Let us assume that $f(\cdot)$ is a sigmoid activation faction (the shape of the sigmoid function is illustrated in Figure \ref{fig3:actfun}.\textbf{a}. Note, its first derivative plays an active role in Eq.~\ref{eq:derErrw} and Eq.~\ref{eq:derErrb}. This fact affects the learning speed by slowing it down. The reason is that every time the neuron's output $f(x)$ is close to $1$ or $0$, the sigmoid curve becomes very flat, and so $f'(x)$ becomes very small ($\approx 0$) resulting in a learning slowdown Cross-entropy cost function is not affected by this issue. Indeed, it is defined as:
\begin{equation}\label{eq:crossEntr}
  C =  -\frac{1}{n} \sum_x [y lna + (1-y)ln(1-a)], 
\end{equation}
where the partial derivative of the cross-entropy cost with respect to the weights is given by
\begin{equation}\label{eq:crossEntrDerW}
 \frac{\partial C}{\partial w_j} =  -\frac{1}{n} \sum_x  \Big(  \frac{y}{f(z)} - \frac{(1-y)}{1-f(z)}  \Big)\frac{\partial f}{\partial w_j} , 
\end{equation}
\[
  \quad \quad =  -\frac{1}{n} \sum_x  \Big(  \frac{y}{f(z)} - \frac{(1-y)}{1-f(z)}  \Big) f'(z)x_j, 
\]
Putting everything over a common denominator and simplifying  
\[
\frac{\partial C}{\partial w_j} =  \frac{1}{n} \sum_x    \frac{f'(z) x_j}{f(z) (1-f(z) )} (f(z) - y) , 
\]
Using the definition of the sigmoid function (in Eq.~\ref{eq:nonlinearity}), the first derivative of $f(z)$ is given by $f'(z) = f(z)(1-f(z))$ it implies:
\begin{equation}\label{eq:crossEntrDerWFinal}
 \frac{\partial C}{\partial w_j} =  \frac{1}{n} \sum_x  \Big(  x_j (f(z) - y)  \Big).
\end{equation}
In this equation the terms $f'(z)$ are not present anymore, therefore the rate at which the weight learns is controlled by $f(z) - y$, i.e., by the error in the output. According to this solution, the larger the error, the faster the neuron will learn. As a result, it avoids the learning slowdown caused by the $f'(z)$ term in the analogous equation for the quadratic cost,  Eq.~\ref{eq:derErrw}. In a similar way, we can compute the partial derivative for the bias: 
\begin{equation}\label{eq:crossEntrDerbFinal}
 \frac{\partial C}{\partial b} =  \frac{1}{n} \sum_x  \Big(  f(z) - y \Big).
\end{equation}
Again, this avoids the learning slowdown caused by the $f'(z)$ term in the analogous equation for the quadratic cost, Eq.~\ref{eq:derErrb}.

In conclusion, it is worth mentioning that there is a standard way of interpreting the cross-entropy that comes from the field of \textit{information theory} \cite{cover2012elements}. Generally speaking, the idea is that the cross-entropy is a measure of \textit{surprise}. In particular, our neuron is trying to compute the function $x \to y = y(x)$. But instead it computes the function $x \to a = a(x)$. Suppose we think of $a$ as our neuron's estimated probability that $y$ is $1$, and $1-a$ is the estimated probability that the right value for $y$ is $0$. Then the cross-entropy measures how surprised we are, on average, when we learn the true value for $y$. We get low surprise if the output is what we expect, and high surprise if the output is unexpected.

\subsection{The Back-Propagation Algorithm}

The back-propagation algorithm was introduced in the 1970s, but its importance was not fully appreciated until a famous 1986 paper \cite{Rumelhart1988}. That paper shows that back-propagation works far faster than earlier approaches to learning, allowing several neural networks to solve problems which had previously been unsolvable. Today, the back-propagation algorithm is a standard of learning in neural networks. \\
This section is more mathematically involved than the rest of the chapter, and it assumes that the reader is familiar with traditional neural networks (see \cite{Bishop2006} for further details). 

At the heart of back-propagation is an expression for the partial derivative $\frac{\partial C}{\partial w}$ of the cost function $C$ with respect to any weight $w$ (or bias $b$) in the network. Therefore, back-propagation gives detailed insights into how changing the weights and biases changes the overall behaviour of the network. The back-propagation algorithm is based on common linear algebraic operations. In particular, suppose $s$ and $t$ are two vectors of the same dimension. We can use $s \odot t$  to denote the elementwise product of the two vectors, known in literature as Hadamard product. Thus the components of $s \odot t$ are just $(s \odot t)_j = s_jt_j$.

Let us introduce an intermediate quantity $\delta_t^l$ which stands for the error in the $j^{th}$ neuron in the $l^{th}$ layer. Back-propagation will give a procedure to compute the error $\delta_t^l$ and then will relate $\delta_t^l$ to $\frac{\partial C}{\partial w^l_{jk}}$ and $\frac{\partial C}{\partial b^l_j}$. Where $w^l_{jk}$ denotes the weight for the connection from the $k^{th}$ neuron in the $(l-1)^{th}$ layer to the $j^{th}$ neuron in the $l^{th}$ layer. The equation for the error in the output layer $\delta_t^l$ is defined as
\begin{equation}\label{eq:DeltaErrors}
 \delta_t^l = \frac{\partial C}{\partial a^l_j} f(z^l_j),
\end{equation}
 where $\frac{\partial C}{\partial a^l_j}$ measures how fast the cost is changing as a function of the $j^{th}$ output activation and its form depends on the form of the cost function. $f(z^l_j)$ accounts for how much the activation function $f(\cdot)$ is changing at $z^l_j$.
$\delta_t^l$ can be vectorized in a matrix-based form as follows (i.e., it allows us to remove $j$ and employ the Hadamard product)
\begin{equation}\label{eq:BP1}
 \delta^l = \bigtriangledown_aC \odot f'(z^l),
\end{equation}
where $\bigtriangledown_aC$ is a matrix-based form whose components are exactly the partial derivatives $\frac{\partial C}{\partial a^l_j}$. This first equation is in terms of the output layer $a$. A second equation for $\delta^l$ is used which takes into account the error in the next layer $l+1$, given by
\begin{equation}\label{eq:BP2}
 \delta^l =  ((w^{l+1})^T\delta^{l+1}) \odot f'(z^l).
\end{equation}
This operation mimics the back-propagation of the error through the network, measuring the error at the output of the $l^{th}$ layer. Eq.~\ref{eq:BP1} and Eq.~\ref{eq:BP2} are used to compute the error $\delta^l$ at any layer in the network. 

In order to estimate the rate of change of the cost with respect to any bias in the network, the partial derivatives $\frac{\partial C}{\partial b^l_j}$ are computed
\begin{equation}\label{eq:BP3}
 \frac{\partial C}{\partial b^l_j} = \delta^l_j.
\end{equation}
The result shows that the rate is equal to the error $\delta^l_j$. While the rate of change of the cost with respect to any weight is given by:
\begin{equation}\label{eq:BP4}
 \frac{\partial C}{\partial w^l_{jk}} = a^{l-1}_k\delta^l_j.
\end{equation}
The important consequence of this fact is that when the activation $a \to 0$, the gradient term $\frac{\partial C}{\partial w}$ will also tend to be small insignificant. This is the case in which the learning turns out to be slow. This fact clearly explains why certain activation functions perform better than others (e.g., a ReLU rather than of a sigmoid function). Procedure ~\ref{algorithm} reports the sketch of the back-propagation algorithm.

\begin{algorithm}[H]
\caption{The back-propagation algorithm}
\label{algorithm}
\begin{algorithmic}
\REQUIRE{$x$ (\textit{Training examples for the first input layer})}\\
\ENSURE{$\frac{\partial C}{\partial w^l_{jk}}$ and $\frac{\partial C}{\partial b^l_j}$ (\textit{The gradient of the cost function})}\\
\STATE \textbf{Feedforward}:
\FOR{$l = 1 : L$}
\STATE $z^l = w^la^{l-1} + b^l$ \\
\STATE $a^l = f(z^l)$
\ENDFOR \\
\STATE $\delta^L = \bigtriangledown_aC \odot f'(z^l)$\\
\STATE \textbf{Backpropagate the error}:\\
\FOR{$l = L : -1 : 2$}
\STATE $\delta^l = ((w^{l+1})^T\delta^{l+1}) \odot f'(z^l)$ \\
\ENDFOR \\
\RETURN $a^{l-1}_k\delta^l_j$ and $\delta^l_j$
\end{algorithmic}
\end{algorithm}

\subsection{Minibatch Stochastic Gradient Descent}\label{sec3:MinSGD}

Optimization methods, such as the limited-memory Broyden-Fletcher-Goldfarb-Shannon (BFGS) algorithm, which use the full training data to estimate the next update to parameters tend to converge conform well to local optima. However, in practical applications the cost and gradient for the entire training set can be very slow and sometimes intractable on a single machine. Moreover, a second limitation of such methods is that they do not give a straightforward way to incorporate new data in an ``online" setting. Stochastic Gradient Descent (SGD) addresses both of these issues by following the negative gradient of the objective after seeing only a single or a few training examples. The use of SGD in the deep neural network setting is motivated by the high cost of running back propagation over the full training set, since SGD can overcome this cost and still lead to fast convergence. 

In SGD, the true gradient of the cost function $C(\theta)$ is approximated by updating the parameters $\theta$ of the objective $C$ as
\begin{equation}\label{eq:SGD}
 \theta =  \theta - \lambda \bigtriangledown_{\theta} C(\theta; \textbf{x}, \textbf{y}).
\end{equation}
where the pair of vectors [\textbf{x},\textbf{y}], from the training set, is called \textit{minibatch} (i.e., a small set of training examples $\approx 100$). Actually, the classic SGD method performs the above update for each training example, but in deep neural networks each parameter update is computed with regards to a \textit{minibatch} as opposed to a single example. The reason for this is twofold: first this reduces the variance in the parameter update which can lead to more stable convergence, second this allows the computation to take advantage of highly optimized matrix operations that should be used in a well vectorized computation of the cost and gradient.  

Let us consider the cross-entropy cost function as an objective (see section ~\ref{sec3:crossEntropyCost}, Eq.~\ref{eq:crossEntr}). Without loss of generality, since $a = f(w^T x + b)$, we can rewrite Eq.~\ref{eq:crossEntr} as follows:
\begin{equation}\label{eq:CrossEntCostGeneral}
 C =  -\frac{1}{n} \sum_x [y ln(f(w^T x + b)) + (1-y)ln(1-(f(w^T x + b)))].
\end{equation}
This is a more explicit form and it makes the connection with the SGD method effortless in Eq.\ref{eq:SGD} where $\theta := [w, b]$. As a result, the gradient of the objective function to be minimized is
\[ 
 \bigtriangledown_{\theta} C = \Big[ \frac{\partial C}{\partial w^l_{jk}} , \frac{\partial C}{\partial b^l_j} \Big],
\] 
where the two partial derivatives are outputs of the back-propagation algorithm (see Procedure \ref{algorithm}). Finally, when SGD is used to minimize the above function, the minibatch gradient descent method would perform the following iterations:
\begin{equation}\label{eq:SGD}
 [w, b]_{E_{t+1}} =  [w, b]_{E_{t}} - \lambda \Big[ \frac{\partial C}{\partial w^l_{jk}} , \frac{\partial C}{\partial b^l_j} \Big],
\end{equation}
where $E_t$ stands for the epoch at time $t$, and $\lambda$ is the learning rate. $\lambda$ is typically much smaller than a corresponding learning rate in batch gradient descent (e.g., BFGS) because there is much more variance in the update. 

It is important to note that SGD is the order in which we present the data to the algorithm. If the data is given in some meaningful order, this can bias the gradient and lead to poor convergence. Generally a good method to avoid this is to randomly shuffle the data prior to each epoch of training.

\begin{figure*}[t]
\centering
\includegraphics[width=0.55\textwidth]{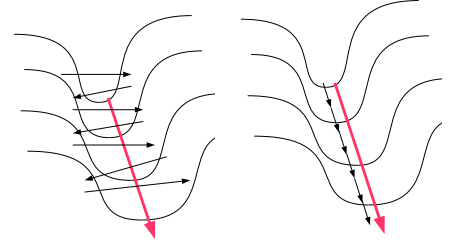}
\caption{\textbf{LEFT}: shows a long shallow ravine leading to the optimum and steep walls on the sides. Standard SGD will tend to oscillate across the narrow ravine. \textbf{RIGHT}: Momentum is one method for pushing the objective more quickly along the shallow ravine. }\label{fig3:momentum}
\end{figure*}

Further proposals include the \textit{momentum} method, which appeared in \cite{Rumelhart1988}. SGD with momentum remembers the update at each iteration, and determines the next update as a convex combination of the gradient and the previous update (see Figure \ref{fig3:momentum}). 
The objectives of deep architectures have the form in Figure \ref{fig3:momentum} near local optima and thus standard SGD can lead to very slow convergence particularly after the initial steep gains. Momentum overcomes this issue and its update is given by
\begin{equation}\label{eq:Momentum1}
 \Delta \theta = \gamma \Delta \theta + \lambda \bigtriangledown_{\theta} C(\theta; \textbf{x}, \textbf{y}),
\end{equation}
\begin{equation}\label{eq:Momentum2}
 \theta = \theta - \Delta \theta,
\end{equation}
In the above equation $\Delta \theta$ is the current velocity vector which is of the same dimension as the parameter vector $\theta$. $\gamma \in (0,1]$ determines for how many iterations the previous gradients are incorporated into the current update. Generally $\gamma$ is set to $\frac{1}{2}$ until the initial learning stabilizes and then is increased to $\frac{3}{4}$ or higher.

\subsection{Overfitting and Regularization}\label{sec3:Overfitting}

Overfitting occurs when a model is excessively complex, for example, when having too many parameters relative to the number of observations. A model that has been overfit has poor predictive performance, as it overreacts to minor fluctuations in the training data. Increasing the amount of training data is one way of reducing overfitting. Another possible solution is to reduce the number of network layers $L$ so as to obtain far fewer parameters. However, large networks have the potential to be more powerful than small networks, and so this is an option adopted reluctantly. Other interesting techniques to reduce overfitting with fixed network depth and fixed training data are known as \textit{regularization} techniques. In this section we describe one of the most frequently used regularization approaches known as \textit{L2 regularization} or \textit{weight decay} \cite{NowlanHinton1992}. 

The idea of L2 regularization is to add an extra term to the cost function, a term known as the \textit{regularization term}.
\begin{equation}\label{eq:RegCrossEntropy}
    C = -\frac{1}{n} \sum_{x_j}  \Big[ y_{j} ln(a^L_{j}) + ( 1 - y_{j}) ln(1 - a^L_{j})  \Big] + \frac{\delta}{2n} \sum_w w^2,
\end{equation}
where $L$ denotes the number of layers in the network, and $j$ stands for the $j^{th}$ neuron in the last $L^{th}$ layer. The first term is the common expression for the cross-entropy in $L$-layer multi-neuron networks, the regularization term is a sum of the squares of all the weights in the network. The regularization term is scaled by a factor $\frac{\delta}{2n}$, where $\delta > 0$ is known as the \textit{regularization parameter}, and $n$ is the size of the training set (see Table~\ref{tab:notation11} for the used notation). In a similar way it is possible to regularize other cost function, such as the quadratic cost.

Intuitively, regularization can be viewed as a way of compromising between finding small weights and minimizing the original cost function. The relative importance of the two elements of the compromise depends on the value of $\delta$. This kind of compromise helps reduce overfitting.

Taking the partial derivatives of Eq. \ref{eq:RegCrossEntropy} gives
\begin{equation}\label{eq:derErrWReg}
 \frac{\partial C}{\partial w} = \frac{\partial C_0}{\partial w} + \frac{\delta}{n}w,
\end{equation}
\begin{equation}\label{eq:derErrbReg}
 \frac{\partial C}{\partial b} =  \frac{\partial C_0}{\partial b}.
\end{equation}
The terms $\frac{\partial C_0}{\partial w}$ and $\frac{\partial C_0}{\partial b}$ can be computed using back-propagation, as previously described. Hence, the gradient of the regularized cost function can be obtained by adding the term $\frac{\delta}{n}w$ to the partial derivative of all the weight terms, while the partial derivatives with respect to the biases are unchanged.
 
Regularization is useful to prevent overfitting. The regularization term $\sum_w w^2$ is a constraint which forces the optimization procedure to maintain small weights. This, in turn, affects the behavior of the neural network in a way that the network is more robust against noise in data. In contrast, a network with large weights may strongly change its behaviour in response to small changes in the input. And so an unregularized network can use large weights to learn a complex model that carries a lot of information about the noise in the training data. In other words, regularized networks are resistant to learning peculiarities of the noise in the training data.

There are many regularization techniques other than L2 regularization. We briefly describe two other approaches to reducing overfitting: \textit{L1 regularization} and \textit{dropout}. 

\subsubsection{L1 regularization}

Similarly to $L2$ regularization, $L1$ regularization helps in penalizing large weights and tends to make small weights preferential to the network \cite{NowlanHinton1992}. In $L1$ regularization, the regularization term is the sum of the absolute values of the weights, as follows: 
\begin{equation}\label{eq:RegCrossEntropyL1}
    C = C_0 + \frac{\delta}{2n} \sum_w |w|,
\end{equation}
$L1$ regularization gives a different behavior respect to $L2$, the reason is given by the partial derivatives of the cost function
\begin{equation}\label{eq:derErrWRegL1}
 \frac{\partial C}{\partial w} = \frac{\partial C_0}{\partial w} + \frac{\delta}{n}sgn(w),
\end{equation}
where $sgn(w)$ is the sign of $w$. Using this expression, the resulting rule for the stochastic gradient descent by back-propagation using L1 regularization is given by
\begin{equation}\label{eq:derErrWRegL1}
 w_{E_{t+1}} = w_{E_t} - \lambda \frac{\partial C}{\partial w} - \lambda  \frac{\delta}{n}sgn(w),
\end{equation}
where $\lambda$ is the learning rate, and $\delta$ regulates the relative importance of the two elements of the compromise. Let us compare that to the update rule for L2 regularization
\begin{equation}\label{eq:derErrWRegL1}
 w_{E_{t+1}} = w_{E_t} - \lambda \frac{\partial C}{\partial w} - ( 1 - \frac{\lambda\delta}{n}) w,
\end{equation}
In both expressions the effect of regularization is to shrink the weights, but the way the weights shrink differs. In $L1$ regularization, the weights shrink by a constant amount toward $0$. In $L2$ regularization, the weights shrink by an amount which is proportional to $w$. As a result, when a particular weight has a large magnitude, $|w|$, $L1$ regularization shrinks the weight much less than $L2$ regularization does. By contrast, when $|w|$ is small, $L1$ regularization shrinks the weight much more than $L2$ regularization.

\subsubsection{Dropout}\label{sec3:dropout}

\begin{figure*}[t]
\centering
\includegraphics[width=0.99\textwidth]{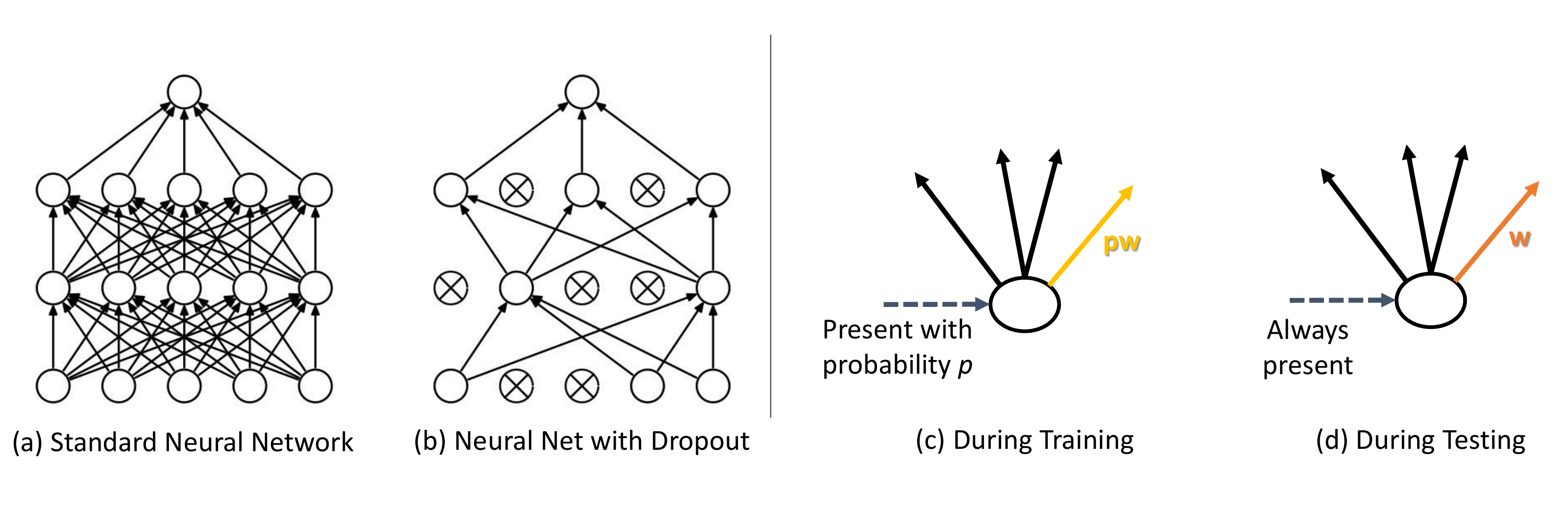}
\caption{An illustration of the dropout mechanism within a multi-layer neural network. (a) Shows a standard neural network with $2$ hidden layers. (b) Shows an example of a thinned network produced by applying dropout, where crossed units have been dropped. (c) At training time each unit has a probability to be present in the network. (d) At test time all units are used and contribute to the output.}\label{fig3:Dropout}
\end{figure*}

Dropout prevents overfitting and provides a way of approximately combining many different neural network architectures exponentially and efficiently. Indeed, dropout can be viewed as a form of \textit{ensemble learning}. The term dropout refers to dropping out units (hidden and visible) in a neural network (see Figure \ref{fig3:Dropout}). By dropping a unit out, we mean temporarily removing it from the network, along with all its incoming and outgoing connections.
In particular, given a training input $x$ and corresponding desired output $y$. Dropout, randomly and temporarily deletes the $h\%$ of the hidden units in the network, while leaving the input and output units untouched. After this step, the training procedure starts by forward-propagating $x$ through the network ($x$ as a mini-batch of examples), and then back-propagating to determine the contribution to the gradient, therefore updating the appropriate weights and biases. As shown in Figure \ref{fig3:Dropout}.\textbf{b}, the neurons which have been temporarily deleted, are still ghosted in the network. The described process is repeated, first restoring the dropout neurons, then choosing a new random subset of hidden neurons to delete, estimating the gradient for a different mini-batch $x$, and updating the weights and biases in the network. 

Dropout allows the network to learn weights and biases under conditions in which $h\%$ the hidden neurons were dropped out. According to Figure \ref{fig3:Dropout}.\textbf{c-d}, at test time all neurons will be active, this fact mimics the \textit{ensemble learning} methods. As a consequence, the main reason why dropout helps with regularization directly comes from this similarity. Indeed, applying dropout on the network can be viewed as taking a number of weaker classifiers (i.e., thinned networks produced), training them separately and then, at test time, using all of them by averaging the responses of all ensemble members.

\section{Transfer Learning by Fine-tuning}\label{sec3:FineTuning}

Transfer learning by fine-tuning deep nets offers a way to take full advantage of existing datasets to perform well on new tasks. One of the main reasons for the success of the CNN models is that it is possible to directly use the pre-trained model (e.g., AlexNet 2012 \cite{KrizhevskyNIPS2012}, VGG Oxford 2014 \cite{Simonyan14c}, GoogleNet 2015 \cite{Szegedy2015}, ResNet 2016 \cite{he2016deep}, DenseNets 2016 \cite{huang2016densely}, etc.) to do various other tasks for which it was not originally intended. For example, given a model trained for a particular task (e.g., image classification), it is possible to use the trained weights as an initialization and run SGD for a new, but related, task. 

Fine-tuning can be performed on a new dataset considering two important factors: the size of the new dataset (small or big), and its similarity to the original dataset (e.g. a network trained on the Image-Net for image classification can be used for localization, face identification, pose estimation, etc.). According to these two factors, four different scenarios may emerge:
\begin{itemize}
    \item \textbf{New dataset is large and similar to the original dataset.} In this first case, fine-tuning can be used to generate a more specific model for the new dataset.
     \item \textbf{New dataset is large and slightly different from the original.} It is very often beneficial to initialize with weights from a pre-trained model. In this case, there is enough confidence in the data to fine-tune through the entire network.
    \item \textbf{New dataset is small and similar to original.} Since the new dataset consists of few training sample, fine-tuning could lead to overfitting. On one hand, fine-tuning could be applied by re-training only the few last layers of the network while leaving the first layers untouched. To avoid overfitting, the learning rate is usually much lower than that used for learning the original net. On the other hand, since the data is similar to the original, it is expected that higher-level features in the CNN are relevant to this dataset as well. As a result, it is possible to use CNNs as feature extractors and train a linear classifier directly on generic feature descriptors produced by the pre-trained network. 
    \item \textbf{New dataset is small but slightly different from the original.} Since the new dataset is small and different it is not suitable for fine-tuning. However, in these cases CNNs are often used as feature extractors. These features have been found to be better than hand-crafted features (e.g., SIFT or HoG for various computer vision tasks). As the two datasets are different a good solution is to use CNN activations from layers somewhere earlier in the network (generic features rather than very semantic attributes form the top of the network). 
\end{itemize}

When fine-tuning is applied, the learning rate is much lower than that used for learning the original net. In this way, the earlier layers can be fixed and only the later, more semantic layers need to be re-learnt. 

\section{Conclusions}

Some open research problems in deep learning, particularly of interest to computer vision include, but not limited to, (i) the tuning of a large number of hyper-parameters,  (ii) CNNs are robust to small geometric transforms. However, they are not invariant. The study of invariance for extreme deformation is largely missing.  (iii) CNNs are presently trained in a one-shot way. The formulation of an online method of training would be
desirable for robotics applications. (iv) Unsupervised learning is one more area where it is expected to deploy deep learning models. This would enable researchers to leverage the massive amounts of unlabelled image data on the web. Classical deep networks like auto-encoders and restricted Boltzmann machines were formulated as unsupervised models.

In this chapter, we have surveyed the use of modern deep learning networks, in particular we provide an intuitive introduction to convolutional neural networks that relies on both biological and theoretical constructs. We showed that CNNs enabled complicated hand-tuned algorithms being replaced by single monolithic algorithms trained in an end-to-end manner. We reviewed the basics of supervised learning for deep architectures. Firstly, we looked briefly into the history of biologically inspired algorithms exploring the connections between neuroscience and computer science enabled by recent rapid advances in both biologically-inspired computer vision and in experimental neuroscience methods. Secondly, we introduced the convolutional neural network architectures, where we discussed the choice of the cross-entropy cost function and presented the back-propagation algorithm. Thirdly, the mini-batch stochastic gradient descent algorithm was addressed along with one of its popular variants known as Momentum. Then, we addressed the regularization methods (such as $L_2$ and $L_1$ regularization and dropout) which make the deep networks better at generalizing beyond the training data, and considerably reducing the effects of overfitting. The last part focused on transfer learning by fine-tuning deep  networks, which offers a way  to take full advantage of existing datasets to perform well on new tasks. In conclusion, we provided a discussion on the need to be deep, highlighting pros and cons of very deep architectures. 


\part{Data Collection}\label{part:one}

\chapter{Benchmarking}\label{ch4:DATASETS}

The collection and organization of data are an integral and critical part of the research process. A formal data collection process is necessary as it ensures that the data gathered are both defined and accurate and that subsequent decisions based on arguments embodied in the findings are valid. The process provides both a baseline from which to measure and in certain cases a target on what to improve. Research data is a set of values that is collected, observed, or created, for purposes of analysis to produce original research results. Data can be generated or collected from interviews, observations, surveys, experiments, or even from previous literature. Often, data are irreplaceable. For this reason, it is important for researchers to be accurate and precise with their data collection and storage techniques and processes. Depending on the source and methodology involved, data collection may be time consuming and expensive. 

In this thesis we collected and make available four datasets: the Skype Chat dataset ($C$-$Skype$), the Typing Biometrics Keystroke Dataset (TBKD), the Social Signal Processing Chat corpus (SSP-Chat), and the ADS-16 for computational advertising. For some of them interviews and surveys are conducted on people, or human subjects. For data collection we used an homogeneous experimental protocol which consists of 4 main steps. \\
- \textbf{Step 1}: A web platform has been developed for the purpose of collecting data. This platform also manage data organization and storage, it stores data in text files using comma-separated values (CSV) format.\\
- \textbf{Step 2}: All the participants signed a consent form which allows us to use the provided information for research purposes. \\
- \textbf{Step 3}: A payment was guaranteed for the participation, but participants were encouraged to do their best work by offering the possibility to significantly increase their reward.\\
- \textbf{Step 4}: All participants filled in a form providing, anonymously, several information about them (e.g., demographic information, personal preferences and habits).\\
Based on the goal for which data is collected, different other forms have been submitted to the subjects such as the Big Five Inventory-10 to measure personality traits~\cite{rammstedt:2007} or others. 

This chapter introduces the four datasets we collected, presenting their features and peculiarities. In section \ref{ch4:skype} the C-Skype dataset is introduced. It is a corpus of dyadic textual chat conversations which consists of $94$ subjects in total. Section \ref{ch4:TBK} presents the Typing Biometrics Keystroke Dataset (TBKD). This dataset takes into account the temporal aspects of typing behaviour, since it includes the timestamps of each keystroke typed from any individual during the experiment. Some examples of research data include measured values (keystrokes, timestamps) and responses to questionnaires (by a web-based survey). Then, in Section \ref{ch4:SSPChat} the \textit{Social Signal Processing Chat} (SSP-Chat) corpus is introduced. This work is supported by the School of Computing Science, University of Glasgow. The experimental protocol adopted allows to preserve the context of collection and interpretation. Finally, in section \ref{ch4:ADS16} the ADS-16 dataset is presented in the context of ranking for recommendation. 

\section{The Skype Chat Dataset}\label{ch4:skype}

Text chats are an intriguing type of data, representing crossbreeds of literary text and spoken conversations. The $C$-$Skype$ dataset is a collection of text chats representing dyadic conversations extracted from the $Skype$ application. It was released to the social media community in $2012$-$2013$ for academic research purposes. The corpus has been collected for the first time in 2012 \cite{Cristani:2012} and extended in  2013 \cite{Roffo2013,Cristani:Skype:AVSS:2013}. In its first version, the dialogs covered up to 1 year of spontaneous conversations performed by 77 different subjects. Extensions in \cite{Roffo2013,Cristani:Skype:AVSS:2013}, provided both more subjects ($94$ individuals) and meta-data, for example regarding the time of messages being delivered (i.e., a timestamp for each ENTER key). 

Textual conversations within the corpus can be modeled as sequences of turns. In face-to-face (f2f) conversations, turns are intervals of time during which only one person talks. As well as in f2f, a turn is intended as a block of text written by one participant during an interval of time in which none of the other participants writes anything. In other words, a ``turn" is a stream of symbols and words (possibly including ``ENTER" characters) typed consecutively by one subject without being interrupted by the interlocutor. Turns is the unit of analysis over which the experiments on this dataset are performed. Noteworthy, the definition of turns may change on other chat datasets.

The corpus \cite{Cristani:Skype:AVSS:2013} involves $94$ Italian native speakers between 22 an 50 years old. The median of the participants age is 28 years. Participants were recruited at the Italian Institute of Technology and University of Verona. The cultural background of the participants is from all sorts of sources. Note, for cultural background we encompass many aspects of society, such as socioeconomic status, race and ethnicity, and other factors that contribute to an individual's cultural background include gender, age, religion, traditions, language and hobbies. In terms of gender, 42 are females and 52 males, resulting into 22 male-female dyads, 15 male-male dyads and 12 female-female dyads.

\begin{table}[!]
\centering
  \begin{tabular}{ | l | l | c |}
    \hline
    \textbf{Plain-text} & \textbf{Template} & \textbf{Timestamps} \\ \hline
    Hello John, & Xxxx Xxxx, & 1334230039 \\ 
    thank you for calling me :) & xxxxx xxx xxx xxxxxxx xx :) & 1334230251\\
    How may I help you? & Xxx xxx X xxxx xxx?  & 1334230633\\ 
    \hline
  \end{tabular}
  \caption{An example of the process of encoding messages in such a way that only stylometric cues can be read (e.g., number of words, characters, punctuation marks and emoticons).}\label{tab:encrypt}
\end{table}
For ethical and privacy issues, any cue which involves the content of the conversation has been discarded. Indeed, the content of the conversation is encrypted to hide the topic of discussion according to Table \ref{tab:encrypt}. At the same time, this solution allows the collection of many other features related to the style of writing and non-verbal schemes in data which can help in identifying unequivocally the subjects involved in a text chat conversation. For example:
\begin{itemize}
\item {\bf Turn duration}: the time spent in completing a turn (in hundredth of seconds);
\item {\bf Number of sentences}: \# of sentences (i.e., phrases ended by a return) produced in a turn; since in general each turn is composed by a single sentence, this trait could mirror the tendency of a speaker of maintaining its turn;
\item {\bf Writing speed}: \#characters in a turn divided by \emph{turn duration}; this element explains an intrinsic characteristic of a user;
\item {\bf Mimicry tendency}: the ratio \#\emph{words written by the other speaker in the previous turn}$/$\#\emph{words written by the $n-$th speaker in the current turn}; this characteristic would model the tendency of a subject in following the conversation style of the other participant, intended as the number of words written. A high tendency of mimicking in general means a good social skill, so we exploit this cue for characterizing the social attitude of a user;
\end{itemize}
\begin{table*}[!t]
\begin{center}
  \begin{tabular}{|c|l|l|}
  \hline
\textbf{No.}&\textbf{Feature} &\textbf{Range}  \\\hline
1  &\# words							& [0,260]       \\\hline
2  &\# emoticons       				    & [0,40]       \\\hline
3  &\# emoticons per word				& [0,1]       \\\hline
4  &\# emoticons per characters		    & [0,0.5]       \\\hline
5  &\# exclamation marks 		        & [0,12]       \\\hline
6  &\# question marks					& [0,406]       \\\hline
7  &\# characters						& [0,1318]       \\\hline
8  &   words length			            & [0,20]       \\\hline
9  &\# three points					    & [0,34]       \\\hline
10 &\# uppercase letters				& [0,94]       \\\hline
11 &\# uppercase letters/\#words        & [0,290       \\\hline
\textbf{\textcolor{BrickRed}{12}} &   \textbf{\textcolor{BrickRed}{turn duration}}					& [0,1800(sec.)]       \\\hline
\textbf{\textcolor{BrickRed}{13}} &\textbf{\textcolor{BrickRed}{\# sentences}} 					    & [1,20]       \\\hline
\textbf{\textcolor{BrickRed}{14}} & \textbf{\textcolor{BrickRed}{writing speed}}				        & [0,20(chars/second)]\\\hline
\textbf{\textcolor{BrickRed}{15}} &\textbf{\textcolor{BrickRed}{\# words per second}}					& [0,260]       \\\hline
\textbf{\textcolor{BrickRed}{16}} &  \textbf{\textcolor{BrickRed}{mimicry degree}}					& [0,1115]		  \\\hline
   \hline
   \end{tabular}
  \caption{Example of stylometric features. Stylometric features can be extracted by turns. The \emph{turn-taking based} features are those written in bold red. }
  \label{table4:FeatsPossible}
  \end{center}
\end{table*}
In Table \ref{table4:FeatsPossible} is reported a list of stylometric features, the term ``stylometry''  was coined in 1851 \cite{Holm:1998} in the authorship attribution context. Indeed stylometry is related to the statistical analysis of variations in literary style between one writer or genre and another. Along with the list of features, Table \ref{table4:FeatsPossible} proposes the ranges of each feature and in red those features which account for the turn-taking dynamics within the social interaction.

\begin{figure}[hbt]
    \centerline{
      \includegraphics[width=0.80\linewidth]{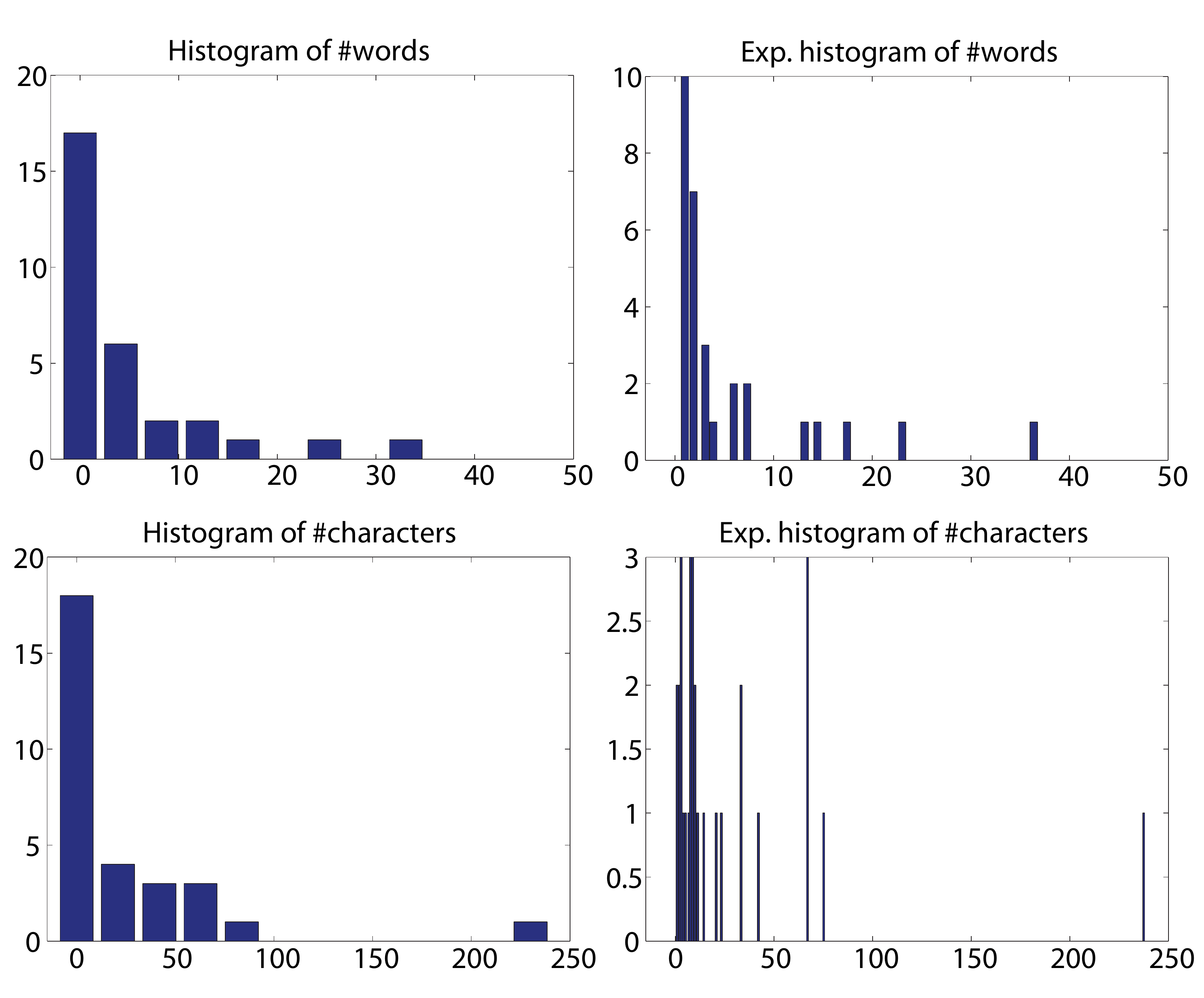}}
      \caption{Distributions of some lexical features: linear histogram (left), exponential histogram (right).}\label{fig4:DistribCompa}
\end{figure}

In Figure \ref{fig4:DistribCompa} four plots reproducing the distributions of two lexical features: number(\#) of words, \#characters. Distributions have been reported according to normal and exponential histograms. Exponential histograms produce a more effective representation when small-sized bin ranges are located toward zero. 

This dataset allows different tasks such as user re-identification and verification discussed in Chapter \ref{ch7:CHATS} on learning to rank.

\section{Typing Biometrics Keystroke Dataset}\label{ch4:TBK}

According to the previous section, the peculiarity of text chats stands in the fact that they show both aspects of literary text and of spoken conversation, due to the turn-taking dynamics that characterize text delivery. Therefore, it is interesting to investigate which aspects of these two communication means are intertwined in chat exchanges. The Typing Biometrics Keystroke Dataset (TBKD) changes radically the analysis protocol used for the C-Skype. In this case, we developed a chat service, embedded into the $Klimble$ web platform, designed with key logging capabilities; this in practice allows us to retain the timing of each single hit of a key, recording at the finest level the behavior of users while they chat. The TBKD has been proposed for the first time in 2014 \cite{Roffo:HBU2014,Roffo:icmi2014}. The TBKD consists of semi-structured chats between $50$ subjects for a total of 16 hours of conversation. It has been designed to discover whether the manifestation of keystroke dynamics (i.e., speed of typing or writing rhythms) produce a recognizable/unique chatting style that helps an identification system. 

Under the perspective of pragmatics and social signal processing, it is interesting to verify the presence of non verbal cues in chats. Non verbal signals enrich the spoken conversation by characterizing how sentences are uttered by a speaker~\cite{knapp2012nonverbal}, forging a unique style that characterizes the latter among many other subjects. In the same vein, they express personal beliefs, thoughts, emotions and personality~\cite{degroot2009}. As a result, this dataset includes personality traits of the subjects involved in the experiments. These traits, intended as individual differences that endure over time, can be measured by well-known self-administered questionnaires such as 1) the \textit{10 Item Big Five Inventory} (Big5) \cite{Rammstedt2007.20140805,rammstedt:2007},2) the Barratt Impulsiveness Scale \cite{patton1995}, focused on 3 different impulsiveness factors (attentional, motor and non-planning impulsiveness), 3) BIS-BAS \cite{carver1994}, decoding human motivations to behavioral inhibition (BIS) and activation (BAS), and 4) PANAS \cite{watson1988}, analyzing positive and negative affectivity traits.

\subsection{Experimental Protocol}\label{sec4:Data}

The data collection has been based on a public website, where the text chatting interface has been equipped with keylogging functionalities. At the moment of the subscription, users have been informed that the chat platform was equipped with a key logging application, but that the produced keylogs would be kept private and would not be shared. The data collection has been realized on this platform by setting up a campaign of chats; the subscribers who met certain requirements have been invited to participate to an experiment involving a text chat session along with the filling in of some questionnaires. The explanation of the experiment was given in general terms, and the time required to complete the experiment was not specified. The selected subjects are in the age interval of 18-30 years. The cultural background of the participants is uniform. Most of the participants have a university education (the most represented subjects are Master students in Computer Science) and were recruited at the University of Verona. 

The chats were conducted by a single operator, not having friendship ties with the subjects. The chats were at least 20 minutes long, and the generic arguments suggested by the operator were ``holidays'' and ``friends''. The visual interface employed in the dialogs mirrors the common instant-messaging platforms like Facebook, Skype and the like. Due to the implementation of the chat software, the key logging mechanism allows to get the timings of each button hit with a precision of milliseconds, synchronized with the timing of the other participant. A total of 50 subjects participated to the experiment, with an average age of 24 years, standard deviation $\rho=1.5$.
At the end of the text chat, the  participants were asked to fill in the following three questionnaires, aimed at evaluating psychological factors:

\begin{enumerate}
\item \textbf{The Barratt Impulsiveness Scale, version 11 } (Bis-11 \cite{patton1995}) to measure the levels of impulsiveness, based on three different sub-scales: ``attentional impulsiveness'', indicating
a lack of attention and cognitive instability, ``motor impulsiveness'', indicating a lack of control in motor behavior, and ``non-planning impulsiveness'', indicating a deficit in planning their own behavior. The total number of items is 30, answered on a 4-points scale, ranging from ``never'' to ``always''.

\item  \textbf{The Behavioral Inhibition and the Behavioral Activation scales }(BIS/BAS \cite{carver1994}) were used to analyze participants' likelihood of approaching rewards (BAS) or avoiding punishments (BIS) when making a choice. There are 20 items, answered on a 5-points scale, from ``very true for me'' to ``very false for me''.

\item  \textbf{The Positive Affect and Negative Affect Scales} (PANAS \cite{watson1988}) were used in order to analyze positive and negative affectivity. There are 20 items in total, inquiring ``to which extent subjects feel this way'' (indicated by each affective word) and answered on a 5-point scale, from ``very slightly or not at all'' to ``extremely''.
\end{enumerate}
\begin{figure*}[t]
\centering
\includegraphics[width=1\textwidth]{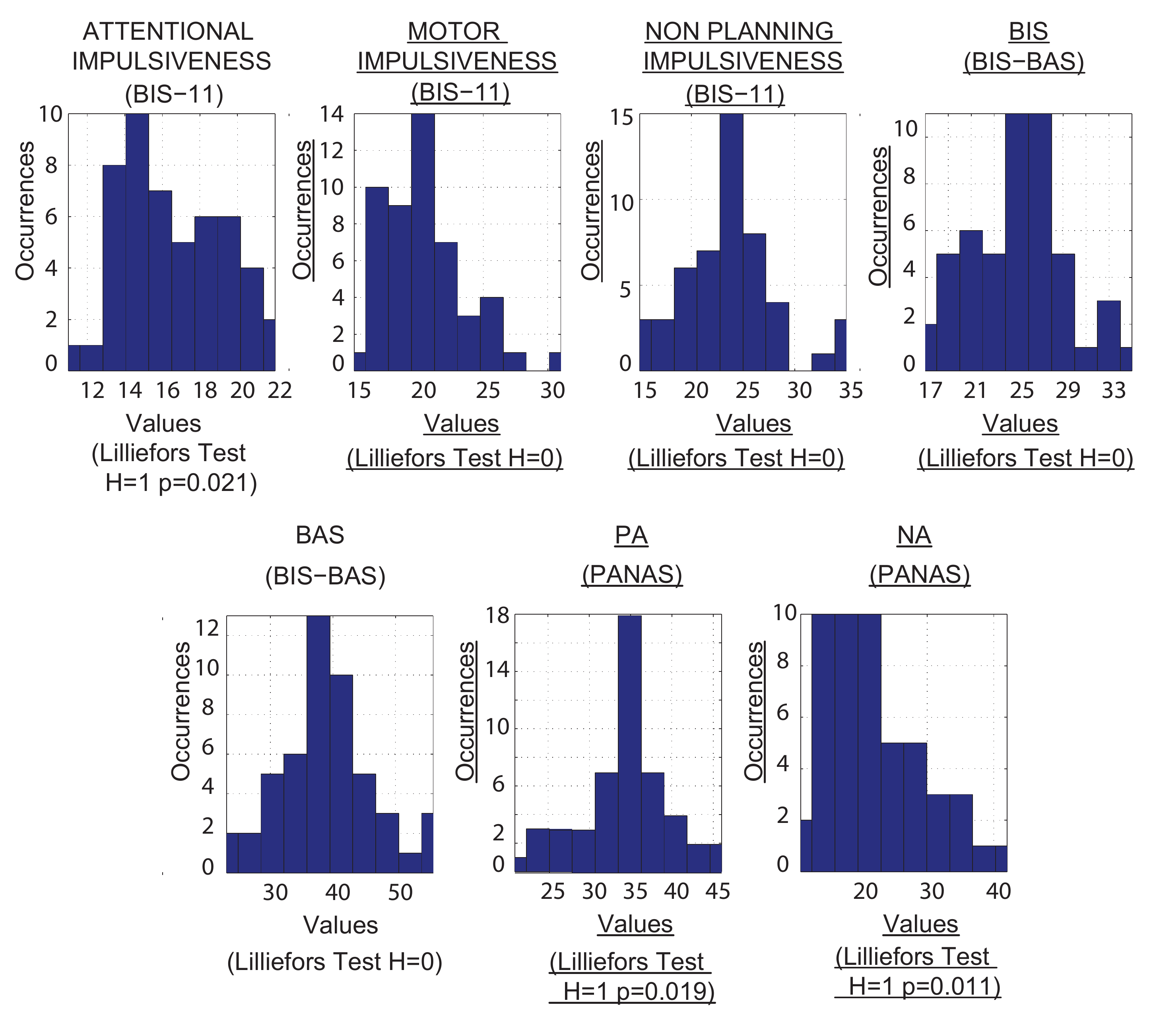}
\caption{Traits distribution. On the bottom, the output of the Lilliefors test for distribution normality (H=0 indicates data normally distributed). Underlined, those personality traits which correlate with at least one stylometric feature.}\label{fig4:traits}
\end{figure*}
In total the tests provide a quantitative indication for 7 personality traits, whose cumulative statistics over the 50 subjects are reported in Fig.~\ref{fig4:traits}.

Differently from the previous study, chats were all performed with the same operator, posing the same questions to everyone. The TBKD allows two different analyses.

A first analysis that could be performed is about the relation between stylometric cues and personality traits. A second analysis could be the investigation of the contribute of each stylometric feature in terms of recognition capability. Interesting insights into the writing behavior can be derived while connecting these two analyses. Our hypothesis is that some personality traits lead people to chat in a particular style, which makes them very recognizable. For example, if the use of short words is a discriminative characteristic between users in chats, and this feature is significantly correlated with higher non planning impulsiveness, it would highlight that people who score higher in this factor of impulsiveness are prone to use a greater number of short words, that is, the more subjects are impulsive, the shorter the time they stop on each single word while writing.

\section{The SSP-Chat Corpus}\label{ch4:SSPChat}

The $SSP$-$Chat$ $Corpus$ is a collection of $30$ dyadic chat conversations between fully unacquainted individuals, for a total  of $60$ subjects ($35$ females and $25$ males). These conversations were collected via a chatting platform provided by an ad-hoc web platform developed within a collaboration with the \emph{School of Computing Science, University of Glasgow}. As for the TBKD, the chatting platform was equipped with a keylogging functionality, which associates a timestamp with each key pressed.

The conversations revolve around the Winter Survival Task (WST), a scenario that is often adopted in social psychology research \cite{Joshi2005}, in which the subjects are asked to identify items that increase the chances of survival after a plane crash in Northern Canada~\cite{Joshi2005} (see Appendix \ref{app:B} for further details). In particular, the subjects are given a list of $12$ items (steel wool, axe, pistol, butter, newspaper, lighter without fuel, clothing, canvas, air-map, whisky, compass, chocolate) and they have to make a consensual decision for each of them (``\emph{yes}'' if it increases the chances of survival and ``\emph{no}'' if it does not). The main advantage of the scenario is that people, on average, do not hold any expertise relevant to the topic. Thus, the outcome of the conversations depends on social dynamics rather than on actual knowledge about the problem. Moreover, the WST requires that the items are discussed sequentially and that the participants do not move to the next item until agreeing on a decision for the current one. The decision agreed on for any item cannot be changed after moving to the next item.

A chat can be thought of as a stream of keys that are typed sequentially by the subjects. The sequence can be split into \emph{tokens}, i.e., strings of non-blank characters delimited by blank spaces. The \emph{rationale} behind this choice is that such strings should correspond to semantically meaningful units. Overall, the data includes a total of 33,085 tokens, 21,019 typed by the female subjects and 12,066 typed by the male ones. In every dyadic conversation, one of the subjects is \emph{proactive} (the person that starts the chat) and the other is \emph{reactive} (that person that joins the chat). Every subject has been randomly assigned one of these two roles.

During the chats, the subjects must press the ``\emph{Enter}'' key to send a sequence of tokens to their interlocutor. Some subjects press the Enter key every few words while others do it only after having written long and articulated messages. In both cases, the chat can be segmented into \emph{chunks}, i.e., sequences of tokens delimited by Enter keys. Therefore, in contrast with the $TBKD$ and $C$-$Skype$ datasets, the chunks are the analysis units of the $SSP$-$Chat$ $Corpus$. The experimental protocol has been detailed in Appendix \ref{app:B}.

At the end of the text chat, the  participants were asked to fill in the following two questionnaires, aimed at evaluating psychological factors and conflict handling style:

\begin{enumerate}
\item The \textbf{10 Item Big Five Inventory} (Big5) \cite{Rammstedt2007.20140805} is a psychological tool that describes an individual's personality according to five different traits (Openness, Conscientiousness, Extraversion, Agreeableness, Neuroticism). 
\begin{itemize}
\item \textbf{Openness}. This trait features characteristics such as imagination and insight, and those high in this trait also tend to have a broad range of interests. People who are high in this trait tend to be more adventurous and creative. People low in this trait are often much more traditional and may struggle with abstract thinking.
It is important to note that each of the five personality factors represents a range between two extremes.
For example, extraversion represents a continuum between extreme extraversion and extreme introversion. In the real world, most people lie somewhere in between the two polar ends of each dimension (people with a high score in openness are artistic, curious, imaginative, etc.).
\item \textbf{Conscientiousness}. Standard features of this dimension include high levels of thoughtfulness, with good impulse control and goal-directed behaviors. Those high on conscientiousness tend to be organized and mindful of details (people with a high score in conscientiousness are responsible, organised, etc.).
\item \textbf{Extraversion}. Extraversion is characterized by excitability, sociability, talkativeness, assertiveness and high amounts of emotional expressiveness.
People who are high in extroversion are outgoing and tend to gain energy in social situations. People who are low in extroversion (or introverted) tend to be more reserved and have to expend energy in social settings (people with a high score in extraversion are energetic, talkative, sociable, etc.).
\item \textbf{Agreeableness}: This personality dimension includes attributes such as trust, altruism, kindness, affection and other prosocial behaviors.
People who are high in agreeableness tend to be more cooperative while those low in this trait tend to be more competitive and even manipulative (people with a high score in agreeableness are trustful, kind, etc.).
\item \textbf{Neuroticism}. It is a trait characterized by sadness, moodiness, and emotional instability. Individuals who are high in this trait tend to experience mood swings, anxiety, moodiness, irritability and sadness. Those low in this trait tend to be more stable and emotionally resilient (people with a high score in neuroticism are nervous, emotionally unstable, etc.).
\end{itemize}
A personality profile can be obtained by completing a questionnaire that includes questions regarding the above traits, with answers ranging from totally disagree to totally agree. The score of each trait is then calculated based on the individual's answers. The questionnaire used in this work to measure the personality traits was a short version of the Big Five Inventory (BFI-44) with just 10 items (BFI-10) \cite{rammstedt:2007}.

\item The \textbf{Rahim Organizational Conflict Inventory II}  (ROCI-II) \cite{Rahim1983}

Conflict is a social phenomena occurring between two or more interacting individuals that have different individual goals. In this work, the Rahim Organizational Conflict Inventory II (ROCI - II) \cite{Rahim1983} has been used to measure individual attitudes in regard to handling conflict and disagreement. In particular, the ROCI–II measures the 5 conflict management styles (integrating, avoiding, dominating, obliging, and compromising) identified by Rahim.

It consist of 28 statements on a 5–point Likert scale measuring five independent dimensions of the styles of handling interpersonal conflict: 7 statements for Integrating (IN), 6 statements for Obliging (OB), 5 statements for Dominating (DO), 6 statements for Avoiding (AV), and 4 statements for Compromising (CO). The instrument contains Forms A, B, and C to measure how an organizational member handles conflict with supervisor, subordinates, and peers, respectively. A higher score represents greater use of a conflict style.
Sample items of the instrument are reported as follows:
\begin{itemize}
\item \textbf{Compromising}:  I try to find a middle course to resolve an impasse.
\item \textbf{Avoiding}: I attempt to avoid being “put on the spot” and try to keep my conflict with my supervisor/subordinates/peers to myself.
\item \textbf{Obliging}: I generally try to satisfy the needs of my supervisor/subordinates/peers.
\item \textbf{Dominating}: I use my influence to get my ideas accepted.
\item \textbf{Integrating}: I try to investigate an issue with my supervisor/subordinates/peers to find a solution acceptable to us.
\end{itemize}
A summary of the 5 styles of handling conflicts is reported below:
\begin{itemize}
\item \textbf{Compromising}: Tendency to find a compromise between different positions or interests.
\item \textbf{Avoiding}: involves low concern for self as well as the other party. It is associated with withdrawal, attribution of responsibility for one's own actions to others, sidestepping, ignoring of undesirable information. Tendency to avoid conflict.
\item \textbf{Obliging}: involves of low concern for self and high concern for the other party involved in the conflict. It is expression of attempts to play down the differences and emphasise the commonalities to satisfy the other party. Tendency to accept the position of others.
\item \textbf{Dominating}: involves high concern for self and low concern for the other party. It is a I win- you lose orientation and forces behaviour to affirm one’s position. Tendency to impose one's views in case of disagreement.
\item \textbf{Integrating}: expression of high concern for self as well as the other party involved
in the conflict. It involves collaboration between parties to reach a solution. Tendency to attract others towards one's own positions.
\end{itemize}
From the questionnaire, scores measuring each of the above dimensions can be obtained to form the individual's conflict handling style.
\end{enumerate}

Summarizing, the SSP-Chat provides a new text chat corpus which consists of 60 subjects (30 conversations), for each subject demographic information, conflict handling style, and personality traits have been acquired. Data contains the timestamps of each character pressed by the subjects during the experiment. Moreover, the subjects were involved in a negotiation task, where all the negotiation outcomes have also been recorded. This corpus promises a high scientific potential, it allows the  investigation of social cues and behaviors that may affect a conversation or a negotiation outcome. A pilot work on this dataset is presented in Chapter \ref{ch7:CHATS}.

\section{The ADS-16 Dataset}\label{ch4:ADS16}

\begin{figure*}[!]
 \centering
\includegraphics[width=0.5\textwidth]{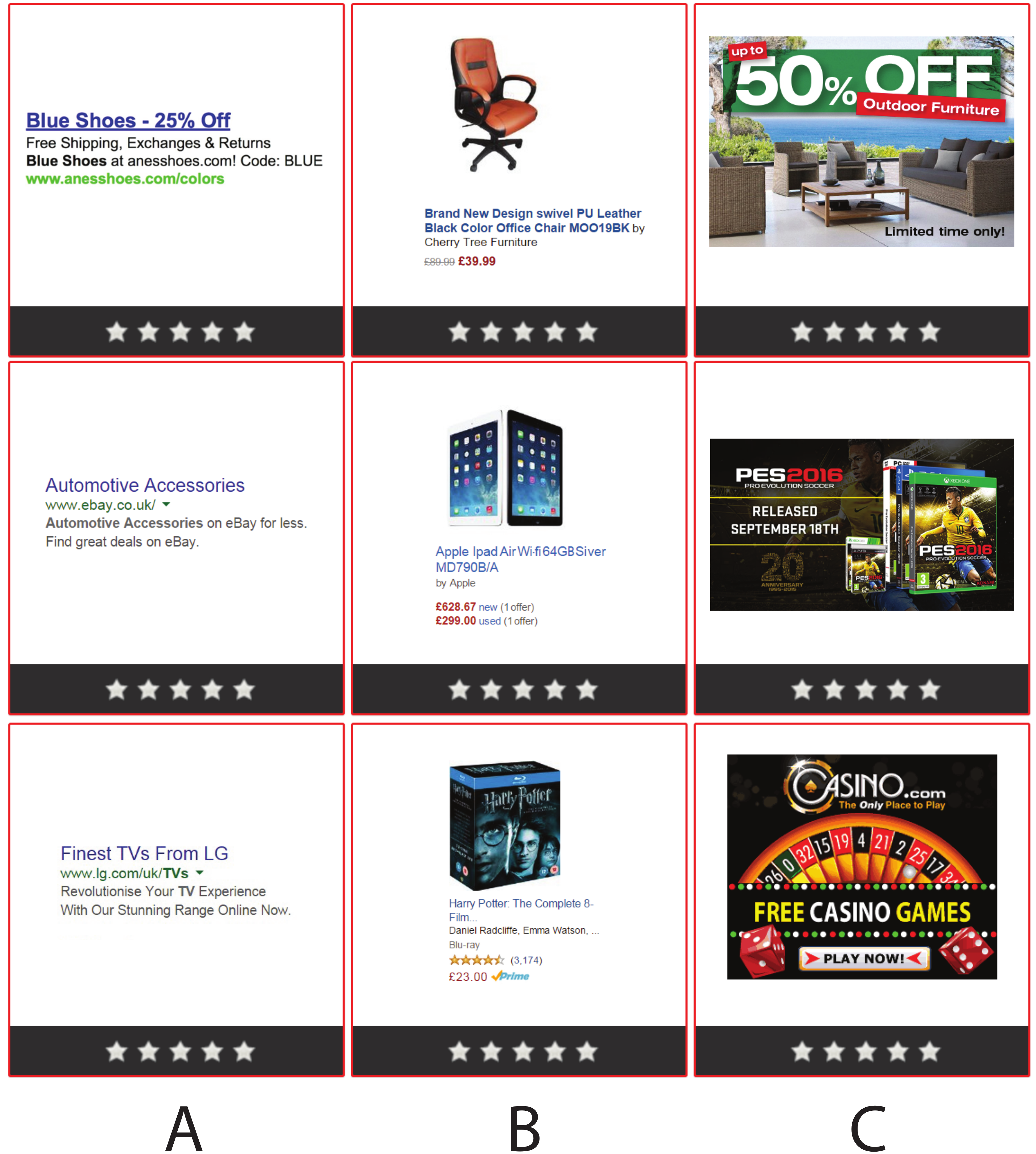}
\caption{The figure shows three different examples for each display format. (A) Shows Text Ads that received 26.5\% of the total amount of clicks. (B) Image Ads (32.7\% of clicks), and (C) Rich Media Ads (40.8\% of clicks).}\label{fig4:ads}
\end{figure*}

In the last decade, new ways of shopping online have increased the possibility of buying products and services more easily and faster than ever. In this new context, personality is a key determinant in the decision making of the consumer when shopping. A person's buying choices are influenced by psychological factors like impulsiveness; indeed some consumers may be more susceptible to making impulse purchases than others. Since affective metadata are more closely related to the user's experience than generic parameters, accurate predictions reveal important aspects of user's attitudes, social life, including attitude of others and social identity. This section presents a highly innovative research that uses a personality perspective to determine the unique associations among the consumer's buying tendency and advert recommendations. The ADS-16 includes $300$ advertisements voted by unacquainted individuals (120 subjects in total.  Note, the data collection process is still running). Adverts equally cover three display formats: Rich Media Ads, Image Ads, Text Ads (i.e., $100$ ads for each format). Participants rated (from 1-star to 5-stars) each recommended advertisement according to if they would or would not click on it (some examples are shown in the Fig.\ref{fig4:ads}). We labeled adverts as ``clicked''  (rating
greater or equal to four), otherwise ``not clicked'' (rating less than four). The distribution of the ratings across the adverts that were scored by the users turns out to be unbalanced (4,841 clicked vs 31,159 unclicked). 

Advert content is categorized in terms of $20$ main product/service categories. For each one of the categories $15$ real adverts are provided. Table~\ref{table4:adsCat} reports the full list of the categories used with the associated class annotations and the percentage of clicks received. At the category level, the distribution of the ratings results to be balanced (1,229 clicked vs 1,171 unclicked), where a category is considered to be clicked whenever it contains at least one clicked advert. 
\begin{table}[!t]
\centering
\begin{tabular}{c|l | c}
\textbf{Class Labels} & \textbf{Category Name}   &    \textbf{\% Clicks}            \\\hline
1  & \color{PineGreen}\textbf{Clothing \& Shoes}     &    \textbf{ 6.2\%}  \\
2  & Automotive        &       3.3\%    \\
3  & Baby             &        3.3\%   \\
4  & Health \& Beauty    &      6.0\%  \\
5  & Media                &     6.6\%   \\
6  & \color{PineGreen}\textbf{Consumer Electronics}      &   \textbf{9.2\%}\\
7  & \color{PineGreen}\textbf{Console \& Video Games}  &    \textbf{8.5\%} \\
8  & \color{red}Tools \& Hardware     &     3.0\%  \\
9  & Outdoor Living       &    5.6\%   \\
10 & \color{PineGreen}\textbf{Grocery}               &     \textbf{7.3\%} \\
11 & Home                &     4.7\%   \\
12 & \color{red}Betting               &    1.6\%   \\
13 & Jewelery \& Watches       &  5.9\%\\
14 & Musical instruments        &  3.6\%\\
15 & Stationery \& Office Supplies & 5.4\%\\
16 & Pet Supplies       &     3.1\%  \\
17 & Computer Software       &  5.6\%\\
18 & Sports          &       5.0 \%    \\
19 & Toys \& Games         &  5.1\% \\
20 & \color{red}Social Dating Sites    & 1.0\% \\         
\end{tabular}
\caption{ADS Dataset provides a set of $15$ real adverts categorized in terms of $20$ product/service categories. The most clicked categories are highlighted in green and the less clicked in red.}\label{table4:adsCat}
\end{table}
Inspired from recent findings which investigate the effects of personality traits on online impulse buying~\cite{Bosnjak2007,soBob,Turkyilmaz201598}, and many other approaches based upon behavioral economics, lifestyle analysis, and merchandising effects~\cite{Bosnjak2007,Mowen2000}, the proposed dataset supports a trait theory approach to study the effect of personality on user's motivations and attitudes toward online in-store conversions. The trait approach was selected because it encourages the use of scientifically sound scale construction methods for developing reliable and valid measures of individual differences. As a result, the corpus includes the Big Five Inventory-10 to measure personality traits~\cite{rammstedt:2007}, the five factors have been defined as \textit{openness to experience}, \textit{conscientiousness}, \textit{extraversion}, \textit{agreeableness}, and \textit{neuroticism}, often listed under the acronyms \textit{OCEAN}. \\

Recent soft-biometric approaches have shown the ability to unobtrusive acquire these traits from social media~\cite{Roffo:HBU2014,Roffo:icmi2014}, or infer the personality types of users from visual cues extracted from their favorite pictures from a social signal processing perspective~\cite{Vinciarelli_JIVC_2009}. While not necessarily corresponding to the actual traits of an individual, attributed traits are still important because they are predictive of important aspects of social life, including attitude of others and social identity. 

\begin{table*}[!t]
\centering
  \resizebox{1\textwidth}{!}{%
\begin{tabular}{|m{2.4cm}|m{3.3cm}|m{8cm}|m{2.2cm}|}
\hline
\textbf{Group}  &\textbf{Type} & \textbf{Description} & \textbf{References}\\\hline
\textbf{Users' \newline Preferences} 
& Websites, Movies, Music, TV Programmes, Books, Hobbies &  Categories of: websites users most often visit (WB), watched films (MV), listened music (MS), watched T.V. Programmes (TV), books users like to read (BK), favourite past times, kinds of sport, travel destinations.& \cite{He:2014,Lee20102142,NzVald20121186}\\
 
\hline
\textbf{Demographic} & Basic information  & Age, nationality, gender, home town, CAP/zip-code, type of job, weekly working hours, monetary well-being of the participant &\cite{NzVald20121186}\\

\hline
 \textbf{Social\newline Signals} &  Personality Traits&  BFI-10: Openness to experience, Conscientiousness, Extraversion, Agreeableness, and Neuroticism (OCEAN) &~\cite{Chen201657,rammstedt:2007}\\
 &  Images/Aesthetics  & Visual features from a gallery of 1.200 \textit{positive / negative} pictures and related meta-tags &~\cite{lovato2012tell}\\ 
\hline
 \textbf{Users' \newline Ratings} &  Clicks & 300 ads annotated with Click / No Click by 120 subjects &~\cite{He:2014,Avila16,wang2010click}\\
 &  Feedback  & From 1-star (Negative) to 5-stars (Positive) users' feedback on 300 ads &~\cite{He:2014,Avila16,wang2010click}\\ 

\hline	
\end{tabular}}
  \caption{The table reports the type of raw data provided by the ADS Dataset. Data of the first and last group can be considered as historical information about the users in an offline user study. }
  \label{tab:features}
\end{table*}

As a result, the proposed benchmark includes 1,200 spontaneously uploaded images that hold a lot of meaning for the participants and their related annotations: \textit{positive/negative} (see Table~\ref{tab:features} for further details). The images are personal (i.e., family, friends etc.) or just images participants really like/dislike. The motivations for labeling a picture as favorite are multiple and include social and affective aspects like, e.g., positive memories related to the content and bonds with the people that have posted the picture. Moreover, they are provided with a set of TAGS describing the content of each of them. 

Finally, many other users' preference information are provided. Table~\ref{tab:features} lists the raw data provided with the dataset, such as users' past behavior selected from a pre-defined list (e.g., watches movies, listen songs, read books, travel destinations, etc.), demographic information (like age, nationality, gender, etc.). Note, all data is anonymized (i.e., name, surname, private email, etc.), ensuring the privacy of all participants.  
 
For further analyses related to the adverts' quality, this benchmark also provides the entire set of 300 rated advertisements (500 x 500 pixels) in PNG format.

%
%
\subsection{Participant Recruitment}\label{sec4:protocol}

The subjects involved in the data collection, performed all the steps of the following protocol:\\
- \textbf{Step 1}: All participants have filled in a form providing, anonymously, several information about their preferences (e.g., demographic information, personal preferences).\\
- \textbf{Step 2}: All participants have filled the Big Five Inventory-10 to measure personality traits~\cite{rammstedt:2007}.\\
- \textbf{Step 3}: The participants voted each advert according with if they would or not click on the recommended ad. Ads have been displayed in the same order to all the participants.\\
- \textbf{Step 4}: The participants submitted some images that they like (Positives) and some others that disgust or repulse them (Negatives). Once they have uploaded their images, they also added some TAGS that describe the content of each image. \\

%
%
\subsection{The Subjects}\label{sec4:subjects}
\begin{figure*}[!]
 \centering
\includegraphics[width=0.999\textwidth]{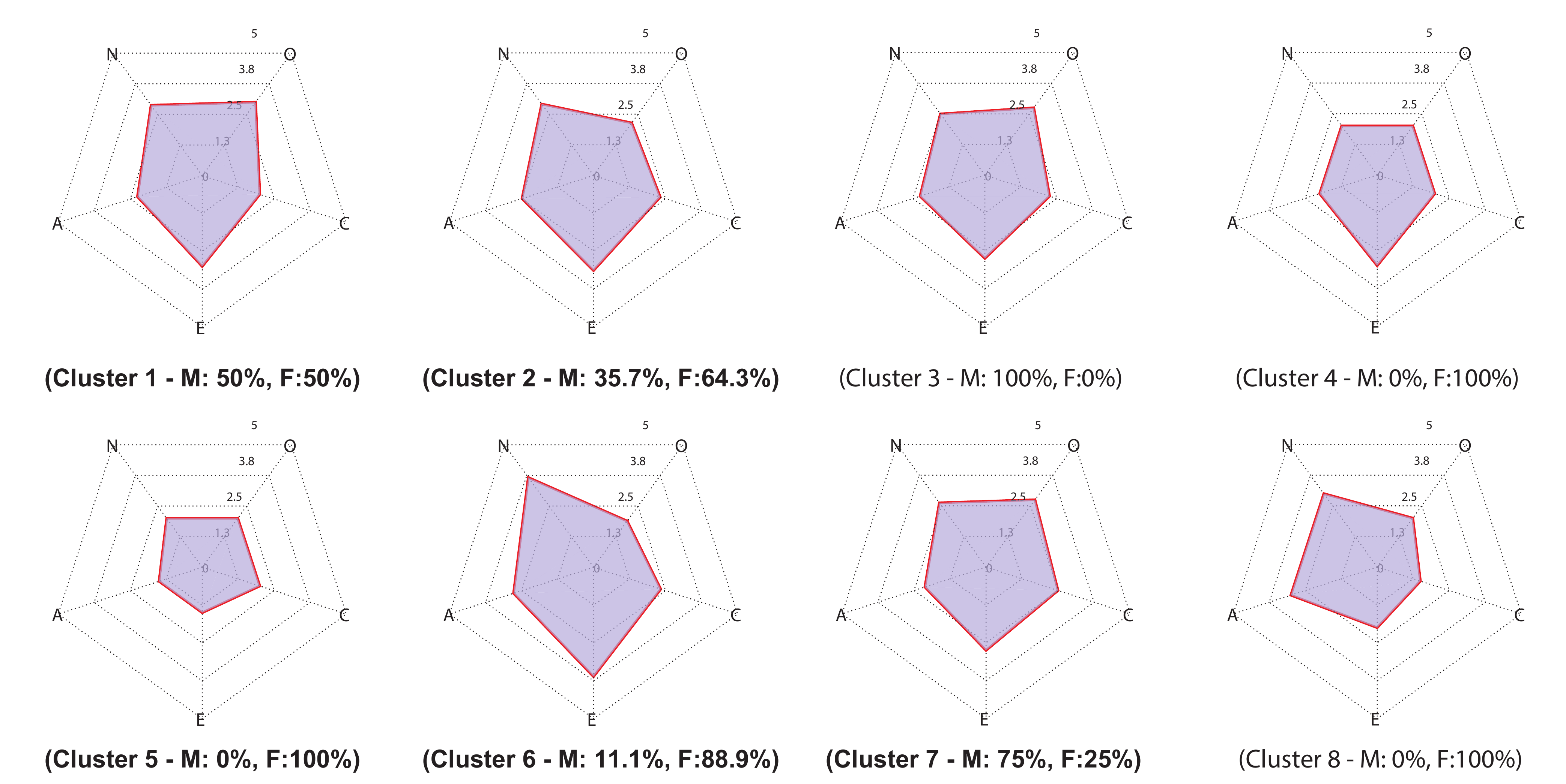}
\caption{Spider-Diagrams for \textit{O-C-E-A-N} Big-Five traits. The percentage of Males (M) and Females (F) belonging to each cluster is reported. We indicate in bold each instance where a statistacal significant effect (i.e., Pearson correlation at the 5\% level) was found between ranks and personality factors.}\label{fig4:Traits}
\end{figure*}
This corpus involves 120 English native speakers between 18 and 68. The median of the participants age is 28 ($\mu$=31.7, $\sigma$=12.1). Most of the participants have a university education. In terms of gender, 77 are females and 43 males. The percentage distribution of household income within the sample is: 23\% less or equal to 11K USD per year, 48\% from 11K to 50K USD, 21\% from 50K to 85K USD, and 8\% more than 85K USD. The median income is between 11K and 50K USD. 

In analyzing this complex data, one can observe that users' preferences are not independent of each other, they are likely to be co-expressed. Hence, it is of great significance to study groups of preferences rather than to perform a single analysis. This fact is also true for personality factors, analyzing subsets of data yields crucial information about patterns inside the data. Thus, clustering users' preferences can provide insights into personality of individuals which share the same preferences. We performed a statistical analysis of personality and users' preferences, linking the $5$ personality factors and the most favorite users' product categories (i.e., most clicked) by means of the affinity propagation (AP) clustering algorithm~\cite{frey07affinitypropagation}.  

AP is an algorithm that takes as input measures of similarity between pairs of data points and simultaneously considers all data points as potential exemplars. We calculated a similarity input matrix between each individual $u_i$ considering as feature vectors $v_i$ a binary sequence of click/no-click (i.e., $v_i$ is $1 \times 300$). AP exchanges real-valued messages between data points until a high-quality set of exemplars and corresponding clusters gradually emerges. Hence, the number of clusters is automatically detected, and when applied on ADS data, AP grouped the data into 8 different clusters.  Figure~\ref{fig4:Traits} illustrates 8 spider-diagrams, one for each cluster. Each diagram shows the average of the big-five factors regarding the subjects within the group (reported in figure as O-C-E-A-N). Then, we ranked the most clicked categories according with samples within the group in order to compare these two variables by means of correlation obtaining interesting clues. 

For instance, let us consider the cluster number $6$ where 88.9\% of the members are females, and 11.1\% males and the average of the group members age is 28. The first 5 most clicked categories are \textit{Baby}, followed by \textit{Consumer Electronics}, \textit{Stationery \& Office Supplies}, \textit{Home}, and \textit{Jewelery \& Watches}. This group is characterized by high neuroticism (see the diagram in Figure~\ref{fig4:Traits}.(Cluster 6)), those who score high in neuroticism are often emotionally reactive and vulnerable to stress, high neuroticism causes a reactive and excitable personality, often very dynamic individuals. This group also share the highest levels of extroversion, high extraversion is often perceived as attention-seeking, and domineering.

Cluster 5 shows a subset of individuals which scores low for all the types (see the plot in Figure~\ref{fig4:Traits}.(Cluster 5)). For instance, those with low openness seek to gain fulfillment through perseverance, and are characterized as pragmatic sometimes even perceived to be dogmatic. Some disagreement remains about how to interpret and contextualize the openness factor. The first 5 most clicked categories are \textit{Clothing \& Shoes}, \textit{Health \& Beauty}, \textit{Jewelery \& Watches}, \textit{Outdoor Living}, and then \textit{Consumer Electronics}. In this case the average of the group members age is 68, and the cluster contains 100\% females.

\begin{table}[!ht]
\small
\centering
\begin{tabular}{c |  c | c   c   c   c   c     }
 Cluster Id& Avg. Age & r.1 & r.2 & r.3 & r.4 & r.5 \\\hline
\textbf{1}&  \textbf{32} & \textbf{6}   & \textbf{15}   & \textbf{13}   & \textbf{19}   & \textbf{ 4}\\
\textbf{2}& \textbf{31} &  \textbf{6}  &   \textbf{5}  &  \textbf{ 7}  &  \textbf{10 }  & \textbf{17}\\
3 &22 &1  &   7   & 10   &  4   &  6\\
4&57    &6&     9  &  10 &    2   &  1\\
\textbf{5}& \textbf{ 68 } & \textbf{1} &     \textbf{4 } & \textbf{ 13}  &  \textbf{9 } &  \textbf{6}\\
\textbf{6}& \textbf{28 }& \textbf{3 }  &  \textbf{6}  &  \textbf{15 }&   \textbf{11}  &  \textbf{13}\\
\textbf{7}&  \textbf{20 } &\textbf{7 } &   \textbf{6 }&   \textbf{10} &   \textbf{11}  &  \textbf{17}\\
8&     52 &3  &   9  &  19  &   7   &  4\\
\end{tabular} 
\caption{Top-5 ranked categories. For each cluster the table reports the average age, and the ordered list of the most clicked categories. We indicate in bold each instance where a statistical significant effect (i.e., Pearson correlation at the 5\% level) was found between ranks and personality factors.}
\label{tab:table0}\vspace{-0.1cm}
\end{table} 
Cluster 7 is characterized by good levels of conscientiousness that is the tendency to be organized and dependable, aim for achievement, and prefer planned rather than spontaneous behavior. This cluster scores low in agreeableness, which is related to personalities often competitive or challenging people. The openness factor ($>$2.5) reflects the degree of intellectual curiosity, creativity and a preference for novelty and variety a person has. Interestingly, among the most preferred categories there are \textit{Console \& Video Games}, \textit{Consumer Electronics}, \textit{Grocery} and \textit{Computer Software}.

\section{Conclusions}

Regardless of the field of study, accurate data collection is essential to maintaining the integrity of research. Both the selection of appropriate data collection instruments (existing, modified, or newly developed) and clearly delineated instructions for their correct use reduce the likelihood of errors occurring. 
This chapter presented four different datasets ( $C$-$Skype$ sec. \ref{ch4:skype}, $TBKD$ sec. \ref{ch4:TBK}, $SSP$-$Chat$ $Corpus$ sec. \ref{ch4:SSPChat}, and $ADS$-$1$6 sec. \ref{ch4:ADS16}). For each of them, data acquisition, experimental protocol, and the process of analyzing data and summarizing it in useful information have been discussed. Data collection has been carried out in order to maintain integrity, accuracy and completeness.

\part{Ranking to Learn}\label{part:two}

\chapter{Feature Ranking and Selection for Classification}\label{ch5:FS}

In an era where accumulating data is easy and storing it inexpensive, feature selection plays a central role in helping to reduce the high-dimensionality of huge amounts of otherwise meaningless data. Advances over the past decade have highlighted the fact that feature selection leads to many potential benefits. Selecting subset of discriminant features helps in data visualization and understanding, in reducing the measurement and storage requirements, and in reducing training and utilization times. Moreover, in many modern classification scenarios, the number of adopted features is usually very large, and it is continuously growing due to several causes. Data sets grow rapidly, in part because they are increasingly gathered by cheap and numerous information-sensing mobile devices, aerial (remote sensing), software logs, cameras, microphones, radio-frequency identification readers and wireless sensor networks \cite{Dobre:2014,segaran2009beautiful}. The world's technological per-capita capacity to store information has roughly doubled every 40 months since the 1980s as of 2012, every day 2.5 exabytes ($2.5 \times 10^{18}$)) of data are generated  \cite{Hilbert:8805204}. The management of high-dimensional data requires a strong feature selection to individuate irrelevant and/or redundant features and avoid overfitting \cite{guyon2006feature}. 
In particular, feature selection has become crucial in classification settings like object recognition, recently faced with feature learning strategies that originate thousands of cues. Feature selection is also used in mining complex patterns in data. The rationale behind this fact is that feature selection is different from dimensionality reduction. Both approaches seek to reduce the number of attributes in the dataset, but a dimensionality reduction method do so by creating new combinations of attributes, where as feature selection methods include and exclude attributes present in the data without changing them. Examples of dimensionality reduction methods include Principal Component Analysis (PCA), Singular Value Decomposition (SVD) and Sammon’s Mapping. On the other hand, feature selection is itself crucial and it mostly acts as a filter, muting out features that are not useful in addition to the existing features (i.e., redundancy). 

Ranking to learn is inserted in the above scheme, and it aids one to create an accurate predictive model by firstly ranking all the the features available and then selecting a subset of the most relevant ones. In particular, ranking to learn in classification settings helps in choosing features that will lead to as good or better accuracy of the system whilst requiring less data. Feature selection methods can be used to identify and remove unneeded, irrelevant and redundant attributes from data that do not contribute to the accuracy of a predictive model or may in fact decrease the accuracy of the model. Fewer attributes is desirable because it reduces the complexity of the model, and a simpler model is simpler to understand and explain. Therefore, \textit{the objective of variable selection is three-fold: improving the prediction performance of the predictors, providing faster and more cost-effective predictors, and providing a better understanding of the underlying process that generated the data} \cite{guyon2006feature}.

In this chapter we propose two feature selection methods exploiting the convergence properties of power series of matrices, and introducing the concept of graph-based feature selection \cite{Roffo_2015_ICCV,RoffoECML16,roffomelzi2016}. The most appealing characteristic of the approaches is that they evaluate the importance of a given feature while considering all the other cues, in such a way that the score of each feature is influenced by all the other features of the set. The idea is to map the feature selection problem to an affinity graph, and then to consider a subset of features as a path connecting them. The cost of the path is given by the combination of pairwise relationships between the features. 

The first approach we are about to present (in Section \ref{ch5:infFS11}) evaluates analytically the redundancy of a feature with respect to all the other ones taken together, considering paths of a length that tends to infinity. For this reason, we dub our approach \textit{infinite feature selection} (Inf-FS). The Inf-FS is extensively tested on 13 benchmarks of cancer classification and prediction on genetic data (\emph{Colon}~\cite{alon}, \emph{Lymphoma}~\cite{Golub99}, \emph{Leukemia}~\cite{Golub99}, \emph{Lung181}~\cite{Gordon02},  DLBCL~\cite{citeulike:165595}), handwritten recognition (USPS~\cite{USPS,ChaSchZie06}, GINA~\cite{GINA}, \emph{Gisette}~\cite{guyon2004result}), generic feature selection (MADELON~\cite{guyon2004result}), and more extensively, object recognition (Caltech 101-256~\cite{FeiFeiFergusPeronaPAMI}, PASCAL VOC 2007-2012~\cite{pascal-voc-2007,pascal-voc-2012}). We compare the proposed method on these datasets, against eight comparative approaches, under different conditions (number of features selected and number of training samples considered), overcoming all of them in terms of stability and classification accuracy, and setting the state of the art on 8 benchmarks, notably all the object recognition datasets. Additionally, Inf-FS also allows the investigation of the importance of different kinds of features, and in this study, the supremacy of deep-learning approaches for feature learning has been shown on the object recognition tasks.

The second approach (Section \ref{ch5:EigVec21}) gives the solution of the problem by assessing the importance of nodes through some indicators of centrality, in particular, the \emph{Eigenvector Centrality (EC)}. The gist of EC-FS is to estimate the importance of a feature as a function of the importance of its neighbors. Ranking central nodes individuates candidate features, which turn out to be effective from a classification point of view, as proved by a thoroughly experimental section. The EC-FS is of valuable interest as it helps in understanding various factors influencing the Inf-FS. The main difference between these two approaches is that the EC-FS does not account for all the partial solutions as the Inf-FS does. In other words, the Inf-FS integrates each partial solution (e.g., $sol_1+sol_2+...+sol_l, l \to \infty$), that together help the method to react in a different way in presence of noise in data or many redundant cues.

%
%
\section{Infinite Feature Selection}\label{ch5:infFS11}

In the Inf-FS formulation, each feature is a node in the graph, a path is a selection of features, and the higher the score, the most important (or most different) the feature. Therefore, considering a selection of features as a path among feature distributions and letting these paths tend to an infinite number permits the investigation of the importance (relevance and redundancy) of a feature when injected into an arbitrary set of cues. Ranking the importance individuates candidate features, which turn out to be effective from a classification point of view, as proved by a thoroughly experimental section. The Inf-FS has been tested on thirteen diverse benchmarks, comparing against filters, embedded methods, and wrappers; in all the cases we achieve top performances, notably on the classification tasks of PASCAL VOC 2007-2012. The novelty of Inf-FS in terms of the state of the art is that it assigns a score of ``importance" to each feature by taking into account \emph{all the possible feature subsets} as paths on a graph, bypassing the combinatorial problem in a methodologically sound fashion.

\subsection{The Method}\vspace{-0.05cm}

Given a set of feature distributions $F = \{  f^{(1)}, ...,  f^{(n)} \}$ and $x\in \mathcal{R}$ representing a sample of the generic distribution $f$, we build an undirected fully-connected graph $G = (V, E)$; $V$ is the set of vertices corresponding, one by one, to each feature distribution, while $E$ codifies (weighted) edges, which model pairwise relations among feature distributions.  Representing $G$ as an adjacency matrix $A$, we can specify the nature of the weighted edges: each element $a_{ij}$ of $A$, $1\leq i,j \leq n$, represents a pairwise energy term. Energies have been represented as a weighted linear combination of two simple pairwise measures linking $f^{(i)}$ and $f^{(j)}$, defined as:
\begin{equation}\label{eq5:partzero}
  a_{ij} = \alpha \varrho_{ij} + (1-\alpha) c_{ij},
\end{equation}
where $\alpha$ is a loading coefficient $\in [0,1]$, $\varrho_{ij} = max\left(\sigma^{(i)},\sigma^{(j)}\right)$,
with $\sigma^{(i)}$ being the standard deviation over the samples $\{x\}\in f^{(i)}$, and the second term is  $c_{ij} = 1 - \left|Spearman(f^{(i)},f^{(j)}) \right|$,
with $Spearman$ indicating Spearman's rank correlation coefficient. In practice, $a_{ij}$ connects two feature distributions, accounting for the maximal feature dispersion and their correlation. Note that the standard deviation is normalized by the maximum standard deviation over the set $F$ and that $|Spearman(\cdot,\cdot)|\in [0,1]$, so the two measures are comparable in terms of magnitude. The idea is that, suppose $\alpha=0.5$, a high $a_{ij}$ indicates at least one feature among $f^{(i)}$ and $f^{(j)}$ could be discriminant since it covers a large feature space, and $f^{(i)}$ and $f^{(j)}$ are not redundant~\cite{Gennari:1989}.

After this pairwise analysis of features, we want to individuate the energy associated to sets larger than two feature distributions. Let $\gamma = \{ v_{0}=i,v_{1}, ..., v_{l-1}, v_{l}=j \}$ denote a path of length $l$ between vertices $i$ and $j$, that is, features $f^{(i)}$ and $f^{(j)}$, through other features $v_{1},...,v_{l-1}$. For simplicity, suppose that the length $l$ of the path is lower than the total number of features $n$, and the path has no cycles, so no features are visited more than once. In this setting, a path is simply a subset of the available features that come into play.
We can then define the  \emph{energy} of $\gamma$ as
\begin{equation}\label{eq5:two}
\mathcal{E}_{\gamma } = \prod_{k=0}^{l-1} a_{v_{k},v_{k+1}},
\end{equation}
where $\mathcal{E}_{\gamma }$ essentially accounts for the pairwise energies of all the features' pairs that compose the path, and it can be assumed as the joint energy of the subset of features. Now we relax the assumption of the presence of cycles, and
we define the set $\mathbb{P}_{i,j}^l$ as containing all the paths of length $l$ between $i$ and $j$; to account for the energy of all the paths of length $l$, we sum them as follows:
\begin{equation}\label{eq5:three}
R_{l}(i,j) =\sum_{\gamma \in \mathbb{P}_{i,j}^l }  \mathcal{E}_{\gamma},
\end{equation}
which, following standard matrix algebra, gives:
$$ R_{l}(i,j) = A^{l}(i,j) , $$
that is, the power iteration of $A$. Much attention should be paid to $R_{l}$, which now contains cycles; in terms of feature selection, it is like if a single feature is considered more than once, possibly associated to itself (a self-cycle), or if two or more features are repeatedly used (e.g., the path $<1,2,3,1,2,3,4>$ connects feature 1 and 4 by a 3-variable cycle). Anyway, by extending the path length to infinity, the probability of being part of a cycle is uniform for all the features and is actually taken into account by the construction of $R_{l}$, so a sort of normalization comes into play. Given this, we can evaluate the \emph{single feature energy score} for the feature $f^{(i)}$ at a given path length $l$ as

\begin{equation}\label{eq5:four}
s_{l}(i) = \sum_{j \in V} R_{l}(i,j) = \sum_{j \in V} A^{l}(i,j),
\end{equation}
In practice, Eq.\ref{eq5:four} models the role of the feature $f^{(i)}$ when considered in whatever selection of $n$ features; the higher $s_{l}(i)$ is, the more energy is related to the $i$-th feature.
Therefore, a first idea of feature selection strategy could be that of ordering the features decreasingly by $s_{l}$, taking the first $m$ for obtaining an effective, nonredundant set. Unfortunately, the computation of $s_{l}$ is expensive ($\mathcal{O}((l-1)\cdot n^3)$): as a matter of facts, $l$ is of the same order of $n$, so the computation turns out to be $\mathcal{O}(n^4)$ and becomes impractical for large sets of features to select ($>10K$); Inf-FS addresses this issue 1) by expanding the path length to infinity $l\rightarrow \infty$ and 2) using algebra notions to simplify the calculations in the infinite case.

\subsection{Infinite sets of features}\vspace{-0.05cm}

The passage to infinity implies that we have to calculate a new type of single feature score, that is,
\begin{equation}\label{eq5:fourPOINTfive}
s(i) = \sum_{l=1}^\infty s_{l} (i) =\sum_{l=1}^\infty \big( \sum_{j \in V} R_{l}(i,j) \big).
\end{equation}
Let $S$ be the geometric series of matrix $A$:
\begin{equation}\label{eq5:fourPOINTfive2}
 S=\sum _{l=1}^{\infty} A^{l},
\end{equation}
It is worth noting that $S$ can be used to obtain $s(i)$ as
\begin{equation}\label{eq5:preliminaries2}
s(i) = \sum _{l=1}^{\infty} s_{l}(i)=[(\sum ­_{l=1}^{\infty}A^{l})\textbf{e}]_{i}=[S \textbf{e}]_{i},
\end{equation}
where $\textbf{e}$ indicates a $1D$ array of ones. As it is easy to note, summing infinite $A^l$ terms brings to divergence; in such a case, regularization is needed, in the form of generating functions~\cite{Graham:1994}, usually employed to assign a consistent value for the sum of a possibly divergent series. There are different forms of generating functions~\cite{Bergshoeff}. We define the generating function for the $l$-path as
\begin{equation}\label{eq5:five}
\check{s}(i) = \sum_{l=1}^\infty  r^{l} s_{l} (i) = \sum_{l=1}^\infty\sum_{j \in V} r^{l} R_{l}(i,j),
\end{equation}
where $\textit{r}$ is a real-valued regularization factor, and $\textit{r}$ raised to the power of $l$ can be interpreted as the weight for paths of length $l$. Thus, for appropriate choices of $\textit{r}$, we can ensure that the infinite sum converges. From an algebraic point of view, $\check{s}(i)$ can be efficiently computed by using the convergence property of the geometric power series of a matrix~\cite{HubHub01}:
\begin{equation}\label{eq5:six}
          \check{S} = (\textbf{I} - \textit{r}A)^{-1} - \textbf{I},
\end{equation}
Matrix $\check{S}$ encodes all the information about the energy of our set of features, the goodness of this measure is strongly related to the choice of parameters that define the underlying adjacency matrix $A$. We can obtain final energy scores for each feature simply by marginalizing this quantity:
\begin{equation}\label{eq5:seven}
          \check{s}(i) = [\check{S} \textbf{e}]_{i},
\end{equation}
and by ranking in decreasing order the $\check{s}(i)$  energy scores, we obtain a rank for the feature to be selected. It is worth noting that so far, no label information has been employed. The ranking can be used to determine the number $m$ of features to select, by adopting whatever classifier and feeding it with a subset of the ranked features, starting from the most energetic one downwards, and keeping the $m$ that ensures the highest classification score. \red{This last operation resembles the works on graph centrality~\cite{bonacich1987power} (see for an example \cite{zhu2007improving}), whose goal was to rank nodes in social networks that would be visited the most, along whatever path in the structure of the network. In our case, the top entries in the rank are those features more different w.r.t all the other ones, irrespective on the subsets of cues they lie. Procedure \ref{algorithm} reports the sketch of the Inf-FS algorithm.}

\begin{algorithm}[H]
\caption{Infinite Feature Selection}
\label{algorithm}
\begin{algorithmic}
\REQUIRE{$F = \{ f^{(1)}, ..., f^{(n)} \}$ , \textbf{$\alpha$}}\\
\ENSURE{$\check{s}$ energy scores, for each feature }\\
Building the graph \\
\FOR{$i = 1 : n$}
\FOR{$j = 1 : n$}
\STATE $\varrho_{ij}=max( \sigma(f^{(i)}) ,\sigma(f^{(j)}) )$\;\\
\STATE $c_{ij}= 1-|Spearman( f^{(i)},f^{(j)})|$\;\\
\STATE $A(i,j) = \alpha \varrho_{ij} + (1 - \alpha ) c_{ij} $
\ENDFOR
\ENDFOR \\
Letting paths tend to infinite \\
\STATE $r =\frac{ 0.9}{\rho (A)}$\\
\STATE $\check{S} = (\textbf{I} - rA)^{-1} - \textbf{I}$\\
\STATE $\check{s} = \check{S} \thinspace \textbf{e}$
\RETURN $\check{s}$
\end{algorithmic}
\end{algorithm}

%
%
\subsection{Convergence analysis}\label{sec5:discussion}\vspace{-0.05cm}

In this section, we want to justify the correctness of the method in terms of convergence. The value of $r$ (used in the generating function) can be determined by relying on linear algebra~\cite{HubHub01}. Consider $\left\lbrace \lambda _{0},..., \lambda _{n-1} \right\rbrace $ eigenvalues of the matrix $\textbf{A}$, drawing from linear algebra, we can define the spectral radius $\rho (\textbf{A})$ as:
$$\rho(\textbf{A}) = \max_{\lambda_{i} \in \left\lbrace \lambda _{0},..., \lambda _{n-1} \right\rbrace} \Big( \vert \lambda _{i} \vert \Big) .$$
For the theory of convergence of the geometric series of matrices, we also have::
$$ \lim_{l \to \infty} \textbf{A} ^{l} = 0 \ \Longleftrightarrow \ \rho(\textbf{A})<1 \ \Longleftrightarrow \  \sum_{l=1}^{\infty} \textbf{A} ^{l} = (\textbf{I} - \textbf{A})^{-1} - \textbf{I} .$$
Furthermore, Gelfand`s formula ~\cite{powers2015mathematical} states that for every matrix norm, we have:
$$ \rho (\textbf{A}) = \lim _{k\longrightarrow \infty } \vert \vert \textbf{A} ^{k} \vert \vert ^{\frac{1}{k}}. $$
This formula leads directly to an upper bound for the spectral radius of the product of two matrices that commutes, given by the product of the individual spectral radii of the two matrices, that is, for each pair of matrices $\textbf{A}$ and $\textbf{B}$, we have:
$$ \rho ( \textbf{AB} ) \leq \rho (\textbf{A}) \rho (\textbf{B}) . $$
Starting from the definition of $ \check{s}(i) $ and from the following trivial consideration:
\begin{center}
$r^{l}\textbf{A}^{l} = \left( r^{l}\textbf{I} \right) \textbf{A}^{l} = \left[ \left(r\textbf{I}\right) \textbf{A} \right] ^{l}$,
\end{center}
we can use Gelfand`s formula on the matrices $r\textbf{I}$ and $\textbf{A}$ and thus obtain:
\begin{equation}
\rho \Big( \left( r\textbf{I} \right) \textbf{A} \Big) \leq \rho (r\textbf{I}) \rho (\textbf{A}) = r \rho (\textbf{A}).
\end{equation}
For the property of the spectral radius:
$\lim_{l \to \infty}\left( r\textbf{A} \right) ^{l} = 0 \Longleftrightarrow \rho (r\textbf{A}) < 1 $.
Thus, we can choose $r$, such as $0 <r<\dfrac{1}{\rho(\textbf{A})}$; in this way we have:
\begin{eqnarray}
0 <\rho (r\textbf{A}) &=& \rho \Big( \left( r\textbf{I} \right) \textbf{A}\Big) \leq \rho (r\textbf{I}) \rho (\textbf{A})\nonumber\\
 &=& r \rho (\textbf{A}) <\dfrac{1}{\rho(\textbf{A})} \rho (\textbf{A}) = 1 \,
\end{eqnarray}
that implies $\rho (r\textbf{A}) < 1 $, and so:
$$ \check{\textbf{S}} = \sum _{l=1}^{\infty} (rA)^{l} = (\textbf{I} - \textit{r}\textbf{A})^{-1} - \textbf{I}$$
This choice of $r$ allows us to have convergence in the sum that defines $\check{s}(i)$. Particularly, in the experiments, we use $r=\dfrac{0.9}{\rho(A)}$, leaving it fixed for all the experiments. To avoid the computation of the spectral radius it is possible to use the upper bound for the spectral radius of a nonnegative matrix by using its average 2-row sums \cite{duan2013sharp}. For the computational complexity of Inf-FS, see the next section.

\subsection{Experiments}\label{sec5:exp}\vspace{-0.05cm}

The experimental section has three main goals. The first is to explore the strengths and weaknesses of Inf-FS, also considering eight comparative approaches: four filters, three embedded methods, and one wrapper (see Table~\ref{table:ch5compmethods44}). The Inf-FS overcomes all of them, although it ignores class membership information, being completely unsupervised.  The second goal is to show how Inf-FS, when associated to simple classification models, allows the definition of top performances on benchmarks of cancer classification and prediction on genetic data (\emph{Colon}~\cite{alon}, \emph{Lymphoma}~\cite{Golub99}, \emph{Leukemia}~\cite{Golub99}, \emph{Lung181}~\cite{Gordon02},  DLBCL~\cite{citeulike:165595}), handwritten recognition (USPS~\cite{USPS,ChaSchZie06}, GINA~\cite{GINA}, \emph{Gisette}~\cite{guyon2004result}), generic feature selection (MADELON~\cite{guyon2004result}) and, more extensively, object recognition (Caltech 101-256~\cite{FeiFeiFergusPeronaPAMI}, PASCAL VOC 2007-2012~\cite{pascal-voc-2007,pascal-voc-2012}). Regarding the object recognition tasks, the third goal is to present a study that indicates which of the features commonly used in the recognition literature (bags of words and convolutional features) are the most effective, essentially confirming the supremacy of the deep learning approaches.

For setting the best parameters (the mixing $\alpha$, the $C$ of the linear SVM, and the number of features to consider on the object recognition datasets), we use only the training set (or the validation sets in the PASCAL series), implementing a 5-fold cross validation. For a fair comparative evaluation, we adopt the same protocol used in the selected comparative approaches (partition of the dataset, cross-validation, and other settings). Other specific validation protocols are explained in the following subsections.

\subsubsection{Datasets}\vspace{-0.05cm}
Datasets are chosen for letting Inf-FS deal with diverse feature selection scenarios, as shown on Table~\ref{table:ch5datasets}. In the details, we consider the problems of dealing with few training samples and many features (\emph{few train} in the table), unbalanced classes (\emph{unbalanced}), or classes that severely overlap (\emph{overlap}), or whose samples are noisy (\emph{noise}) due to: a) complex scenes where the object to be classified is located (as in the VOC series) or b) many outliers (as in the genetic datasets, where samples are often \emph{contaminated}, that is, artifacts are injected into the data during the creation of the samples). Lastly we consider the \emph{shift} problem, where the samples used for the test are not congruent (coming from the same experimental conditions) with the training data.

Table \ref{table:ch5datasets} also reports  the best classification performances so far (accuracy or average precision depending on the task), referring to the studies that produced them. 

\begin{table*}[ht]
\small
\centering
\resizebox{1\textwidth}{!}{%
\begin{tabular}{l c c c c c c c c c }
\hline \hline
Name &   \# samples & \# classes & \# feat.  & \emph{few train}& \emph{unbal. (+/-)} & \emph{overlap} & \emph{noise} & \emph{shift} & SoA \\\hline
USPS~\cite{USPS,ChaSchZie06} & 1.5K &2& 241   &   &  & X & & & 97.4\% ~\cite{maji2009fast} \\
GINA~\cite{GINA} & 3153 &2& 970   &   &  & X & & & 99.7\%~\cite{GINA}\\ 
\emph{Gisette}~\cite{NIPS2003} & 13.5K &2& 5K   &   &  & X & & & 99.9\% ~\cite{NIPS2003}\\
\hline
\emph{Colon}~\cite{alon} & 62 &2& 2K   &  X & (40/22) & & X  & & 89.6\%~\cite{LovatoBCJP12} \\
\emph{Lymphoma}~\cite{Golub99} & 45 &2& 4026   &  X & & & & & 93.8\%*~\cite{wang2014improved} \\
\emph{Leukemia}~\cite{Golub99} & 72 &2& 7129   &  X & (47/25) & & X & X & 97.2\%*~\cite{LeiYi10.1109}\\
\emph{Lung181}~\cite{Gordon02} & 181 &2& 12533   &  X & (31/150) & & X & X & 98.8\%*~\cite{LeiYi10.1109}\\
DLBCL~\cite{citeulike:165595} & 77 &2& 7129   &  X & (19/58) & & X & & 93.3\%*~\cite{seo2013novel} \\
\hline
MADELON~\cite{NIPS2003} & 4.4K  &2& 500   &   &  & X &  & & 98.0\%~\cite{NIPS2003}\\
\hline \vspace{0.02cm}
Caltech-101~\cite{FeiFeiFergusPeronaPAMI} & ~8K  &102& n.s.   & X &  &  &  & & 91.4\%*~\cite{HeZR014}\\
Caltech-256~\cite{FeiFeiFergusPeronaPAMI} & ~30K  &257& n.s.   & X & & X & X  & & 77.6\%*~\cite{Chatfield14}\\
VOC 2007~\cite{pascal-voc-2007} & ~10K &20& n.s.   & & X & & X & & 82.4\%*~\cite{Chatfield14} \\
VOC 2012~\cite{pascal-voc-2012} & ~20K &20& n.s.   & & X & & X & & 83.2\%*~\cite{Chatfield14}\\
\hline
\end{tabular}}
\caption{Panorama of the used datasets, together with the challenges for the feature selection scenario, and the state of the art so far. The abbreviation \emph{n.s.} stands for \emph{not specified} (for example, in the object recognition datasets, the features are not given in advance). We indicate with an asterisk each instance where the Inf-FS approach, together with a linear SVM, defines the new top performance.}
\label{table:ch5datasets}
\end{table*}

\subsubsection{Comparative approaches}\label{sec5:compAppr}\vspace{-0.05cm}

Table~\ref{table:ch5compmethods44} lists the methods compared, whose details can be found in Chapter ~\ref{ch2:classification}. Here we just note their \emph{type}, that is, \emph{f} = filters, \emph{w} = wrappers, \emph{e} = embedded methods, and their \emph{class}, that is, \emph{s} = supervised or \emph{u} = unsupervised (using or not using the labels associated with the training samples in the ranking operation).  Additionally, we report their computational complexity (if it is documented in the literature); finally, we report their timing when applied to a randomly generated dataset consisting of 20 classes, 10k samples, and 1k features (features follow a uniform distribution (range [0,1000])), on an Intel i7-4770 CPU 3.4GHz 64-bit, 16.0 GB of RAM, using MATLAB ver. 2015a. Note that only four of them have publicly available codes (that is, Relief-F~\cite{liu2008} , FSV~\cite{Bradley98featureselection,Grinblat:2010}, Fisher-G ~\cite{Quanquanjournals}, and MutInf~\cite{Hutter:02feature}), while in the other cases, we refer to the results reported in the literature.\\

\begin{table}[ht]
\small
\centering
\resizebox{0.6\textwidth}{!}{%
\begin{tabular}{p{3.5cm} C{0.5cm} C{0.2cm} C{2.5cm} C{0.8cm}}
\hline \hline
Acronym &   \small{Type} & \small{Cl.} & Compl. & Time (sec.) \\\hline
\small{SVM-RFE \cite{Guyon:2002}} & e& s& \small{$\mathcal{O}(T^2 n log_2n )$} & N/A\\
Ens.SVM-RFE~\cite{LeiYi10.1109} &e&s& $\mathcal{O}(K T^2  n log_2n )$ & N/A \\
SW SVM-RFE~\cite{LeiYi10.1109} &e& s&$\mathcal{O}(T^2 n log_2n )$ & N/A \\
Relief-F \cite{liu2008} &f&s& $\mathcal{O}(iTnC)$ & 656.9 \\
SW Relief-F \cite{LeiYi10.1109} &f&s& $\mathcal{O}(iTnC)$ &  N/A\\
FSV \cite{Bradley98featureselection,Grinblat:2010} & w& s& N/A & 1414.6 \\
Fisher-G \cite{Quanquanjournals} &f& s& $\sim\mathcal{O}(iCT)$ & 0.12 \\
MutInf \cite{Hutter:02feature} &f& s&$\sim\mathcal{O}( n^2T^2)$ & 8.61 \\
\textbf{Inf-FS}  &f&u& $\mathcal{O}(n^{2.37}(1+T) )$ & 4.05 \\
\end{tabular}}
\caption{List of the feature selection approaches considered in the experiments, specified according to their \emph{Type}, class (\emph{Cl.}), complexity (\emph{Compl.}), and \emph{Time} spent on a standard feature selection task (see Sec.~\ref{sec5:compAppr}). As for the complexity, $T$ is the number of samples, $n$ is the number of initial features, $K$ is a multiplicative constant, $i$ is the number of iterations in the case of iterative algorithms, and $C$ is the number of classes. The complexity of FSV cannot be specified since it is a wrapper (it depends on the chosen classifier).}
\label{table:ch5compmethods44}
\end{table}

The complexity of the Inf-FS approach is $\mathcal{O}(n^{2.37}+\frac{n^2}{2}T)$, the calculation of the matrix inversion for an $n\times n$ matrix requires $O(n^{2.37})$ \cite{matrixInversion2}, and the second term $O(\frac{n^2}{2}T)$ comes from the estimate of $a_{i,j}$ energies. This complexity allows the Inf-FS approach to obtain the timing that is the second best among the ones whose codes are publicly available.

\subsubsection{Exp. 1: Varying the cardinality of the selected features}\vspace{-0.05cm}

In the first  experiment, we consider the protocol of~\cite{LeiYi10.1109}, which starts with a pool of features characterizing the training data. These features are selected, generating different subsets of different cardinalities. The training data described by the selected features is then used to learn a linear SVM, subsequently employed to classify the test samples. The dataset used in~\cite{LeiYi10.1109} and considered here is the \emph{Colon}. The experiment serves to understand how well important features are ranked high by a feature selection algorithm. Table~\ref{table:ch5Lung181} presents the results in terms of AUC.

\begin{table}[ht]
\small
\centering
\resizebox{0.5\textwidth}{!}{%
\begin{tabular}{| l  |C{0.6cm}|C{0.6cm} |C{0.6cm}| C{0.6cm}| C{0.6cm}|}
\hline
\multicolumn{6}{|c|}{\textbf{Colon}} \\
\hline
& \multicolumn{5}{c|}{\# Features} \\
\hline
\textbf{Sel. Method} &10 & 50 &100 &150&200 \\\hline
SVM-RFE 			&76.4&77.5&	79.2&	79.4	&80.1  \\\hline
Ens. SVM-RFE	&80.3&	79.4&	78.6&	78.6&	79.4\\\hline
SW SVM-RFE	&79.5&	81.2&	78.4	&76.2&	76.2 \\\hline
ReliefF			&78.8&	80.1&	78.5&	77.5	&76.1\\\hline
SW ReliefF		&78.3&	79.6&	78.1&	76.4&	75.4 \\\hline
Fisher-G & 84.2   & 86.2  &  87.1 &   86.0  & 86.9  \\\hline
MutInf & 80.1  &  83.0  &  82.9 &   83.3  &  83.4 \\\hline
FSV &  81.3  &   83.2 & 84.0  &   83.9   &  84.7  \\\hline
\textbf{Inf-FS}	&\textbf{86.4}&\textbf{89}	&\textbf{89.4}	&\textbf{89.3}	&	\textbf{89} \\\hline
\end{tabular}}
\caption{Average accuracy results while varying the cardinality of the selected features.}
\label{table:ch5Lung181}
\end{table}

The Inf-FS outperforms all the competitors at all the features cardinalities, being very close to the absolute state of the art. On all the other datasets, Table~\ref{table:ch5OtherResults} lists the scores obtained by averaging the results of the different cardinalities of the features considered. Even in this case, the results show Inf-FS as overcoming the other competitors.

\begin{table}[ht]
\small
\centering
\resizebox{0.7\textwidth}{!}{%
\begin{tabular}{| l |C{0.9cm} |C{1.5cm} |C{1.2cm}| C{1.2cm}|C{1.2cm}|}
\hline
\multicolumn{6}{|c|}{\textbf{Varying the \# of the selected features - other datasets}} \\
\hline
\textbf{Dataset}  & FSV  &Fisher-G  &MutInf  &ReliefF & \textbf{Inf-FS} \\\hline
GINA& 84.2 & 87.1&77.7&87.7&  \textbf{89.3}\\\hline
USPS  &  91.2  &88.6  &92.1 & 92.0& \textbf{94.1}\\\hline
Lymphoma& 92.6 & 97.7&88.7 & 97.5& \textbf{97.9*} \\\hline
Leukemia & 98.2  & 99.7 &  91.9& 95.0 & \textbf{100*} \\\hline
Lung181 &  99.7 & 99.7 & 97.0 & 96.8 & \textbf{99.8*} \\\hline
DLBCL& 92.5 & 97.7  & 88.7 &97.5  & \textbf{98.0*} \\\hline
MADELON & 66.7 & 71.3 & 59.9&  66.6 & \textbf{74.6} \\\hline
GISETTE & 61.6 & 73.9  & 51.7 & 62.9 & \textbf{87.3} \\\hline
\end{tabular}}
\caption{Varying the cardinality of the selected features. AUC (\%) on different datasets of SVM classification, averaging the performance obtained with the first 10, 50, 100, 150, and 200 features ordered by our Inf-FS algorithm. Each asterisk indicates a new top score  (being an average of scores, the genuine top score for Lymphoma is 98\% - 100 features and for DLBCL is 98.3\% - 150 features).}
\label{table:ch5OtherResults}
\end{table}

\subsubsection{Exp. 2: CNN on object recognition datasets}\vspace{-0.05cm}

This section starts with a set of tests on the object recognition datasets, that is, Caltech 101-256 and PASCAL VOC 2007-2012. The Caltech benchmarks have been taken into account due to their high number of object classes. 

The second experiment considers as features the cues extracted with convolutional neural network (CNN) \cite{DBLPZeilerF13}. 
The CNN features have been pre-trained on ILSVRC (we adopt the MatConvNet distribution \cite{arXivVedaldi}), using the 4,096-dimension activations of the penultimate layer as image features, L2-normalized afterwards. We do not perform fine tuning, so the features are fixed, given a dataset. This will help future comparisons.

The classification results on the Caltech series have been produced by considering three random splits of training and testing data, averaging the results. For the PASCAL series, mean average precision (mAP) scores have been reported, while in the Caltech case, we show the average accuracies. Table~\ref{table:ch5cnnResults} presents the results, where the percentages of the selected features are enclosed in parentheses. As for the comparative approaches, we evaluate only those whose codes are publicly available.

\begin{table}[ht]
\small
\centering
\resizebox{0.5\textwidth}{!}{%
\begin{tabular}{| l  |c |c| c| c|}
\hline
\multicolumn{5}{|c|}{\textbf{Object recognition by CNN features}} \\
\hline
\centering{\textbf{Methods}}& \multicolumn{4}{c|}{\textbf{Datasets} } \\
\hline
 & \textbf{VOC'07} & \textbf{VOC'12}& \textbf{Cal.101} & \textbf{Cal.-256}\\\hline
Relief-F  & 80.4  &    82.7  & 90.8 & 79.8 \\
& (81\%) & (96\%) &  (81\%) &  (81\%) \\\hline
Fisher-G  & 80.7& 82.9  & 90.9& 79.9 \\
& (81\%) & (87\%) &  (81\%) &  (81\%) \\\hline
MutInf  &  80.6 & 82.8    &90.9 & 79.9 \\
& (88\%) & (92\%) &  (81\%) &  (81\%) \\\hline
FSV  & 80.8 &  81.6&89.7& 79.6\\
& (86\%) & (89\%) &  (81\%) &  (81\%) \\\hline
\textbf{Inf-FS}  & \textbf{ 83.5}& \textbf{84.0} & \textbf{91.8} & \textbf{81.5}\\
 & (88\%) & (89\%) &  (81\%) &  (81\%) \\\hline
\end{tabular}}
\caption{Feature selection on the object recognition datasets. The numbers in parentheses are the percentages of features kept by the approach after the cross-validation phase.}
\label{table:ch5cnnResults}
\end{table}
As visible, the combination of our Inf-FS method and a simple linear SVM classifier gives the state of the art in all the datasets (see Table~\ref{table:ch5datasets} for the current top scores). Notably, the top scores so far have been implemented by CNN features plus SVM, which can be considered the framework we adopt \emph{without} the feature selection. As for the percentage of the selected features, Inf-FS is somewhat in line with the other comparative approaches.

\subsubsection{Exp. 3: Varying the number of input samples}\vspace{-0.05cm}
The availability of training samples for the feature selection operation is an important aspect to consider: actually, in some cases, it is difficult to deal with consistent quantities of data, especially in biomedical scenarios. For this sake, we consider the PASCAL VOC 2007 dataset (with plenty of data), and we evaluate our approach (and the comparative ones of the previous section) while diminishing the cardinality of the training + validation dataset, uniformly removing images from all the 20 classes, going from 5K samples to 600 circa. In all these cases, we keep 1K features for the final classification. Other than calculating the accuracy measures, we investigate how stables are the partial ranked lists produced, that is, how often the same subsets of features are selected with the same ordering. For this reason, we employ the stability index based on Jensen-Shannon Divergence $D_{JS}$, proposed by~\cite{guzman2011feature}, with a [0,1] range, where 0 indicates completely random rankings and 1 means stable rankings. Interestingly, the index accounts for both the ability of having subsets of features a) with the same elements and b) ordered in the same way, where the differences at the top of the list are weighted more than those at the bottom. 
\begin{table}[!t]
\small
\centering
\resizebox{0.4\textwidth}{!}{%
\begin{tabular}{| l  |c | c| c| }
\hline
\multicolumn{4}{|c|}{\textbf{Stability analysis - PASCAL VOC 2007}} \\
\hline
 \textbf{Method} & \textbf{\#Images}& \textbf{mAP} & $\mathbf{D_{JS}}$ \\\hline
\textbf{Relief-F} & & &\\
    & 5,011 & 80.4\%&  1.0\\\hline
    & 2,505 & 80.3\% &  0.81 \\\hline
    & 1,252 & 78.2\%&  0.64\\\hline
    & 626 & 74.4\% &  0.44\\\hline
\hline
\textbf{Fisher-G} & & &\\
    & 5,011 & 80.7\% & 1.0\\\hline
    & 2,505 & 80.3\%&  0.95\\\hline
    & 1,252 & 78.2\%& 0.84\\\hline
    & 626 & 74.6\% &  0.69\\\hline
\hline
\textbf{MutInf} & &&\\
    & 5,011 & 80.6\%&  1.0\\\hline
    & 2,505 & 80.3\% & 0.88\\\hline
    & 1,252 & 78.2\% & 0.64\\\hline
    & 626 & 74.5\%& 0.34\\\hline
\hline
\textbf{FSV} & & &\\
    & 5,011 & 80.8\%& 1.0\\\hline
    & 2,505 & 80.1\% &0.90\\\hline
    & 1,252 & 78.1\% &0.87\\\hline
    & 626 & 74.4\%&0.86\\\hline
\hline
\textbf{Inf-FS} & & &\\
    & 5,011 & 83.5\% &  1.0\\\hline
    & 2,505 & 81.9\% & 0.99\\\hline
    & 1,252 & 79.8\% &0.97\\\hline
    & 626 & 76.5\% & 0.94\\\hline
\hline
\end{tabular}}
\caption{Stability analysis: mAP scores by reducing the number of training images, and the $D_{JS}$ index taking into account the first 1K ranked features.}
\label{table:ch5stability}
\end{table}
Table~\ref{table:ch5stability} presents interesting results since Inf-FS is the more stable even with 626 images, but at the same time, especially going from 5,011 to 2,505 images, it lowers more the final accuracy. This is probably  because the pruned-away images could be those that the classifier uses to discriminate among the classes.

\subsubsection{Exp. 4: Evaluating the mixing $\alpha$}\label{sec5:mixAlpha}
\begin{figure}[t!]
        \centering
                \includegraphics[width=0.75\linewidth]{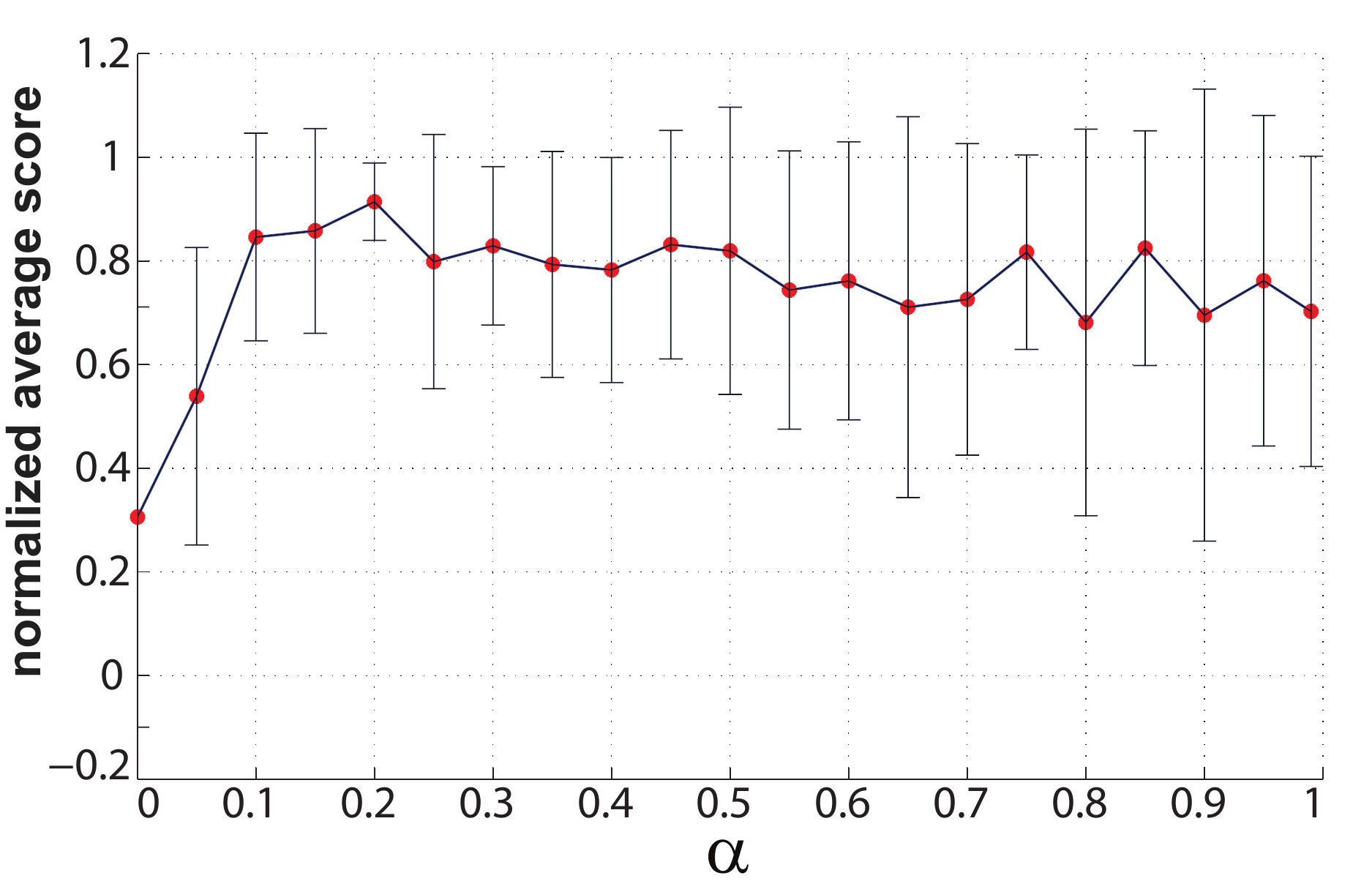}
        \caption{Evaluating the mixing $\alpha$: normalized average score and related error bar}
\label{fig5:alfa}
\end{figure}

\vspace{-0.05cm}
The coefficient $\alpha$ of Eq.~\ref{eq5:partzero} drives the algorithm in weighting the maximum variance of the features and their correlation. As previously stated, in all the experiments, we select $\alpha$ by cross-validation on the training set. In this section, we show how different values of $\alpha$ are generally effective for the classification. For this purpose, we examine all the datasets analyzed so far, fixing the percentage of the selected features to 80\% of the starting set, and we keep the $C$ value of the SVM that gave the best performance. We then vary the value of  $\alpha$ in the interval [0,1] at a 0.1 step, keeping the classification accuracies/mAP obtained therein; given a dataset, all the classification scores with the different $\alpha$ values are normalized by the maximum performance. These normalized scores of all the experiments are then averaged, so having a \emph{normalized averaged score} close to 1 means that with that $\alpha$ value, all the methods performed at their best. Fig.~\ref{fig5:alfa}  illustrates the resulting curve, showing interesting and desirable characteristics.
At $\alpha=0$ (only correlation is accounted), the performance is low, rapidly increasing until $\alpha=0.2$ where a peak is reached. After that, the curve is decreasing, even if close to 1, there are some oscillations. At $\alpha=1$ (only maximum variance is considered), the approach works at 70\% of its capabilities, on average. Analyzing the error bars (showing the $2.5$ standard deviation intervals) is very informative, as it tells that the value of $\alpha=0.2$ represents the best mix of the two feature characteristics.

\subsubsection{Exp. 5: Augmenting the features}\vspace{-0.05cm}

For our final study, we extend the kinds of features used for the object recognition datasets, including a 1,024-dimension BoW. The idea is to see if augmenting the descriptions of the images will improve the classification performance; at the same time, analyzing the kept features can give insights into the relevance of the features that come into play. Specifically, four word dictionaries of 256 entries have been calculated on a subset of 10\% of the datasets VOC07/12 respectively, extracting dense PHOW features (SIFT have been extracted on 7-pixel squared cells with a 5-pixel step). Subsequently, 4-cell spatial histograms have been computed, ending with a 1,024-dimension representation for each image. Each histogram bin is thus a feature. BoWs have been concatenated to CNN features, resulting in a 5,120 feature set. As for the protocol, we have fixed the number of features to be selected at 85\%, representing a valid compromise among the percentages chosen by the different approaches on the sole CNN (see Table~\ref{table:ch5cnnResults}). Table~\ref{table:ch5Augmenting} shows the results. 

\begin{table}[!t]
\small
\centering
\resizebox{0.4\textwidth}{!}{%
\begin{tabular}{| l  |c |c|}
\hline
\multicolumn{3}{|c|}{\textbf{CNN + BoW }} \\
\hline
\multicolumn{3}{|c|}{\textbf{Datasets} } \\
\hline
 & \textbf{VOC'07} & \textbf{VOC'12}\\\hline
\textbf{Methods}  & (mAP) & (mAP)\\\hline
Relief-F    & 81.6 &   83.5   \\
 & ([76\%,24\%]) & ([75\%,25\%] )  \\\hline
Fisher-G    &81.9& 83.9  \\
& ([93\%,7\%]) &([95\%,5\%] ) \\\hline
MutInf    & 80.7&   83.8 \\
& ([97\%,3\%]) &([92\%,8\%] )  \\\hline
FSV     & 81.1 &   83.9  \\
 & ([98\%,2\%]) &([98\%,2\%] )   \\\hline
\textbf{Inf-FS}  & \textbf{ 83.6*}& \textbf{84.1*}\\
 & ([91\%,9\%] ) & ([93\%,7\%] ) \\\hline
\end{tabular}}
\caption{Feature selection on augmented feature descriptions. See the text.}
\label{table:ch5Augmenting}
\end{table}
It is evident that adding a further kind of cue is generally useful. With inf-FS the increase is minimal, probably because we are close to an intrinsic upper bound, given the features and the classifier. The numbers enclosed in square brackets (Table~\ref{table:ch5Augmenting}) show the percentages of the kept features, with CNN in the first position and BoW in the second position. In all the cases (except the relief-F method), CNN tends to be preferred to BoW, witnessing its expressivity. Moreover, the ordering of the features (not shown here) indicates that in almost all the cases, most of the CNN features (95\% circa) are ranked ahead of the BoW ones.

\subsection{Summary and Discussions}
 
In this first part we presented the idea of considering feature selection as a regularization problem, where features are nodes in a graph, and a selection is a path through them. The application of the Inf-FS approach to all the 13 datasets against 8 competitors, at most, (employing simple linear SVM) contributes to top performances, notably setting the absolute state of the art on 8 benchmarks. The Inf-FS is also robust with a few sets of training data, performs effectively in ranking high the most relevant features, and has a very competitive complexity. 

\begin{figure}[t!]
        \centering
                \includegraphics[width=0.8\linewidth]{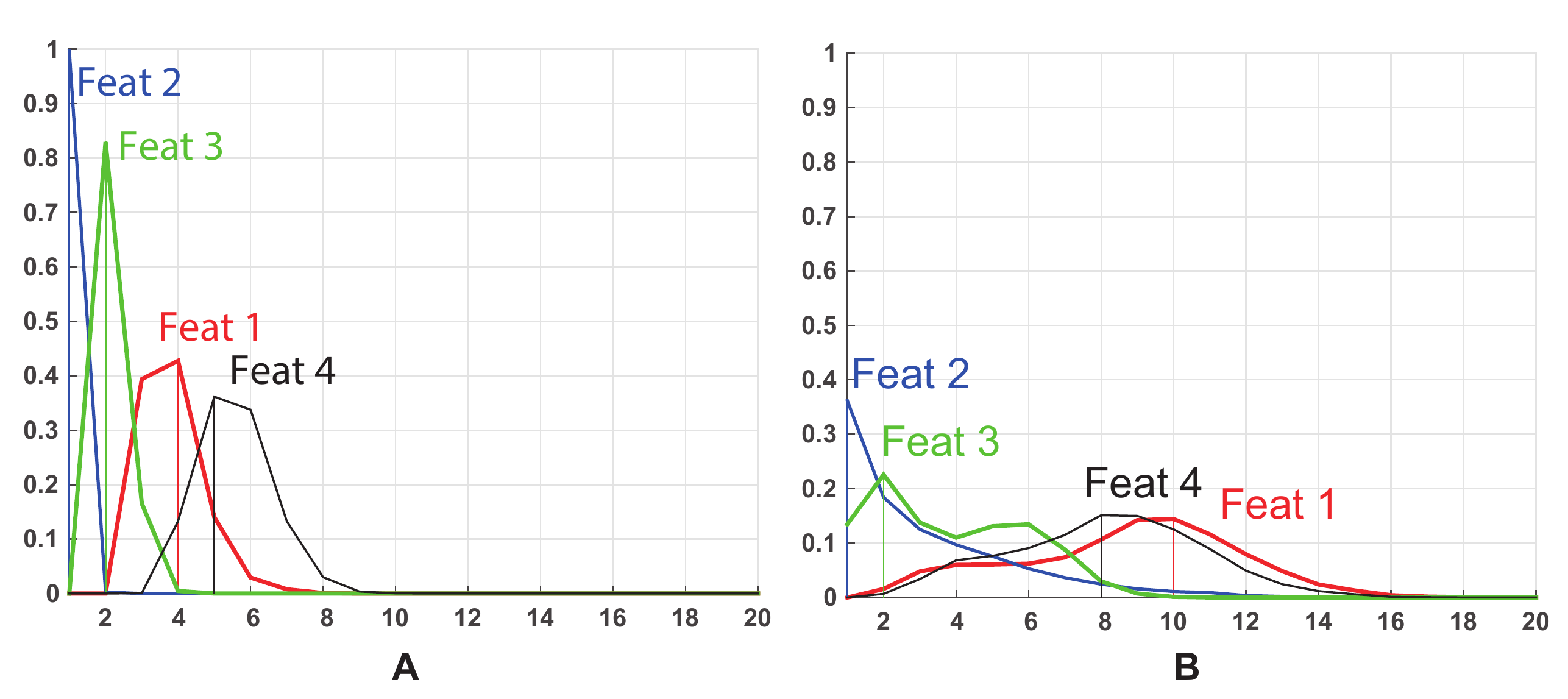}
        \caption{Ranking tendency of Inf-FS in the linear and periodic cases. The curve represents the density profile of a given feature (only the first four original features are reported) of being in a particular ranking position.}
\label{fig5:synth}
\end{figure}

We use synthetic data to gain insights into when the Inf-FS can correctly rank the representative features, resembling the analysis done in \cite{Zhu2015438}. The Inf-FS allows dealing with the set of initial features as if they constitute a weighted graph, with the weights modeling a similarity relation between features. Therefore, the design of the similarity relation suggests the scenarios where the approach has to be preferred. In our case, we use Spearman's rank correlation plus a variance score (see Eq.~\ref{eq5:partzero}). Spearman individuates when features are connected with linear or nonlinear monotonic correlations, and in such cases, the approach is expected to work well. In case the features are connected in a more complicated manner (that is, via periodic relations), the approach does not work nicely. To show this, we first extract from the IRIS dataset 150 samples and 4 (independent) features.
In a first case, 16 features are artificially generated as the linear convex combination of the 4 original ones. In the second case,  the 16 features are generated by using a periodic function, that is, the linear convex combination of the \emph{sin} of the features. On these data, Inf-FS is expected to rank the four original features first, followed by the other ones. For the sake of generalization, we repeat the experiment 20 times, each one with diverse mixing coefficients. Results are shown in Fig.\ref{fig5:synth}.
\red{As expected, the Inf-FS works definitely better in the first case, as shown in Fig.\ref{fig5:synth}\textbf{A}, keeping the first 4 features in the top position, while in the second (see Fig.\ref{fig5:synth}\textbf{B}), it starts to produce very different orderings.}

As a result, nonlinearities between the features can be encoded by theoretical information measures, instead of simple correlations. 

At the end of this chapter an interesting and deep discussion about the Inf-FS is provided. We report its source of inspiration from \textit{Quantum Mechanics} and \textit{Path Integrals} along with its interpretation from two different perspectives.

%
%
\section{Feature Selection via Eigenvector Centrality}\label{ch5:EigVec21}

In this section, we propose a novel graph-based feature selection algorithm that ranks features according to a graph centrality measure (Eigenvector centrality~\cite{Bonacich}). The main idea behind the method is to map the problem to an affinity graph, and to model pairwise relationships among feature distributions by weighting the edges connecting them. The novelty of the proposed method in terms of the state of the art is that it assigns a score of ``importance" to each feature by taking into account all the other features mapped as nodes on the graph, bypassing the combinatorial problem in a methodologically sound fashion. Indeed, eigenvector centrality differs from other measurements (e.g., degree centrality) since a node - feature - receiving many links does not necessarily have a high eigenvector centrality. The reason is that not all nodes are equivalent, some are more relevant than others, and, reasonably, endorsements from important nodes count more (see section~\ref{sec5:varphi}). Noteworthy, another important contribution of this work is the scalability of the method. Indeed, centrality measurements can be implemented using the Map Reduce paradigm ~\cite{Kang_centralitiesin,Lerman:2010,MapReduceBetweenness}, which makes the algorithm prone to a possible distributed version~\cite{Rawat2015}.

Our approach (EC-FS) is extensively tested on 7 benchmarks of cancer classification and prediction on genetic data (\emph{Colon}~\cite{alon}, \emph{Prostate}~\cite{Golub99}, \emph{Leukemia}~\cite{Golub99}, \emph{Lymphoma}~\cite{Golub99}), handwritten recognition (GINA~\cite{GINA}), generic feature selection (MADELON~\cite{guyon2004result}), and object recognition (PASCAL VOC 2007~\cite{pascal-voc-2007}). We compare the proposed method on these datasets, against seven comparative approaches, under different conditions (number of features selected and number of training samples considered), overcoming all of them in terms of ranking stability and classification accuracy.

Finally, we provide an open and portable library of feature selection algorithms, integrating the methods with uniform input and output formats to facilitate large scale performance evaluation. The \textit{Feature Selection Library} (FSLib 4.2 \footnote{The FSLib is publicly available on File Exchange - MATLAB Central at:\\ \url{https://it.mathworks.com/matlabcentral/fileexchange/56937-feature-selection-library}}) and interfaces are fully documented. The library integrates directly with MATLAB, a popular language for machine learning and pattern recognition research.

\subsection{Building the Graph}\label{sec5:Graph}

Given a set of features $X = \{  x^{(1)}, ..., x^{(n)} \}$ we build an undirected graph $G = (V, E)$; where $V$ is the set of vertices corresponding, one by one, to each variable $x$. $E$ codifies (weighted) edges among features. Let the adjacency matrix $A$ associated with $G$ define the nature of the weighted edges: each element $a_{ij}$ of $A$, $1\leq i,j \leq n$, represents a pairwise potential term. Potentials can be represented as a binary function $\varphi(x^{(i)},x^{(j)})$ of the nodes $x^{(k)}$ such as:

\begin{equation}\label{eq5:partzero2}
  a_{ij} = 	\varphi(x^{(i)},x^{(j)}).
\end{equation}

The graph can be weighted according to different heuristics, therefore the function $\varphi$ can be handcrafted or automatically learned from data.

\subsubsection{$\varphi$-Design}\label{sec5:varphi}

The design of the $\varphi$ function is a crucial operation. In this work, we weight the graph according to good reasonable criteria, related to class separation, so as to address the classification problem. In other words, we want to rank features according to how well they discriminate between two classes. Hence, we draw upon best-practice in FS and propose an ensemble of two different measures capturing both relevance (supervised) and redundancy (unsupervised) proposing a kernelized-based adjacency matrix. Before continuing with the discussion, note that each feature distribution $x^{(i)}$ is normalized so as to sum to $1$. 

Firstly, we apply the Fisher criterion:
 \[
	 f_i = \frac{\left | \mu_{i,1} - \mu_{i,2} \right |^2}{ \sigma_{i,1}^2+\sigma_{i,2}^2},
\]
where $\mu_{i,\textit{C}}$ and $\sigma_{i,\textit{C}}$ are the mean and standard deviation, respectively, assumed by the $i$-th feature when considering the samples of the $\textit{C}$-th class. The higher $f_i$, the more discriminative the $i$-th feature.

Because we are given class labels, it is natural that we want to keep only the features that are related to or lead to these classes. Therefore, we use mutual information to obtain a good feature ranking that score high features highly predictive of the class.  
\[
	m_{i} = \sum_{y \in Y } \sum _{z \in x^{(i)}} p(z, y)log \Big( \frac{p(z,y)}{p(z)p(y)} \Big),
\]
where $Y$ is the set of class labels, and $p(\cdot,\cdot)$ the joint probability distribution. A kernel $k$ is then obtained by the matrix product 
 \[
	 k =  (f \cdot m^\top),
\]
where $f$ and $m$ are $n \times 1$ column vectors normalized in the range $0$ to $1$, and $k$ results in a $n \times n $ matrix. To boost the performance, we introduce a second feature-evaluation metric based on standard deviation~\cite{Guyon:2002} -- capturing the amount of variation or dispersion of features from average -- as follows:
\[
	\Sigma(i,j) = max\left(\sigma^{(i)},\sigma^{(j)}\right) ,
\]
where $\sigma$ being the standard deviation over the samples of $x$, and $\Sigma$ turns out to be a $n \times n $ matrix with values $\in$ [$0$,$1$]. Finally, the adjacency matrix $A$ of the graph $G$ is given by
\begin{equation}\label{eq5:alpha2}
	A  = \alpha k + (1-\alpha) \Sigma,
\end{equation}
where $\alpha$ is a loading coefficient $\in [0,1]$. The generic entry $a_{ij}$ accounts for how much discriminative are the feature $i$ and $j$ when they are jointly considered; at the same time, $a_{ij}$ can be considered as a weight of the edge connecting the nodes $i$ and $j$ of a graph, where the $i$-th node models the $i$-th feature distribution (we report EC-FS method in Procedure \ref{algorithm3}). 

\begin{algorithm}[!t]
\caption{Eigenvector Centrality Feature Selection (EC-FS)}
\label{algorithm3}
\begin{algorithmic}
\REQUIRE{$X = \{ x^{(1)}, ..., x^{(n)} \}$ , $Y = \{ y^{(1)}, ..., y^{(n)} \}$}\\
\ENSURE{$v_{0}$ ranking scores for each feature }\\
\textbf{- Building the graph} \\
\STATE	$C_1$ positive class, $C_2$ negative class
\FOR{$i = 1 : n$}
\STATE	Compute $\mu_{i,1}$, $\mu_{i,2} $, $\sigma_{i,1} $, and $\sigma_{i,2} $
\STATE	Fisher score: $f(i) = \frac{(\mu_{i,1} - \mu_{i,2})^2}{ \sigma_{i,1}^2 + \sigma_{i,2}^2} $
\STATE	Mutual Information: $m(i) = \sum_{y \in Y } \sum _{z \in x^{(i)}} p(z, y)log \Big( \frac{p(z,y)}{p(z)p(y)} \Big)$
\ENDFOR
\FOR{$i = 1 : n$}
\FOR{$j = 1 : n$}
\STATE  $k(i,j) =  f(i)m(j)$,
\STATE $\Sigma(i,j) = max\left(\sigma^{(i)},\sigma^{(j)}\right)$ ,
\STATE $A(i,j) = \alpha k(i,j) + (1-\alpha) \Sigma(i,j)$
\ENDFOR
\ENDFOR
\STATE \textbf{- Ranking}
\STATE Compute eigenvalues $ \left\lbrace \Lambda\right\rbrace $ and eigenvectors $ \left\lbrace V \right\rbrace $ of $A$\\
\STATE $\lambda_{0} = \underset{\lambda \in \Lambda }{max} (abs(\lambda))$
\RETURN $v_{0}$ the eigenvector associated to $\lambda_{0}$
\end{algorithmic}
\end{algorithm}

\subsection{Experiments and Results}\label{sec5:exp2}

\subsubsection{Datasets and Comparative Approaches}

We consider the problems of dealing with unbalanced classes (\emph{unbalanced}), or classes that severely overlap (\emph{overlap}), or few training samples and many features (\emph{few train} in Table~\ref{table:ch5datasets2}), or whose samples are noisy (\emph{noise}) due to: a) complex scenes where the object to be classified is located (as in the VOC series) or b) many outliers (as in the genetic databases, where samples are often \emph{contaminated}, that is, artifacts are injected into the data during the creation of the samples). Finally, we consider the \emph{shift} problem, where the samples used for the test are not congruent (coming from the same experimental conditions) with the training data.
\begin{table*}[t!]
\small
\centering
\resizebox{1\textwidth}{!}{%
\begin{tabular}{l c c c c c c c c }
\hline \hline
Name &   \# samples & \# classes & \# feat.  & \emph{few train}& \emph{unbal. (+/-)} & \emph{overlap} & \emph{noise} & \emph{shift} \\\hline
GINA~\cite{GINA} & 3153 &2& 970   &   &  & X & & \\ 
MADELON~\cite{NIPS2003} & 4.4K  &2& 500   &   &  & X &  & \\
\hline
\emph{Colon}~\cite{alon} & 62 &2& 2K   &  X & (40/22) & & X  &  \\
\emph{Lymphoma}~\cite{Golub99} & 45 &2& 4026   &  X & (23/22)& & & \\
\emph{Prostate}~\cite{citeulike:1624492} & 102& 2& 6034   &  X & (50/52)  &  & &\\
\emph{Leukemia}~\cite{Golub99} & 72 &2& 7129   &  X & (47/25) & & X & X \\
\hline \vspace{0.02cm}
VOC 2007~\cite{pascal-voc-2007} & ~10K &20& n.s.   & & X & & X &  \\
\hline
\end{tabular}}
\caption{This table reports several attributes of the datasets used. The abbreviation \emph{n.s.} stands for \emph{not specified} (for example, in the object recognition datasets, the features are not given in advance).}
\label{table:ch5datasets2}
\vspace{-5.5mm}
\end{table*}

Table~\ref{table:ch5compmethods2} lists the methods in comparison, whose details can be found in Chapter ~\ref{ch2:classification}. Here we just note their \emph{type}, that is, \emph{f} = filters, \emph{w} = wrappers, \emph{e} = embedded methods, and their \emph{class}, that is, \emph{s} = supervised or \emph{u} = unsupervised (using or not using the labels associated with the training samples in the ranking operation).  Additionally, we report their computational complexity (if it is documented in the literature). The computational complexity of the EC-FS approach is $O(Tn +n^2)$.
\begin{table*}[t!]
\small
\centering
\begin{tabular}{p{2.9cm} C{0.7cm} C{0.7cm} C{2.5cm}}
\hline \hline
Acronym &   \small{Type} & \small{Cl.} & Compl.  \\\hline
Fisher-S~\cite{Bishop2006}   &f&s& $\mathcal{O}(Tn)$   \\
FSV~\cite{Bradley98featureselection,Grinblat:2010} & e& s& N/A \\
Inf-FS~\cite{Roffo_2015_ICCV}  &f&u& $\mathcal{O}(n^{2.37}(1+T) )$ \\
MI~\cite{Hutter:02feature} &f& s&$\sim\mathcal{O}( n^2T^2)$ \\
LS~\cite{HCN05a} &f&u& $N/A$   \\
Relief-F~\cite{liu2008} &f&s& $\mathcal{O}(iTnC)$  \\
RFE \cite{Guyon:2002} & w/e& s& \small{$\mathcal{O}(T^2 n log_2n )$}\\
\textbf{EC-FS}  &f&s& $\mathcal{O}(Tn +n^2)$  \\
\end{tabular}
\caption{List of the FS approaches considered in the experiments, specified according to their \emph{Type}, class (\emph{Cl.}), and complexity (\emph{Compl.}). As for the complexity, $T$ is the number of samples, $n$ is the number of initial features, $K$ is a multiplicative constant, $i$ is the number of iterations in the case of iterative algorithms, and $C$ is the number of classes. N/A indicates that the computational complexity is not specified in the reference paper.}
\label{table:ch5compmethods2}
\vspace{-1.5mm}
\end{table*}
The term $Tn$ is due to the computation of the mean values among the $T$ samples of every feature $(n)$. The $n^2$ concerns the construction of the matrix $A$. As for the computation of the leading eigenvector, it costs $O(m^2n)$, where $m$ is a number much smaller than $n$ that is selected within the algorithm~\cite{lehoucq1998arpack}. In the case that the algorithm can not be executed on a single computer, we refer the reader to~\cite{Kang_centralitiesin,Lerman:2010,Rawat2015,MapReduceBetweenness} for distributed algorithms.

\subsection{Exp. 1: Deep Representation (CNN) with pre-training}

This section proposes a set of tests on the PASCAL VOC-2007~\cite{pascal-voc-2007} dataset. 
\begin{table}[t!]
\small
\centering
\resizebox{1\textwidth}{!}{%
\begin{tabular}{|C{0.44cm}|C{0.62cm} |C{0.62cm}|C{0.62cm} |C{0.62cm}| C{0.62cm}| C{0.62cm}| C{0.62cm}| C{0.52cm}| C{0.02cm}| C{0.62cm}| C{0.62cm}| C{0.62cm}| C{0.52cm} |C{0.62cm}| C{0.62cm}| C{0.62cm}| C{0.62cm} |  }
\hline
\multicolumn{18}{|c|}{\textbf{PASCAL VOC 2007
}} \\
\hline
 & \multicolumn{8}{c|}{ \textbf{First 128/4096 Features Selected}} & &\multicolumn{8}{c|}{ \textbf{First 256/4096 Features Selected}} \\
\hline
& \tiny{Fisher-S} &\tiny{FSV} &\tiny{Inf-FS} & \tiny{LS} & \scriptsize{MI} & \tiny{ReliefF} &\tiny{RFE} & \scriptsize{\textbf{EC-FS}}  & &\tiny{Fisher-S} &\tiny{FSV} & \tiny{Inf-FS} & \tiny{LS} & \tiny{MI} & \tiny{ReliefF} &\tiny{RFE} & \scriptsize{\textbf{EC-FS}}  \\
 \hline
\centering
	$ \vcenter{\includegraphics[scale=0.1]{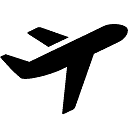}} $ & \scriptsize{52.43} & \scriptsize{	87.90} & \scriptsize{	88.96} & \scriptsize{	\textbf{89.37}} & \scriptsize{	12.84} & \scriptsize{	57.20} & \scriptsize{	86.42} & \scriptsize{	88.09} & 
 & 
\scriptsize{82.65} & \scriptsize{	90.22} & \scriptsize{	\textbf{91.16}} & \scriptsize{	90.94} & \scriptsize{	73.51} & \scriptsize{	81.67} & \scriptsize{	88.17} & \scriptsize{	90.79	}
\\
\hline 
\centering
	$ \vcenter{\includegraphics[scale=0.1]{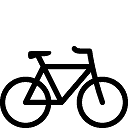}} $ & \scriptsize{ 13.49} & \scriptsize{	80.74} & \scriptsize{	80.43} & \scriptsize{	80.56} & \scriptsize{	13.49} & \scriptsize{	49.10} & \scriptsize{	\textbf{82.14}} & \scriptsize{	80.94} &  & 
\scriptsize{83.21} & \scriptsize{	80.07} & \scriptsize{	83.36} & \scriptsize{	84.21} & \scriptsize{	75.04} & \scriptsize{	71.27} & \scriptsize{	83.30} & \scriptsize{	\textbf{84.72}}
\\
\hline 
\centering
	$ \vcenter{\includegraphics[scale=0.1]{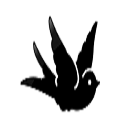}} $ &  \scriptsize{ 85.46} & \scriptsize{	86.77} & \scriptsize{	87.04} & \scriptsize{	86.96} & \scriptsize{	80.91} & \scriptsize{	75.42} & \scriptsize{	83.16} & \scriptsize{	\textbf{88.74}} &  & 
\scriptsize{ 89.14} & \scriptsize{	86.15} & \scriptsize{	88.88} & \scriptsize{	\textbf{89.31}} & \scriptsize{	85.48} & \scriptsize{	83.54} & \scriptsize{	86.12} & \scriptsize{	89.15	}
\\
\hline 
\centering
	$ \vcenter{\includegraphics[scale=0.1]{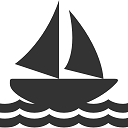}} $ & \scriptsize{ 79.04} & \scriptsize{	83.58} & \scriptsize{	85.31} & \scriptsize{	83.51} & \scriptsize{	61.50} & \scriptsize{	63.75} & \scriptsize{	78.55} & \scriptsize{	\textbf{86.90}} &   & 
\scriptsize{87.05} & \scriptsize{	80.68} & \scriptsize{	87.24} & \scriptsize{	\textbf{87.84}} & \scriptsize{	75.25} & \scriptsize{	73.30} & \scriptsize{	86.13} & \scriptsize{	87.42}
\\
\hline 
\centering
	$ \vcenter{\includegraphics[scale=0.1]{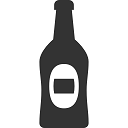}} $ &\scriptsize{ 46.61} & \scriptsize{	39.80} & \scriptsize{	44.83} & \scriptsize{	\textbf{49.36}} & \scriptsize{	35.39} & \scriptsize{	18.33} & \scriptsize{	46.24} & \scriptsize{	47.37} & & 
\scriptsize{52.54} & \scriptsize{	49.00} & \scriptsize{	52.65} & \scriptsize{	49.44} & \scriptsize{	48.94} & \scriptsize{	35.67} & \scriptsize{	47.28} & \scriptsize{	\textbf{53.20}	}
\\
\hline 
\centering
	$ \vcenter{\includegraphics[scale=0.1]{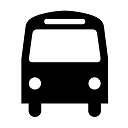}} $ & \scriptsize{ 12.29} & \scriptsize{	72.89} & \scriptsize{	76.69} & \scriptsize{	\textbf{76.98}} & \scriptsize{	12.29} & \scriptsize{	31.54} & \scriptsize{	74.68} & \scriptsize{	76.27} &  & 
\scriptsize{77.32} & \scriptsize{	78.69} & \scriptsize{	79.23} & \scriptsize{	79.97} & \scriptsize{	59.23} & \scriptsize{	63.83} & \scriptsize{	79.38} & \scriptsize{	\textbf{80.57	}}
\\
\hline 
\centering
	$ \vcenter{\includegraphics[scale=0.1]{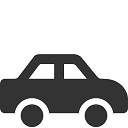}} $ & \scriptsize{ 82.09} & \scriptsize{	78.61} & \scriptsize{	85.78} & \scriptsize{	85.82} & \scriptsize{	63.58} & \scriptsize{	74.95} & \scriptsize{	83.94} & \scriptsize{	\textbf{85.92}} &  & 
\scriptsize{ 85.86} & \scriptsize{	84.01} & \scriptsize{	86.74} & \scriptsize{	\textbf{87.06}} & \scriptsize{	85.27} & \scriptsize{	82.76} & \scriptsize{	85.61} & \scriptsize{	86.56	}
\\
\hline 
\centering
	$ \vcenter{\includegraphics[scale=0.1]{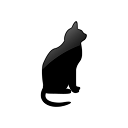}} $ & \scriptsize{ 75.29} & \scriptsize{	82.25} & \scriptsize{	\textbf{83.34}} & \scriptsize{	81.81} & \scriptsize{	40.96} & \scriptsize{	66.95} & \scriptsize{	81.02} & \scriptsize{	83.29} &  & 
\scriptsize{83.46} & \scriptsize{	83.49} & \scriptsize{	\textbf{85.61}} & \scriptsize{	84.98} & \scriptsize{	79.16} & \scriptsize{	76.78} & \scriptsize{	84.50} & \scriptsize{	85.57	}
\\
\hline 
\centering
	$ \vcenter{\includegraphics[scale=0.1]{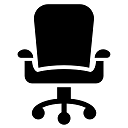}} $  & \scriptsize{ 54.81} & \scriptsize{	52.37} & \scriptsize{	58.62} & \scriptsize{	60.07} & \scriptsize{	16.95} & \scriptsize{	29.07} & \scriptsize{	59.84} & \scriptsize{	\textbf{60.57}} & & 
\scriptsize{63.14} & \scriptsize{	62.54} & \scriptsize{	63.93} & \scriptsize{	64.23} & \scriptsize{	63.20} & \scriptsize{	48.19} & \scriptsize{	62.16} & \scriptsize{	\textbf{64.53}	}
\\
\hline 
\centering
	$ \vcenter{\includegraphics[scale=0.1]{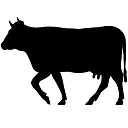}} $ & \scriptsize{ 47.98} & \scriptsize{	61.68} & \scriptsize{	59.23} & \scriptsize{	\textbf{65.50}} & \scriptsize{	11.42} & \scriptsize{	11.42} & \scriptsize{	62.96} & \scriptsize{	60.55} &  & 
\scriptsize{66.51} & \scriptsize{	70.18} & \scriptsize{	67.96} & \scriptsize{	\textbf{71.54}} & \scriptsize{	22.96} & \scriptsize{	51.28} & \scriptsize{	64.20} & \scriptsize{	69.71	}
\\
\hline 
\centering
	$ \vcenter{\includegraphics[scale=0.1]{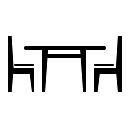}} $ & \scriptsize{49.68} & \scriptsize{	63.50} & \scriptsize{	67.69} & \scriptsize{	63.86} & \scriptsize{	12.62} & \scriptsize{	12.62} & \scriptsize{	67.05} & \scriptsize{	\textbf{67.70}} &   & 
\scriptsize{68.42} & \scriptsize{	69.27} & \scriptsize{	\textbf{71.78}} & \scriptsize{	71.01} & \scriptsize{	65.77} & \scriptsize{	52.24} & \scriptsize{	71.43} & \scriptsize{	70.95	}
\\
\hline 
\centering
	$ \vcenter{\includegraphics[scale=0.1]{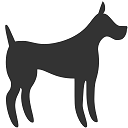}} $ & \scriptsize{ 81.06} & \scriptsize{	80.57} & \scriptsize{	83.16} & \scriptsize{	\textbf{83.21}} & \scriptsize{	70.70} & \scriptsize{	68.12} & \scriptsize{	80.07} & \scriptsize{	83.00} &  & 
\scriptsize{84.24} & \scriptsize{	84.15} & \scriptsize{	85.08} & \scriptsize{	\textbf{85.20}} & \scriptsize{	82.03} & \scriptsize{	74.85} & \scriptsize{	83.52} & \scriptsize{	\textbf{85.20}	}
\\
\hline 
\centering
	$ \vcenter{\includegraphics[scale=0.1]{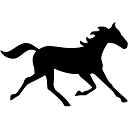}} $& \scriptsize{ 74.91} & \scriptsize{	\textbf{83.33}} & \scriptsize{	81.23} & \scriptsize{	81.75} & \scriptsize{	14.13} & \scriptsize{	63.06} & \scriptsize{	81.55} & \scriptsize{	82.79} & & 
\scriptsize{\textbf{85.68}} & \scriptsize{	83.13} & \scriptsize{	85.28} & \scriptsize{	85.41} & \scriptsize{	71.36} & \scriptsize{	75.53} & \scriptsize{	83.47} & \scriptsize{	85.28	}
\\
\hline 
\centering
	$ \vcenter{\includegraphics[scale=0.1]{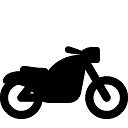}} $ & \scriptsize{ 13.18} & \scriptsize{	71.42} & \scriptsize{	81.32} & \scriptsize{	80.24} & \scriptsize{	13.18} & \scriptsize{	34.43} & \scriptsize{	76.57} & \scriptsize{	\textbf{82.20}} &  & 
\textbf{\scriptsize{84.29}} & \scriptsize{	81.16} & \scriptsize{	84.20} & \scriptsize{	83.81} & \scriptsize{	81.01} & \scriptsize{	70.68} & \scriptsize{	82.97} & \scriptsize{	84.12	}
\\
\hline 
\centering
	$ \vcenter{\includegraphics[scale=0.1]{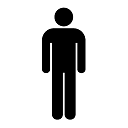}} $ & \scriptsize{ \textbf{91.33}} & \scriptsize{	90.03} & \scriptsize{	89.10} & \scriptsize{	89.33} & \scriptsize{	91.08} & \scriptsize{	88.85} & \scriptsize{	89.03} & \scriptsize{	91.27} &  & 
\scriptsize{91.95} & \scriptsize{	89.99} & \scriptsize{	90.65} & \scriptsize{	90.64} & \scriptsize{	91.77} & \scriptsize{	90.38} & \scriptsize{	90.64} & \scriptsize{	\textbf{91.99}	}
\\
\hline 
\centering
	$ \vcenter{\includegraphics[scale=0.1]{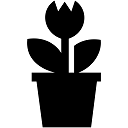}} $ &\scriptsize{ 47.89} & \scriptsize{	39.40} & \scriptsize{	45.38} & \scriptsize{	47.94} & \scriptsize{	13.23} & \scriptsize{	13.30} & \scriptsize{	48.61} & \scriptsize{	\textbf{49.05}} &  & 
\scriptsize{54.94} & \scriptsize{	47.95} & \scriptsize{	53.86} & \scriptsize{	54.31} & \scriptsize{	48.98} & \scriptsize{	34.74} & \scriptsize{	50.18} & \scriptsize{	\textbf{55.88}	}
\\
\hline 
\centering
	$ \vcenter{\includegraphics[scale=0.1]{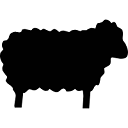}} $ & \scriptsize{ 10.87} & \scriptsize{	68.82} & \scriptsize{	73.35} & \scriptsize{	\textbf{74.05}} & \scriptsize{	10.87} & \scriptsize{	10.87} & \scriptsize{	66.86} & \scriptsize{	73.80} &  & 
\scriptsize{73.43} & \scriptsize{	75.84} & \scriptsize{	79.01} & \scriptsize{	\textbf{81.57}} & \scriptsize{	10.87} & \scriptsize{	11.73} & \scriptsize{	75.47} & \scriptsize{	78.85	}
\\
\hline 
\centering
	$ \vcenter{\includegraphics[scale=0.1]{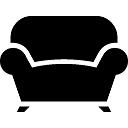}} $ & \scriptsize{ 45.87} & \scriptsize{	56.08} & \scriptsize{	58.94} & \scriptsize{	58.92} & \scriptsize{	13.30} & \scriptsize{	13.31} & \scriptsize{	\textbf{62.06}} & \scriptsize{	61.32} & &
\scriptsize{66.46} & \scriptsize{	59.77} & \scriptsize{	63.07} & \scriptsize{	63.92} & \scriptsize{	58.78} & \scriptsize{	44.74} & \scriptsize{	\textbf{66.68}} & \scriptsize{	64.86	}
\\
\hline 
\centering
	$ \vcenter{\includegraphics[scale=0.1]{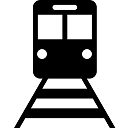}} $ & \scriptsize{ 63.51} & \scriptsize{	88.52} & \scriptsize{	91.42} & \scriptsize{	\textbf{91.48}} & \scriptsize{	58.62} & \scriptsize{	73.32} & \scriptsize{	88.46} & \scriptsize{	91.30} & & 
\scriptsize{84.05} & \scriptsize{	90.61} & \scriptsize{	\textbf{93.21}} & \scriptsize{	93.16} & \scriptsize{	81.33} & \scriptsize{	82.93} & \scriptsize{	90.24} & \scriptsize{	92.31	}
\\
\hline 
\centering
	$ \vcenter{\includegraphics[scale=0.1]{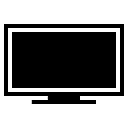}} $  & \scriptsize{ 64.29} & \scriptsize{	65.61} & \scriptsize{	66.79} & \scriptsize{	62.99} & \scriptsize{	47.25} & \scriptsize{	24.96} & \scriptsize{	67.10} & \scriptsize{	\textbf{67.30}} &  & 
\scriptsize{71.44} & \scriptsize{	69.19} & \scriptsize{	70.56} & \scriptsize{	70.75} & \scriptsize{	71.39} & \scriptsize{	55.59} & \scriptsize{	\textbf{73.17}} & \scriptsize{	72.49}
\\
\hline 
\centering
	$ \vcenter{\includegraphics[scale=0.1]{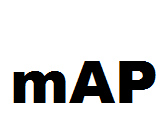}} $  &  \scriptsize{ 54.60} & \scriptsize{	71.69} & \scriptsize{	74.43} & \scriptsize{	74.69} & \scriptsize{	34.72} & \scriptsize{	44.03} & \scriptsize{	73.32} & \scriptsize{	\textbf{75.42}} &  & 
\scriptsize{76.79} & \scriptsize{	75.80} & \scriptsize{	78.17} & \scriptsize{	78.47} & \scriptsize{	66.57} & \scriptsize{	63.09} & \scriptsize{	76.73} & \scriptsize{	\textbf{78.71}	}
\\
\hline 
\end{tabular}}
\caption{Varying the cardinality of the selected features. The image classification results achieved in terms of average precision (AP) scores while selecting the first $128$ (3\%) and $256$ (6\%) features from the total $4,096$.}
\label{table:ch5voc07_2562}
\vspace{-0.5mm}
\end{table}
In object recognition VOC-2007 is a suitable tool for testing models, therefore, we use it as reference benchmark to assess the strengths and weaknesses of using the EC-FS approach regarding the classification task. For this reason, we compare the EC-FS approach against 8 state-of-the-art FS methods reported in Table~\ref{table:ch5compmethods2}, including the Inf-FS introduced in the previous section. 
This experiment considers as features the cues extracted with a deep convolutional neural network architecture (CNN). We selected the pre-trained model called very deep ConvNets~\cite{Simonyan14c}, which performs favorably to the state of the art for classification and detection in the ImageNet Large-Scale Visual Recognition Challenge 2014 (ILSVRC). We use the 4,096-dimension activations of the last layer as image descriptors (i.e., 4,096 features in total). The VOC-2007 edition contains about 10,000 images split into train, validation, and test sets, and labeled with twenty object classes. A one-vs-rest SVM classifier for each class is learnt (where cross-validation is used to find the best parameter C and $\alpha$ mixing coefficient in Eq. \ref{eq5:alpha2} on the training data) and evaluated independently and the performance is measured as mean Average Precision (mAP) across all classes. 

Table~\ref{table:ch5voc07_2562} serves to analyze and empirically clarify how well important features are ranked high by several FS algorithms. The amount of features used for the two experiments is very low: $\approx$3\% and $\approx$6\% of the total. The results are significant: EC-FS method achieved the best performance in terms of mean average precision (mAP) followed by the unsupervised filter methods LS and Inf-FS.    
As for the methods in comparison, one can observe the high variability in classification accuracy; indeed, results show that EC-FS method is robust to classes (i.e., by changing the testing class its performance is always comparable with the top scoring method). According to Table \ref{table:ch5cnnResults}, the Inf-FS achieved the best performance by selecting the $88\%$ of CNN features, in this experiment it is third in the list while working on $\approx$3\% and $\approx$6\% of the total.  

\subsection{Exp. 2: Testing on Microarray Databases}

\begin{figure*}[!t]
\centerline{\includegraphics[width=0.8\columnwidth]{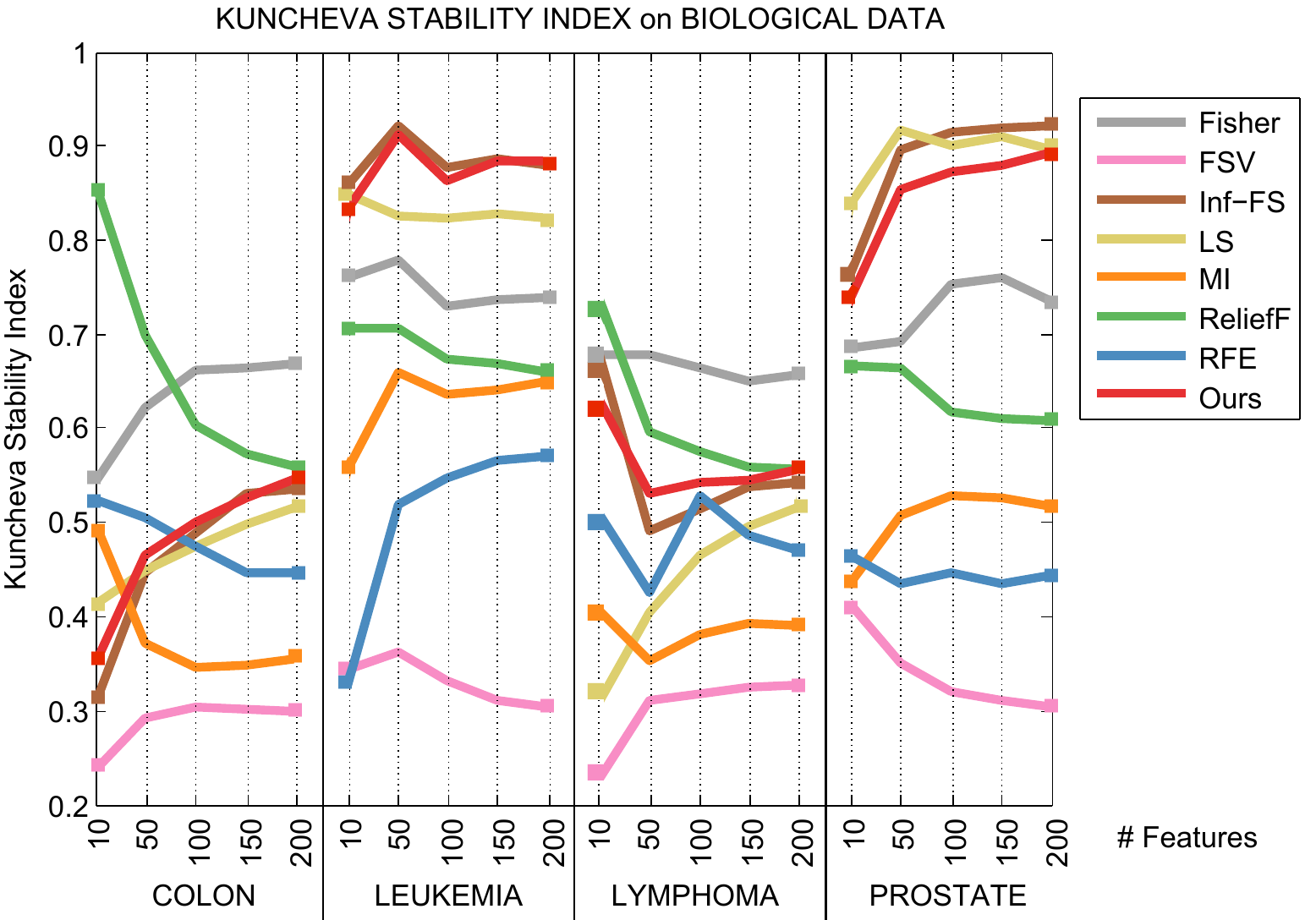}}
\caption{The Kuncheva stability indices for each method in comparison are presented. The figure reports the stability while varying the cardinality of the selected features from 10 to 200 on different benchmarks. The EC-FS is referred to as \textit{Ours}.}
\label{fig5:stability2}\vspace{-0.5cm}
\end{figure*}

In application fields like biology is inconceivable to devise an analysis procedure which does not comprise a FS step. A clear example can be found in the analysis of expression microarray data, where the expression level of thousands of genes is simultaneously measured. Within this scenario, we tested the proposed approach on four well-known microarray benchmark datasets for two-class problems. Results are reported in Table~\ref{tab_BIO2}. 
\begin{table}[t!]
\begin{center}
\resizebox{1\textwidth}{!}{%
\begin{tabular}{|l|C{0.62cm}|C{0.62cm}|C{0.62cm}|C{0.62cm}|C{0.82cm}|C{0.82cm}|C{0.02cm}|C{0.62cm}|C{0.62cm}|C{0.62cm}|C{0.62cm}|C{0.82cm}|C{0.82cm}|}
\hline
 \multicolumn{14}{|c|}{\large{\textbf{Microarray Databases}}} \\
 \hline
 & \multicolumn{6}{c|}{\textbf{COLON}}                                                                                        & & \multicolumn{6}{c|}{\textbf{LEUKEMIA}}                                                                                         \\ \hline
 & \multicolumn{4}{c|}{\# Features} & \multicolumn{2}{c|}{} & & \multicolumn{4}{c|}{\# Features} & \multicolumn{2}{c|}{} \\ \hline
\multicolumn{1}{|c|}{\scriptsize{\textbf{Method}}}   & 50      & 100     & 150     & 200     & \scriptsize{\textbf{Average}} & \scriptsize{\textbf{Time}} &  & 50      & 100     & 150     & 200     & \scriptsize{\textbf{Average}} & \scriptsize{\textbf{Time}}\\ \hline
\scriptsize{Fisher-S} 
&  \scriptsize{91.25} & \scriptsize{88.44} & \scriptsize{89.38} & \scriptsize{87.81}  & \scriptsize{89.22} & \scriptsize{0.02} & &
\scriptsize{99.33} & \scriptsize{99.78} & \scriptsize{99.78} & \scriptsize{99.78}  & \scriptsize{99.66} & \scriptsize{0.01}  \\ \hline
\scriptsize{FSV} 
&  \scriptsize{85.00} & \scriptsize{88.12} & \scriptsize{89.38} & \scriptsize{89.69}  & \scriptsize{88.04} & \scriptsize{0.18} & &
 \scriptsize{98.22} & \scriptsize{98.44} & \scriptsize{99.11 } & \scriptsize{99.33}  & \scriptsize{98.77} & \scriptsize{0.37}  \\ \hline
\scriptsize{Inf-FS}
&  \scriptsize{88.99 } & \scriptsize{89.41} & \scriptsize{89.32} & \scriptsize{89.01 }  & \scriptsize{89.18} & \scriptsize{0.91} & &
 \scriptsize{99.91} & \scriptsize{\textbf{99.92 }} & \scriptsize{\textbf{99.97}} & \scriptsize{\textbf{99.98}}  & \scriptsize{\textbf{99.95}} & \scriptsize{5.49}  \\ \hline
\scriptsize{LS}
&  \scriptsize{90.31} & \scriptsize{89.06} & \scriptsize{89.38} & \scriptsize{90.00}  & \scriptsize{89.68} & \scriptsize{0.03} & &
\scriptsize{98.67} & \scriptsize{99.33} & \scriptsize{99.56} & \scriptsize{99.56}  & \scriptsize{99.28} & \scriptsize{0.07}  \\ \hline
\scriptsize{MI}
&  \scriptsize{89.38} & \scriptsize{90.31} & \scriptsize{90.63} & \scriptsize{\textbf{90.94}}  & \scriptsize{90.31} & \scriptsize{0.31} & &
 \scriptsize{99.33} & \scriptsize{99.33} & \scriptsize{99.56} & \scriptsize{99.33}  & \scriptsize{98.38} & \scriptsize{0.21}  \\ \hline
\scriptsize{ReliefF}
& \scriptsize{80.94} & \scriptsize{84.38} & \scriptsize{85.94} & \scriptsize{87.50}  & \scriptsize{84.69} & \scriptsize{0.52} & &
 \scriptsize{99.56} & \scriptsize{99.78} & \scriptsize{99.78} & \scriptsize{99.78}  & \scriptsize{99.72} & \scriptsize{1.09}  \\ \hline
\scriptsize{RFE }   
& \scriptsize{89.06} & \scriptsize{85.00} & \scriptsize{86.88} & \scriptsize{85.62}  & \scriptsize{86.64} & \scriptsize{0.18} & &
 \scriptsize{\textbf{100}} & \scriptsize{99.78 } & \scriptsize{99.56} & \scriptsize{99.78}  & \scriptsize{99.78} & \scriptsize{0.14}  \\ \hline
\scriptsize{\textbf{EC-FS} }  
&  \scriptsize{\textbf{91.40}} & \scriptsize{\textbf{91.10}} & \scriptsize{\textbf{91.11}} & \scriptsize{90.63}  & \scriptsize{\textbf{91.06}} & \scriptsize{0.45} & &
 \scriptsize{99.92} & \scriptsize{\textbf{99.92}} & \scriptsize{99.77} & \scriptsize{99.85}  & \scriptsize{99.86} & \scriptsize{1.50}  \\ \hline
\end{tabular}}
\resizebox{1\textwidth}{!}{%
\begin{tabular}{|l|C{0.62cm}|C{0.62cm}|C{0.62cm}|C{0.62cm}|C{0.82cm}|C{0.82cm}|C{0.02cm}|C{0.62cm}|C{0.62cm}|C{0.62cm}|C{0.62cm}|C{0.82cm}|C{0.82cm}|}
\hline
 & \multicolumn{6}{c|}{\textbf{LYMPHOMA}}                                                                                        & & \multicolumn{6}{c|}{\textbf{PROSTATE}}                                                                                         \\ \hline
 & \multicolumn{4}{c|}{\# Features} & \multicolumn{2}{c|}{} & & \multicolumn{4}{c|}{\# Features} & \multicolumn{2}{c|}{} \\ \hline
\multicolumn{1}{|c|}{\scriptsize{\textbf{Method}}}   & 50      & 100     & 150     & 200     & \scriptsize{\textbf{Average}} & \scriptsize{\textbf{Time}} &    & 50      & 100     & 150     & 200     & \scriptsize{\textbf{Average}} & \scriptsize{\textbf{Time}}\\ \hline
\scriptsize{Fisher-S} 
& \scriptsize{98.75} & \scriptsize{98.38} & \scriptsize{98.38} & \scriptsize{100}  & \scriptsize{98.87} & \scriptsize{0.01} & &
 \scriptsize{96.10} & \scriptsize{96.20} & \scriptsize{96.30} & \scriptsize{97.30}  & \scriptsize{96.47} & \scriptsize{0.02}  \\ \hline
\scriptsize{FSV} 
& \scriptsize{98.22} & \scriptsize{98.44} & \scriptsize{99.11} & \scriptsize{99.33}  & \scriptsize{98.77} & \scriptsize{0.18} & &
\scriptsize{96.70} & \scriptsize{96.70} & \scriptsize{96.50} & \scriptsize{96.30}  & \scriptsize{96.55} & \scriptsize{0.63}  \\ \hline
\scriptsize{Inf-FS}
&  \scriptsize{98.12} & \scriptsize{98.75} & \scriptsize{98.75} & \scriptsize{99.38}  & \scriptsize{98.75} & \scriptsize{7.61} & &
 \scriptsize{\textbf{96.80}} & \scriptsize{\textbf{96.90}} & \scriptsize{\textbf{97.10}} & \scriptsize{96.70}  & \scriptsize{96.87} & \scriptsize{26.85}  \\ \hline
\scriptsize{LS}
&\scriptsize{90.00} & \scriptsize{96.88} & \scriptsize{99.38} & \scriptsize{98.75}  & \scriptsize{96.25} & \scriptsize{0.04} & &
 \scriptsize{85.80} & \scriptsize{94.60} & \scriptsize{96.90} & \scriptsize{97.00}  & \scriptsize{93.57} & \scriptsize{0.24}  \\ \hline
\scriptsize{MI}
&  \scriptsize{97.50} & \scriptsize{98.75} & \scriptsize{99.38} & \scriptsize{99.38}  & \scriptsize{98.75} & \scriptsize{0.59} & &
 \scriptsize{96.00} & \scriptsize{\textbf{96.90}} & \scriptsize{96.00} & \scriptsize{96.20}  & \scriptsize{96.27} & \scriptsize{1.01}  \\ \hline
\scriptsize{ReliefF}
&  \scriptsize{96.80} & \scriptsize{97.00} & \scriptsize{98.80} & \scriptsize{98.80}  & \scriptsize{97.85} & \scriptsize{0.74} & &
 \scriptsize{92.72} & \scriptsize{93.46} & \scriptsize{93.62} & \scriptsize{93.85}  & \scriptsize{93.41} & \scriptsize{2.68}  \\ \hline
\scriptsize{RFE }   
&  \scriptsize{96.00} & \scriptsize{98.00} & \scriptsize{98.80} & \scriptsize{99.00}  & \scriptsize{97.95} & \scriptsize{0.02} & &
 \scriptsize{93.40} & \scriptsize{96.40} & \scriptsize{\textbf{97.10}} & \scriptsize{96.32}  & \scriptsize{95.80} & \scriptsize{0.3}  \\ \hline
\scriptsize{\textbf{EC-FS} }  
&  \scriptsize{\textbf{99.40}} & \scriptsize{\textbf{99.20}} & \scriptsize{\textbf{99.60}} & \scriptsize{\textbf{99.20}}  & \scriptsize{\textbf{99.20}} & \scriptsize{1.50} & &
\scriptsize{96.28} & \scriptsize{\textbf{96.90}} & \scriptsize{96.80} & \scriptsize{\textbf{98.10}}  & \scriptsize{\textbf{97.02}} & \scriptsize{2.81}  \\ \hline
\end{tabular}}
\caption{The tables show results obtained on the expression microarray scenario. Tests have been repeated 100 times, and the means of the computed AUCs are reported for each dataset.}
\label{tab_BIO2}
\end{center}
\vspace{-1.5mm}
\end{table} 
The testing protocol adopted in this experiment consists in splitting the dataset up to 2/3 for training and 1/3 for testing. In order to have a fair evaluation, the feature ranking has been calculated using only the training samples, and then applied to the testing samples. The classification is performed using a linear SVM. For setting the best parameters (C of the linear SVM, and $\alpha$ mixing coefficient) we used a 5-fold cross validation on the training data. This procedure is repeated several times and results are averaged over the trials. Results are reported in terms of the Receiver Operating Characteristic or ROC curves. A widely used measurement that summarizes the ROC curve is the Area Under the ROC Curve (AUC)~\cite{BAMBER1975} which is useful for comparing algorithms independently of application. Hence, classification results for the datasets used show that the proposed approach produces superior results in all the cases. The overall performance indicates that EC-FS approach is more robust than the others on microarrays data, by changing the data it still produces high quality rankings. 
We assessed the stability of the selected features using the Kuncheva index~\cite{Kuncheva:2007}. This stability measure represents the similarity between the set of rankings generated over the different splits of the dataset. The similarity between sequences of size $N$ can be seen as the number of elements $n$ they have in common (i.e. the size of their intersection). The Kuncheva index takes values in [-1, 1], and the higher its value, the larger the number of commonly selected features in both sequences. The index is shown in Figure~\ref{fig5:stability2}, comparing EC-FS approach and the other methods. 
The proposed method shows, in most of the cases, a high stability whereas the highest performance is achieved.

\subsection{Exp. 2: Other Benchmarks} 

GINA has sparse input variables consisting of 970 features. It is a balanced data set
with 49.2\% instances belonging to the positive class. Results obtained on GINA indicate that the proposed approach overcomes the methods in comparison, and select the most useful features from a data set with high-complexity and dimensionality.
\begin{table} 
\begin{center}
\resizebox{1\textwidth}{!}{%
\begin{tabular}{|l|C{0.62cm}|C{0.62cm}|C{0.62cm}|C{0.62cm}|C{0.82cm}|C{0.82cm}|C{0.02cm}|C{0.62cm}|C{0.62cm}|C{0.62cm}|C{0.62cm}|C{0.82cm}|C{0.82cm}|}
\hline
 \multicolumn{14}{|c|}{\large{\textbf{FS Challenge Datasets}}} \\
 \hline
 & \multicolumn{6}{c|}{\textbf{GINA - Handwritten Recognition}} & & \multicolumn{6}{c|}{\textbf{MADELON - Artificial Data}}                                                                                         \\ \hline
 & \multicolumn{4}{c|}{\# Features} & \multicolumn{2}{c|}{} & & \multicolumn{4}{c|}{\# Features} & \multicolumn{2}{c|}{} \\ \hline
\multicolumn{1}{|c|}{\scriptsize{\textbf{Method}}}      & 50      & 100     & 150     & 200     & \scriptsize{\textbf{Average}} & \scriptsize{\textbf{Time}} &  & 50      & 100     & 150     & 200     & \scriptsize{\textbf{Average}} & \scriptsize{\textbf{Time}}\\ \hline
\scriptsize{Fisher-S} 
& \scriptsize{ 89.8} & \scriptsize{ 89.4} & \scriptsize{ 90.2} & \scriptsize{ \textbf{90.4}} & \scriptsize{89.9} & \scriptsize{ 0.05} & &
 \scriptsize{	61.9} & \scriptsize{	63.0} & \scriptsize{62.3} & \scriptsize{64.0} & \scriptsize{   62.5} & \scriptsize{  0.02}  
\\ \hline
\scriptsize{FSV} 
 & \scriptsize{ 81.9} & \scriptsize{ 83.7} & \scriptsize{ 82.0} & \scriptsize{ 83.6} & \scriptsize{ 82.7} & \scriptsize{ 138 } & &
 \scriptsize{59.9} & \scriptsize{ 60.6} & \scriptsize{	61.0} & \scriptsize{ 61.0} & \scriptsize{ 60.7} & \scriptsize{732}  
\\ \hline
\scriptsize{Inf-FS}
 & \scriptsize{ 89.0} & \scriptsize{ 88.7} & \scriptsize{ 89.1} & \scriptsize{ 89.0} & \scriptsize{ 88.9} & \scriptsize{ 41} & &
 \scriptsize{	62.6} & \scriptsize{	\textbf{63.8}} & \scriptsize{	\textbf{65.4}} & \scriptsize{	60.8} & \scriptsize{	63.2} & \scriptsize{  0.04}  \\ \hline
\scriptsize{LS}
& \scriptsize{ 82.2} & \scriptsize{ 82.4} & \scriptsize{ 83.4} & \scriptsize{ 83.2} & \scriptsize{ 82.7} & \scriptsize{ 1.30} & &
 \scriptsize{	62.8} & \scriptsize{	62.9} & \scriptsize{	63.3} & \scriptsize{	64.7} & \scriptsize{	63.4} & \scriptsize{       8.13}  \\ \hline
\scriptsize{MI}
& \scriptsize{ 89.3} & \scriptsize{ 89.7} & \scriptsize{ 89.8} & \scriptsize{ 90.1} & \scriptsize{ 89.6} & \scriptsize{ 1.13} & &
\scriptsize{	63.0} & \scriptsize{	63.7} & \scriptsize{	63.5} & \scriptsize{ 64.7} & \scriptsize{	63.6} & \scriptsize{ 0.4}  \\ \hline
\scriptsize{ReliefF}
 & \scriptsize{ 77.9} & \scriptsize{ 76.3} & \scriptsize{ 77.3} & \scriptsize{ 76.9} & \scriptsize{ 77.2} & \scriptsize{ 0.12} & &
 \scriptsize{ 62.9 } & \scriptsize{ 63.1 } & \scriptsize{	63.2 } & \scriptsize{ \textbf{64.9} } & \scriptsize{ 63.5 } & \scriptsize{ 10.41}
\\ \hline
\scriptsize{RFE }   
 & \scriptsize{ 82.2} & \scriptsize{ 82.4} & \scriptsize{ 83.4} & \scriptsize{ 83.2} & \scriptsize{ 82.7} & \scriptsize{ 6.60} & &
 \scriptsize{  55.0} & \scriptsize{ 61.2} & \scriptsize{ 57.1} & \scriptsize{ 60.2} & \scriptsize{     56.5} & \scriptsize{    50163}  \\ \hline
\scriptsize{\textbf{EC-FS} }  
 & \scriptsize{ \textbf{90.9}} & \scriptsize{ \textbf{90.3}} & \scriptsize{ \textbf{90.4}} & \scriptsize{ 89.5} & \scriptsize{  \textbf{90.3}} & \scriptsize{ 1.56} & &
\scriptsize{	\textbf{63.6}} & \scriptsize{\textbf{63.8}} & \scriptsize{	63.7} & \scriptsize{	63.3} & \scriptsize{	\textbf{63.7}} & \scriptsize{  0.57}  \\ \hline
\end{tabular}}
\caption{Varying the cardinality of the selected features. (ROC) AUC (\%) on different datasets by SVM classification. Performance obtained with the first 50, 100, 150, and 200 features.}
\label{table:ch5NIPS2}
\end{center}
\vspace{-1.5mm}
\end{table}
MADELON is an artificial dataset, which was part of the NIPS $2003$ feature selection challenge. It represents a two-class classification problem with continuous input variables. The difficulty is that the problem is multivariate and highly non-linear. Results are reported in Table~\ref{table:ch5NIPS2}.
This gives a proof about the classification performance of the EC-FS approach that is attained on the test sets of GINA and MADELON.

FS techniques definitely represent an important class of preprocessing tools, by eliminating uninformative features and strongly reducing the dimension of the problem space, it allows to achieve high performance, useful for practical purposes in those domains where high speed is required.

\subsection{Reliability and Validity}

In order to assess if the difference in performance is statistically significant, t-tests have been used for comparing the accuracies. Statistical tests are used to determine if the accuracies obtained with the proposed approach are significantly different from the others (whereas both the distribution of values were normal). The test for assessing whether the data come from normal distributions with unknown, but equal, variances is the \emph{Lilliefors} test. Results have been obtained by comparing the results produced by each method over 100 trials (at each trial corresponds a different split of the data). Given the two distributions $x_p$ of the proposed method and $x_c$ of the current competitor, of size $1 \times 100$, a \textit{two-sample t-test} has been applied obtaining a test decision for the null hypothesis that the data in vectors $x_p$ and  $x_c$ comes from independent random samples from normal distributions with equal means and equal but unknown variances. Results (highlighted in Table~\ref{tab_BIO2} and Table \ref{table:ch5NIPS2}) show a statistical significant effect in performance (p-value $<$ 0.05, Lilliefors test H=0).

\subsection{Summary}\label{sec5:conc}

In this section, we presented the idea of solving feature selection via the Eigenvector centrality measure. We designed a graph -- where features are the nodes -- weighted by a kernelized adjacency matrix, which draws upon best-practice in feature selection while assigning scores according to how well features discriminate between classes. The method (supervised) estimates some indicators of centrality identifying the most important features within the graph. The results are remarkable: the proposed method, called EC-FS, has been extensively tested on $7$ different datasets selected from different scenarios (i.e., object recognition, handwritten recognition, biological data, and synthetic testing datasets), in all the cases it achieves top performances against $7$ competitors selected from recent literature in feature selection.  EC-FS approach is also robust and stable on different splits of the training data, it performs effectively in ranking high the most relevant features, and it has a very competitive complexity. This study also points to many future directions; focusing on the investigation of different implementations for parallel computing for big data analysis or focusing on the investigation of different relations among the features. Finally, we provide an open and portable library of feature selection algorithms, integrating the methods with uniform input and output formats to facilitate large scale performance evaluation. The \textit{Feature Selection Library} (FSLib 4.2 \footnote{Available on File Exchange - MATLAB Central \url{https://it.mathworks.com/matlabcentral/fileexchange/56937-feature-selection-library}}) and interfaces are fully documented. The library integrates directly with MATLAB, a popular language for machine learning and pattern recognition research (see Appendix \ref{app:A} for further details).

\section{Discussions and Final Remarks}\label{sec5:FinalRemarks} 
The most appealing characteristics of the Inf-FS are i) all possible subsets of features are considered in evaluating the rank of a given feature and ii) it is extremely efficient, as it converts the feature ranking problem to simply calculating the geometric series of an adjacency matrix. Although it outperforms most of the state-of-the-art feature selection methods in image classification and gene expression problems, the algorithm suffers from two important deficiencies. Firstly, the methodology used to weight the graph is data-driven and data-dependent. In other words, the Inf-FS may be not always robust across all datasets. If a certain feature has zero dispersion (i.e. variance) over the examples in the dataset, then that feature does not have any information and can be discarded. For a feature with non-zero dispersion, although we can not definitively relate its relevance to its dispersion magnitude, our experiments also show that keeping features that have large standard deviation, improves the classification accuracy (it has also been shown previously that using dispersion measures improves the performance \cite{Guyon03}). Secondly, we used the Spearman's rank correlation coefficient as a measure
of redundancy of a feature. However, this measure is not able to individuate complex nonlinear dependencies between features (e.g. nonmonotonic non-linear dependencies). 
These deficiencies can be overcame developing novel ways for the graph weighting, such as the application of machine learning techniques (e.g., structured SVMs) which are able to adaptively weight the graph according to the data and/or the problem.

\subsection{Discrete Path Integrals}\label{sec5:Quantum} 

A considerable amount of the theoretical developments in physics would not be understandable without the use of \textit{path integrals}. A fundamental difference between classical physics and quantum theory is the fact that, in the quantum world, certain predictions can only be made in terms of probabilities.

As an example, take the question whether or not a particle that starts at the time $t_i$ at the location $x_i$ will reach location $x_j$ at the later time $t_j$. Classical physics can give a definite answer. Depending on the particle's initial velocity and the forces acting on it, the answer is either yes or no. In quantum theory, it is merely possible to give the probability that the particle in question can be detected at location $x_j$ at time $t_j$. The path integral formalism, which was introduced by the physicist Richard Feynman~\cite{Feynman:1948aa}, is a tool for calculating such quantum mechanical probabilities. According to~\cite{Feynman:1948aa}, the Feynman's recipe applied to a particle travelling from $x_i$ to $x_j$, considers all the possibilities for the particle travelling from $x_i$ to $x_j$. Not only the straight-line approach, but also the possibility of the particle turning loopings and making diverse detours. Given a path from $x_i$ to $x_j$, the first part of the particle's trajectory may be travelled at high speed and the final part at a lower one.

Let us consider one possible trajectory. Imagine that we make an enormous number of successive position measurements of the coordinate $x$, such as $x_1,x_2,x_3,...$, at successive times $t_1,t_2,t_3,...$ separated by a small time interval $\xi$ (where $t_{i+1} = t_i + \xi$). From the classical point of view, the successive values, $x_1,x_2,x_3,...$ of the coordinate practically define a path $x(t)$. In quantum theory, it is merely possible to give the probability of such a path, which is a function of $x_1,x_2,x_3,...$, say $P(x_1,x_2,x_3,...)$. The probability that the path lies in a particular region $R$ of space-time is obtained classically by integrating $P$ over that region. In quantum mechanics, this procedure is correct for the case that $x_1,x_2,x_3,...$ were actually all measured, and then only those paths lying within $R$ were taken.

Since in quantum mechanics each particle acts also as a wave, there is the need to introduce the complex number $\varphi$ (wave function) which is a way to link classical mechanics to the quantum one. 

\begin{equation}
P(x_1,x_2) = |\varphi(x_1,x_2)|^2
\end{equation}

What is expected is that the probability that the particle is found by the measurement above in the region $R$ is the square of this complex number: $|\varphi(R)|^2$. The number $\varphi(R)$ is the probability amplitude for the region $R$ which is given by:
\begin{equation}\label{eq1}
\varphi(R) = \lim_{\xi\to0} \int_R \Phi(...x_i,x_{i+1}...) \,dx_i\,dx_{i+1}....
\end{equation}
where the complex number $\Phi(...x_i,x_{i+1}...)$ is a function of the variable $x_i$, defining the path. $\Phi$ is a product of contributions from successive sections of the path, this is why this quantity has the properties of a wave function.

\begin{equation}\label{eq2}
\Phi(...x_i,x_{i+1}...) = ...\Phi(x_{i-1})\Phi(x_i)\Phi(x_{i+1})...
\end{equation}

The absolute square of this quantity gives the probability of that path. Thus:
\begin{equation}\label{eq3}
P(R) = |\varphi(R)|^2 = \lim_{\xi\to0} |\int_R \Phi(...x_i,x_{i+1}...) \,dx_i\,dx_{i+1}....|^2
\end{equation}

Summarizing, in order to determine whether a particle has a path lying in a region $R$ of space-time, the probability that the result will be affirmative is the absolute square of a sum of a product of contributions from successive sections of the path (complex contributions), one for each path in the region~\cite{Feynman:1948aa}. Path integrals are given by sum over all paths satisfying some boundary conditions and can be understood as extensions to an infinite number of integration variables of usual multi-dimensional integrals. Path integrals are powerful tools for the study of quantum mechanics. Indeed, in quantum mechanics, physical quantities can be expressed as averages over all possible paths weighted by the exponential of a functional term related to the action principle of classical mechanics. 

The result has a probabilistic interpretation. The sum over all paths of the exponential factor can be seen as the sum over each path of the probability of selecting that path. The probability is the product over each segment of the probability of selecting that segment, so that each segment is probabilistically independently chosen. 

\subsubsection{A Step Towards the Inf-FS}

According to the Feynman formulation~\cite{Feynman:1948aa}, we derive a discrete form for path integral. Before linking this process to feature selection, we mapped the space-time manifold to a simplified form: a graph.  
Let \textbf{G} be a graph, where each vertex represents a location of the space-time $x_i = x_1,x_2,x_3,...$, at successive times $t_i = t_1,t_2,t_3,...$ separated by a small finite time interval $\epsilon$. 

Let be $\varphi_t(x_i,x_j)$ a complex number, which is a function of variable $x$, whose absolute module expresses the probability of travelling from vertex $x_i$ to $x_j$ (that is to say, from location $x_i$ to $x_j$). 

\[
     p_{ij} = |\varphi_t(x_i,x_j)|^2 
\]

Since $\varphi_t$ is a wave function, the probability is calculated as its amplitude.

Generally, $\varphi_t$ has the following form:

\[
     \varphi_t =  e^{- \Delta t} 
\]

In physics $\Delta$ is a functional term related to the action principle of classical mechanics. In order to calculate Eq.\ref{eq2} on a finite graph let $\gamma = \{ x_{0}=i, x_{1}, ..., x_{l-1}, x_{l}=j \}$ denote a path of length $l$ between vertices $i$ and $j$ through the vertices $x_{1},...,x_{l-1}$. 

\begin{equation}\label{four}
\Phi_{\gamma_l}  = \prod_{k=0}^{l-1} \Phi(x_{k},x_{k+1}),
\end{equation}

We can rewrite this equation by explicating the probabilities of paths.

\begin{equation}\label{five}
P_{\gamma_l}  = \prod_{k=0}^{l-1} |\Phi(x_{k},x_{k+1})|^2,
\end{equation}

In order to apply the path integral process out of the quantum mechanics, it is possible to directly start from probabilities. The weighted adjacency matrix of the graph will contain the probability to travel from a vertex $i$ to vertex $j$ on the edge connecting them.

Eq.\ref{eq1} sums all the paths in the region $R$ connecting two nodes. In the discrete case, the region is the set of the edges of \textbf{G}. Thus, we have to consider for each pairs $i$ and $j$, all the possible paths which connect them. Therefore, we have also to consider all the possible length of these paths, $l=1,..,\infty$, and repeat this process for each pairs of vertices $i$ and $j$. We define the set $\mathbb{P}_{i,j}^l$ as containing all the paths of length $l$ between $i$ and $j$; we recall that a path is $\gamma = \{ x_{0}=i, x_{1}, ..., x_{l-1}, x_{l}=j \}$.

\begin{equation}\label{eq33}
P(R) = \sum_{l=1}^\infty \big( \sum_{\gamma \in \mathbb{P}_{i,j}^l } \prod_{k=0}^{l-1} |\Phi(x_{k},x_{k+1})|^2 \big).
\end{equation}

Eq.\ref{eq33} expresses this concept, considering all the possible paths connecting each pairs of vertices. It multiplies the contributions of each segment of a path of a certain length $l$, and sum all the possible paths of the the length $l$ together. finally, it lets the length $l$ go to infinity $l\rightarrow\infty$, so as to actually integrate over all possible paths connecting the points.

Summarizing, the path integral formalism, is a tool for calculating quantum mechanical probabilities. The basic idea behind path integrals is that of measuring the quantum probability of a space-time event. In particular, if a particle is measured to be at a particular position at a particular time, to calculate the quantum amplitude (or its probability) to be in a different position at a later time, all the possible space-time paths the particle can take between those two space-time events must be considered. Thus, we derived a discrete form for path integral. Then, we mapped the space-time to a simplified discrete form, without time, that is: \textit{a graph}. Finally, the framework used to estimate the most likely  position where to find a particle has been switched to the novel problem of finding the most relevant feature (see Sec. \ref{sec5:Quantum} for details). According to this, we derived an algorithm to rank features by mapping each node to a particular feature and each weighted edge to an arbitrary measurement expressing a degree of relevance (note, the weight can be designed in many different way or automatically learned from data).

\subsection{Markov Chains and Random Walks}\label{sec:markovProcesses}\vspace{-0.2cm}

This section provides a probabilistic interpretation of the proposed algorithm based on Absorbing Random Walks. 
Here, we reformulate the problem in terms of Markov chains and random walks. The set of nodes in a Markov chain are called \textit{states} and each move is called a \textit{step}. 
Let $T$ be the \textit{matrix of transition probabilities}, or the \textit{transition matrix} of the Markov chain. If the chain is currently in state $v_i$, then it moves to state $v_j$ at the next step with a probability denoted by $t_{ij}$, and this probability does not depend upon which states the chain was in before the current state. The probabilities $t_{ij}$ are called transition probabilities.
The process can remain in the state it is in, and this occurs with probability $t_{ii}$. 
An absorbing Markov chain is a special Markov chain which has absorbing states, i.e., states which once reached cannot be transitioned out of (i.e., $t_{ii} = 1$). A Markov chain is absorbing if it has at least one absorbing state, and if from every state it is possible to go to an absorbing state in a finite number of steps. In an absorbing Markov chain, a state that is not absorbing is called transient.
The transition matrix for any absorbing chain can be written in the \textit{canonical} form
\[
T =
  \begin{bmatrix}
    \textbf{I} & \textbf{0}  \\
    R & A
  \end{bmatrix}
\]
where $R$ is the rectangular submatrix giving transition probabilities from non-absorbing to absorbing states, $A$ is the square submatrix giving these probabilities from non-absorbing to non-absorbing states, $\textbf{I}$ is an identity matrix, and \textbf{0} is a rectangular matrix of zeros.

Note that $R$ and $\textbf{0}$ are not necessarily square. More precisely, if there are $m$ absorbing states
and $n$ non-absorbing states, then $R$ is $n \times m$, $A$ is $n \times n$ , \textbf{I} is $m \times m$, and \textbf{0} is $m \times n$. Iterated multiplication of the $T$ matrix yields

\[
T^2 =
  \begin{bmatrix}
    \textbf{I} & \textbf{0}  \\
    R & A
  \end{bmatrix}
\begin{bmatrix}
    \textbf{I} & \textbf{0}  \\
    R & A
  \end{bmatrix}
=
\begin{bmatrix}
    \textbf{I} & \textbf{0}  \\
    R+AR & A^2
  \end{bmatrix}
\]
\[
T^3 =
 \begin{bmatrix}
    \textbf{I} & \textbf{0}  \\
    R+AR & A^2
  \end{bmatrix}
\begin{bmatrix}
    \textbf{I} & \textbf{0}  \\
    R & A
  \end{bmatrix}
=
\begin{bmatrix}
    \textbf{I} & \textbf{0}  \\
    R+AR+A^2R & A^3
  \end{bmatrix}
\]
and hence by induction we obtain
\[
T^l =
\begin{bmatrix}
    \textbf{I} & \textbf{0}  \\
   (\textbf{I}+A+A^2+...+A^{l-1})R & A^l
  \end{bmatrix}
\]
The preceding example illustrates the general result that $A^l \to 0$ as $l \to \infty$. Thus
\[
T^{\infty} =
\begin{bmatrix}
    \textbf{I} & \textbf{0}  \\
   SR & \textbf{0}
  \end{bmatrix}
\]
where the matrix
\[
S = \textbf{I} + A + A^2 + ... + A^{\infty}= (I-A)^{-1}
\]
is called the \textit{fundamental matrix} for the absorbing chain. Note that $S$, which is a square matrix with rows and columns corresponding to the non-absorbing states, is derived in the same way of Eq.\ref{eq5:six}. $S(i, j)$ is the expected number of periods that the chain spends in the $j^{th}$ non-absorbing state given that the chain began in the $i^{th}$ non-absorbing state. Perhaps this interpretation comes from the specification of the matrix $S$ as the infinite sum, since $A^l(i, j)$ is the probability that the process which began in the $i^{th}$ non-absorbing state will occupy the $j^{th}$ non-absorbing state in period $l$. However, $A^l(i, j)$ can also be understood as the expected proportion of period $l$ spent in the $j^{th}$ state. Summing over all time periods $l$, we thus obtain the total number of periods that the chain is expected to occupy the $j^{th}$ state.

\subsection{A Graph Theory Perspective}\label{sec5:graphtheory}

An interesting fact emerges when we compare the two techniques Inf-FS and EC-FS from a methodological point of view. Indeed, from a graph theory perspective identifying the most important nodes corresponds to individuate some indicators of centrality within a graph (e.g., the relative importance of nodes). A first way used in graph theory is to study accessibility of nodes, see~\cite{Garrison1960,Pitts1965} for example. The idea is to compute $A^l$ for some suitably large $l$ (often the diameter of the graph), and then use the row sums of its entries as a measure of accessibility (i.e. $scores(i) = [A^l\textbf{e}]_i$, where $\textbf{e}$ is a vector with all entries equal to $1$). The accessibility index of node $i$ would thus be the sum of the entries in the $i$-th row of $A^l$, and this is the total number of paths of length $l$ (allowing stopovers) from node $i$ to all nodes in the graph. One problem with this method is that the integer $l$ seems arbitrary. However, as we count longer and longer paths, this measure of accessibility converges to a index known as eigenvector centrality measure (EC)~\cite{Bonacich}. 

The basic idea behind the EC is to calculate $v_0$ the eigenvector of $A$ associated to the largest eigenvalue. Its values are representative of how strongly each node is connected to the other nodes. Since the limit of $A^l$ as $l$ approaches a large positive number $L$ converges to $v_0$,
\begin{equation}\label{eq5:extra2}
          \lim_{l \to L} [A^l\textbf{e}] = v_0, 
\end{equation}
the EC index makes the estimation of indicators of centrality free of manual tuning over $l$, and computationally efficient.

Let us consider a vector, for example $\textbf{e}$, that is \emph{not} orthogonal to the principal vector $v_0$ of $A$. It is always possible to decompose $\textbf{e}$ using the eigenvectors as basis with a coefficient $\beta_{0} \neq 0 $ for $v_{0}$. Hence:
\begin{equation}
	\textbf{e} = \beta_0 v_0+\beta_1 v_1+ \ldots +\beta_n v_n,  \quad (\beta_0 \neq 0).
	\label{eq5:a12}
\end{equation}
Then
\begin{equation}
\begin{split}
	A \textbf{e} = A(\beta_0 v_0+\beta_1 v_1+ \ldots +\beta_n v_n) = \beta_0 A v_0+\beta_1 A v_1+ \ldots +\beta_n A v_n =	\\
	 = \beta_0 \lambda _0 v_0 +\beta_1 \lambda  _1 v_1+ \ldots +\beta_n \lambda  _n v_n.
\end{split}
	\label{eq5:a22}
\end{equation}
So in the same way:
\begin{equation}
\begin{split}
	A^l \textbf{e} = A^l(\beta_0 v_0+\beta_1 v_1+ \ldots +\beta_n v_n) = \beta_0 A^{l} v_0+\beta_1 A^{l} v_1+ \ldots +\beta_n A^{l} v_n =	\\
	 = \beta_0 \lambda _0 ^{l}v_0 +\beta_1 \lambda ^{l}_1 v_1+ \ldots +\beta_n \lambda   ^{l}_n v_n , \quad (\beta_0 \neq 0).
\end{split}
	\label{eq5:a32}
\end{equation}
Finally we divide by the constant $\lambda_0^l \neq 0$ (see Perron-Frobenius theorem~\cite{Meyer:2000:MAA:343374}),  
 \begin{equation}
	\frac{A^l \textbf{e}}{\lambda_0^l } = \beta_0 v_0+ \frac{\lambda_1^l \beta_1 v_1}{\lambda_0^l}+ \ldots + \frac{\lambda_n^l \beta_n v_n}{\lambda_0^l}, \quad (\beta_0 \neq 0).
	\label{eq5:a4}
\end{equation}
The limit of $\frac{A^l \textbf{e}}{\lambda_0^l }$ as $l$ approaches infinity equals $\beta_0 v_0$ since $ \lim_{l \to \infty} \frac{\lambda_1^l}{\lambda_0^l} = 0 $, $\forall l > 0 $. What we see here is that as we let $l$ increase, the ratio of the components of $A^l\textbf{e}$ converges to $v_0$. Therefore, marginalizing over the columns of $A^l$, with a sufficiently large $l$, corresponds to calculate the principal eigenvector of matrix $A$~\cite{Bonacich}. 

\begin{figure}[t!]
\centerline{\includegraphics[width=1\columnwidth]{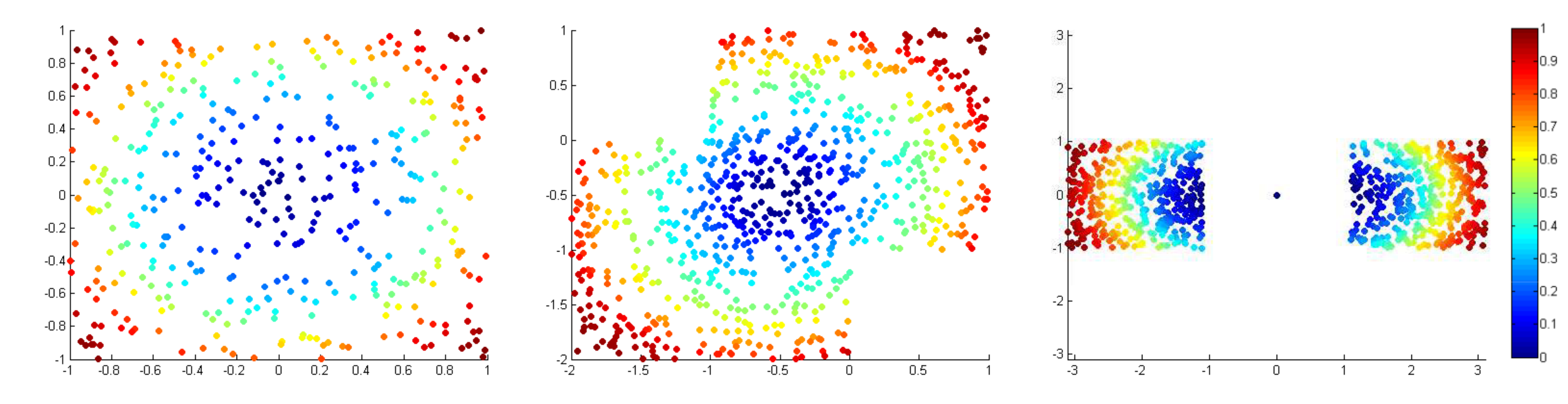}}
\caption{Eigenvector centrality plots for three random planar graphs. On the left, a simple Gaussian distribution where central nodes are at the peripheral part of the distribution as expected. The central and right plots, some more complicated distributions, a node receiving many links does not necessarily have a high eigenvector centrality.}
\label{fig5:ex2}
\end{figure}
This is a really interesting result that gives valuable insights on the Inf-FS method. Indeed, the main difference between the EC-FS and the Inf-FS is that the EC-FS does not account for all the contributions given by the power series of matrices (or sub-solutions). The Inf-FS integrates each partial solution (i.e., $S = A^1+A^2+...+A^l, l\to \infty$), that together help the method to react in a different way in presence of noise in data or many redundant cues. 

Figure~\ref{fig5:ex2} illustrates a toy example for the EC-FS taking three random planar graphs. Graphs are made of $700$ nodes and they are weighted by the Euclidean distance between each pair of points. In the example, high scoring nodes are those ones farther from the mean (i.e., the distance is conceived as quantity to maximize), the peculiarity of the eigenvector centrality is that a node is important if it is linked to by other important nodes (higher scores). 

Summarizing, the gist of eigenvector centrality is to compute the centrality of a node as a function of the centralities of its neighbors. The Inf-FS algorithm can be seen as a centrality measure that characterizes the \emph{global} (as opposed to local) prominence of a node in the graph while taking into account incremental contributions to the final solution.

\chapter{Online Feature Ranking for Visual Object Tracking}\label{ch6:TRACKING}

Visual tracking remains a highly popular research area of computer vision, with the number of motion and tracking papers published at high profile conferences exceeding 40 papers annually ~\cite{Danelljan_2014_CVPR,struck,Frag,jia2012visual}. The significant activity in the field over last two decades is reflected in the abundance of review papers. Most of the tracking systems employ a set of features which are used statically across successive frames~\cite{Henriques:2012}. It is well known that tracking deals with image streams that change over time~\cite{CollinsLL05}, therefore, data will easily produce frames in which the object has low-contrast or the target region is blurred (e.g. due to the motion of the target or camera), in these cases to obtain robust tracking a great deal of prior knowledge about scene structure or expected motion is imposed~\cite{bao2012real,wu2012online,zhang2012robust}, and thus tracking success is bought at the price of reduced generality.  Selecting the right features plays a critical role in tracking~\cite{CollinsLL05,Collins_2005_5110,OAB,BSBT}. Trackers which comprises feature selection strategies into their inner engines can be flexible enough to handle gradual changes in appearance in real scenarios~\cite{Danelljan_2014_CVPR}. The degree to which a tracker can discriminate a target from its surroundings is directly related to this flexibility. Since foreground and background appearance can also drastically change as the target object moves from place to place, tracking features should also need to adapt. 

Many feature selection methods used in off-line settings (e.g., bioinformatics, data mining~\cite{Peng05featureselection,Roffo_2015_ICCV,Hutter:02feature,Quanquanjournals}) have been so far largely neglected, to the best of our knowledge, at the level of online visual tracking, especially under the hard constraint of speed required for target tracking. 
\begin{figure}
\center
\includegraphics[width=0.98\textwidth]{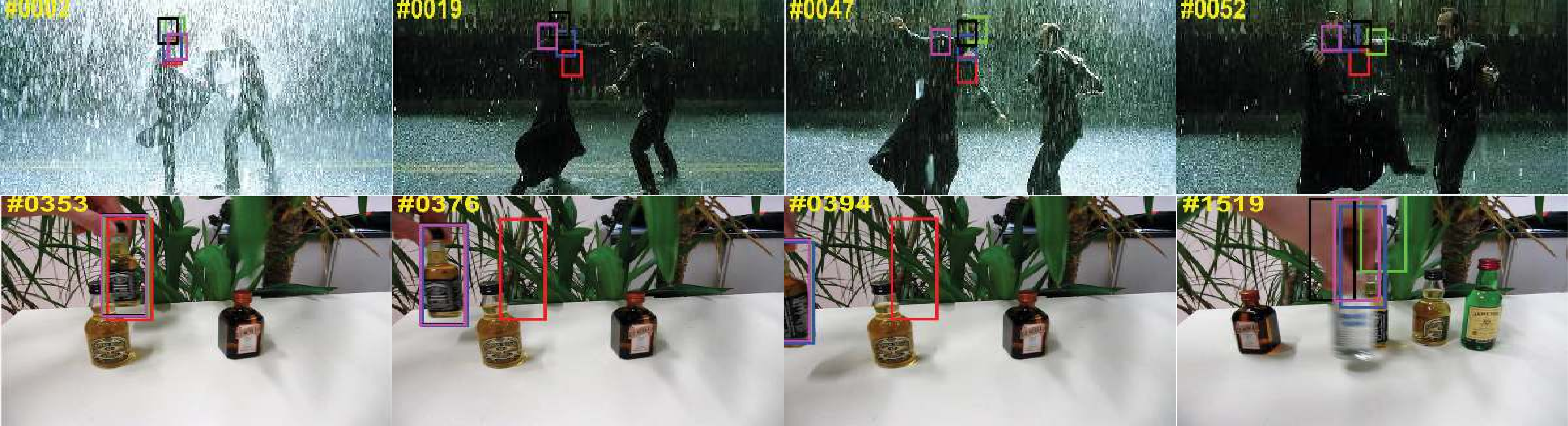}
\caption{Comparison of feature selection approaches embedded on the ACT tracker in challenging situations such as fast motion, out-of-view, occlusion, illumination variation, motion blur. The results of ACT, ACTFisher, ACTMI, ACT-mRMR and ACTinffs are represented by red, green, blue, black and pink boxes respectively.}
\label{fig6:frames}
\end{figure}
This chapter demonstrates the importance of feature selection in realtime applications,  resulted in what is clearly a very impressive performance. Figure~\ref{fig6:frames}.\textbf{A} presents tracking results in an environment of illumination variations, occlusions, out-of-views, fast motion among other challenging aspects. The example frames are from the \textit{Matrix} and \textit{liquor} sequences. All the feature selection approaches performs favorably against the baseline ACT tracker (red box). The contribution of the work presented in this chapter is threefold. Firstly, we evaluate a pool of modern feature selection algorithms (among filter, embedded, and wrapper methods) selected to meet the requirements of a real-time application. Secondly, we investigate the strengths and weaknesses of these methods for a classification task in order to identify the right candidates for visual tracking. Finally, the selected candidates are embedded on the Adaptive Color Tracking system~\cite{Danelljan_2014_CVPR} (ACT). We extensively test the solutions on 50 test sequences from the Online Object Tracking~\cite{OOT.CVPR.2013} (OTB-50) benchmark. In section \ref{sec6:OnlineFS} our solutions performance has been evaluated with the same protocol and methodology provided by the OTB-50 benchmark. The baseline ACT and its variants, with different feature selection mechanisms, have been also compared against 29 state-of-the-art trackers (it is worth noting that 10 of them use feature selection as part of their framework). In section \ref{sec6:VOT16} we present our contribution to the Visual Object Tracking Challenge, VOT 2016, an initiative established to address performance evaluation for short-term visual object trackers. 

In detail, this chapter is organized as follows: Section~\ref{sec6:SoAx} presents a brief look at online feature selection for visual tracking, mostly focusing on the comparative approaches we consider in this study. In Section~\ref{sec6:ACTtracker} we present the ACT tracker and in Section~\ref{sec6:Embedded} how we embedded feature selection on it. Extensive experiments on the OTB-50 are reported in Section~\ref{sec6:Exp}, where we also analyze the strengths and weaknesses of the proposed method regarding the classification task and the OTB-50 dataset is discussed, defining the evaluation methodology and reporting results. In Section~\ref{sec6:Concl}, a summary of this first part is given.  Section \ref{sec6:VOT16} presents the Visual object tracking challenge, VOT 2016, that aims at comparing short-term single-object visual trackers that do not apply prelearned models of object appearance. In section \ref{sec6:votResults} results of 70 trackers are presented, with a large number of trackers being published at major computer vision conferences and journals in the recent years. The number of tested state-of-the-art trackers makes the VOT 2016 the largest and most challenging benchmark on short-term tracking to date. The VOT 2016 has received an astonishingly large number of submissions, many of these can arguably considered a state-of-the-art. As our further contribution to the tracking community, we collaborated to the development of the repository of submitted trackers. This resulted in a library of roughly 40 trackers which is now publicly available from the VOT page (\url{http://www.votchallenge.net/vot2016/trackers.html}).

\section{Related Work}\vspace{-0.05cm}
\label{sec6:SoAx}

The problem of real-time feature selection has, to the best of our knowledge, rarely been addressed in the literature, especially under the hard constraint of speed required for target tracking.This section presents the related literatures of real-time feature selection for target tracking, mainly focusing on the comparative approaches used in the experimental section. The Variance Ratio (VR)~\cite{CollinsLL05} tracker is an online feature ranking mechanism based on applying the two-class variance ratio to log likelihood distributions computed for a given feature from samples of object and background pixels. This feature ranking approach is embedded in a tracking system that selects top-ranked features for tracking. Other recent variants of VR are provided by the Video Verification of Identity (VIVID) testbed~\cite{Collins_2005_5110} which includes the mean shift (MS-V), template matching (TM-V), ratio shift (RS-V), and peak difference (PD-V) methods. Another robust tracker is the Sparsity-based Collaborative Model~\cite{zhong2012robust} (SCM), it is an object tracking algorithm which uses a robust appearance model that exploits both holistic templates and local representations, it is based on the collaboration of generative and discriminative modules where feature selection is part of the framework. Moreover, the On-line AdaBoost (OAB)~\cite{OAB} tracker, by using fast computable features like Haar-like wavelets, orientation histograms, local binary patterns, selects the most discriminating features for tracking resulting in stable tracking results. OAB method does both - adjusting to the variations in appearance during tracking and selecting suitable features which can learn object appearance and can discriminate it from the surrounding background.

Semi Boosting Tracker (SBT) and Beyond Semi Boosting Tracker (BSBT)~\cite{BSBT} are multiple object tracking approaches which extend semi-supervised tracking by object specific and adaptive priors. Valuable information which would be ignored in a pure semi-supervised approach is safely included in the prior using a detector for validation and a tracker for sampling. The prior is interpreted as recognizer of the object as similar objects are distinguished. The tracked objects are used to train local detectors to simplify detection in the specific scene. By using feature selection the classifier framework is able to track various objects, even under appearance changes and partial occlusions, in challenging environments.

Finally, the Scale of the Mean-Shift~\cite{SMS} (SMS) is another efficient technique for tracking 2D blobs through an image. Lindeberg's theory of feature scale selection based on local maxima of differential scale-space filters is applied to the problem of selecting kernel scale for mean-shift blob tracking. \vspace{-0.3cm}

\section{Online Feature Selection for Visual Tracking}\label{sec6:OnlineFS}

In this section we analyse a pool of modern feature selection methods with the goal in mind to choose some good candidates among them for a final application in real-time visual tracking.

\subsection{The Adaptive Color Tracking System}\label{sec6:ACTtracker}\vspace{-0.05cm}

The ACT system~\cite{Danelljan_2014_CVPR} is one of the most recent solutions for tracking, which extends the CSK tracker~\cite{Henriques:2012} with color information. ACT exploits color naming (CNs), proposed in ~\cite{VSVL09} (i.e., the action of assigning linguistic color labels to image pixels), to target objects and learn an adaptive correlation filter by mapping multi-channel features into a Gaussian kernel space. Schematically, the ACT tracker contains three improvements to CSK tracker: (i) A temporally consistent scheme for updating the tracking model is applied instead of training the classifier separately on single samples, (ii) CNs are applied for image representation, and (iii) ACT employs a dynamically adaptive scheme for selecting the most important combinations of colors for tracking. 

In the ACT framework, for the current frame $p$, CNs are extracted and used as features for visual tracking. Moreover, a grayscale image patch is preprocessed by multiplying it with a Hann window~\cite{Danelljan_2014_CVPR}, then, the final representation is obtained by stacking the luminance and color channels. 
The ACT algorithm considers all the extracted appearances $x^p$ to estimate the associated covariance matrix $C_p$. A projection matrix $B_p$, with orthonormal column vectors, is calculated by eigenvalue decomposition (EVD) over $C_P$.  Let $x^p_1$ be the $D_1$-dimensional learned appearance. Thus, the projection matrix $B_p$ is used to compute the new $D_2$-dimensional feature map $x^p_2$ of the appearance by the linear mapping $x^p_1(m,n) = B^T_px^p_2(m,n), \forall m,n$. The projection matrix $B_p$ is updated by selecting the $D_2$ normalized eigenvectors of $R_p$ (see Eq.\ref{eq:ATC1}), that corresponds to the largest eigenvalues. 
\begin{equation}\label{eq:ATC1}
	R_p = C_p + \sum^{p-1}_{j=1}B_j \Lambda_jB^T_j,
\end{equation}
where $C_p$ is the covariance matrix of the current appearance and $\Lambda_j$ is a $D_2 \times D_2$ diagonal matrix of weights needed for each basis in $B_p$. Finally, $D_2$ represents the number of dimensions where the $D_1$ features are projected on.

Summarizing, $B_p$ is used to transform the original feature space to yield a subspace by performing dimensionality reduction. Finally, the projected features are used to compute the detection scores and the target position in in the new frame $p+1$ (see ~\cite{Henriques:2012,Danelljan_2014_CVPR} for further details).

\subsection{Embedding Feature Selection}\label{sec6:Embedded}\vspace{-0.05cm}

We present a collection of modern algorithms in Sec.~\ref{sec5:compAppr}, suitable to be embedded on the ACT system. For all of them, the following steps are taken in the embedding procedure. Firstly, the ACT appearances $x$ are extracted for object and background classes, and computed using samples taken from the most recently tracked frame. Secondly, feature ranking is applied in a supervised/unsupervised manner depending on the selection method. This important step can be interpreted as ranking by relevance the dimensions of the feature vector $x$ (10-D), where features in the first ranked positions are tailored to the task of discriminating object from background in the current frame (note: in Section~\ref{sec6:SOT} the first $4$ features have been selected). Finally, the most discriminative features are used to estimate the $C$ covariance matrix used to feed the ACT.
This procedure enables ACT to continually update the set of features used, which turns out to better separate the target from its immediate surroundings.
 



\subsection{Experiments}\vspace{-0.05cm}
\label{sec6:Exp}

\subsubsection{Experiment 1: Classification Task}
\label{sec6:OR}

This section proposes a set of tests on the PASCAL VOC-2007~\cite{pascal-voc-2007} dataset. In object recognition VOC-2007 is a suitable tool for testing models, therefore, we use it as reference benchmark to assess the strengths and weaknesses of using feature selection approaches regarding the classification task while taking care at their execution times. For this reason, we compare seven modern approaches where their details are reported in Table~\ref{table:ch5compmethods44}. 

This experiment considers as features the cues extracted with a deep convolutional neural network architecture (CNN). We selected the recent pre-trained model called GoogLeNet~\cite{googLeNet} (see Section \ref{sec3:needToDeep} for further details), which achieves the new state of the art for classification and detection in the ImageNet Large-Scale Visual Recognition Challenge 2014 (ILSVRC14). We use the 1,024-dimension activations of the last five layers as image descriptors (5,120 features in total). The VOC-2007 edition contains about 10,000 images split into train, validation, and test sets, and labeled with twenty object classes. 
\begin{figure*}[!t]
\centering
\includegraphics[width=0.95\textwidth]{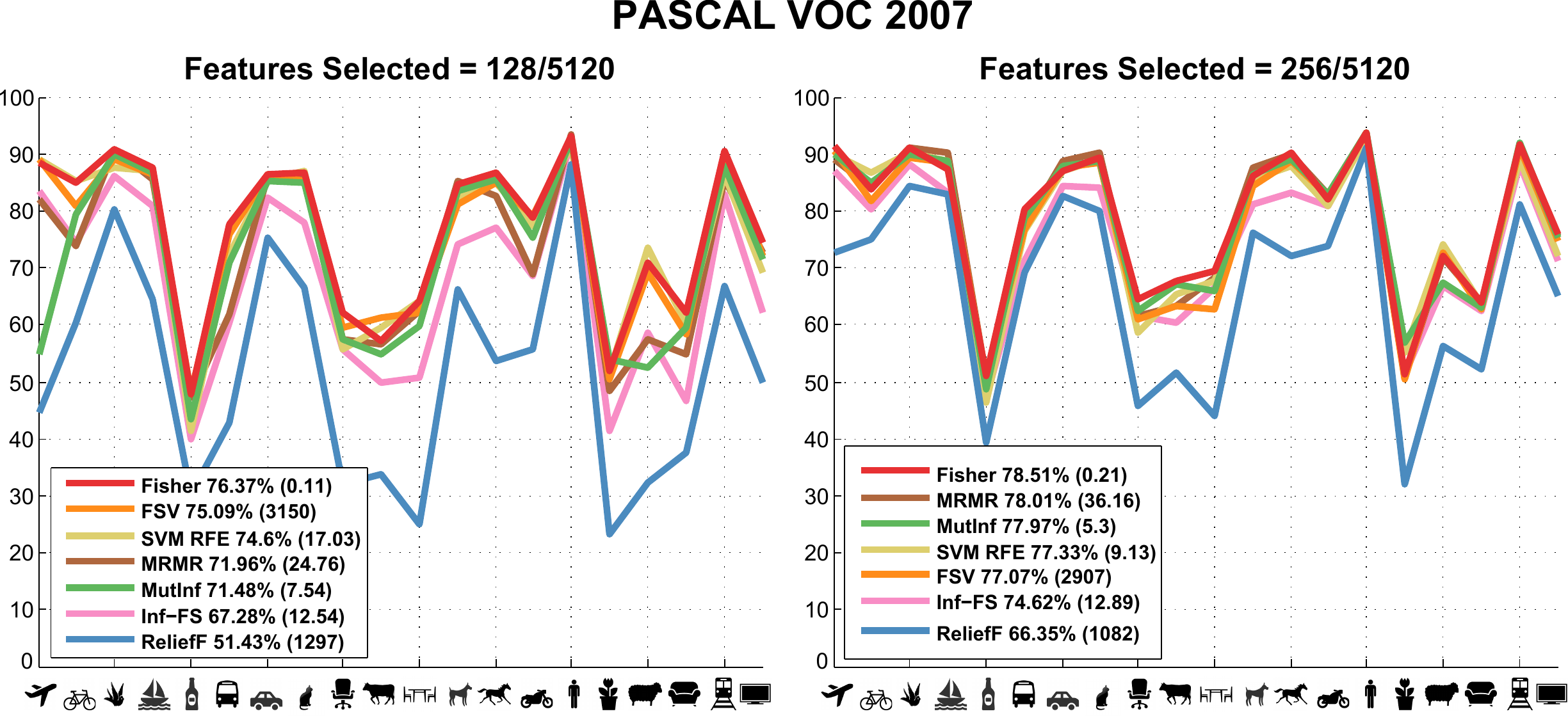}
\caption{Varying the cardinality of the selected features. The image classification results achieved in terms of mean average precision (mAP). The execution times for each method (in seconds) are reported within the brackets, while selecting the first $128$ and $256$ features from the total $5,120$.}
\label{fig6:voc07_256}\vspace{-0.2cm}
\end{figure*}
A one-vs-rest SVM classifier for each class is learnt and evaluated independently and the performance is measured as mean Average Precision (mAP) across all classes. 

Figure~\ref{fig6:voc07_256} serves to analyze and empirically clarify how well important features are ranked high by several feature selection algorithms, selecting 128 features on the left and 256 features on the right from the 5,120 available.
We report as curves the performance in terms of mean average precision (mAP) on the twenty object classes represented by intuitive icons on the abscissa axis. The legends report the mAP in percentage and the average execution time for each method. 
It is worth noting that Fisher method achieved the best performance ($77.44\%$ as mean mAP on the two tests) followed by the wrapper method FSV ($76.08\%$) and the embedded SVM-RFE ($75.97\%$). The quality of the ranking is compared with the execution time of each method. The average time spent by each method to produce the output for each class shows the supremacy of the Fisher approach which thanks to its low computational complexity free of hidden constants completes the task in less than a second ($0.16$ as mean time on the two tests).
The fastest approaches after Fisher, are MutInf ($6.42$), inf-FS ($12.72$), SVM-RFE ($13.08$), and mRMR ($30.46$) that are all under the minute while ReliefF ($1,189.50$) and FSV ($3,028.50$) are not comparable in terms of time.

\subsubsection{Experiment 2: Feature selection for single object tracking}\vspace{-0.05cm}
\label{sec6:SOT}
Taking advantage from the results reported in the previous section, we decided to use the following four candidate methods: \textit{MutInf}, \textit{Fisher}, \textit{Inf-FS}, and \textit{mRMR}. In particular, we take care that execution times of these methods meet the requirements for a real-time application. What is remarkable is that most of these methods achieved good results in terms of average precision for classification and may be examples worth using in tracking. We discarded SVM-RFE because it becomes unstable at some values of the feature filter-out factor ~\cite{mundra2010svm}, i.e., the number of features eliminated in each iteration. We selected the mRMR since in embedded methods, the classifier characteristics such as SVM weights in SVM-RFE provide a criterion to rank features based on their relevancy, but they do not account for the redundancy among the features, while mRMR takes into account both relevancy and redundancy. As a result, our pool consists of $4$ filter methods which evaluate a feature subset by looking at the intrinsic characteristics of data with respect to class labels and do not incorporate classifier operation into the selection process.

In line with Section~\ref{sec6:Embedded}, we embedded these methods on the ACT system obtaining four variants: ACTMI for MutInf, ACTFisher, ACTinffs, and ACT-mRMR. We compare them against the baseline ACT, and also against 29 different state-of-the-art trackers shown to provide excellent results in literature. Some trackers include feature selection within their framework such as the VIVID tracker suite (VR-V, PD-V, RS-V, MS-V, and TM-V),  SCM, OAB, SBT and BSBT, and SMS (see Sec.~\ref{sec6:SoA} for further details). 
Other trackers used for the comparison are the following: CSK~\cite{Henriques:2012}, CPF~\cite{CPF}, Struck~\cite{struck},  CXT~\cite{CXT}, VTD~\cite{VTD}, VTS~\cite{VTS}, LSK~\cite{LSK},  KMS~\cite{KMS}, Frag~\cite{Frag},  MIL~\cite{MIL}, CT~\cite{Zhang:2012}, TLD~\cite{kalal2010pn}, IVT~\cite{ross2008incremental}, DFT~\cite{sevilla2012distribution}, ASLA~\cite{jia2012visual}, L1APG~\cite{bao2012real}, ORIA~\cite{wu2012online}, MTT~\cite{zhang2012robust}, and LOT~\cite{oron2015locally}.\vspace{-0.3cm}

%
\subsubsection{Datasets}\label{sec6:Datasets}\vspace{-0.1cm}

The OTB-50 benchmark is a tracking dataset with 50 fully annotated sequences to facilitate tracking evaluation. The sequences used in our experiments pose challenging situations such as motion blur, illumination changes, scale variation, heavy occlusions, in-plane and out-plane rotations, deformation, out of view, background clutter and low resolution (see Table~\ref{table6:datasets2}). 
\begin{table}[!t]
\small
\centering
\begin{tabular}{ c || l}
 \emph{Attr}& Description \\
\hline \vspace{0.02cm}
1&	Illumination Variation, the illumination in the target region is significantly changed.\\
2& Scale Variation, the ratio of the bounding boxes of the first frame and the current frame\\ 
& is out of the range [$1/t_s$, $t_s$], $t_s > 1$ ($t_s$=2).\\
3& Occlusion, the target is partially or fully occluded.\\
4& Deformation, non-rigid object deformation.\\
5&	Motion Blur, the target region is blurred due to the motion of target or camera.\\
6&	Fast Motion, the motion of the ground truth is larger than tm pixels (tm=20).\\
7&	In-Plane Rotation, the target rotates in the image plane.\\
8&	Out-of-Plane Rotation, the target rotates out of the image plane.\\
9&	Out-of-View, some portion of the target leaves the view.\\
10&Background Clutters, the background near the target has the similar color or texture\\ 
&as the target.\\
11&Low Resolution, the number of pixels inside the ground-truth bounding box is less\\ 
&than tr (tr =400).\\
\end{tabular} 
\caption{The OTB-50 benchmark includes 50 test sequences from recent literatures and 11 attributes, which represents the challenging aspects in visual tracking (table from~\cite{OOT.CVPR.2013}).}
\label{table6:datasets2}
\end{table}\vspace{-0.3cm}

\subsubsection{Results}\label{sec6:res}\vspace{-0.1cm}
\begin{figure}[!t]
\centering
\includegraphics[width=0.98\textwidth]{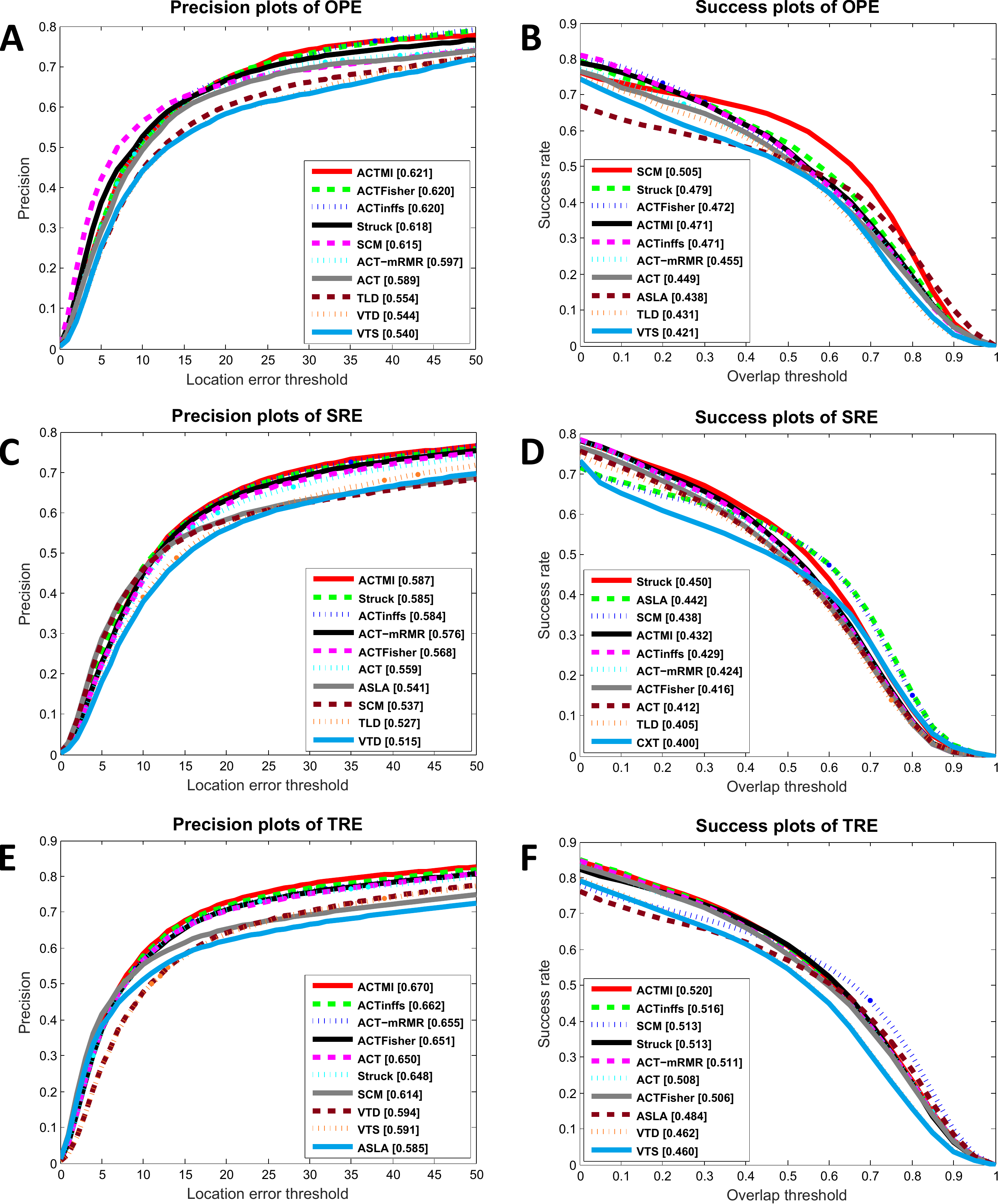}
\caption{Precision and success plots over all 50 sequences provided by the OTB-50 benchmark. Only the top 10 (out of 34) trackers are displayed for clarity. The mean precision and success scores for each tracker are reported in the legend.}
\label{fig6:ALL}\vspace{-0.05cm} 
\end{figure}
Generally, trackers evaluation performance is done by running them throughout a test sequence with the initialization from the ground truth position in the first frame and reporting the average precision and success rate. Therefore, we use the same evaluation methodology used in OTB-50, where precision and success rate have been used for quantitative analysis. The former, is a widely used metric, it is the average Euclidean distance between the center locations of the tracked targets and the manually labeled ground truth. As a result, a precision plot shows the percentage of frames whose estimated locations is within the given threshold distance of the ground truth. 
The latter, success rate, measures the bounding box overlap between the tracked bounding box $r_t$ and the ground truth bounding box $r_a$. The success plot shows the ratio of successful frames as the thresholds varied from 0 to 1. When we consider the initialization from the ground truth in the first frame we refer this as one-pass evaluation (OPE). Figure~\ref{fig6:ALL}.(A-B) shows precision and success plots for OPE criterion. 

Note that ACTMI, ACTFisher and ACTInfFS improve the baseline ACT tracker more than 3\% in mean distance precision, and more than 2\% on bounding box overlaps. Feature selection methods used in this work perform variable rankings, for such a reason we decided to reduce the problem dimensionality by 60\%, so the amount of selected features is up to $4$. Since trackers may be sensitive to the initialization, in \cite{OOT.CVPR.2013} other two ways to analyze a tracker's robustness to initialization have been proposed. These tests are referred as temporal robustness evaluation (TRE) and spatial robustness evaluation (SRE). 
\begin{figure*}[!t]
\centering
\includegraphics[width=0.99\textwidth]{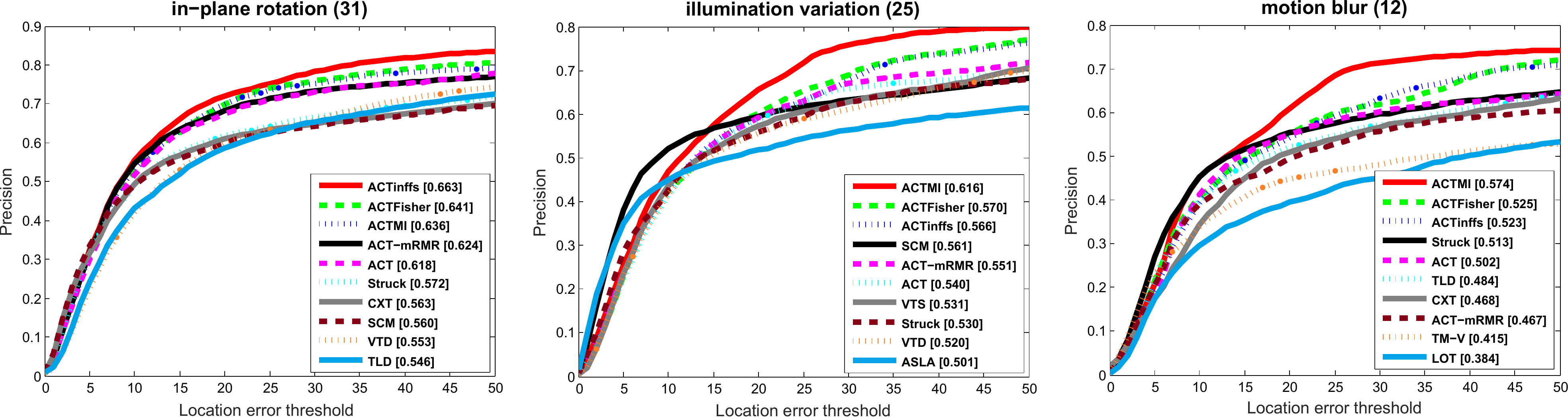}
\caption{This figure presents three challenging tracking problems that illustrate the benefits of combining online feature selection with object tracking.}
\label{fig6:atributesTrack}\vspace{-0.2cm}
\end{figure*}
As for SRE, the initial bounding box in the first frame is sample by shifting or scaling the ground truth. Figure~\ref{fig6:ALL}.(C-D) reports tracking results for the SRE criterion, we used 8 spatial shifts and 4 scale variations. Thus we evaluated each tracker 12 times for SRE. In TRE tests, the tracker is initialized at different $t$ time instances. At each time, the initial bounding box location is set according with the ground truth, then the tracker run to the end of the sequence. We applied perturbations at 20 different times to each tracker TRE. Figure~\ref{fig6:ALL}.(E) shows how ACTMI improves the baseline ACT by 2\% in mean distance precision. In both cases, Figure~\ref{fig6:ALL}.(E-F), ACTMI, ACTinffs, and ACT-mRMR perform favorably to the baseline ACT. Noteworthy, averaging precision scores across the three criteria, ACTMI (62.6\% / 19.0 fps) define in general the new top-score on this benchmark, followed by ACTinffs (62.2\% / 111.4 fps) where both overcome the ACT (59.9\% , 196 fps). As for the other trackers, SCM (58.9\% / 0.5 fps) and Struck (61.7\% / 20.2). As a result, ACTinffs turns out to be the best trade-off between accuracy (62.2\%) and speed (111.4 fps). We say ACTinffs has the same order of magnitude of the ACT in terms of fps.

Figure~\ref{fig6:atributesTrack} shows example precision plots of different attributes. Note, only the top 10 trackers out of 34 are displayed. From left to right: in-plane rotation (comprises 31 sequences), illumination variations (25 sequences) and motion blur (12 sequences). As for in-plane rotation, Figure~\ref{fig6:atributesTrack} shows a scenario where the ACT overcomes the competitors, in such a case feature selection allows to improve its performance by $4.5\%$. Illumination variation and motion blur scenarios represent a more challenging task, where the appearance of the target changes drastically due to extrinsic factors, the ability to identify the most discriminative features for the current frame allows to obtain an improvement in precision up to $7.6\%$ and $7.2\%$ respectively, and at the same time to overcome the methods in comparison.

For all the attributes the ACTMI, ACT-Fisher, and ACTinffs provide superior results compared to existing methods. This is due to the fact that feature selection allows a photometric invariance while preserving discriminative power. Even in those  situations where the baseline ACT does not perform top performance, feature selection permits to overcome the other methods (e.g., for motion blur ACTMI improves the ACT by 7\% in precision). \vspace{-0.4cm}
 

\subsection{Summary}\vspace{-0.3cm}
\label{sec6:Concl} 
In this first part we evaluated a collection of seven modern feature selection approaches, used in off-line settings so far. We investigated the strengths and weaknesses of these algorithms in a classification setting to identify the right candidates for a real-time task. We selected four candidates who meet the requirements of speed and accuracy for visual tracking. Finally, we showed how these feature selection mechanisms can be successfully used for ranking features combined with the ACT system, and, at the same time, maintaining high frame rates (ACTinffs operates at over 110 FPS). Results show that our solutions improve by 3\% up to 7\% their baseline. Moreover, ACTMI resulted in a very impressive performance in precision, providing superior results compared to 29 state-of-the-art tracking methods. We hope that this work motivates researchers to take into account the use of fast feature selection methods as an integral part of their tracker systems. For the sake of repeatability, the code library is posted on the project page (official VOT2016 repository) to provide the material needed to replicate our experiments.

\section{The Visual Object Tracking Challenge (VOT2016)}\label{sec6:VOT16}

In 2013 the Visual object tracking, VOT, initiative was established to address performance evaluation for short-term visual object trackers. The initiative aims at establishing datasets, performance evaluation measures and toolkits as well as creating a platform for discussing evaluation-related issues. Since its emergence in 2013, three workshops and challenges have been carried out in conjunction with the ICCV2013 (VOT2013 \cite{Kristan2013VOT}), ECCV2014 (VOT2014 \cite{Kristan2015}) and ICCV2015 (VOT2015 \cite{Kristan_2015_ICCV_Workshops}). This section discusses the VOT2016 \cite{KristanLMFPCVHL16} challenge, organized in conjunction with the ECCV2016 Visual object tracking workshop, and the results obtained. Like VOT2013, VOT2014 and VOT2015, the VOT2016 challenge considers single-camera, single-target, model-free, causal trackers, applied to short-term tracking. The model-free property means that the only training example is provided by the bounding box in the first frame. The short-term tracking means
that trackers are assumed not to be capable of performing successful re-detection after the target is lost and they are therefore reset after such event. The causality means that the tracker does not use any future frames, or frames prior to reinitialization, to infer the object position in the current frame. 

\subsection{The VOT2016 challenge}

VOT2016 follows VOT2015 challenge and considers the same class of trackers. The dataset and evaluation toolkit are provided by the VOT2016 organizers. The evaluation kit records the output bounding boxes from the tracker, and if it detects tracking failure, re-initializes the tracker. The authors participating in the challenge were required to integrate their tracker into the VOT2016 evaluation kit, which automatically performed a standardized experiment. The results were analyzed by the VOT2016 evaluation methodology. In addition to the VOT reset-based experiment, the toolkit conducted the main OTB experiment in which a tracker is initialized in the first frame and left to track until the end of the sequence without resetting. The performance on this experiment is evaluated by the average overlap measure. 

Participants were expected to submit a single set of results per tracker. Participants who have investigated several trackers submitted a single result per tracker. Changes in the parameters did not constitute a different tracker. The tracker was required to run with fixed parameters on all experiments. The tracking method itself was allowed to internally change specific parameters, but these had to be set automatically by the tracker, e.g., from the image size and the initial size of the bounding box, and were not to be set by detecting a specific test sequence and then selecting the parameters that were hand-tuned to this sequence. The organizers of VOT2016 were allowed to participate in the challenge, but did not compete for the winner of VOT2016 challenge title.

The advances of VOT2016 over VOT2013, VOT2014 and VOT2015 are the following: (i) The ground truth bounding boxes in the VOT2015 dataset have been re-annotated. Each frame in the VOT2015 dataset has been manually perpixel segmented and bounding boxes have been automatically generated from the segmentation masks. (ii) A new methodology was developed for automatic
placement of a bounding box by optimizing a well defined cost function on manually per-pixel segmented images. (iii) The evaluation system from VOT2015 is extended and the bounding box overlap estimation is constrained to image region. The toolkit now supports the OTB no-reset experiment and their main performance measures.

\subsection{The VOT2016 dataset}

The VOT2016 dataset thus contains all 60 sequences from VOT2015, where each sequence is per-frame annotated by the following visual attributes: (i) occlusion, (ii) illumination change, (iii) motion change, (iv) size change, (v) camera motion. In case a particular frame did not
correspond to any of the five attributes, it is denoted as (vi) unassigned. In VOT2015, the rotated bounding boxes have been manually placed in each frame of the sequence by experts and cross checked by several groups for quality control. To enforce a consistency, the annotation rules have been specified. Nevertheless, it has been noticed that human annotators have difficulty following the annotation rules, which makes it impossible to guarantee annotation consistency. For this reason, VOT2016 team has developed a novel approach for dataset annotation. The new approach takes a pixel-wise segmentation of the tracked object and places a bounding box by optimizing a well-defined cost function. For further details on per-frame segmentation mask construction and the new bounding box generation approach we refer to \cite{KristanLMFPCVHL16}. 

\subsection{Performance evaluation methodology}

The OTB-related methodologies evaluate a tracker by initializing it on the first frame and letting it run until the end of the sequence, while the VOT-related methodologies reset the tracker once it drifts off the target. Performance is evaluated in all of these approaches by overlaps between the bounding boxes predicted from the tracker with the ground truth bounding boxes. The OTB initially considered performance evaluation based on object center estimation as well, but the center-based measures are highly brittle and overlap-based measures should be preferred. The OTB introduced a success plot which represents the percentage of frames for which the overlap measure exceeds a threshold, with respect to different thresholds, and developed an ad-hoc performance measure computed as the area under the curve in this plot. This measure remains one of the most widely used measures in tracking papers. It was later analytically proven that the ad-hoc measure is equivalent to the average overlap (AO), which can be computed directly without intermediate success plots, giving the measure a clear interpretation.  The VOT2013 introduced a ranking-based methodology that accounted for statistical significance of the results, which was extended with the tests of practical differences in the VOT2014. The notion of practical differences is unique to the VOT challenges and relates to the uncertainty of the ground truth annotation. The VOT ranking methodology treats each sequence as a competition among the trackers. Trackers are ranked on each sequence and ranks are averaged over all sequences. This is called the sequence normalized ranking. Accuracy-robustness ranking plots were proposed \cite{Kristan2013VOT} to visualize the results. A drawback of the AR-rank plots is that they do not show the absolute performance. In VOT2015 \cite{Kristan2015}, the AR-rank plots were adopted to show the absolute average performance. A high average rank means that a tracker was well-performing in accuracy as well as robustness relative to the other trackers. While ranking converts the accuracy and robustness to equal scales, the averaged rank cannot be interpreted in terms of a concrete tracking application result.  To address this, the VOT2015 introduced a new measure called the expected average overlap (EAO) that combines the raw values of perframe accuracies and failures in a principled manner and has a clear practical interpretation. The EAO measures the expected no-reset overlap of a tracker run on a short-term sequence. The VOT2015 noted that state-of-the-art performance is often misinterpreted as requiring a tracker to score as number one on a benchmark, often leading authors to creatively select sequences and experiments and omit related trackers in scientific papers to reach the apparent top performance. To expose this misconception, the VOT2015 computed the average performance of the participating trackers that were published at top recent conferences. This value is called the VOT2015 state-of-the-art bound and any tracker exceeding this performance on the VOT2015 benchmark should be considered state-of-the-art according to the VOT standards.

In VOT2016, three primary measures are used to analyze tracking performance: accuracy (A), robustness (R) and expected average overlap (AEO). Note, the VOT challenges apply a reset-based methodology. Whenever a tracker predicts a bounding box with zero overlap with the ground truth, a failure is detected and the tracker is re-initialized five frames after the failure. The \textit{accuracy} is the average overlap between the predicted and ground truth bounding boxes during successful tracking periods. On the other hand, the \textit{robustness} measures how many times the tracker loses the target (fails) during tracking. The potential bias due to resets is reduced by ignoring ten frames after re-initialization in the accuracy measure, which is quite a conservative margin. Stochastic trackers are run 15 times on each sequence to obtain reduce the variance of their results. The per-frame accuracy is obtained as an average over these runs. Averaging per-frame accuracies gives per-sequence accuracy, while per-sequence robustness is computed by averaging failure rates over different runs. The third primary measure, called the \textit{expected average overlap} (EAO), is an estimator of the average overlap a tracker is expected to attain on a large collection of short-term sequences with the same visual properties as the given dataset. 

Apart from accuracy, robustness and expected overlaps, the tracking speed is also an important property that indicates practical usefulness of trackers in particular applications. To reduce the influence of hardware, the VOT2014 introduced a new unit for reporting the tracking speed called equivalent filter operations (EFO) that reports the tracker speed in terms of a predefined filtering operation that the toolkit automatically carries out prior to running the experiments. The same tracking speed measure is used in VOT2016.

\subsection{Dynamic Feature Selection Tracker}\label{sec6:votResults}

Together 48 valid entries have been submitted to the VOT2016 challenge. Each submission included the binaries/source code that was used by the VOT2016 committee for results verification. The VOT2016 committee and associates additionally contributed 22 baseline trackers. For these, the default parameters were selected, or, when not available, were set to reasonable values. Thus in
total 70 trackers were tested in the VOT2016 challenge.

We propose an optimized visual tracking algorithm based on the real-time selection of locally and temporally discriminative features. According to the previous section, a novel feature selection mechanism is embedded in the Adaptive Color Names~\cite{Danelljan_2014_CVPR} (ACT) tracking system that adaptively selects the top-ranked discriminative features for tracking. The Dynamic Feature Selection Tracker (DFST) provides a significant gain in accuracy and precision allowing the use of a dynamic set of features that results in an increased system flexibility. Our ranking solution is based on the Inf-FS~\cite{Roffo_2015_ICCV}. The Inf-FS is an unsupervised method, it ranks features according with their ``redundancy" (for further details on the original method see ~\cite{RoffoBMVC2016}). For the sake of foreground/background separation, we propose a supervised variant that is able to score high features with respect to class ``relevancy", that is, how well each feature discriminates between foreground (target) and background. Therefore, we design the input adjacency matrix of the Inf-FS in a supervised manner by significantly reducing the time needed for building the graph. The ACT tracking system does not fit the size of the bounding box. Indeed, in the original framework, the bounding box remains of the same size during the tracking process. We propose a simple yet effective way of adapting the size of the box by using a fast online algorithm for learning dictionaries~\cite{Mairal:2009}. At each update, we use multiple examples around the target (at different positions and scales), we find tight bounding boxes enclosing the target by selecting the one that minimizes the reconstruction error. Thus, we also improved the ACT by adding micro-shift at the predicted position and bounding box adaptation. The interested reader is referred to \cite{roffo2016object} for details.

\subsubsection{The Proposed Method}

The proposed solution can be mainly divided into two parts. Firstly, the DFST tracker ranks the set of features dynamically at each frame. Secondly, it selects a subset and updates the current set of features used for tracking.

The first step of our approach is feature ranking and selection. The ACT system uses a set of $10$ color names for each pixel of the target box, as described in the previous section. Our ranking solution is based on the Inf-FS~\cite{Roffo_2015_ICCV}. Each feature is mapped on an affinity graph, where nodes represent features, and weighted edges the relationships between them. In the original version, the graph is weighted according to a function which takes into account both correlations and standard deviations between feature distributions in an unsupervised manner. The computation of all the weights results to be the bottleneck of this method, at least when the speed is a primary requirement. The inf-FS method is unsupervised, we propose a supervised solution which is able to score high features with respect to how well each feature discriminate between foreground (target) and background. Therefore, we propose a supervised version of the Inf-FS algorithm, where we significantly reduced the time needed for building the graph. Firstly, we labeled all the pixels into the bounding box of the target as positive (+1), then we select the immediate background of the target box and we label each pixel as belonging to the negative class (-1). Given the set of $10$-dimensional samples for classes $Target = C_1$ and $Background = C_2$, and the set of features $F = \{  f^{(1)}, ..., f^{(10)} \}$, where $f^{(i)}$ is the distribution of the $i$-th feature over the samples, we measure the tendency of the $i$-th feature in separating the two classes by considering three different measurements. The Fisher criterion:
 \[
	 p_i = \frac{\left | \mu_{i,1} - \mu_{i,2} \right |^2}{ \sigma_{i,1}^2+\sigma_{i,2}^2}
\]
where $\mu_{i,k}$ and $\sigma_{i,k}$ are the mean and standard deviation, respectively, assumed by the $i$-th feature when considering the samples of the $k$-th class. The higher $p_i$, the more discriminative is the $i$-th feature. As for a second measure for class separation, we perform a t-test of the null hypothesis that foreground samples and background samples are independent random samples from normal distributions with equal means and equal but unknown variances, against the alternative that the means are not equal. We consider a rejection of the null hypothesis at the 5\% significance level. The t-test is the following:
 \[
	 tt_i = \frac{ \mu_{i,1} - \mu_{i,2}  }{ \sqrt{ \frac{\sigma_{i,1}^2}{n_1}  +  \frac{\sigma_{i,2}^2}{n_2} }}
\]
where $n_k$ is the sample size, and $tt_i$ is the p-value, that is, the probability, under the null hypothesis, of observing a value as extreme or more extreme of the test statistic. In our case, the test statistic, under the null hypothesis, has Student's t distribution with $n_1 + n_2 - 2$ degrees of freedom. Finally, we use the Pearson's linear correlation criterion $c_i$. The central hypothesis is that good feature sets contain features that are highly correlated with the class. Since we are interested in identifying \emph{subsets} of features which are maximally discriminative, we continue our analysis by considering pairs of features. To this sake, we firstly create a final vector $s$ by averaging all the previous feature evaluation metrics, then we compute the pairwise matrix $A = s \cdot s^\top$. The generic entry $a_{ij}$ accounts for how much discriminative are the feature $i$ and $j$ when they are jointly considered; at the same time, $a_{ij}$ can be considered as a weight of the edge connecting the nodes $i$ and $j$ of a graph, where the $i$-th node models the $i$-th feature distribution. Under this perspective, the matrix $A$ models a fully connected graph, where paths of arbitrary length can encode the discriminative power of a set of features by simply multiplying the weights of the edges that form them.

The matrix $A$ is fed to the inf-FS approach. Therefore, the Inf-FS evaluates each path on the graph defined by $A$ as a possible selection of features. Letting these paths tend to an infinite length permits the investigation of the relevance of each possible subset of features to class labels. The supervised Inf-FS assigns a final score to each feature of the initial set; where the score is related to how much a given feature is a good candidate for foreground/background separation. Therefore, ranking the outcome of the Inf-FS in descendant order allows us to reduce the number of features and to select only the most relevant for the current frame.   

\subsubsection{Step 2: Embedding Feature Selection}

The supervised Inf-FS algorithm is suitable to be embedded on the ACT system Fig.~\ref{fig6:frames}. Firstly, the ACT appearances are extracted for object $x_{pos}$ and background $x_{neg}$ classes, and computed using samples taken from the most recently tracked frame. Secondly, feature ranking is applied. This important step can be interpreted as ranking by relevance the dimensions of the feature vector $x_{pos}$, where features in the first ranked positions are the ones that better discriminate the object from the background in the current frame. Finally, these features are used to estimate the  covariance matrix $C$ of the ACT, resulting in an impressive improvement of the baseline tracker.
\begin{figure}
\center
\includegraphics[width=0.98\textwidth]{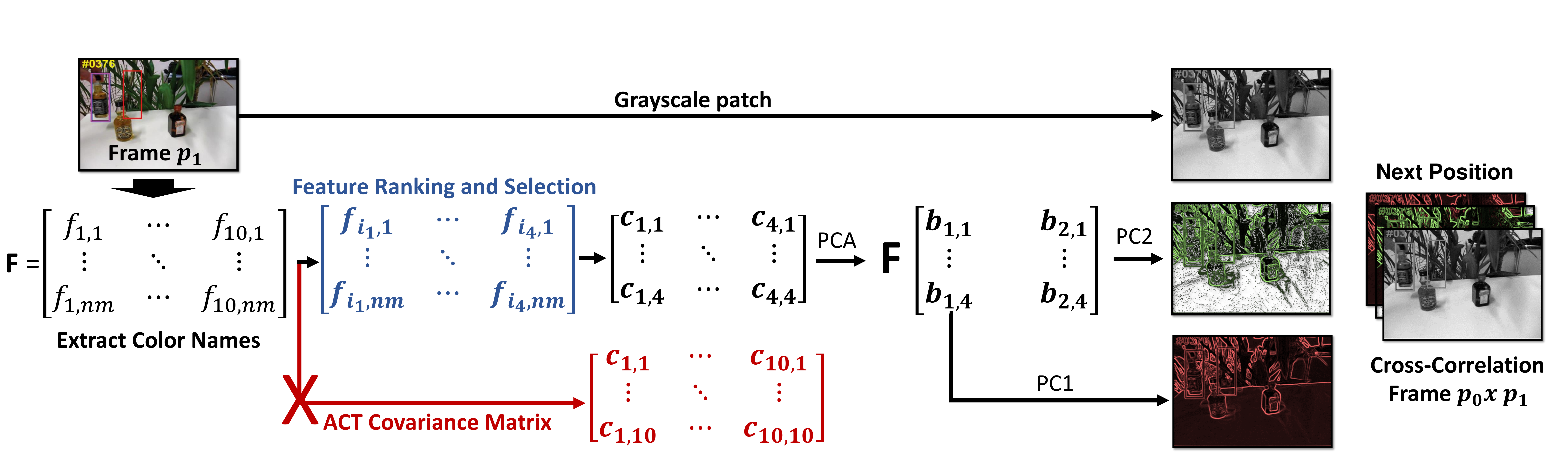}
\caption{Pipeline: Embedding Feature Selection on the ACT System }
\label{fig6:frames}
\end{figure}
Tracking success or failure depends on how a tracking system can efficiently distinguish between the target and its surroundings. ACT uses a fixed set of features that results in a reduced system flexibility. Intrinsic and extrinsic factors affect the target appearance by introducing a strong variability. Our solution allows to deal with drastic changes in appearance and to maintain good execution time (i.e., high frame rates). The ACT tracker does not take into account the appearance of the background, thus, does not appropriately manage the presence of strong distractors. ACT enhanced by feature selection penalizes features that produces spatially-correlated background clutter or distractors. The Inf-FS also allows to deal with strong illumination variation, by selecting those features which remain different when affected by the light.

\subsubsection{Experimental environment}

During the experiments, the tracker was using neither GPU nor any kind of distributed processing.\newline

  \begin{itemize}
    \item The platform used is a PCWIN64 (Microsoft Windows 10), on an Intel i7-4770 CPU 3.4GHz 64-bit, 16.0 GB of RAM, using MATLAB ver.2015a.
    \item The processing speed measured by the evaluation kit is up to $16.53$ (baseline), $5.06$ (unsupervised).
    \item  Parameters: we set the learning rate for the appearance model update scheme up to 0.005 which gives more importance to the appearance learnt in the first frames. The extra area surrounding the target box (padding) is automatically estimated according with the ration between the frame size and target bounding box size. We also reduce the learning rate for the adaptive dimensionality reduction to 0.1. The number of selected features is up to 8, and the dimensionality of the compressed features is 4. As for the dictionary learning, DFST learns a dictionary with 250 elements, by using at most 200 iterations. 
  \end{itemize}
 
 \subsection{Results}
 
\begin{figure}
\center
\includegraphics[width=0.98\textwidth]{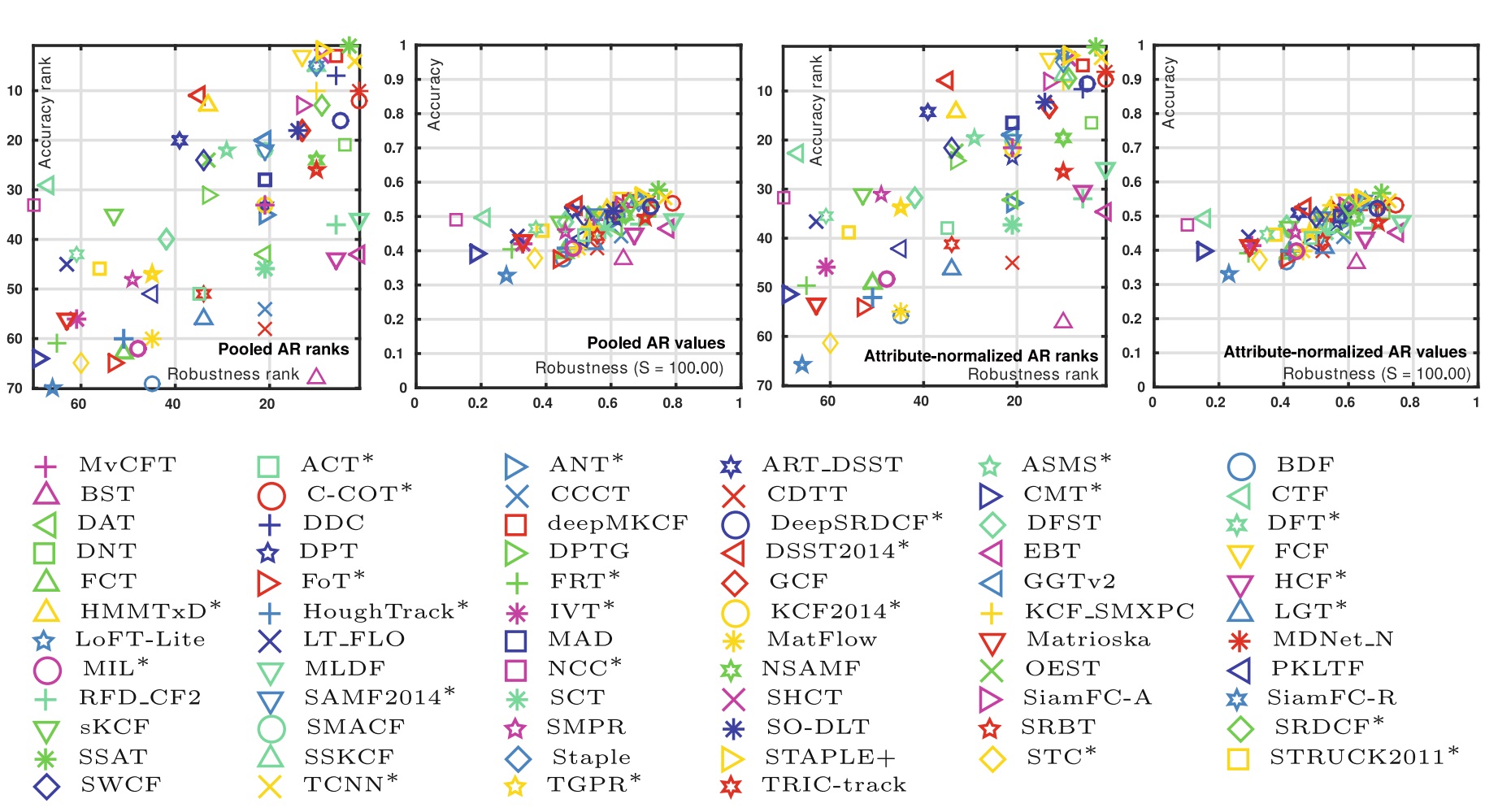}
\caption{The AR-rank plots and AR-raw plots generated by sequence pooling (left) and attribute normalization (right).}
\label{fig6:resultsVOT0}
\end{figure}

\begin{figure}
\center
\includegraphics[width=0.98\textwidth]{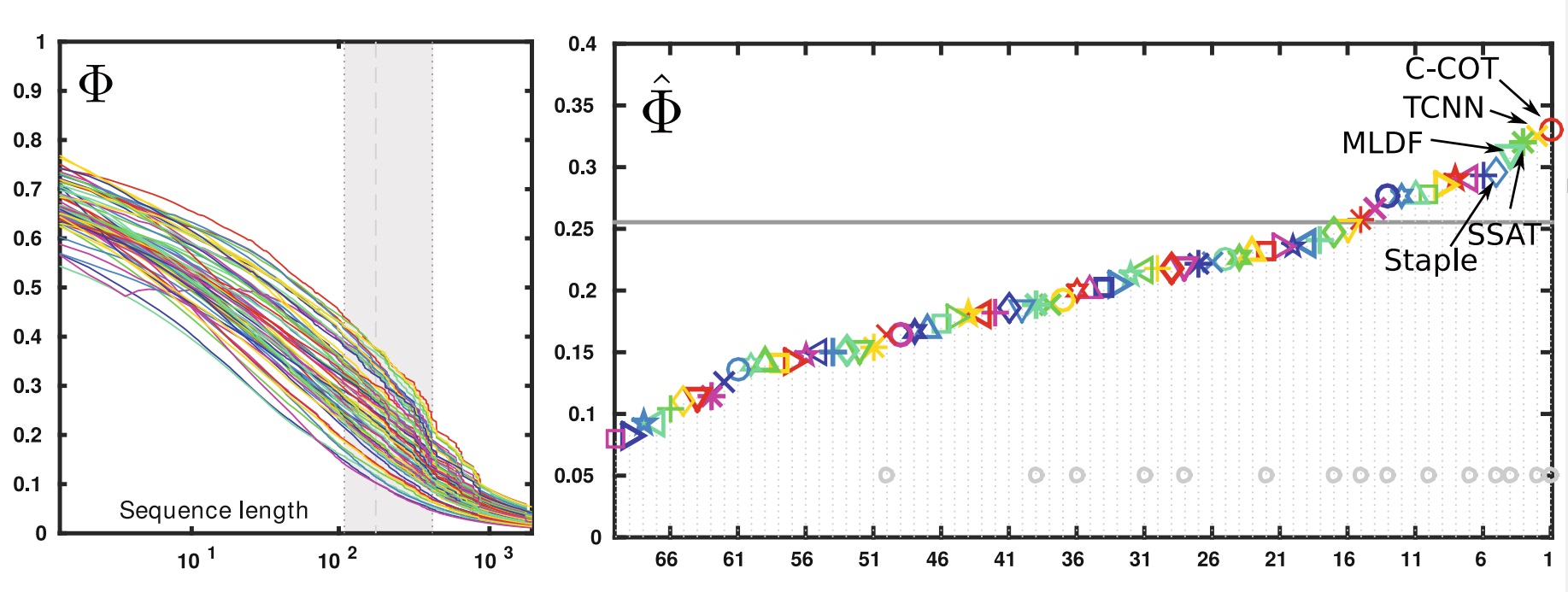}
\caption{Expected average overlap curve (left) and expected average overlap graph (right) with trackers ranked from right to left. The right-most tracker is the top-performing according to the VOT2016 expected average overlap values. See Fig. \ref{fig6:resultsVOT0} for legend. The dashed horizontal line denotes the average performance of fourteen state-of-the-art trackers published in 2015 and 2016 at major computer vision venues. These trackers are denoted by gray circle in the bottom part of the graph.}
\label{fig6:resultsVOT02}
\end{figure}

 The results are summarized in sequence-pooled and attribute-normalized AR-raw plots in Fig. \ref{fig6:resultsVOT0}. The sequence-pooled AR-rank plot is obtained by concatenating the results from all sequences and creating a single rank list, while the attribute-normalized AR-rank plot is created by ranking the trackers over each attribute and averaging the rank lists. The AR-raw plots were constructed in similar fashion. The expected average overlap curves and expected average overlap scores are shown in Fig. \ref{fig6:resultsVOT02}. The raw values for the sequence-pooled results and the average overlap scores are also given in Table \ref{fig6:resultsVOT1}.

It is worth pointing out some EAO results appear to contradict AR-raw measures at a first glance. For example, the Staple obtains a higher EAO measure than Staple+, even though the Staple achieves a slightly better average accuracy and in fact improves on Staple by two failures, indicating a greater robustness. The reason is that the failures early on in the sequences globally contribute more to penalty than the failures that occur at the end of the sequence. For example, if a tracker fails once and is re-initialized in the sequence, it generates two sub-sequences for computing the overlap measure at sequence length N. The first sub-sequence ends with the failure and will contribute to any sequence length N since zero overlaps are added after the failure. But the second sub-sequence ends with the sequence end and zeros cannot be added after that point. Thus the second sub-sequence only contributes to the overlap computations for sequence lengths N smaller than its length. This means that re-inits very close to the sequence end (tens of frames) do not affect the EAO.
 \begin{figure}
\center
\includegraphics[width=0.98\textwidth]{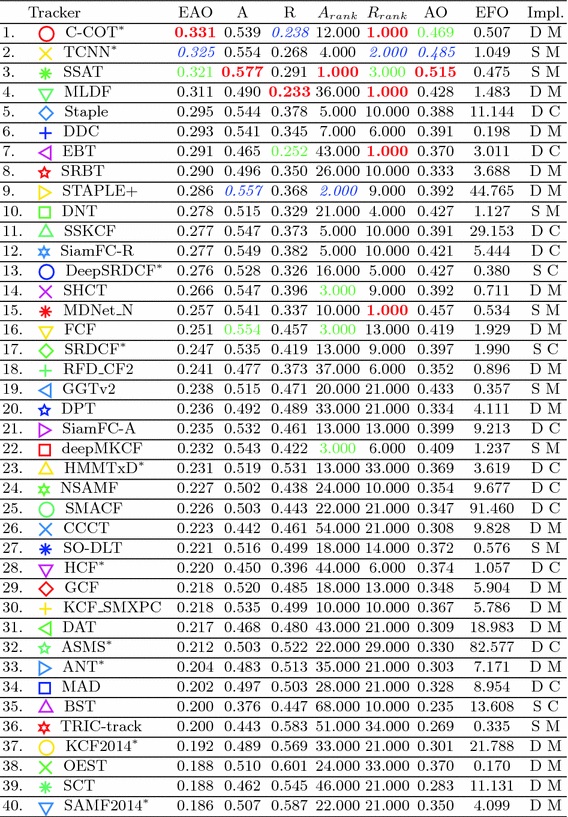}
\end{figure}
 \begin{figure}
\center
\includegraphics[width=0.98\textwidth]{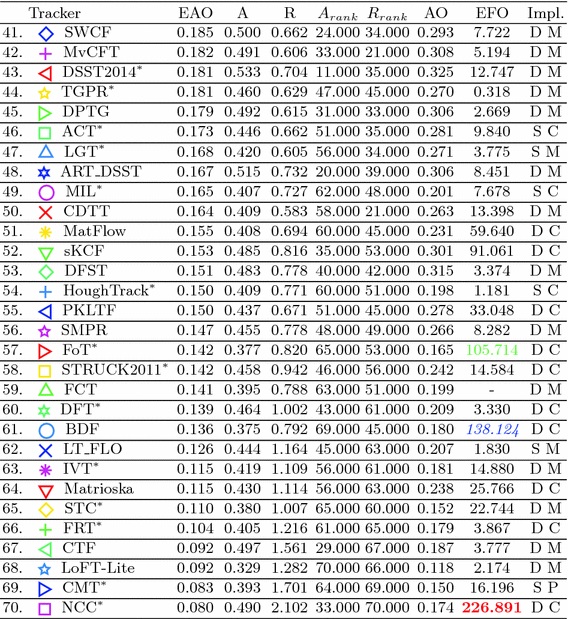}
\caption{The table shows expected average overlap (EAO), accuracy and robustness raw values (A,R) and ranks ($A_{rank}$,$A_{rank}$), the no-reset average overlap AO, the speed (in EFO units) and implementation details (M is Matlab, C is C or C++, P is Python). Trackers marked with * have been verified by the VOT2015 committee. A dash ``-" indicates the EFO measurements were invalid.}
\label{fig6:resultsVOT1}
\end{figure}
Note that the trackers that are usually used as baselines, i.e., MIL, and IVT are positioned at the lower part of the AR-plots and the EAO ranks, which indicates that majority of submitted trackers are considered state-of-the-art. In fact, fourteen tested trackers have been recently (in 2015 and 2016) published at major computer vision conferences and journals. These trackers are indicated in Fig. \ref{fig6:resultsVOT1}, along with the average state-of-the-art performance computed from the average performance of these trackers, which constitutes a very strict VOT2016 state-of-the-art bound. Approximately 22\%  of submitted trackers exceed this bound.

\subsection{Summary}

This last part reviewed the VOT2016 challenge and its results. The challenge contains an annotated dataset of sixty sequences in which targets are denoted by rotated bounding boxes to aid a precise analysis of the tracking results. All the sequences are the same as in the VOT2015 challenge and the per-frame visual attributes are the same as well. A new methodology was developed to automatically place the bounding boxes in each frame by optimizing a well-defined cost function. In addition, a rule-of-thumb approach was developed to estimate the uniqueness of the automatically placed bounding boxes under the expected bound on the per-pixel annotation error. A set of 70 trackers have been evaluated. A large percentage of trackers submitted have been published at recent conferences and top journals, including ICCV, CVPR, TIP and TPAMI, and some trackers have not yet been published (available at arXiv). For example, fourteen trackers alone have been published at major computer vision venues in 2015 and 2016 so far.

The results of VOT2016 indicate that the top performing tracker of the challenge according to the EAO score is the C-COT tracker. This is a correlation-filter-based tracker that applies a number of state-of-the-art features. The tracker performed very well in accuracy as well as robustness and trade-off between the two is reflected in the EAO. The C-COT tracker is closely followed by TCNN and SSAT which are close in terms of accuracy, robustness and the EAO. These trackers come from a different class, they are pure CNN trackers based on the winning tracker of VOT2015, the MDNet. It is impossible to conclusively decide whether the improvements of C-COT over other top-performing trackers come from the features or the approach. Nevertheless, results of top trackers conclusively show that features play a significant role in the final performance. All trackers that scored the top EAO perform below real-time. Among the realtime trackers, the top performing trackers were Staple+ and SSKCF that implement a simple combination of the correlation filter output and histogram backprojection.

The main goal of VOT is establishing a community-based common platform for discussion of tracking performance evaluation and contributing to the tracking community with verified annotated datasets, performance measures and evaluation toolkits.  As our further contribution to the visual tracking community, we posted our tracker online to the VOT page (\url{http://www.votchallenge.net/vot2016/trackers.html}). 

\part{Learning to Rank}\label{part:trhee}

\chapter{Biometric Identification and Verification}\label{ch7:CHATS}

Biometric technologies are used to secure several electronic communications, including access control and attendance, computer and enterprise network control, financial and health services, government, law enforcement, and telecommunications. A biometric system consists of two different phases: the enrollment phase and authentication/verification phase. During the enrollment phase user biometric data is acquired, processed and stored as reference file in a database. This is treated as a template for future use by the system in subsequent authentication operations. In order to authenticate a user, at least two verifications are required. Indeed, verification is the process of \textit{verifying} if the information provided is accurate. Verification does not verify the actual user (or their identity), just the information provided. For example, in the information security world, user authentication is performed by verifying \textit{username} and \textit{password} of a user, which turns out to determine whether a user is, in fact, who they are declared to be. In Biometric Authentication (BA) a user is uniquely identified through one or more distinguishing biological traits, such as fingerprints, hand geometry, earlobe geometry, retina and iris patterns, voice waves, keystroke dynamics, DNA and signatures. 

Among the many types of biometric authentication technologies this chapter focuses on \textit{keystroke biometrics}. An interesting example of application is given by the new authentication mechanism added to \textit{Coursera}, a social entrepreneurship company that partners with universities to offer free courses online. It introduced a new feature aimed at verifying online student identity with their typing behavior. This typing measurement, called keystroke dynamics, is the detailed timing information that describes exactly when each key was pressed and when it was released as a person types on a keyboard. Keystroke biometrics \cite{Deng2013,A2004,Shepherd1995} refers to the art and science of recognizing an individual based on an analysis of his typing patterns. 
The concept of keystroke biometrics has arisen as a hot topic of research only in the past two decades. Researchers at MIT \cite{lau2004enhanced} looked at the idea of authentication through keystroke biometrics in 2004 and identified a few major advantages and disadvantages to the use of this biometric for authentication. For example, in \cite{lau2004enhanced} the authors conclude that measuring keystroke dynamics is an accessible and unobtrusive biometric as it requires very little hardware besides a keyboard. Also, as each keystroke is captured entirely by key pressed and press time, data can be transmitted over low bandwidth connections. The authors also identified disadvantages to the use of keystroke dynamics as an authentication tool. First of all, typing patterns can be erratic and inconsistent as something like cramped muscles and sweaty hands can change a person's typing pattern significantly. Also, they found that typing patterns vary based on the type of keyboard being used, which could significantly complicate verification.

To overcome these disadvantages, it is necessary to reinforce the set of features by adding other different cues related to the writing style of the user, known as \emph{stylometric} features. Stylometric cues have been introduced in literature in the Authorship Attribution (AA) domain. AA is the science of recognizing the author of a piece of text, analyzing features that measure the style of writing. Five groups of writing traits have been proposed in literature that focus on \emph{lexical, syntactic, structural, content-specific and idiosyncratic} aspects of a document \cite{AbbW:2008}. Earlier computer-aided AA attempts focused on textbooks, exploiting mainly lexical cues (statistical measures of lexical variation as character/word frequency) and syntactic features (punctuation, function words) \cite{Holm:1998}.
Later on, the diffusion of Internet had a huge impact on the AA community, delivering novel authorship challenges, since online texts (emails, web pages, blogs) exhibit distinctive qualities: structured layout, diverse fonts etc. To this aim, structural features were introduced \cite{DeVelACM:2001}. Content-specific and idiosyncratic cues mine the text through topic models and grammar checking tools, unveiling deliberate stylistic choices \cite{Argamon:2009}. Table~\ref{table2:FeatTaxonomy} is a synopsis of the features applied so far in the literature.

In the last years, the rise of the social web requested a dramatic update of the stylometry, especially concerning one of its most relevant communication channel, i.e., the chat or Instant Messaging (IM). Standard stylometric features have been employed to categorize the content of
a chat \cite{Oreb:2009} or the behavior of the participants \cite{ZhouProc:2004}; AA on chats (i.e., recognizing each participant of a dialog) is still at its infancy. This is probably due to the peculiar nature of the instant messaging, which mixes together multimedia communication aspects: a very short text, whose lexicon is unstable, that follows turn-taking mechanisms inherited from the spoken language realm.

In this chapter we focus on biometric identification and verification on text chat conversations. We start with proposing a set of stylometric features aimed at measuring the writing style of a user involved into a chat session. Then we move on Keystroke biometrics, where we enrich the feature set by adding timing information which describes exactly when each key was pressed and when it was released as a person is typing at a computer keyboard. Experiments have been performed on the datasets already introduced in Part \ref{part:one} Chapter \ref{ch4:DATASETS}.

\section{A Novel Set of Stylometry Features}

In this section, we propose a pool of novel stylometric markers, especially suited for the AA of dyadic chats. Such features are privacy-preserving, ignoring the semantic content of the messages. In particular, some of them can be fitted in the taxonomy reported in Table~\ref{table2:FeatTaxonomy}, while some other require the definition of a novel group; in facts, they are based on the concept of \emph{turn-taking}, encoding aspects that characterize the spoken exchanges. All the features are calculated considering the \emph{turn} as most significant chunk of information, instead of working on the overall conversation, resembling the analysis performed on conversational corpora. In addition, the features are complementary and minimal, in a feature selection sense; finally, they are expressive, ensuring a compelling AA performance on the $C$-$Skype$ dataset (see section \ref{ch4:skype}).

We focus on a set of $N$ subjects, each one involved in a dyadic exchange with another single individual. For each person involved in a conversation, we examine his stream of turns (suppose $T$), completely ignoring the input of the other subject. This implies that we assume the chat style (as modeled by our features) independent from the interlocutor: this has been validated experimentally. From these data, a personal signature is extracted, that is composed by different cues: some of them could be associated to particular classes of the taxonomy of Table~\ref{table2:FeatTaxonomy}, while some others need a new categorization, being them tightly connected with the "verbal'' nature of a chat. Therefore, we define a new class of ``turn-taking'' features. In all the cases, it is very important to note that in standard AA approaches, the features are counted over entire conversations, obtaining a single quantity. In our case, we consider the turn as a basic analysis unit, obtaining $T$ numbers for each feature. For ethical and privacy isues, we decide to discard whatever cue which involves the content of the conversation.

In the following, we will list the proposed features: in cursive bold, we indicate the novel features\footnote{In some sense, all the features are novel, since they are collected on turns instead of the whole text; still here we want to highlight ``structurally'' novel features.}, together with a brief explanation of them.

\paragraph{\bf{Lexical Features}}\vspace{-0.4cm}
\begin{itemize}
\itemsep0em
\item Number of words, chars, mean word length, number of uppercase letters;
\item {\bf Number of Uppercase / Number of Chars}; usually, entire words written in capital letters indicate a strong emotional message. This feature records such communicative tendency.

\item {\bf n-order Length Transitions (noLT)}; These features resemble the n-grams of \cite{Grauman}; the strong difference here is in the fact that we consider solely the length of the words, and not their content. In practice, for a noLT of order $n=1$ (1oLT), we build probability transition matrices that in the entry $i,j$, $1 \leq i,j \leq I$, exhibit the probability of moving from a word of length $i$ to a word of length $j$. In our case, we set $I=15$. noLT of order $n=2$ (2oLT) are modeled by transition matrices of $I^3$. We did not take into account superior order, for sparsity issues.
\end{itemize}

\paragraph{\bf{Syntactic Features}}\vspace{-0.4cm}
\begin{itemize}
\itemsep0em
\item Number of ? and ! marks, three points (...), generic marks (",.:*;), rate of emoticons / words, rate of emoticons / chars;
\end{itemize}

\paragraph{\bf{Turn-taking Features}}\vspace{-0.3cm}
\begin{itemize}
\itemsep0em
\item {\bf Turn duration}; the time spent to complete a turn (in seconds);
this feature accounts for the rhythm of the conversation with faster exchanges typically corresponding to higher engagement.
\item {\bf Writing speed}; number of typed characters -or words- per second (typing rate); these two features indicate whether the duration of a turn is simply due to the amount of information typed (higher typing rates) or to cognitive load (low typing rate), i.e. to the need of thinking about what to write

\item {\bf Emoticons Category \emph{Positive}, \emph{Negative}, \emph{Other}}; these features aim at individuating a particular mood expressed in a turn through emoticons. In particular, we partition 101 diverse emoticons in three classes, portraying positive emotions (happiness, love, intimacy, etc. $-$ 20 emot.), negative emotions (fear, anger, etc. $-$ 19 emot.), and neutral emoticons (portraying actions, objects etc. $-$ 62 emot.), counting their total number of occurrences. We are conscious that our partition is somewhat subjective: still, our attempt was to discover whether emotion-oriented classes of emoticons were more expressive than a unique class, reporting all the possible emoticons. Experimentally, our choice lead to higher recognition performance.
\item {\bf Mimicry}; ratio between number of chars -or words- in current turn and number of chars -or words- in previous turn; this feature models the tendency of a subject to follow the conversation style of the interlocutor (at least for what concerns the length of
the turns). The mimicry accounts for the social attitude of the subjects.
\item {\bf Answer Time}; this feature is the time spent to answer a question in the previous turn of another interlocutor.
\end{itemize}

These quantities are extracted from each turn, as written above, with the exception of the mean word length, the noLT feature, the Emoticons Category: actually, in such cases, the turn does not offer sufficient statistics for a robust description. Therefore, for these cues, we consider all the turns of a subject as they were a unique corpus. Conversely, for all the other cues, we have $T$ numbers; these numbers are then described employing histograms. On our data, we noted that most of the features extracted are strongly collapsed toward small numeric values: for this reason, we adopt exponential histograms, where small-sized bin ranges are located toward zero, increasing their sizes while going to higher numbers. Experimentally, we get much better results than exploiting uniformly binned histograms over the whole range of the features.
For the sake of clarity, the features are numbered in Table~ \ref{table7:Featsx}, reporting also their minimum and maximum values.
\begin{table*}
\begin{center}
\resizebox{0.55\textwidth}{!}{%
\begin{tabular}{|l|l|l|l|l|}
\hline
ID & Name  & Range & nAUC & Rank\\
\hline
1 & \#Words(W)& [0,1706] &75.6\% & 5\\\hline
2 & \#Chars(C)& [0,15920] & 77.3\%& 2\\\hline
3 &  Mean Word Length & [0,11968]& 74.2\%& 7\\\hline
4 & \#Uppercase letters&[0,11968]& 70.7\%& 14\\\hline
5 & \textit{\textbf{\#Uppercase / C}}&[0,1] & 71.7\%& 12\\\hline
6 &  \textit{\textbf{1o\_LT}}&[0,127]& 76.1\%& 4\\\hline
7 &  \textit{\textbf{2o\_LT}}&[0,127]& 70.0\%& 15\\\hline
8 & \# ? and ! marks&[0,21] & 58.8\%& 21\\\hline
9 &  \#Three points (...)&[0,54]& 71.4\%&  13\\\hline
10 &  \#Marks (",.:*;)&[0,1377]& 83.1\%& 1\\\hline
11 &  \textit{\textbf{\#Emoticons / W}}&[0,4]& 77.0\%&  3\\\hline
12 &  \textit{\textbf{\#Emoticons / C}}&[0,1]& 75.0\%& 6\\\hline
13 & \textit{\textbf{Turn Duration}}&[0,1800]& 72.5\%& 11\\\hline
14 & \textit{\textbf{Word Writing Speed}}&[0,562] & 72.9\%& 9\\\hline
15 & \textit{\textbf{ Char Writing Speed}}&[0,5214]& 72.9\%& 10\\\hline
16 &  \textit{\textbf{\#Emo. Pos.}}&[0,48]& 73.0\%& 8\\\hline
17 & \textit{\textbf{\#Emo. Neg.}}&[0,5]& 62.8\%& 17\\\hline
18 &  \textit{\textbf{\#Emo. Oth.}}&[0,20]& 61.2\%& 19\\\hline
19 &  \textit{\textbf{Imitation Rate / C}}&[0,2611]& 65.2\%&  16\\\hline
20 &  \textit{\textbf{Imitation Rate / W}}&[0,1128]& 62.9\%& 18\\\hline
21 &  \textit{\textbf{Answer Time}}&[0,2393]& 59.8\%& 20\\
\hline
\end{tabular}}
\end{center}
\caption{Stylometric features used in the experiments and recognition statistics.}
\end{table*}\label{table7:Featsx}

\subsection{Matching personal descriptions}\label{sec7:Match}\vspace{-0.2cm}
Let us suppose to have collected the features for two subjects, $A$ and $B$. We now have to exploit them for obtaining a single distance, describing the overall similarity between $A$ and $B$. As first step, we derive a plausible distance for each feature separately: in the case we have histograms, we employ the Bhattacharyya distance. For the features represented by mean values, we adopt the Euclidean distance. In the case of the $noLT$ features, we consider the \emph{diffusion distance} \cite{Ling&Okada06cvpr}, which acts similarly to the Pyramid Matching Kernel \cite{Grauman}. In practice, the diffusion distance measures the linear distance among the matrices' entries, applying iteratively ($L$ times) Gaussian kernels of increasing variance: this allows to include cross-entries relations in the final measure,  thus alleviating sparsity problems as well as quantization effects. Briefly speaking, given $M_A$ and $M_B$ the $noLT$ matrices, the diffusion distance $K(M_A,M_B)$ is
\begin{equation}
K(M_A,M_B)=sum_{l=0}^L|d_l(\mathbf{x})|
\end{equation}
where
\begin{eqnarray}
d_0(\mathbf{x}) &=& M_A(\mathbf{x}) - M_B(\mathbf{x})\\
d_l(\mathbf{x}) &=& \left[ d_{l-1}(\mathbf{x})\ast \phi(\mathbf{x},\sigma)\right]\downarrow_2 \;\; l=1,\ldots,L
\end{eqnarray}
with $\mathbf{x}$ the elements of a matrix, with dimension $I\times I$; $\phi(\mathbf{x},\sigma)$ is the 2D Gaussian filter of standard deviation $\sigma$; $L$ indicates the number of levels employed, and $\downarrow_2$ denotes half size downsampling. The parameter $\sigma=4$ and the level $L=4$ have been set by crossvalidation.

Since the aim of this work is explorative on the nature on the features, and not how to fuse them, we do not investigate how such features should be combined together. Therefore, in this work, we adopt a simple average rule, \emph{i.e.}, the final distance is obtained by averaging over the contribute of the single distances, opportunely normalized between 0 and 1.

\subsection{Testing the Stylometric Features}\label{Sec7:Expx}

In the experiments, we evaluate the effectiveness of each feature in performing identity recognition; subsequently, we analyze how the compound of all the features does the same task; finally we consider the identity verification task. In this work, we work on the $C$-$Skype$ corpus made of $94$ dyadic chat conversations collected with Skype. The number of turns per subject ranges between $200$ and $1000$. Hence, the experiments are performed over $110$ turns of each person. The turns of each subject are split into \emph{probe} and \emph{gallery} set, each including $55$ samples. In this way, any bias due to differences in the amount of available material should be avoided.  When possible, we pick different turns selections (maintaining their chronological order) in order to generate different probe/gallery partitions.

The first part of the experiments aims at assessing each feature independently, as a simple ID signature. A particular feature of a single subject is extracted from the probe set, and matched against the corresponding gallery features of all subjects, employing a given metrics. This happens for all the probe subjects, resulting in a $N\times N$ distance matrix. Ranking in ascending order the $N$ distances for each probe element allows one to compute the \emph{Cumulative Match Characteristic} (CMC) curve, i.e., the expectation of finding the correct match in the top $n$ positions of the ranking.

The CMC is an effective performance measure for AA approaches \cite{Bolle:2003}, and in our case is a valid measure for evaluating the task of \emph{identity recognition}: given a test sample, we want to discover its identity among a set of $N$ subjects.
In particular, the value of the CMC curve at position $1$ is the probability that the probe ID signature of a subject is closer to the gallery ID signature of the same subject than to any other gallery ID signature; the value of the CMC curve at position $n$ is the probability of finding the correct match in the first $n$ ranked positions.

\begin{figure}[hbt]
    \centerline{
      \includegraphics[width=0.8\linewidth]{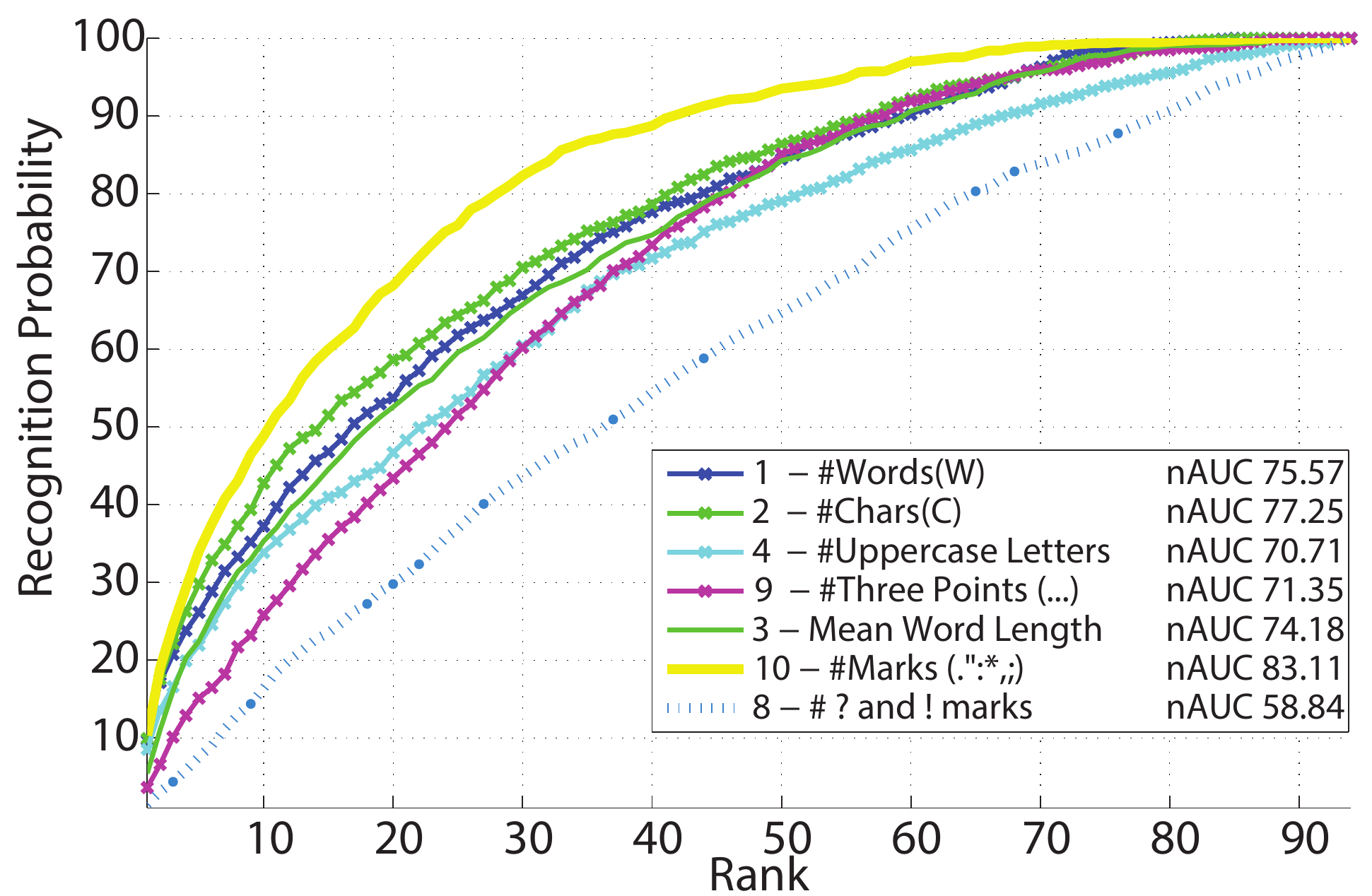}}\vspace{-0.05cm}
      \caption{CMC curve for each ``classical'' feature.}\label{Fig7:FeaturesAnalysisClassical}
\end{figure}

\begin{figure}[hbt]
    \centerline{
      \includegraphics[width=0.8\linewidth]{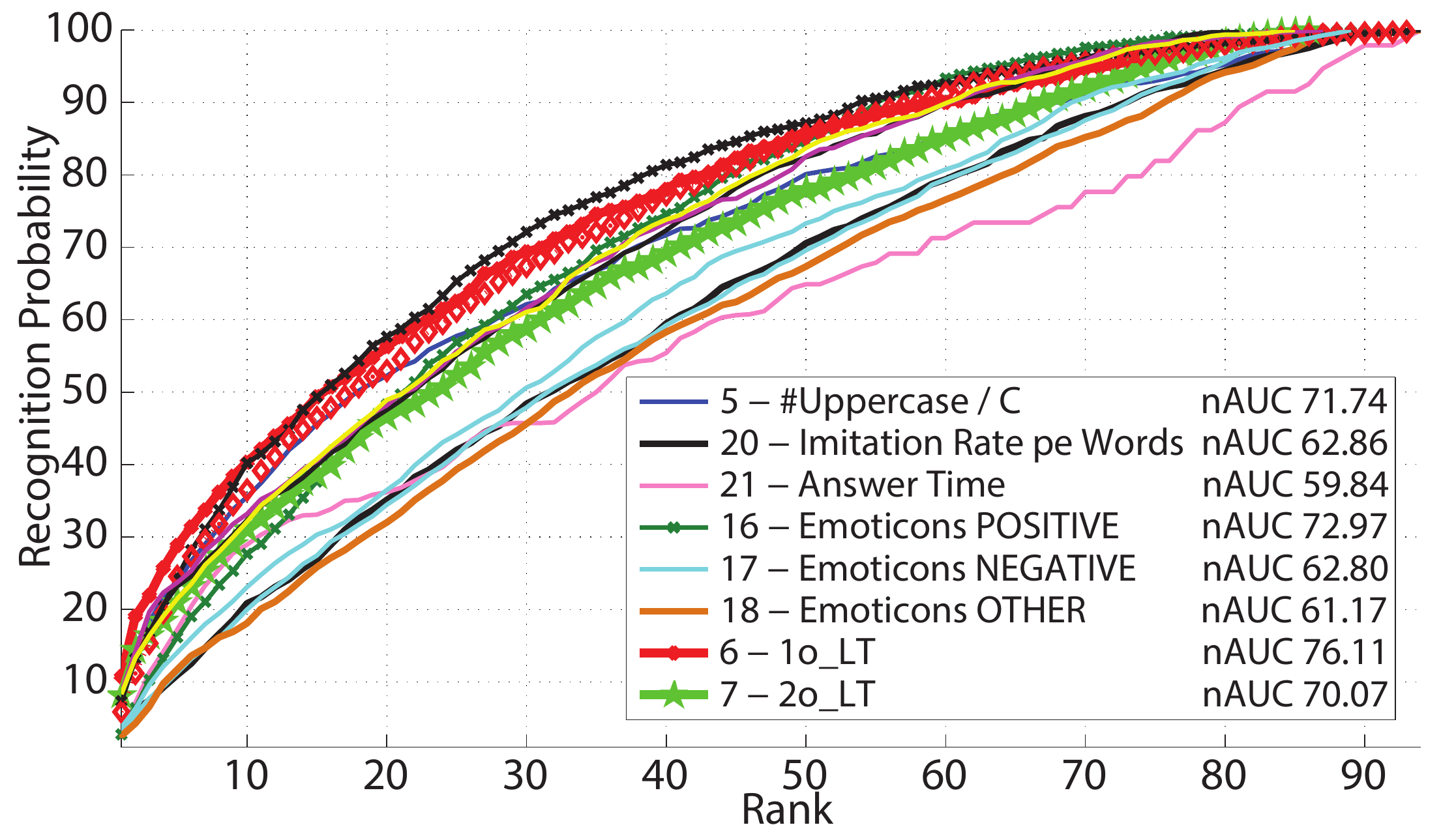}}\vspace{-0.05cm}
      \caption{CMC curve for each novel feature.}\label{Fig7:FeaturesAnalysisNew}
\end{figure}

\begin{figure}[hbt]
    \centerline{
      \includegraphics[width=0.8\linewidth]{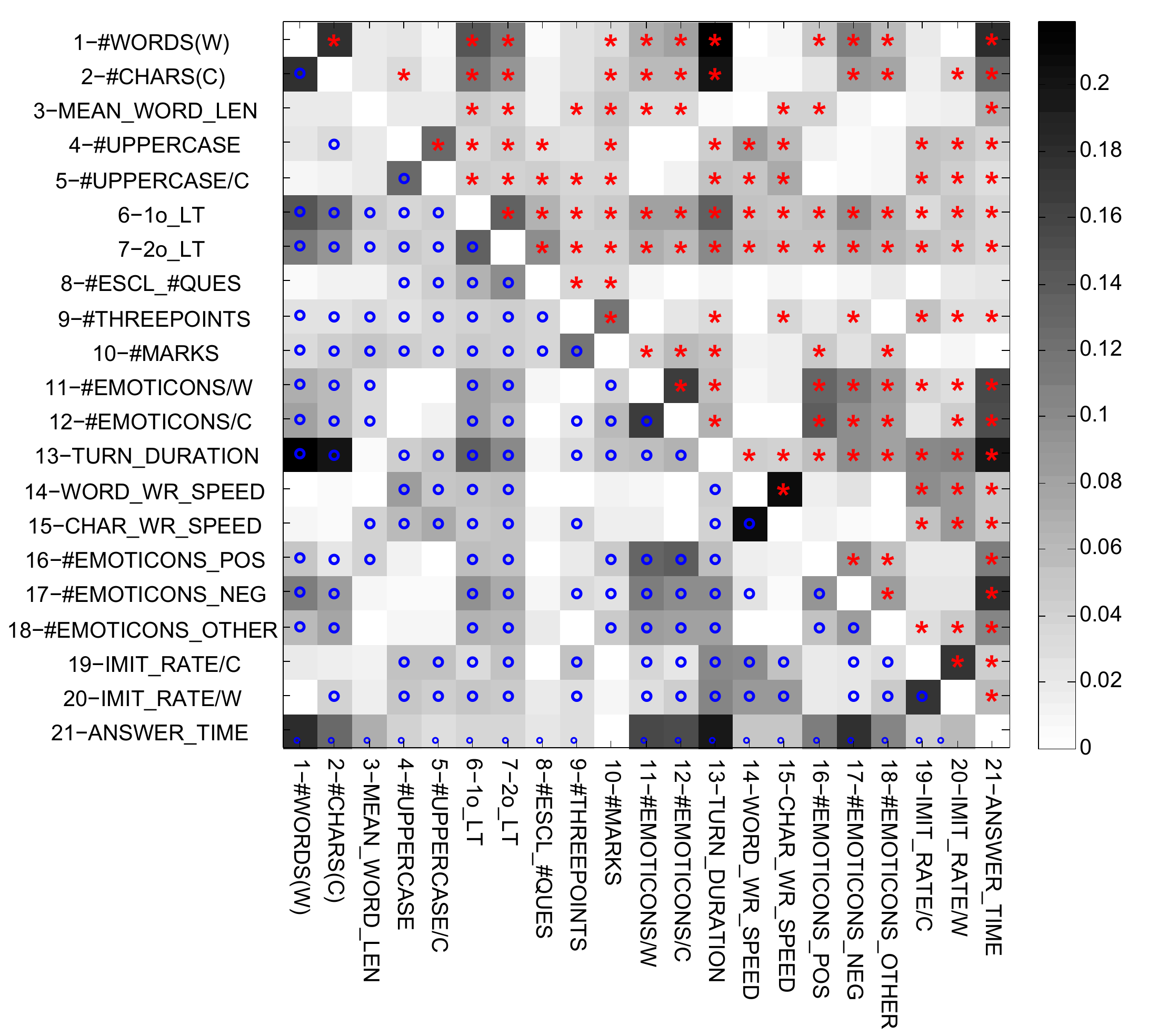}}
      \caption{Correlation analysis between features: red asteriscs report correlations significant with p-value$<1\%$, blue dots correlations with p-value$<5\%$.}\label{Fig7:CorrelationAnalysis}
\end{figure}

Given the CMC curve for each feature (obtained by averaging on 10 trials), the normalized Area Under Curve (nAUC) is
calculated as a measure of accuracy. For the sake of clarity, the features are partitioned in two sets: those resembling the classical AA features, now calculated on turns (Fig.~\ref{Fig7:FeaturesAnalysisClassical}) and the novel ones (Fig.~\ref{Fig7:FeaturesAnalysisNew}). As visible, all our features gives performances above the chance: in Table~\ref{table7:Featsx}, last two columns, are reported the nAUC score and the rank built over the nAUC score.
In order to understand the information contained in all the features, and how they are interrelated, we calculate the Spearman's rank correlation coefficient (see Fig.~\ref{Fig7:CorrelationAnalysis}), highlighting in the upper triangular part statistically significant correlations with p-value$<1\%$, in the bottom triangular part those correlations significant at p-value$<5\%$.
In general, a high level of correlation is existent between features. Quite interesting, $noLT$ features seem to be correlated with all the other cues.

\subsubsection{Identity Recognition and Verification by all the features}\vspace{-0.2cm}

\begin{figure}[hbt]
    \centerline{
      \includegraphics[width=1\linewidth]{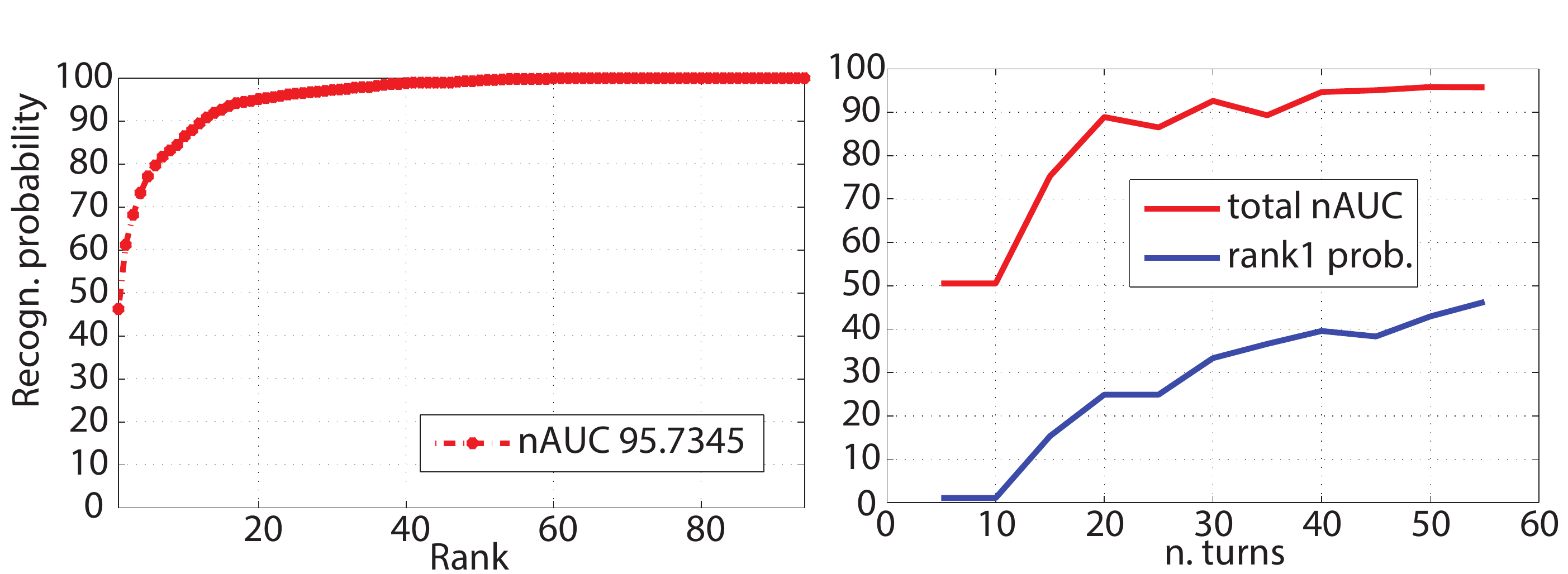}}
      \caption{(left) Global CMC Curve; (right) nAUC and rank1 probability varying the number of turns.}\label{Fig7:overallCMC}
\end{figure}

In this section, we put together all the proposed features as described previously. In Fig.~\ref{Fig7:overallCMC} is reported the CMC curve obtained by joining all the distances, which gives the identity recognition performance.
This curve is strongly superior than all the cues taken separately, realizing an nAUC of 0.9573. This witnesses that, even if the features are strongly correlated, they model highly complementary information. In facts, adopting standard feature selection strategies, as Forward Feature Selection, gives that all the features do increase the recognition rate. It is worth noting that, the probability of guessing the correct user at rank 1 is slighty below 50\% which is quite encouraging (actually, in standard people re-identification tasks, where the features are the images of people, performances with a similar number of subject into play is quite inferior).
\begin{figure}[hbt]
    \centerline{
      \includegraphics[width=0.8\linewidth]{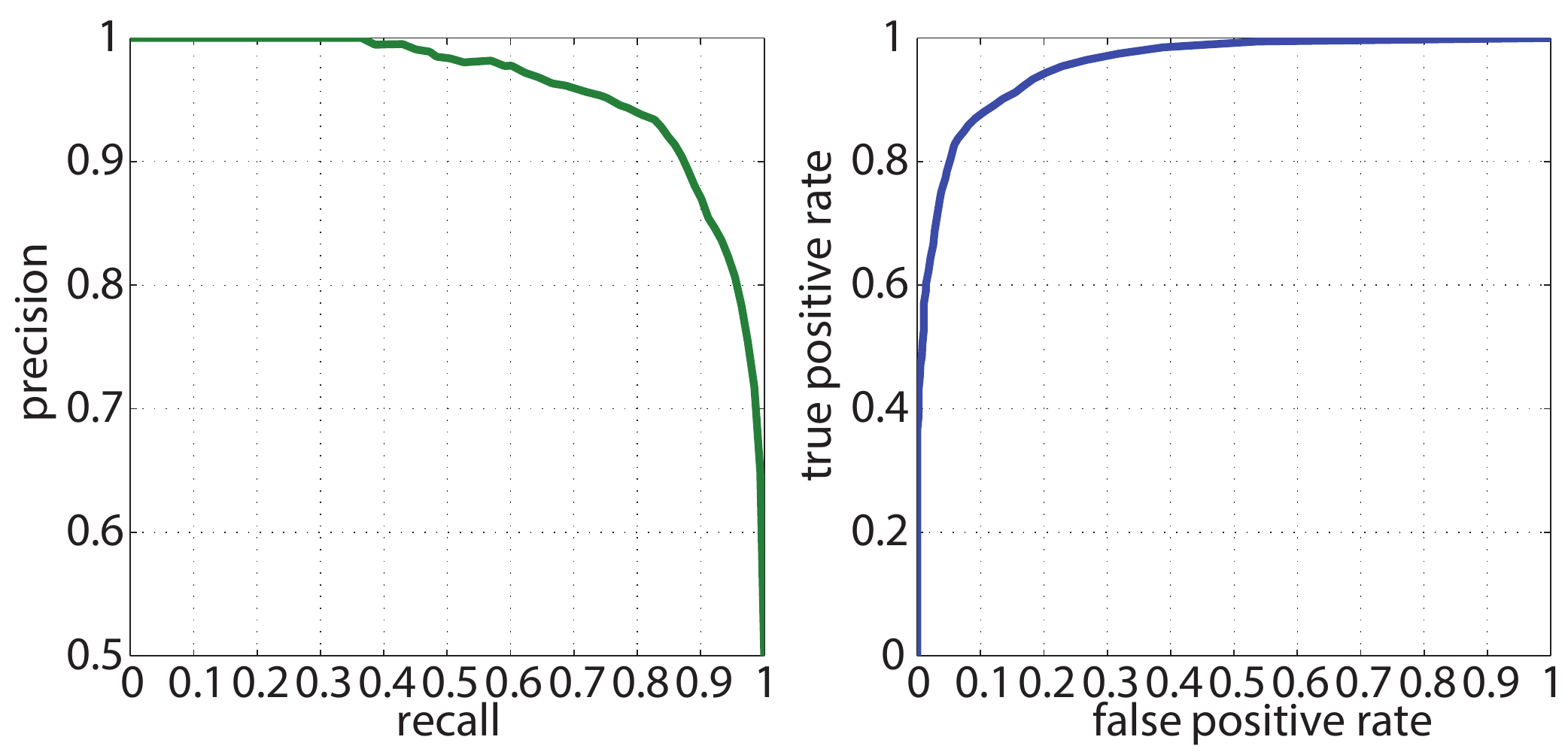}}
      \caption{Precision, Recall and ROC Curve.}\label{Fig7:ROC}
\end{figure}
To investigate how important is the number of turns taken into account for modeling the gallery and probe subjects, we show in Fig.~\ref{Fig7:overallCMC} the nUAC of the CMC curve and the rank1 probability by varying the number of turns. Intuitively, the higher the number of turns, the higher the recognition rate.

Considering the verification task, we adopt the following strategy: given the signature of user $i$, if it matches with the right gallery signature with a matching distance which is ranked below the rank $K$, it is verified. Intuitively, there is a tradeoff in choosing $K$. A high K (for example, 50) gives a 100\% of true positive rate (this is obvious by looking at the global CMC - Fig.~\ref{Fig7:overallCMC}), but it brings in a lot of potential false positives. Therefore, taking into account the number $K$ as varying threshold, we can build ROC and precision/recall curves, portrayed in Fig.\ref{Fig7:ROC}. Considering the nAUC of both the curves, we get 0.9566 and 0.9351, respectively. The best compromise between precision and recall is obtained calculating the F1 value, which gives 0.88 for precision 0.90 and recall 0.87, corresponding to the value of $K=45$.

\subsubsection{Summary}\label{sec7:Conc}\vspace{-0.2cm}
We presented two contributions to the young literature of the AA on IM by applying a \textit{Social Signal Processing} approach to data \cite{Vinciarelli_JIVC_2009,burgoon2017social}. The first is a pool of novel turn-taking based features, imported from the analysis of the spoken dialogs, which characterize the non-verbal behavior of a IM participant while she/he is conversing. The second is about the usage of the turn as key entity where the features have to be extracted: this appears extremely convenient, confirming that exploiting the parallelism with the analysis of spoken dialogs is fruitful. 

We disclosed a new facet for biometrics, considering the chat content as personal blueprint. From tens of turns, we extracted heterogeneous features, which take from the Authorship Attribution and the Conversational Analysis background. On a test set of 94 people, we demonstrate that identification and verification can be performed definitely above the chance; even if our performance are far from letting our strategy be immediately embedded into commercial systems, many improvements may be done.
First of all, fusion strategies for collapsing intelligently the cues should be investigated. Secondly, learning policies should be taken into account, which are expected to boost the accuracy in a consistent fashion. 

\theoremstyle{plain}
\newtheorem{theorem}{Theorem}
\newtheorem{proposition}{Proposition}
\newtheorem{lemma}{Lemma}
\newtheorem{corollary}{Corollary}
\theoremstyle{remark}
\newtheorem{remark}{Remark}
\newtheorem{example}{Example}
\theoremstyle{definition}
\newtheorem{definition}{Definition}
\newcommand{\mapto}{\ensuremath{\rightarrow}}
\newcommand{\argmax}{\operatornamewithlimits{arg\,max}}
\newcommand{\argmin}{\operatornamewithlimits{arg\,min}}
\newcommand{\approach}{\ensuremath{\rightarrow}}
\newcommand{\equivalent}{\ensuremath{\Longleftrightarrow}}

\renewcommand{\c}{\mathbf c}
\renewcommand{\a}{\mathbf a}
\renewcommand{\b}{\mathbf b}

\newcommand{\f}{\mathbf f}
\newcommand{\x}{\mathbf x}
\newcommand{\z}{\mathbf z}
\newcommand{\y}{\mathbf y}
\newcommand{\w}{\mathbf w}
\newcommand{\C}{{\mathbb C}}
\newcommand{\R}{{\mathbb R}}
\newcommand{\N}{{\mathbb N}}
\newcommand{\Z}{{\mathbb Z}}
\newcommand{\F}{{\mathbb F}}
\renewcommand{\d}{{\partial}}
\newcommand{\W}{{\mathcal{Y}}}
\renewcommand{\H}{{\mathcal{H}}}
\newcommand{\X}{{\mathcal{X}}}
\newcommand{\Y}{{\mathcal{Y}}}

\newcommand{\B}{\mathbf{B}}

\newcommand{\la}{\langle}
\newcommand{\ra}{\rangle}

\def\diag{{\rm diag}}
\def\diam{{\rm diam}}
\def\det{{\rm det}}
\def\argmin{{\rm argmin}}

\def\rank{{\rm rank}}
\def\cond{{\rm cond}}

\def\supp{{\rm supp}}
\def\sinc{{\rm sinc}}

\def\Im{{\rm Im}}

\def\proj{{\rm proj}}

\def\trace{{\rm tr}}
\def\loc{{\rm loc}}
\def\vec{{\rm vec}}

\newcommand{\e}{{e}}
\renewcommand{\v}{{v}}
\section{Learning to Rank People in Chat: Recognition and Verification}

This section presents a learning approach which boosts the performance of user recognition and verification, allowing to recognize a subject with considerable accuracy. The proposed method is based on a recent framework of one-shot multi-class multi-view learning, based on Reproducing Kernel Hilbert Spaces (RKHS) theory. During the learning stage, training conversations of different subjects are collected to form the gallery set. The feature descriptors of each individual are extracted from the related conversations (\emph{i.e.}, conversation in which they are involved), forming the user signature for that individual. Then, the similarity between the descriptors is computed for each feature by means of kernel matrices. Multi-view learning consists of estimating the parameters of the model given the training set. Given a probe signature, the testing phase consists of computing the similarity of each descriptor with the training samples and using the learned parameter to classify it.

\subsection{Features and Kernels}\label{ssec7:descr}

In this experimental section, we examine the $C$-$Skype$ dataset. The dataset is made of $94$ users, we retain only 78 users from it, those users that have at least $3$ conversations with at least $10$ turns each (note, it is needed for the multi-view learning approach). The assumption used to split a full text chat in more conversations is that the time interval that elapses between each pair of successive turns does not exceed 30 minutes. For each person involved in a conversation, we
analyze 
his stream of turns (suppose $T$), 
ignoring the input of the other subject. This
means
that we assume that the chat style (as modeled by our features) is independent from the interlocutor - this assumption
has been validated experimentally.
From these data, a personal signature is extracted, that is composed
of
different cues, written in red in Table~\ref{table7:FeatTaxonomy2}.
\begin{table*}[h]
\centering
  \resizebox{0.9\textwidth}{!}{%
\begin{tabular}{|m{2.5cm}|m{3.7cm}|m{9.5cm}|m{3.2cm}|}
\hline

\textbf{Group}  & \textbf{Description} & \textbf{Examples} & \textbf{References}\\\hline


\hline
\multirow{6}{*}{\textbf{Lexical}}
& \color{red}\emph{Word level} & Total number of words (=M), \# short words/M, \# chars in words/C,  \# different words, chars per word, freq. of stop words &  \cite{AbaS:2008,Iqbal:2011,Oreb:2009,Stamatatos_2009,Zheng:2006}\\

 & \color{red} \emph{Character level} & Total number of characters (chars) (=C), \# uppercase chars/C, \# lowercase chars/C, \# digit chars/C, freq. of letters, freq. of special chars & \cite{ AbaS:2008,  Oreb:2009, Stamatatos_2009, Zheng:2006}\\

 & Character|Digit n-grams& Count of letter|digit n-gram (a, at, ath, 1 , 12 , 123) &\cite{ AbaS:2008, Stamatatos_2009, Zheng:2006}\\

 & \color{red}\emph{Word-length distribution} & Histograms, average word length & \cite{ AbaS:2008, Iqbal:2011, Oreb:2009, Stamatatos_2009, Zheng:2006} \\									
 & Vocabulary richness & Hapax legomena, dislegomena & \cite{AbaS:2008,  Iqbal:2011,   Stamatatos_2009, Zheng:2006}\\

&  \color{red}\emph{Length n-grams}& Considers solely the length of the words; $xo\_LT$ is the length n-gram of order $x$. & \cite{Cristani:Skype:AVSS:2013} \\ \hline

\multirow{2}{*}{\textbf{Syntactic}}

 & Function words & Frequency of function words (of,  for,  to ) &\cite{AbaS:2008,  Iqbal:2011, Oreb:2009, Stamatatos_2009, Zheng:2006}\\

 & \color{red}\emph{Punctuation} & Occurrence of punctuation marks (!, ?, : ), multiple !|? &\cite{ AbaS:2008,   Iqbal:2011,    Oreb:2009, Stamatatos_2009, Zheng:2006}\\

 & \color{red}\emph{Emoticons|Acronym} & :-), L8R, Msg, :( , LOL; emoticons categories such as \emph{Positive} that counts the occurrences of happiness, love, intimacy, etc. icons (20 emot. types in total) ; \emph{Negative}: address fear, anger, etc. (19 emot. types in total); and \emph{Other} or neutral emoticons portray actions, objects etc. (62 emot. types in total)  &\cite{ Oreb:2009, Stamatatos_2009, Cristani:Skype:AVSS:2013}\\\hline

\textbf{Structural} & Message level & Has greetings, farewell, signature &\cite{ AbaS:2008, Iqbal:2011,  Oreb:2009,  Stamatatos_2009, Zheng:2006} \\
\hline

\textbf{Content-specific}  &Word n-grams & Bags of word, agreement (ok, yeah, wow),  discourse markers| onomatopee (ohh), \# stop words, \# abbreviations, gender|age-based words,  slang words &\cite{ AbaS:2008,  Iqbal:2011, Oreb:2009,  Stamatatos_2009, Zheng:2006} \\\hline

\textbf{Idiosyncratic} &Misspelled word& Belveier instead of believer &\cite{ AbaS:2008, Iqbal:2011,   Oreb:2009, Stamatatos_2009}\\ \hline	

\multirow{7}{*}{\textbf{Turn-taking}}

& \color{red}\emph{Turn duration} & Time spent to complete a turn (in seconds);   & \cite{Cristani:Skype:AVSS:2013} \\

& \color{red}\emph{Writing speed} & Number of typed characters or words per second; & \cite{Cristani:Skype:AVSS:2013} \\

& \color{red}\emph{Answer Time} &Time spent to answer a question in the previous turn of another interlocutor & \cite{Cristani:Skype:AVSS:2013} \\

&\color{red} \emph{Mimicry}  & Ratio between number of chars -or words- in current turn and number of chars -or words- in previous turn of the opposite subject;& \cite{Cristani:Skype:AVSS:2013} \\

\hline
\end{tabular}} \vspace{-0.02cm}
  \caption{Synopsis of the state-of-the-art features for AA on chats. ``\#" stands for ``number of''. In red we have the features that we used in our approach (best viewed in colors).}
  \label{table7:FeatTaxonomy2}
\end{table*}
Our approach differs from the other standard AA approaches, where the features are counted over entire conversations,
giving
a single quantity.
We consider the turn as a basic analysis unit, obtaining $T$ numbers for each feature. For ethical and privacy issues, we
discard
any
cue which involves the content of the conversation. Even if this choice is very constraining, because it prunes out many features of Table~\ref{table2:FeatTaxonomy}, the results obtained are very encouraging.

Given the descriptor, we extract a kernel from each feature. In particular, we used $\chi^2$ kernels that they have been proved to perform well in practice in different applications.

\subsection{Multi-view Learning} \label{ssec7:formulation}
In this section, we briefly summarize the multi-view learning framework proposed in \cite{Minh:ICML13}, with particular focus on user recognition in chats.
We suppose to have for training a labeled gallery set  $\{(x_i,y_i)\}$,
where $x_i \in \X$ represents the $i$-th signature of the user with label (identity) $y_i \in \Y$.

Given that $P$ is the number of identities in the re-identification problem, let the output space be $\Y = \R^P$.
Each output vector $y_i \in \Y$, $1 \leq i \leq l$, has the form $y_i = (-1, \ldots,1, \ldots, -1)$, with $1$ at the $p$-th location if $x_i$ is in the $p$-th class.
Let $m$ be the number of views/features and $\mathcal{W} = \Y^m = \R^{Pm}$.

We define user recognition as the following optimization problem based on the least square loss function:
\begin{eqnarray}\label{equation7:vector-leastsquare-general}\nonumber
f^{\star} = \argmin_{f \in \H_K} \frac{1}{l}\sum_{i=1}^l||y_i - Cf(x_i)||^2_{\mathcal{Y}}\\
 + \gamma_A||f||^2_{\H_K} + \gamma_I \la \f, M\f\ra_{\mathcal{W}^{l}},
\end{eqnarray}
where
\begin{itemize}
 \item $f$ is a vector-valued function in an RKHS $\H_K$ that is induced by the matrix-valued kernel $K:\X \times \X \mapto \R^{Pm \times Pm}$, with $K(x,t)$ being a matrix of size $Pm \times Pm$ for each pair $(x,t) \in \X \times \X$,

 \item $f(x) = (f^1(x), \ldots, f^m(x))$, where $f^i(x) \in \R^P$ is the value corresponding to the $i$th view,
 \item $\f = (f(x_1), \ldots, f(x_{l}))$ as a column vector in $\mathcal{W}^l$,
 \item $C$ is the combintation operator that fuses the different views as $Cf(x) = \frac{1}{m}(f^1(x) + \cdots + f^m(x)) \in \R^P$,
 \item $\gamma_A > 0$ and $\gamma_I \geq 0$ are the regularization parameters,
 \item $M$ is defined as $M = I_{l} \otimes (M_m \otimes I_P)$, where $M_m = mI_m - \e_m \e_m^T$ \cite{Minh:ICML13}.
\end{itemize}

The first term of Eq.~\ref{equation7:vector-leastsquare-general} is the least square loss function that measures the error between the estimated output $Cf(x_i)$ for the input $x_i$ with the given output $y_i$ for each $i$. Given an instance $x$ with $m$ features, $f(x)$ represents the output values from all the features, constructed by their corresponding hypothesis spaces, that are combined through the combination operator $C$.
The second term is the standard RKHS regularization term.
The last term is the multi-view manifold regularization \cite{Minh:ICML13}, that performs consistency regularization across different features.

The solution of the minimization problem of Eq.~\ref{equation7:vector-leastsquare-general} is unique \cite{Minh:ICML13}: $f^{\star}(x) = \sum_{i=1}^{l}K(x,x_i) \, a_i$, where the vectors $a_i$ are given by the following system of equations:
\begin{equation}\label{equation7:linear-matrix}
(\mathbf{C^{*}C}K[\x] + l \gamma_I MK[\x] + l \gamma_A I)\a = \mathbf{C^{*}}{\y},
\end{equation}
where $\a = (a_1, \ldots, a_{l})$ is a column vector in $\mathcal{W}^{l}$ and ${\y} = ({y}_1, \ldots, {y}_{l})$ is a column vector in $\mathcal{Y}^{l}$.
Here $K[\x]$ denotes the $l \times l$ block matrix whose $(i,j)$ block is $K(x_i,x_j)$;

$\mathbf{C^{*}C}$ is the $l \times l$ block diagonal matrix, with each diagonal block being $C^{*}C$;  $\mathbf{C^{*}}$ is the $l \times l$ block diagonal matrix, with each diagonal block being $C^{*}$.

Assume that each input $x$ is decomposed into $m$ different views, $x$ = ($x^1$, $\ldots$, $x^m$). For
our setting,
the matrix-valued kernel
$K(x,t)$ is defined as a block diagonal matrix,
with the $(i,i)$-th block given by
\begin{equation}
K(x,t)_{i,i} = k^i(x^i, t^i)I_{P},
\end{equation}
where $k^i$ is a kernel of the $i$-th views as defined in Sec.~\ref{ssec7:descr}.
To simply the computation of the solution we define the matrix-valued kernel $G(x,t)$,
which for each pair $(x,t) \in \X \times \X$ is a diagonal $m \times m$ matrix, with
\begin{equation}
(G(x,t))_{i,i}  =   k^i(x^i,t^i),
\end{equation}
The Gram matrix $G[\x]$ is the $l \times l$ block matrix, where each block $(i,j)$ is the respective $m \times m$ matrix $G(x_i, x_j)$.
The matrix $G[\x]$ then contains the Gram matrices $k^i[\x]$ for all the kernels corresponding to all the views. The two matrices $K[\x]$ and $G[\x]$ are related by
\begin{equation}
K[\x] = G[\x] \otimes I_P.
\end{equation}
The system of linear equations \ref{equation7:linear-matrix} is  then equivalent to
\begin{equation}\label{equation7:linear-matrix2}
BA = Y_C,
\end{equation}
where
\begin{eqnarray*}
B = \left(\frac{1}{m^2}(I_l \otimes \e_m\e_m^T) + l \gamma_I (I_{l} \otimes M_m)\right)G[\x]  \\
+ l \gamma_A I_{lm},
\end{eqnarray*}
which is of size $lm \times lm$, $A$ is the matrix of size $lm\times P$ such that $\a = \vec(A^T)$, and
$Y_C$ is the matrix of size $lm \times P$ such that $\mathbf{C^{*}y} = \vec(Y_C^T)$.

Solving the system of linear equations \ref{equation7:linear-matrix2} with respect to $A$ is equivalent to
solving
system
~\ref{equation7:linear-matrix} with respect to $\a$.

\subsection{Verification and Recognition}\label{sec7:experiment}
The testing phase consists of computing $f^{\star}(v_i) = \sum_{j=1}^{l}K(v_i, x_j)a_j$, given the testing set $\v = \{v_1, \ldots, v_t\} \in \X$.
Let $K[\v, \x]$ denote the $t \times l$ block matrix, where block $(i,j)$
is $K(v_i, x_j)$ and similarly, let $G[\v, \x]$ denote the $t \times l$ block matrix, where block $(i,j)$
is the $m \times m$ matrix $G(v_i, x_j)$. Then
\begin{displaymath}
\f^{\star}(\v) = K[\v, \x]\a = \vec(A^TG[\v, \x]^T).
\end{displaymath}
For the $i$-th sample of the $p$-th user, $f^{\star}(v_i)$ represents the vector that is as close as possible to $y_i = (-1, \ldots,1, \ldots, -1)$, with $1$ at the $p$-th location.
The identity of the $i$-th image is estimated \emph{a-posteriori} by taking the index of the maximum value in the vector $f^{\star}(v_i)$.

First of all, we performed identity recognition in order to investigate the ability of the system in recognizing a particular probe user among the gallery subjects. To this
end,
we consider conversations which are
$10$ turns
long, i.e., very short dyads,
in order to modulate
the number of training conversations that we can have for each individual. Then, keeping fixed the number of training conversations for each user
($3$ conversations), we vary the number of turns from $2$ to $10$ to test the accuracy of the proposed method using a limited number of turns.
After this, we analyze the user verification: the verification performance is defined as the ability of the system in verifying if the person that the probe user claims to be is truly him/herself, or if he/she is an impostor.

As a
performance measure
for the identity recognition, we used the Cumulative Matching Characteristic (CMC) curve. The CMC is an effective performance measure for AA approaches \cite{Bolle:2003}: given a test sample, we want to discover its identity among a set of $N$ subjects.
In particular, the value of the CMC curve at position $1$ is the probability (also called \emph{rank1} probability), that the probe ID signature of a subject is closer to the gallery ID signature of the same subject than to any other gallery ID signature; the value of the CMC curve at position $n$ is the probability of finding the correct match in the first $n$ ranked positions. As a single measure to summarize a CMC curve, we use the normalized Area Under the Curve (nAUC), which is the approximation of the integral of the CMC curve. For the identity verification task, we report the standard ROC curves, together with the Equal Error Rate (EER) values.

As a comparative approach, we consider the strategy of \cite{Cristani:Skype:AVSS:2013}, whose code is publicly available, for a fair comparison we repeated the experiments on the same data. Results show that when working on a very small amount of turns, the proposed approach with learning is more efficient.

\subsubsection{Identity Recognition}

In the identity recognition task, we performed two experiments. In the first experiment we
fixed
the number of turns after which we want an answer from the system to $TT=6$ (in the next experiment we varied this parameter also).
After that, we built a training set, which for each person has a particular number of conversations, that is used by the learning algorithm to train the system.
After training, we applied our approach on the testing set, which was composed of a conversation for each subject, performed the recognition,
then
calculated the CMC curve and the related nAUC value.
We did the same with the comparative approach (which simply calculates distances among features, and computes the average distance among the probe conversation and the three training conversations). All the experiments
were
repeated 10 times, by shuffling the training/testing partitions.
The results are better with our proposal both in case on nAUC and rank 1 score.
In all the cases it is evident that augmenting the number of conversation gives a higher recognition score.
\begin{table}
\begin{center}
\small
\begin{tabular}{|c|c|c|}
\hline
{{Gallery}} & {Roffo et al.}\cite{Cristani:Skype:AVSS:2013} & {Our approach} \\
{Size}& \emph{(nAUC)} &  \emph{(nAUC)} \\
\hline
\emph{1 conv.} & $65.3\% ( 8.9\%)$&  $\mathbf{68.7\% (10.0\%)}$\\\hline
\emph{2 conv.} & $64.6\% (10.7\%)$&  $\mathbf{71.2\% (11.4\%)}$\\\hline
\emph{3 conv.} & $64.3\% (11.1\%)$&  $\mathbf{75.4\% (12.6\%)}$ \\
\hline
\end{tabular}
\end{center}
\caption{Comparison between Roffo et al. \cite{Cristani:Skype:AVSS:2013} and the proposed method increasing the number of conversations (\emph{conv.} formed by $6$ turns each). The first number represents the nAUC, while in parenthesis we have the rank1 probability.}\label{table7:results1}
\end{table}

In the second experiment, we kept fixed the number of conversations per gallery to 3, and we gradually increased the number of turns from $2$ to $10$ with stepsize $2$.
The recognition results of \cite{Cristani:Skype:AVSS:2013} along with our method are reported in Table~\ref{table7:results2}.
It is easy to notice that our approach outperforms \cite{Cristani:Skype:AVSS:2013} in all the cases.
and that the increment with respect to \cite{Cristani:Skype:AVSS:2013} is more evident when using a low number of turns.
This result supports the fact that \cite{Cristani:Skype:AVSS:2013} is good only when having a lot of data and a good descriptor.
Instead, the proposed approach can be used also with very few information.

\begin{table}
\begin{center}
\small
\begin{tabular}{|c|c|c|}
\hline
{Turns} & {Roffo et al.}\cite{Cristani:Skype:AVSS:2013}  & {Our approach} \\
\hline
 2 & $53.3\%$ &  $\mathbf{65.8\%}$  \\\hline
 4 & $58.5\%$ &  $\mathbf{70.9\%}$ \\\hline
 6  & $64.3\%$ & $\mathbf{75.4\%}$  \\\hline
 8  & $70.4\%$ & $\mathbf{76.9\%}$ \\\hline
10  & $77.5\%$ & $\mathbf{79.2\%}$ \\
\hline
\end{tabular}
\end{center}
\caption{Comparison between Roffo et al. \cite{Cristani:Skype:AVSS:2013} and the proposed method in terms of nAUC. We kept the number of conversation per subject in the gallery fixed, while we varied the number of turns per conversation.}\label{table7:results2}
\end{table}

\subsubsection{Identity Verification}


Considering the verification task, we adopted the following strategy: given the signature of user $i$, if it matches with the right gallery signature with a matching distance which is ranked below the rank $K$, it is verified. Intuitively, there is a tradeoff in choosing $K$. A high $K$ (for example, 78, all the subjects of the dataset) gives a 100\% of true positive rate (this is obvious by construction), but it brings in a lot of potential false positives. Therefore, taking into account the number $K$ as varying threshold, we can build ROC and precision/recall curves, using the value $K$ as varying parameter to build the curves.

In particular, we report for each method, and for each number of turns taken into account: the nAUC of the ROC curve; the equal error rate (EER), which is the error rate occurring when the decision threshold of a system ($K$) is set so that the proportion of false rejections will be approximately equal to the proportion of false acceptances (less is better); the best F1 value obtained, together with the $K$ value which gave the best results (in terms of F1 score), and the related precision and recall values. For the sake of the clarity, we produce two tables, one for the \cite{Cristani:Skype:AVSS:2013} method (Table~\ref{table7:results3}), one for our approach (Table~\ref{table7:results4}).

\begin{table}
\begin{center}
\small
\begin{tabular}{|c|c|c|c|c|}
\hline
{Turns} & {ROC}  & {EER} & {Best F1 /} & {Best Prec /} \\
                     & \emph{(nAUC)} &                  & {Best K }   &   {Recall} \\
\hline
2 & 52.8\% & 48.7\%& 67.3\% / 4 & 51.7\% / {96.2\%} \\\hline
4 & 57.9\% & 45.4\% & 69.4\% / 5 & 54.6\% / { 95.0\%} \\\hline
6 & 63.8\% & 40.3\% & 70.5\% / 5 & 55.9\% / {95.1\%} \\\hline
8 & 70.0\% & 34.2\% & 71.9\% / 5 & 57.7\% / { 95.1\%} \\\hline
10 & 77.3\% & 30.1\%& { 74.7\%} / 10 & 64.3\% / {88.9\%} \\\hline
\end{tabular}
\end{center}
\caption{Verification performance of Roffo et al. \cite{Cristani:Skype:AVSS:2013} approach.}\label{table7:results3}
\end{table}

Our approach performs better, except in the case of $10$ turns, where the F1 score is higher for \cite{Cristani:Skype:AVSS:2013}. 
It is worth noting that the higher F1 score is due to a very high recall score, definitely superior to the precision value. In our case, precision and recall are better balanced. In general, it is possible to note that the recall values are higher than the precision, and that augmenting the number of turns gives higher performances.%
%

\begin{table}
\begin{center}
\small
\begin{tabular}{|c|c|c|c|c|}
\hline
{Turns} & {ROC}  & {EER} & {Best F1 /} & {Best Prec /} \\
                     & \emph{(nAUC)} &                  & {Best K }   &   {Recall} \\
\hline
2 & { 65.3\%} & {39.4\%}& { 69.2\%} / 6 & {54.8\%} / 93.7\% \\\hline
4 & { 70.5\%} & { 35.5\%}& { 70.2\%} / 6 & {56.1\%} / 93.8\% \\\hline
6 & { 75.0\%} & { 32.6\%}& {72.6\%} / 6 & { 63.4\%} / 85.0\% \\\hline
8 & { 76.6\%} & { 30.3\%}& {73.2\%} /  9& { 61.7\%} / 90.1\% \\\hline
10 & {78.9\%} & {28.9\%}& 73.9\% /  15& {67.0\%} / 82.5\% \\\hline
\end{tabular}
\end{center}
\caption{Verification performance of the proposed approach.}\label{table7:results4}
\end{table}

%
%
\subsection{Summary}\label{sec7:conc22}
The ability to understand the identity of a person by looking at the way they chat is something that we can intuitively feel: we certainly know that some people are used to answering more promptly to our questions, or we know some people who are very fast in writing sentences. This last approach subsumes these abilities, putting them into a learning approach, which is capable of understanding the peculiar characteristics of a person, enabling effective recognition and verification. In particular, this study offers a first analysis of what a learning approach can do, when it comes to minimizing the information necessary to individualize a particular identity. The results are surprisingly promising: with just 2 turns of conversation, we are able to recognize and verify a person strongly above chance.
This demonstrates that a form of behavioral blueprint of a person can be extracted even on a very small portion of chats. We believe therefore that this work has the potential to open up the possibility of a large range of new applications beyond surveillance and monitoring.

\section{Keystrokes Dynamics and Personality Traits}\label{sec7:keystrokes}

Text chatting is probably one of the most typical communication phenomena of the last 20 years, characterized by a massive worldwide diffusion\footnote{\url{http://goo.gl/2NrKFG}}. The peculiarity of text chats stands in the fact that they show both aspects of literary text and of spoken conversation, due to the turn-taking dynamics that characterize text delivery. After the encouraging work on $C$-$Skype$, in this section we have set up our own chat service in which a key-logging functionality has been activated, so that the timings of each key pressing can be measured. Therefore, it is interesting to investigate which aspects of these two communication means are intertwined in chat exchanges. In particular, we will try to discover how personality comes into play in chats: in fact, though connections of personality traits both with speech \cite{markel1972,hassin2000} and with text \cite{pennebaker1999,pennebaker2001linguistic,coltheart1981mrc,mairesse2007using} have been long studied, a computational analysis of the role of personality in chats is missing. At the same time, we are also interested in discovering whether the manifestation of very particular, that is, recognizable chatting styles may correspond to distinguishable personality profiles. If this is the case, we may single out personality traits that are the most evident via chat, and individuate new means for chats to bring about meaning in addition to literary content. While so far many studies have been published on the recognition of writing style in literary texts (commonly known as \emph{authorship attribution})~\cite{AbbW:2008,Stamatatos_2009}, when it comes to chats the research is still moving its early steps~\cite{Cristani:Skype:AVSS:2013}.

A first direction towards these goals has been presented in \cite{Cristani:Chat:HBU:2014}, where a corpus of spontaneous Skype conversations between 45 different couples have been recorded during a period of 3 months and analyzed. From the corpus a pool of ``stylometric'' chat features have been extracted, which have been shown to codify the chatting style of users~\cite{Cristani:Skype:AVSS:2013}, together with some personality traits (Barratt Impulsiveness Scale ~\cite{patton1995}, Interpersonal Reactivity Index ~\cite{IRI_TEST}). A first correlation study has been conducted, which highlighted some weak but significant links among stylistic features and personality cues.

In this section, we move a step ahead, providing a more solid study on the subject, changing radically the analysis protocol. In this case, we build a chat service, embedded into the \emph{Klimble} social network \footnote{\url{http://www.klimble.com/}}, with key logging capabilities; this in practice allows us to retain the timing of each single hit of a key, recording at the finest level the behavior of users while they chat. In the \cite{Cristani:Chat:HBU:2014} paper, timing information was limited to the recording of each return key pressure, introducing some evident approximations and limitations in the information that could be gathered. Conversely, here a novel set of stylometric features can be extracted, capturing a set of cues so far impossible to get, as the number of backspace hits, the typing speed, and so on; in particular, we design 11 novel features that are much more fine-grained than in the previous work. Take speed as an instance: if the style of an individual is characterized by slow turns, this may be caused by long pauses between a word and the other, or by the fact that the individual often erases what he/she has just written. On this chatting platform, we collect a new dataset in Italian containing data of 50 different subjects, collected within a two months period (see Section \ref{ch4:TBK} for details). All subjects (18-27 years old university students) are asked to chat with an operator barely known by them\footnote{Such choice is motivated by the fact that we want on the one hand to prevent the style of the subjects to be influenced by that of different interlocutors and, on the other, to avoid styles that are peculiar to specific kinds of long term interactions.}, for an average duration of 20 minutes. The operator performs semi-structured conversations, in order to reach a certain degree of homogeneity in the verbal content of the conversations and highlighting possible differences in the non-verbal behavior. To get information on some psychological traits of the subjects, we have used three well-known self-administered questionnaires: 1) the Barratt Impulsiveness Scale ~\cite{patton1995}, focused on 3 different impulsiveness factors (attentional, motor and non-planning impulsiveness)~\cite{patton1995}, 2) BIS-BAS~\cite{carver1994}, decoding human motivations to behavioral inhibition (BIS) and activation (BAS), and 3) PANAS~\cite{watson1988}, analyzing positive and negative affectivity traits. Contrarily to the Big Five~\cite{Rammstedt2007}, these traits have shown to be more effective in capturing basic personality aspects~\cite{cohen2008}. In this study we have abandoned the Interpersonal Reactivity Index because, differently from the previous study, chats were all performed with the same (not well-acquainted with the subjects) operator, posing the same questions to everyone, and therefore features investigated by the IRI~\cite{IRI_TEST} , such as those related to the ability to take into account other's perspective or empathic concerns were not so relevant here.

The first analysis was aimed at investigating the relation between stylometric cues and personality traits. To this sake, correlations have been calculated, showing that 6 psychological factors correlate with 10 out of 28 total stylometric features, in a statistically significant way (\emph{p-value}$<0.05$); for example, our data show that subjects with higher positive affectivity tend to use to write slowly and to use a smaller number of long words. In addition, we perform regression on the personality traits using the features, showing that it is possible to generate predictions which correlate significantly with the original score for 4 personality traits, namely non planning impulsiveness, BIS punishment avoidance, negative affectivity and anxious personality.

The second analysis is aimed at verifying how discriminative is the chatting style of a person when compared to that of other subjects. To this aim, a detailed study is conducted analyzing the contribute of each stylometric feature in terms of recognition capability. Highly discriminative features have been found, some of them correlating with personality traits. Connecting these two analyses allows to hypothesize that possibly some personality traits lead people to chat in a particular style, which makes them very recognizable. For example, in the example above, the use of short words is a discriminative
characteristic in chats, and is significantly correlated with higher non planning impulsiveness, highlighting that people who score higher in this factor of impulsiveness use also a greater number of short words, that is, the more subjects are impulsive, the shorter the time they stop on each single word while writing.

\subsection{Related Work} 

\subsubsection{Computational Approaches}
Most computational approaches on inferring personality traits from text depend strongly on a semantic analysis of the content: usually, psychologists individuate semantic features that are then validated through computational approaches. One of the first studies on the subject dates back to 1999 \cite{pennebaker1999}, while the most cited studies concerning the design of features are about the Linguistic Inquiry and Word
Count (LIWC) features~\cite{pennebaker2001linguistic}: these include both syntactic (e.g. past tense verbs) and semantic cues (e.g. anger words, reference to family members), all validated by experts. Another important source of cues is the MRC Psycholinguistic database~\cite{coltheart1981mrc}, containing statistics for over 150K words, and defininig features like ``syllables per words'' and ``concreteness''.
A study that pools together all these features is \cite{mairesse2007using}, which is focused on finding relations with the Big Five personality traits, and where it is shown that MRC features are useful for models of emotional stability, while LIWC features are applicable to all traits. In this study, we neglect semantic analysis of the text, for privacy issues, and we deal with chat conversations corpora, which have never been considered before. Regarding the author recognition issue, the most related field is that of Authorship Attribution (AA), which introduced the term ``stylometric feature''. In this case, the related literature agrees essentially in considering the taxonomy of \cite{AbbW:2008}, which partitions features into five major groups: \emph{lexical, syntactic, structural, content-specific and idiosyncratic}. The typical workflow of an AA approach is that of extracting from a set of training texts (related to some gallery subjects) a set of features, feeding them into discriminative classifiers, and proceeding with the testing on some unseen texts. The application of AA to chat conversations is quite recent, see~\cite{Stamatatos_2009} for a survey. Notably, the taxonomy of stylometric features has been enlarged in~\cite{Cristani:Skype:AVSS:2013} with \emph{turn-taking based} features, which explicitly account for the dynamics of the turns of text delivery in chats. The same paper also contains a summary table of the most important stylometric features.\vspace{-0.2cm}

\subsubsection{Psychological Literature}
On the other hand, the psychological literature 
has shown that people tend to use rapid judgements on personality traits as a guide during interaction. In conversation this occurs in a significant way. Indeed, several studies showed that individuals tend to believe that speakers' speech characteristics are indicative of their personality traits~\cite{hassin2000}. Other researches have effectively demonstrated that speakers' voice type (loud-slow, loud-fast, soft-slow, soft-fast) is related with speakers' personality traits~\cite{markel1972}. As shown by several studies, text analysis has proven to be a useful aid for measuring personality traits. For example,~\cite{pennebaker1999} showed correlation between lexical expressions and basic personality measures assessed by the Big Five questionnaire. Positive correlation between neuroticism and the use of emotional words with a negative valency, and between extroversion and emotional words with a positive valency were reported. Similar results were also found in~\cite{mehl2006}; however, in this study the content of the verbal material was not taken into consideration. A more recent study,~\cite{cohen2008}, instead, analyzed to which degree basic elements of personality can be measured by using lexical analysis of verbal expression of autobiographical and personally relevant material. Two well-validated models of personality aimed at investigating, respectively, positive and negative affectivity traits (PANAS) and human motivations to behavioral inhibition (BIS) and activation (BAS) were used. Findings showed that individuals with high levels of positive affectivity and behavioral activation tended to express more positive emotions in their natural speech, whereas those with high levels of negative affectivity and behavioral inhibition tended to show more negative emotions.\vspace{-0.2cm}

\subsection{Stylometric \& Keystrokes Features}\label{sec7:Features}

In the design of stylometric features we follow the idea of the previous sections ~\cite{Cristani:Skype:AVSS:2013,Cristani:Chat:HBU:2014}, that is, of using the turn length, and not simply the entire conversation, as fundamental unity for computing stylometric statistics. The big limitation of such works was that timings were calculated by considering the duration of each turn, as given by the Skype APIs; this introduced some approximations, like the ones needed for evaluating the writing speed (number of characters/turn duration) and limited the design of the features (for example, it was not possible to calculate the effective speed in writing words, excluding the spaces). All these limitations have been overcome with this new framework. Since the system has been designed to work on social networks, we have paid much attention to privacy issues: the idea is to neglect the content of the conversation, accounting only for the way it is performed. Our features follow this guideline, avoiding any kind of natural language processing, while other features, as length of words, punctuation, emoticons etc. are preserved and analyzed.
For each person involved in a conversation, we analyze his/her stream of turns (suppose $T$), ignoring the content of the input from the other subject. This means
that we assume that the chat style (as modeled by our features) is independent from the interlocutor (hereafter called \emph{operator}) -- who in this case was the same for all participants, and was barely known by the subjects analyzed. 
From each turn, a stylometric feature is extracted, generating a number; therefore, with $T$ turns we obtain $T$ feature values. Depending on the kind of feature and task to perform (measuring correlations or doing person identification), an histogram or the mean/median is computed, and the resulting measure becomes a part of the signature which characterizes a given individual. In the following, the list of the features together with their explanation is presented.
For convenience, and whereas possible, we have kept the name of the features proposed in the literature, with the substantial difference that in this case the feature extraction was performed using a key-logging framework, ensuring a more precise and capillary temporization. Specifically, all the \emph{turn-taking features} are brand new. How features have been treated to calculate correlations with personality traits or similarities for the sake of recognition will be discussed in the following. In the list below, the numbers in parenthesis indicate the feature ID. \vspace{-0.3cm}
\paragraph{\bf{Lexical Features}}
\begin{itemize}\vspace{-0.3cm}
\itemsep0em
\item[(1)] {\bf Number of words (\#W)}: number of words per turn. With ``word'' we intend a string of \emph{characters} (see below);
\item[(2)] {\bf Number of chars (\#C)}: number of characters per turn. With ``character'', we intend every normal key on the QWERTY keyboard, ignoring special keys like the SPACE, CTRL, etc.; 
\item[(3)] {\bf Number of uppercase chars (\#Uppercase)}: number of uppercase characters in a turn;
\item[(4)] {\bf Number of uppercase chars / number of chars (\#Uppercase/\#C)}: usually, entire words written in capital letters indicate a strong emotional message. This feature accounts for such communicative tendency;
\item[(5)] {\bf Mean word length}: average length of the words in a turn;
\item[(6-7)] {\bf 1(2)-order length transitions (1oLT,2oLT)}: these features resemble the n-grams of~\cite{Stamatatos_2009}; the strong difference here is in the fact that we consider solely the length of the words, and not their content. See \cite{Cristani:Skype:AVSS:2013} for further details.\vspace{-0.4cm}
\end{itemize}\vspace{-0.3cm}
\paragraph{\bf{Syntactic Features}}
\begin{itemize}[leftmargin=1.6cm]\vspace{-0.3cm}
\itemsep0em
\item[(8)] {\bf Number of ? and ! marks (\#?!)}: we keep the  ``?'' and the ``!'' marks in the same feature, since taken separately their relevance is very low;
\item[(9)] {\bf Number of suspension points (\#...)};
\item[(10)] {\bf Number of generic marks (\#Marks)}: a high number of generic marks (",.:*;) usually indicates a more accurate writing style;
\item[(11,12,13)] {\bf Number of \emph{positive}, \emph{negative} and \emph{uncategorized} emoticons (\# Emo+, \# Emo-, \# Emo=}, respectively{\bf)}: features related to emoticons aim at individuating a particular mood expressed in a turn. In particular, 101 diverse emoticons have been divided in three classes, portraying positive emotions (happiness, love, intimacy, etc. $-$ 20 emot.), negative emotions (fear, anger, etc. $-$ 19 emot.) and other emoticons (portraying actions, objects etc. $-$ 62 emot.);
   \item[(14)] {\bf  Number of emoticons (\#Emo)}: Number of emoticons in a turn, independently from their type;
\item[(15-16)] {\bf Number of emoticons / number of words (chars) (\#Emo/\#W, \#Emo/\#C}, res\-pe\-cti\-ve\-ly{\bf )}: it considers how often pictorial symbols are used in a sentence considering the number of words (chars) typed;
\item[(17)] {\bf Number of deletions (\#Back)}; it models the number of hits of the backspace key during a turn.\vspace{-0.4cm}
\end{itemize}\vspace{-0.3cm}
\paragraph{\bf{Turn-taking Features}}
For these novel features, the temporization of the key hits is crucial. In the following, we indicate with $t_n$  the instant when the $n-$th hit of the button $b_n$ occurs during a turn.
\begin{itemize}
\itemsep0em
\item[(18)] {\bf Turn time}: We indicate with $T$ the length of the period in which a turn is kept (before pressing the ``return'' key);
\item[(19)] {\bf Typing time}: thanks to the exact timings of key presses, this feature approximates the time spent in writing words,  as the pauses and the time spent in deciding what to say and in reading what the other has written is not included in the counting. A word is assumed to be a string of consecutive alphanumeric characters, which is anticipated and followed by a blank space. Formally, let us define $I_k$,  the interval of time spent in writing the $k-$th word; we have
    \begin{equation}
    I_k = t_n - t_m
    \end{equation}
    such that $n>m$, $ b_{m-1},b_{n+1}$ are space keys and $b_m$, $b_{m+1},\ldots,b_{n}$  are not space keys. In these cases,
    \begin{equation}
    \text{Typing time} = \sum_k^K I_k
    \end{equation}
    assuming to have $K$ words in the turn under analysis. Calculating this feature is different than simply computing the time spent in a turn, as it allows for instance to distinguish a person who is very fast in typing words but uses long pauses between a word and another (he/she thinks a lot what to write) from an individual whose typing speed is slower, but knows what to say and takes fewer and shorter breaks while writing a sentence;
\item[(20)] {\bf Mean typing time}: it divides the total typing time in a conversation of a subject by the number of turns;
 \item[(21)] {\bf Mean writing speed}: it measures the typing time divided by the number $\#C$ of written chars in a turn;
 \item[(22)] {\bf Deviation writing speed}; the standard deviation of the timings required for typing a character. Assuming $\tilde{t}_n=t_n-t_{n-1}$ and $t_{n}$ is the time when a char button is hit, we have
     \begin{equation}
     \text{Dev.writ.speed} = \frac{1}{\#C-1}\sum_{n=1}^{\#C} \left(\tilde{t}_n - \text{{Mean writ.sp.}}\right)^2
     \end{equation}
\item[(23)] {\bf Silence time ($ST$)}: the time spent in pressing the space bar, which is collected by summing all those intervals of the form
 \begin{equation}
    S_k = t_n - t_m
    \end{equation}
     such that $n>m$, $ b_{m},b_{n}$ are not space keys and $b_m$, $b_{m+1},\ldots,b_{n}$  are space keys. Therefore
    \begin{equation}
    \text{Silence time} = \sum_k^K S_k
    \end{equation}
 This feature models the pauses that are taken between a word and another, which could be thought as the counterpart of the silence in spoken interactions;
\item[(24)] {\bf Speech ratio}: the ratio between the time spent typing words and typing spaces. It gives an idea of the balance between being inactive (because one is waiting for the other's reply, or because he/she is thinking what to write) and being actively typing;

\item[(25-26)] {\bf Imitation per word, per char (\#W imit, \#C imit)}: ratio between number of chars -or words- in the current turn and number of chars -or words- in the previous turn of the interlocutor; this feature models the tendency of a subject to imitate the conversation style of the interlocutor (at least for what concerns the length of
the turns); 
\item[(27)] {\bf Answer time}: this feature is the time spent to answer a question presented in the previous turn of the interlocutor. We assume the presence of a question whenever there is a question mark;
\item[(28)] {\bf Typing time length $n$ (Typing time n)}> it models the average time needed for writing a word of a given length $L$ ($L=1,\ldots,15$), where the averaging operation here is built on the entire conversation, and not solely on a single turn.
\end{itemize}
Since these features are collected for each turn (except the Typing time n, 1oLT and 2oLT features), and assuming there are $T$ turns in a conversation, we end up with $T$ numbers for each feature.
These features are subsequently used for two applications: correlation with psychological traits  and user recognition. Depending on the task at hand, the feature values are treated differently, as discussed in the following.\vspace{-0.2cm}

\subsection{Correlations traits-features}\label{sec7:corr}\vspace{-0.1cm}

In order to inquire whether a connection holds between psychological traits and features, we consider the cues extracted from the chat of each single subject and the scores of his/her questionnaires. In particular, for each feature we calculate the mean or the median number over all the turns of the conversation, obtaining one value per feature, per user. As notable exception, for the 1oLT and 2oLT we keep the maximum values contained within. The reason is simple (take as example the 1oLT): having a high entry at the position $i,j$ of the matrix means that a word of $i$ characters, followed immediately by a word of $j$ characters have been written many times by a subject; this indicates a certain stylistic pattern that, if high, indicates a sort of blueprint for that user.
For the estimation of the correlation, we calculate the Pearson correlation coefficient (where both the distribution of features values and traits are normal), the Spearman coefficient otherwise. The test for normality is the Lilliefors test. Results are shown in Table~\ref{table7:Corr}, showing 17 significant correlations (p-value$<0.05$). Note that Attentional Impulsiveness and BAS did not show any correlation.
\begin{table*}[!ht]
\scriptsize
\begin{center}\vspace{-0.45cm}
\resizebox{0.95\textwidth}{!}{%
\begin{tabular}{|m{3.4cm}|>{\centering\arraybackslash}m{1.9cm}|>{\centering\arraybackslash}m{1.9cm}|>{\centering\arraybackslash}m{1.9cm}|>{\centering\arraybackslash}m{1.5cm}|>{\centering\arraybackslash}m{1.5cm}|}
\hline
\multirow{2}{*}{} &  \textbf{Motor impulsiveness} &\textbf{Non planning impulsiveness}  & \textbf{BIS punishment avoidance} & \textbf{PA Positive affectivity} & \textbf{NA Negative affectivity} \\
\hline
\textbf{\#Emo} (n14) \emph{[66.4\%]}& \nullc& \nullc&  \multicolumn{1}{c|}{\textbf{-0.31 }}&\nullc & \nullc  \\\hline
\textbf{\#Emo/\#W} (n15) \emph{[60.7\%]}&  \nullc& \multicolumn{1}{c|}{\textbf{-0.28 }}& \multicolumn{1}{c|}{\textbf{-0.29 } }&\nullc & \multicolumn{1}{c|}{\textbf{-0.30 }}  \\\hline
\textbf{\#Emo/\#C} (n16)	\emph{[61.0\%]}& \nullc& \multicolumn{1}{c|}{\textbf{-0.28 }}& \multicolumn{1}{c|}{\textbf{-0.29 }}  & \nullc & \multicolumn{1}{c|}{\textbf{-0.31 } } \\\hline
\textbf{\#Emo+} (n11)	\emph{[/]}&  \nullc& \nullc&\nullc  & \nullc& \nullc   \\\hline
\textbf{\#W}	(n1) \emph{[60.4\%]} &\nullc&\nullc&\nullc  & \multicolumn{1}{c|}{\textbf{-0.29} }&\nullc  \\\hline
\textbf{Mean word length} (5) \emph{[63.5\%]}&  \multicolumn{1}{c|}{\textbf{-0.41 }}&\multicolumn{1}{c|}{\textbf{-0.38 }}&\nullc & \multicolumn{1}{c|}{\textbf{+0.38 }} & \nullc  \\\hline
\textbf{Mean writing speed} (20) \emph{[65.9\%]} & \nullc& \nullc&\nullc  & \multicolumn{1}{c|}{\textbf{-0.36}} &\nullc  \\\hline
\textbf{1oLT} (n6) \emph{[69.9\%] }& \nullc& \nullc&\nullc & \multicolumn{1}{c|}{\textbf{-0.34}} &\nullc  \\\hline
\textbf{\#C imit} (n26) \emph{[57.1\%]}& \nullc& \multicolumn{1}{c|}{\textbf{+0.36 }}&  \multicolumn{1}{c|}{\textbf{-0.29 }}  & \nullc & \multicolumn{1}{c|}{\textbf{-0.31}} \\\hline
\textbf{\#W imit} (n25) \emph{[55.9\%]} &\nullc& \multicolumn{1}{c|}{\textbf{+0.30 }}& \nullc  &\nullc &\nullc  \\\hline
\end{tabular}}
\end{center}
\caption{Correlations between stylometric features. In brackets, the ID of the features. In squared brackets the nAUC score, witnessing how effective the feature is in distinguishing people (the higher the more effective, see Sec.~\ref{sec7:reco}). Numerical values in the table indicate statistically significant correlations (\emph{p-value}$<0.05$). Attentional Impulsiveness and BAS did not show any correlation, so they are not reported here.}\label{table7:Corr}
\end{table*}
Our findings suggest that subjects who scored higher in motor and non-planning impulsiveness used shorter words (low \emph{mean word length}), indicating that they were taking less time to provide their answer. Moreover, subjects who referred higher deficit in planning their own behavior (non-planning impulsiveness) tended to imitate to a larger extent what the other interlocutor did (high \emph{imitation rate}) and used a lower number of emoticons in relation to the used words/chars, as suggesting that they were, again, available neither to loose time in chatting nor in thinking about what and how to write.
Relatively to traits involved in human motivation to act and write in a way rather than in another, our findings are also of interest. Specifically, subjects with higher levels of behavioral inhibition (that is, punishment avoidance) showed a lower tendency to imitate the other interlocutor's style in chatting. This result is not clearly intuitive as, usually, in order to avoid punishment, people tend to be accommodating and in this framework one would expect them to imitate the interlocutor's style. Nonetheless, it has to be noticed that the specific task did not imply any punishment, therefore subjects could feel free to write spontaneously, without imitating the interlocutor. Interestingly, they used a lower number of emoticons, not only in absolute terms, but also relatively to the number of words or chars used in writing (low \#Emo/\#W and \#Emo/\#C ). Also the latter chatting style can be the manifestation of a tendency to avoid to show emotions, be they positive or negative, and might be driven by a behavior of punishment avoidance, represented by high scores in BIS scale.
The same negative correlation with the number of emoticons used has been found in connection with higher levels of negative affectivity, showing again a higher tendency to avoid to express emotions while chatting. These findings seem to be related to those found by~\cite{cohen2008}, where subjects with behavioral inhibition and negative affectivity tended to show more negative emotions in their natural speech. Also, subjects with higher negative affectivity showed a lower tendency to imitate the interlocutor's chatting style. Subjects with higher positive affectivity showed a higher tendency to use long words and to write slowly, to use a smaller number of words, and to have a higher tendency to change their chatting style (low \emph{1oLT}). These findings, taken together, seem to suggest that positive affectivity leads to emphasize the specific words used, preferring the length rather than the quantity of words, writing them slowly, and feeling free to change their chatting style.
It is worth noting that the 28x7 correlation tests cannot be taken together and compared with the number of significant correlation values, since they refer to highly different cues and diverse personality markers; instead, the 17 correlations found are meaningful, since they concentrate on few stylometric features (10), which represent few distinct aspects of chatting. Even if the correlations are somewhat weak, they resemble numerical results obtained by analogue studies on personality and writing style, see \cite{pennebaker1999} as an example.\vspace{-0.3cm}

\subsection{Prediction by regression}\label{sec7:Prediction}
To assess the reliability of the correlations, we investigate how well an automatic system can predict the personality traits of a person given a portion of his/her chat. To this sake, for each trait we select all the features which show significant correlations with it, and we train a separate multivariate Epsilon Support Vector Regressor ($\epsilon$-SVR) with RBF kernel, in a Leave One Out sense (learning on N-1 people, testing on the Nth one, for all the people). We use $\epsilon$-SVRs after having tried linear regression, which gave worst performances. As for the parameter selection, we set a single $\epsilon$-SVR parameterization, which work reasonably well on all the traits, that is $C=220,\; \gamma = 1.6, \; \epsilon =1$. As evaluation metrics we use the standard $R^2$ measure and we also perform correlation between the predicted values and the original ones, to show if, apart from bias, there is some regularity in the prediction of the regressor.
The results are shown in Table~\ref{table7:Reg}. As visible, we obtain significant correlations with 3 out of 5 traits, even if the R2 score is very weak. Anyway, our aim is not to promote a product which performs such kind of prediction, but only to assess the influence of personality traits in the way a chat is performed.
\begin{table*}[!ht]
\scriptsize
\begin{center}\vspace{-0.1cm}
\resizebox{0.99\textwidth}{!}{%
\begin{tabular}{|>{\centering\arraybackslash}m{1.9cm}|>{\centering\arraybackslash}m{1.9cm}|>{\centering\arraybackslash}m{1.9cm}|>{\centering\arraybackslash}m{1.9cm}|>{\centering\arraybackslash}m{1.5cm}|>{\centering\arraybackslash}m{1.5cm}|}
\hline
\multirow{2}{*}{} &  \textbf{Motor impulsiveness} &\textbf{Non planning impulsiveness}  & \textbf{BIS punishment avoidance} & \textbf{PA Positive affectivity} & \textbf{NA Negative affectivity}\\
\hline
\textbf{$\rho$} &  \multicolumn{1}{c|}{\textbf{0.31 }}& \multicolumn{1}{c|}{\textbf{0.43 }}&  0.23&-0.16 & \multicolumn{1}{c|}{\textbf{0.44 }} \\\hline
\textbf{$p$} &  \multicolumn{1}{c|}{\textbf{0.03 }}& \multicolumn{1}{c|}{\textbf{0.001 }}& 0.09&0.27 & \multicolumn{1}{c|}{\textbf{0.002 }}  \\\hline
\textbf{$R^2$} & \multicolumn{1}{c|}{\textbf{0.09 }}& \multicolumn{1}{c|}{\textbf{0.06 }}& <0  & <0 & \multicolumn{1}{c|}{\textbf{0.15 }} \\\hline
\end{tabular}}
\end{center}
\caption{Regression results: for each trait, we report the Pearson correlation coefficient $\rho$ calculated between the predicted and the original personality traits, its associated p-value and the $R^2$ metrics.}\label{table7:Reg}
\end{table*}

\subsection{Person Recognition by Style}\label{sec7:reco}
\begin{figure*}[!ht]
\centering
\includegraphics[width=0.8\textwidth]{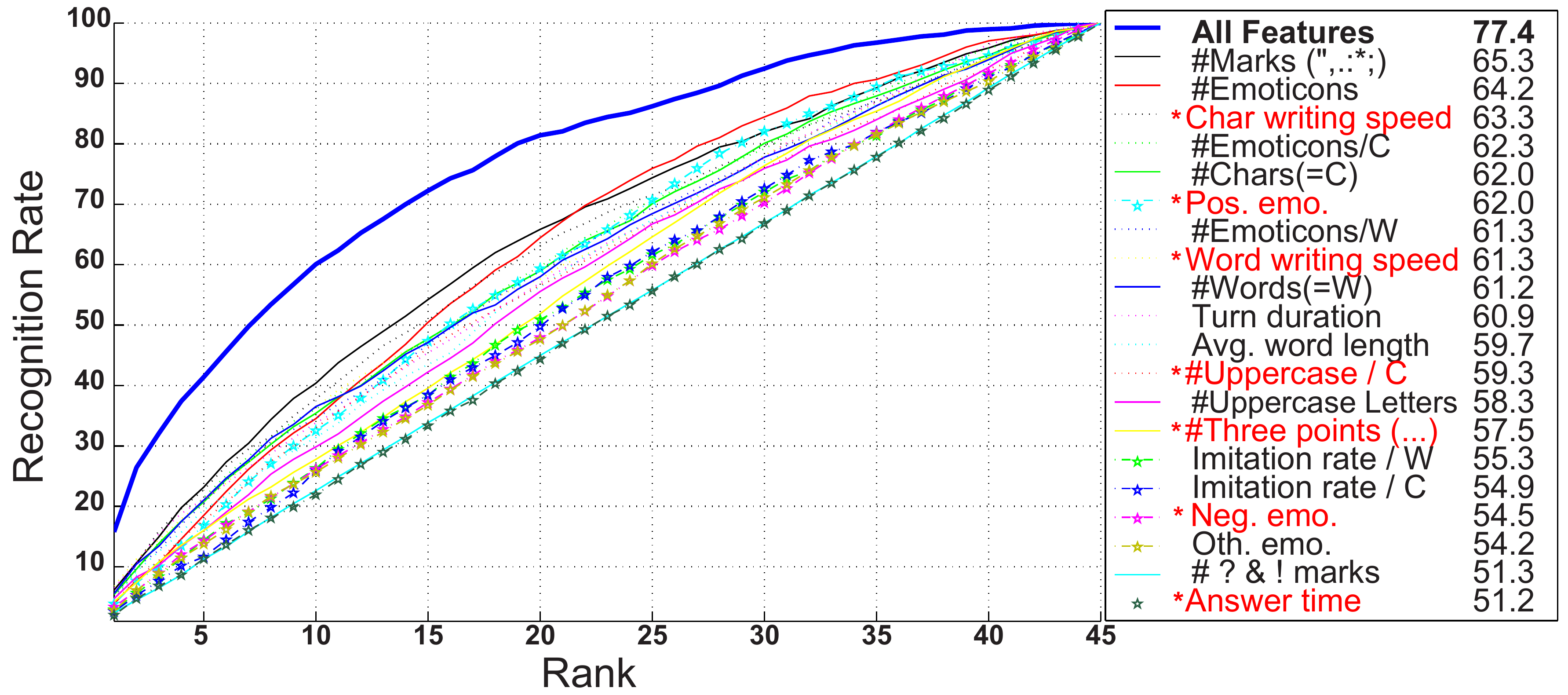}\vspace{-0.05cm}
\caption{ CMC curves for each feature. After each feature, their ID and the value of the correspondent nAUC. With an asterisk, in red, we specify those features which correlate at least with a personality trait. The \emph{All features} CMC indicates the performance of averaging the distance of all the features and calculating the ranking.}\label{fig7:features}
\end{figure*}

Here the aim is to analyze to which extent stylometric features can distinguish a person among many others.
For every subject we collect one conversation and extract the related features. Then, a similarity score is calculated  between each subject and all the others, first by accounting for each single feature separately, then using all the features together. In the former case, the similarity score is obtained by taking the $T$ values of the feature, organizing them into an 8-bin histogram,  where the range of the quantization values had been fixed by considering the whole dataset.  The match between any two histograms is calculated using the Bhattacharyya distance.  This process is applied to all kinds of features, except for the 1oLT and 2oLT features, where the diffusion distance~\cite{Cristani:Skype:AVSS:2013} is employed, and for the Typing time n, where Euclidean distance is adopted.
The turns of each subject are split into \emph{probe} and \emph{gallery} set, so that the probe samples can be given as test and the recognizer can evaluate the matches with the gallery elements. Probe and gallery set contain snaps of $10$ turns of conversation each in order to avoid biases due to quantitatively different available material.  Whenever possible, we have selected different turns (maintaining their chronological order) in order to generate different probe/gallery partitions. For each feature the re-identification has been repeated 30 times, varying probe and gallery partitions.
Then, for each subject we extract a particular feature from the probe set and calculate the distance with the gallery of the corresponding feature for all subjects. If we take  $N$ subjects, we obtain a $N\times N$ distance matrix. At this point we have ranked the $N$ distances for each probe element in ascending order and thus we have computed the \emph{Cumulative Match Characteristic} (CMC) curve, i.e., the expectation of finding the correct match in the top $n$ positions of the ranking. The CMC has been recognized as an effective performance measure for authorship attribution~\cite{Cristani:Skype:AVSS:2013}, and we have then used it to evaluate the task of \emph{identity recognition}, i.e. the ability to discover the identity of a subject among a set of $N$ other subjects. In particular, the value of the CMC curve at position $1$ is the probability that the probe ID signature of a subject is closer to the gallery ID signature of the same subject than to any other gallery ID signature; the value of the CMC curve at position $n$ is the probability of finding the correct match in the first $n$ ranked positions.
Given the CMC curve for each feature (obtained by averaging on 30 trials), the normalized Area Under Curve (nAUC) is
calculated as a measure of accuracy. The results are shown in Fig.~\ref{fig7:features}, where the features are listed in decreasing order of accuracy. The cues which correlate with at least one personality trait are portrayed in red. We have not reported the features producing an nAUC lower than 0.52.
As visible, all the CMC curves related to the different features are similar in expressivity, but not strongly effective; the probability of guessing the right people with only one attempt (corresponding to analyzing the performance of the CMC curve at rank 1) is below the 10\%. Anyway, if we combine all features, mediating the related distances computed among the probe and the gallery subject, we obtain a much more informative curve (see Fig.~\ref{fig7:features}, \emph{All features}). In this case, getting the correct guess after the first attempt is around the 30\%, and with the 80\% of probability we can individuate the correct subject among the first 10 ranked subjects.

\subsection{Linking Personality to Recognizability: Discussion}\label{sec7:cerchio}

Looking jointly at Fig.~\ref{fig7:features} and Table~\ref{table7:Corr}, one can immediately notice that positive affectivity is the trait which is most strongly related with the style of a person (it correlates with 4 stylometric features), making it recognizable with the highest probability (if we use only its correlated features we obtain 72.1\% of nAUC). This result is intuitive,  since positive affectivity may favor the tendency of writing following a well defined structure (the 1oLT feature, capturing the fact that short words follow short words, short words follow long words etc.) and of using few long words, slowly written.
A second important personality trait is the non planning impulsiveness (how much a person refers higher deficit in planning his/her own behavior), whose 5 correlated features, if combined, give 69.4\% of nAUC. The same nAUC score is given by the 4 features correlated with the punishment avoidance (people with higher levels of behavioral inhibition). Negative affectivity has 3 correlated features (65.9\% nAUC). The less indicative trait, among the ones which correlate, is the motor impulsiveness, with only one feature (63.5\% nAUC).
These findings are very interesting as, congruently with the aim of this study, they clearly suggest that specific psychological factors related with impulsivity and involved in human interaction can be predictive of peculiar writing styles people use in writing chat texts.

\subsection{Summary}\label{sec7:Conce}

The analysis of the relations between personality and writing style has been an hot topic in both computational linguistics and psychology fields; recently, the topic has been renowned with the massive diffusion of the chats, where textual artifacts and elements of the spoken interaction coexist. In this scenario, our long term goal is to evaluate whether the influence of personality on chatting style is stronger than in standard text, and in which measure this is manifested by non verbal signals, that is, aspects that go beyond the semantics of the content. This way, algorithms that recognize personality profiles of the speakers may be employed in a privacy respectful manner, facilitating the conversation with diverse chatting layouts, or encouraging contacts between ``compatible'' subjects. At the same time, we are interested in discovering whether the manifestation of a very particular chatting style, that is, very recognizable, may correspond to having also a distinguishable personality profile. In this way we may understand which are the personality traits that are more evident via chat, and in which measure chats can be considered a transparent means of communication. This study goes in this direction, showing that positive affectivity is connected with the usage few long words, slowly written by following a particular rhythm and this turns out to be the most recognizable trait, followed by non planning impulsiveness,  behavioral inhibition, negative affectivity and planning impulsiveness.

%
%
\section{Deep Learning to Rank Users in Text Chats}\label{sec7:GoingDeep}

Deep Neural Networks (DNNs) have shown to be very effective for image classification, speech recognition and sequence modeling in the past few years. Beyond those applications, out recent research outcomes also show the power of DNNs for other various tasks such as re-identification \cite{paisitkriangkrai2015learning}. In this section we present a pilot experiment which makes use of DNNs in order to transform raw data input to a representation that can be effectively exploited in a re-identification task. The novelty of this task stands in the fact that is does not obviate manual feature engineering and allows a machine to both learn at a specific task (using a simple representation of the features) and learn higher level features. Different from previous works, we formulate a unified deep learning framework that jointly tackles both of these key components to maximize their strengths. We start from the principle that the correct match of the probe template should be positioned in the top rank within the whole gallery set. An effective algorithm is proposed to minimize the cost corresponding to the ranking disorders of the gallery. The ranking model is solved with a deep convolutional neural network (CNN). The CNN builds the relation between each input template and its user's identity through representation learning from frequency distributions of a set of data. \\
The contribution of this work is threefold. Firstly, we provided a new CNN architecture with a reduced number of parameters which allows learning from few training examples. Secondly, the designed network consists of 1D convolutional filters that result to be useful when learning from vectorized data. Thirdly, we showed how it is possible to use heterogeneous data sources, combine them together and use this data to feed a CNN. 
Our hypothesis is that deep neural networks with shallow architectures can perform even better that very deep networks when they are fed with pre-processed features rather than raw data. The pros of this strategy is threefold. Firstly, shorter training times. A shallow architecture has much less parameters and learning can be performed also without expensive hardware and GPUs. Secondly, simplification of models to make them easier to interpret by researchers/users. We know that high-level features come from specific cues designed a priori. Finally, enhanced generalization by reducing overfitting. Since the amount of parameters is much less than a deep architecture the risk to incur in overfitting on limited amount of samples is reduced. A comparative evaluation is given, demonstrating that the proposed approach significantly outperforms all the approaches proposed in this chapter.

\subsection{Deep Learning Framework}\label{sec7:Framework}

\begin{figure*}[!]
\centering
\includegraphics[width=0.95\textwidth]{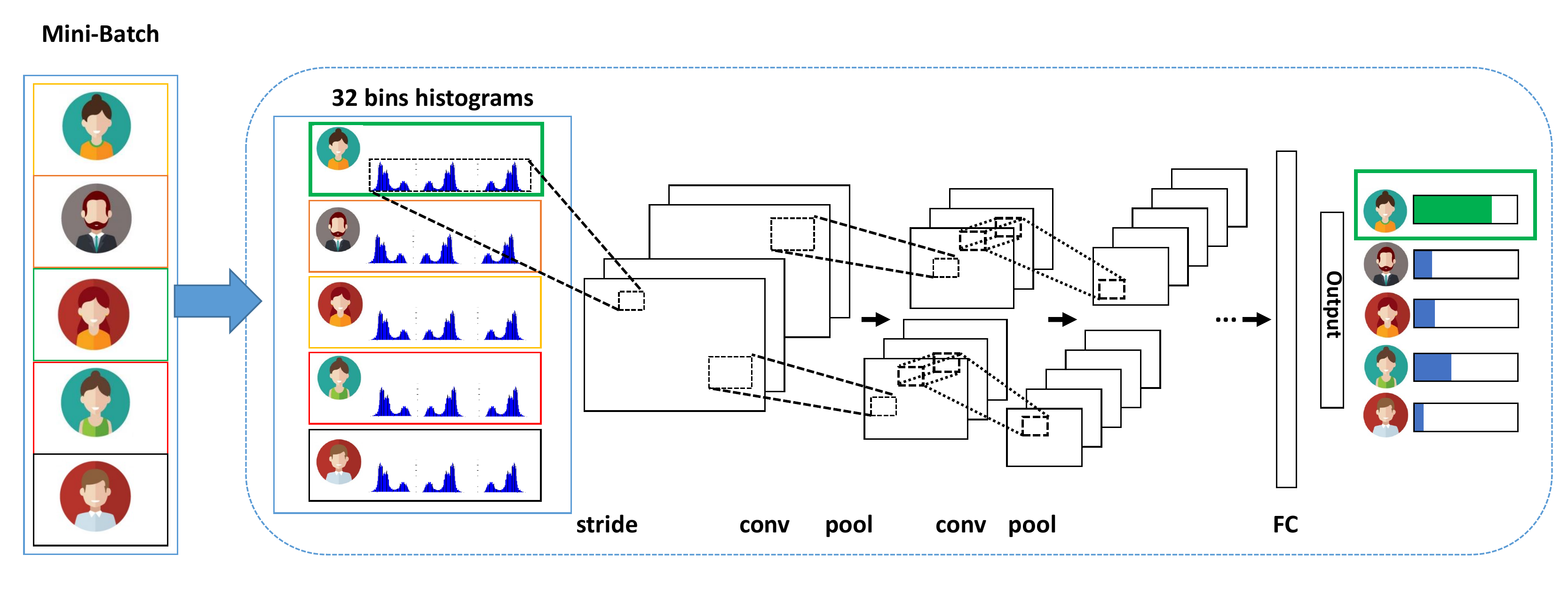}
\caption{Illustration of the proposed deep learning framework, which comprises two key components: a first level representation of features (32 bins histograms) and deep feature learning. We aim to learn a deep CNN that assigns a higher similarity score to the positive pair (marked in green) than any negative pairs in each mini batch. Best viewed in color.}\label{fig7:CNNReIDx}
\end{figure*}

Figure \ref{fig7:CNNReIDx} gives an illustration of the proposed framework. At the training stage, the labeled data are organized into mini batches and then fed into the deep CNN. Since the correct match should be positioned at the top of the gallery, we perform ranking as a classification problem, by minimizing the objective cross-entropy cost function accounting for positive pairs $<template, label>$ in each mini batch. The learnt CNN conducts similarity computing in one shot at the test time. 
Before continuing into the formulation, we describe some of the terminologies associated with our problem that will be used later. Without loss of generality, let us consider solving the following user re-identification problem in a single-shot case. Suppose we are given a training set $X = \{ (x_i,y_i), i = 1,2,...,N \}$, where  $(x_i,y_i)$ is a pair of the template of the $i^{th}$ user in the gallery set and their label from ground truth, $N$ is the number of users. For a probe template $x$ to be matched against gallery set $G$, a ranking list should be generated according to the similarity between $x$ and each template in $G$.  There exists only one correct match $x^+$, which should be placed in the top rank by the learnt ranking model. All other samples in the gallery space are considered to be negative matches, denoted by $G^-$. Intuitively, if the learnt ranking model is perfect, the correctly
matched pair will be assigned a higher similarity score than a mismatched one, which can be expressed as
\[
    S(x^+ | x) > S(z | x ), \forall z \in G^-,
\]
where $S(\cdot): \mathcal{X} \to \Re$ is the learnt similarity metric. The rank of the probe template $x$ with respect to $G$ can be expressed as the list of descending probabilities (i.e., $\approx similarities$) at the output of our CNN network, see Figure \ref{fig7:CNNReID}.

\subsection{The $1D$-CNN Architecture}\label{sec7:Architecture}

\begin{figure*}[!]
\centering
\includegraphics[width=1.0\textwidth]{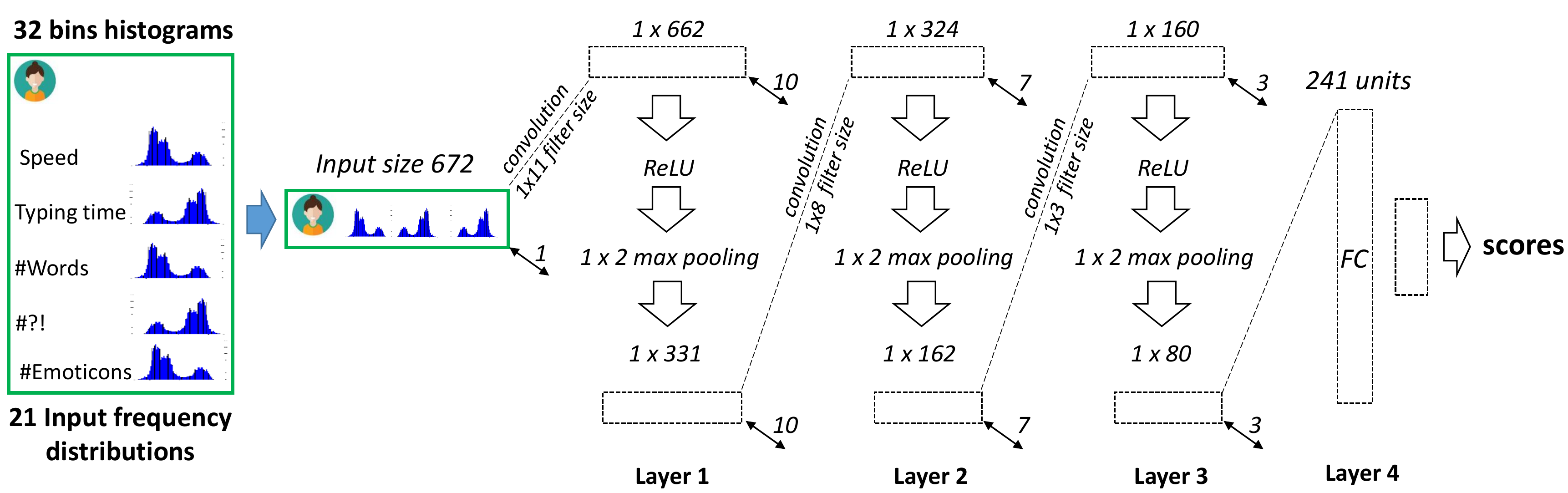}
\caption{Architecture of our deep network. Each conversation in the gallery set is first pre-processed, features are extracted and represented by 32-bins histograms. Then a $ 1 \times 672$ feature vector is presented as the input, which is convolved with 10 different first layer filters, each of size $1 \times 11$. The resulting feature maps are then passed through a rectified linear unit, max-pooled ($1 \times 2$ regions) to give 10 different $1 \times 331$ feature maps. Similar operations are repeated in the second to third layer. The last layer is fully connected, taking features from the top convolutional layer as the input in vector form. Finally, a classification score for each sample is returned.}\label{fig7:CNNReID}
\end{figure*}

Our deep network comprises three convolutional layers to extract features hierarchically, followed by a fully connected layer. The CNN layer configurations are designed using the principles inspired by \cite{KrizhevskyNIPS2012}. Figure\ref{fig7:CNNReID} shows the detailed structure of our network. A notable difference between classic deep learning and the proposed solution is that we propose to learn via 1D convolutions instead of the 2D ones. 1D convolutions naturally account for the 1D structure of input data. Convolutions can also be thought of as regular neural networks with two constraints: Local connectivity and Weight sharing. 
The former comes from the fact that a convolutional filter has much smaller dimensions than the input data on which it operates. The latter comes from the fact that in such a framework the same filter is applied across the input samples. This means we use the same local filters on many locations in the data. In other words, the weights between all these filters are shared. Convolutional layers are followed by a non-linear operation that substitutes the neuron activation function. Non-linearities between layers ensure that the model is more expressive than a linear model. Among the many possible activation functions we use the Rectified Linear Unit (ReLU) which is defined in section \ref{sec3:relu}. ReLU avoids zig-zagging dynamics in the gradient descent optimization algorithms, greatly accelerate the convergence of stochastic gradient descent compared to the sigmoid/tanh functions, and it can be easly implemented by thresholding a matrix of activations at zero. In addition to the convolutional layers and non-linearities just described, this network also contains pooling layers. The pooling layers function is to simplify the output information from the convolutional layer. We propose a $1 \times 2$ max pooling, which partitions the input into a set of non-overlapping sub-vectors and, for each such sub-vector, outputs the maximum value. These pooled features can then be used for the next steps. 

Very deep convolutional networks with hundreds of layers have led to significant reductions in errors on competitive benchmarks. Although, it is worth noting that research on the possibility of using shallow architectures has been conducted over the last few years. This is the case of \cite{BaCaruanaNIPS2014} where the authors provided evidence that shallow feed-forward nets could learn the complex functions, previously learned by deeper networks, and achieved accuracies previously only achievable with deeper models. They empirically showed that single-layer fully connected feedforward nets trained to mimic deep models can perform similarly to well-engineered complex deep architectures. 

One of the goals of this work is to fill, at least partially, the gap above and to investigate whether well balanced deep networks  - meaning the amount of parameters relative to the number of observations - can perform better of complex very-deep convolutional architectures because of overfitting. Indeed, overfitting occurs when a model is excessively complex, for example, when having too many parameters relative to the number of observations. A model that has been overfit has poor predictive performance, as it overreacts to minor fluctuations in the training data. In those situations in which increasing the amount of training data is not possible or difficult to do, a possible solution is to reduce the number of network layers $L$ so as to obtain far fewer parameters. 

As a result, our solution consists of only four layers and few parameters respect to the AlexNet. Indeed, in re-identification (and in many others pattern recognition scenarios as well, for example in bioinformatics applications) the number of training samples is not large as the one in image classification, where tens of thousands of images are available.

\subsection{Vectorized Representations for Heterogeneous Data}\label{sec7:Heterogeneous}

Vectorized data representations frequently arise in many data mining applications. They are easier to handle since each data can be viewed as a point residing in an Euclidean space. Thus, similarities between different data points can be directly measured by an appropriate metric to solve traditional tasks such as classification, clustering and retrieval. Unfortunately, many data sources cannot be naturally represented as vectorized inputs. In our case, a text chat conversation can be seen as a set of turns, where each turn is a sequence of characters and symbols. Many features can be extracted from them, as we showed so far. On the other hand, text chats also contain the detailed timing information which describes exactly when each key was pressed and when it was released as a person is typing at a computer keyboard. Such a kind of data, which describes keystroke dynamics, is represented by lists of timestamps and key-pressed pairs. As a result, there is a need of combining these heterogeneous data sources (text plus meta-data) to come up with meaningful and stronger results. The basic assumption is that, once the vectorized representation is obtained, the mining tasks can be readily solved by deep learning algorithms. To address the aforementioned challenge, we present an interesting idea to representation learning which learn from heterogeneous data. At the training stage, data a firstly represented by \textit{histograms}. According to our previous work (see section \ref{sec7:Features}), we extracted the set of 21 stylometric features, including the ones related to keystroke dynamics, and then we represented these cues as 32-bins histograms which are then used to feed the deep CNN (see Figure\ref{fig7:CNNReID}). As a result from our findings, we noted that some of the features extracted consisted of very smaller values, i.e., many observations around zero, in those situations, we adopted exponential histograms. Small-sized bin ranges are located around zero, and they increase their sizes while going to higher values. It results to give a higher resolution in those locations where most of the data is present.

\subsection{Training Strategies and Optimization}\label{sec7:learning}

In the previous section we presented the details of our network configuration. In this section, we
describe the details of classification 1D-CNN training and evaluation. The CNN training procedure generally follows Krizhevsky et al. \cite{KrizhevskyNIPS2012}. Namely, the training is carried
out by optimising the multinomial logistic regression objective using mini-batch gradient descent with cross entropy cost function. When training neural networks cross entropy is a better choice of cost function than a quadratic cost function. A detailed justification is given in section \ref{sec3:crossEntropyCost}. 
The use of SGD in the deep neural network setting is motivated by the high cost of running back propagation over the full training set, since SGD can overcome this cost and lead to fast convergence.

Interesting techniques to reduce overfitting with fixed network depth and fixed training data are known as regularization techniques described in section \ref{sec3:Overfitting}. We use $L2$ regularization to train our network. The idea of L2 regularization is to add an extra term to the cost function, a term known as the \textit{regularization term}.
\begin{equation}\label{eq:RegCrossEntropy}
    C = -\frac{1}{n} \sum_{x_j}  \Big[ y_{j} ln(a^L_{j}) + ( 1 - y_{j}) ln(1 - a^L_{j})  \Big] + \frac{\delta}{2n} \sum_w w^2,
\end{equation}
where $L$ denotes the number of layers in the network, and $j$ stands for the $j^{th}$ neuron in the last $L^{th}$ layer. The first term is the common expression for the cross-entropy in $L$-layer multi-neuron networks, the regularization term is a sum of the squares of all the weights in the network. The regularization term is scaled by a factor $\frac{\delta}{2n}$, where $\delta > 0$ is known as the \textit{regularization parameter}, and $n$ is the size of the training set (see Table~\ref{tab:notation11} for the used notation). Intuitively, regularization can be viewed as a way of compromising between finding small weights and minimizing the original cost function. The relative importance of the two elements of the compromise depends on the value of $\delta$. This kind of compromise helps reduce overfitting. In our case, the training was regularised by weight decay (the L2 penalty multiplier set to $5 \cdot 10^{-4}$).
\begin{figure*}[!]
\centering
\includegraphics[width=0.65\textwidth]{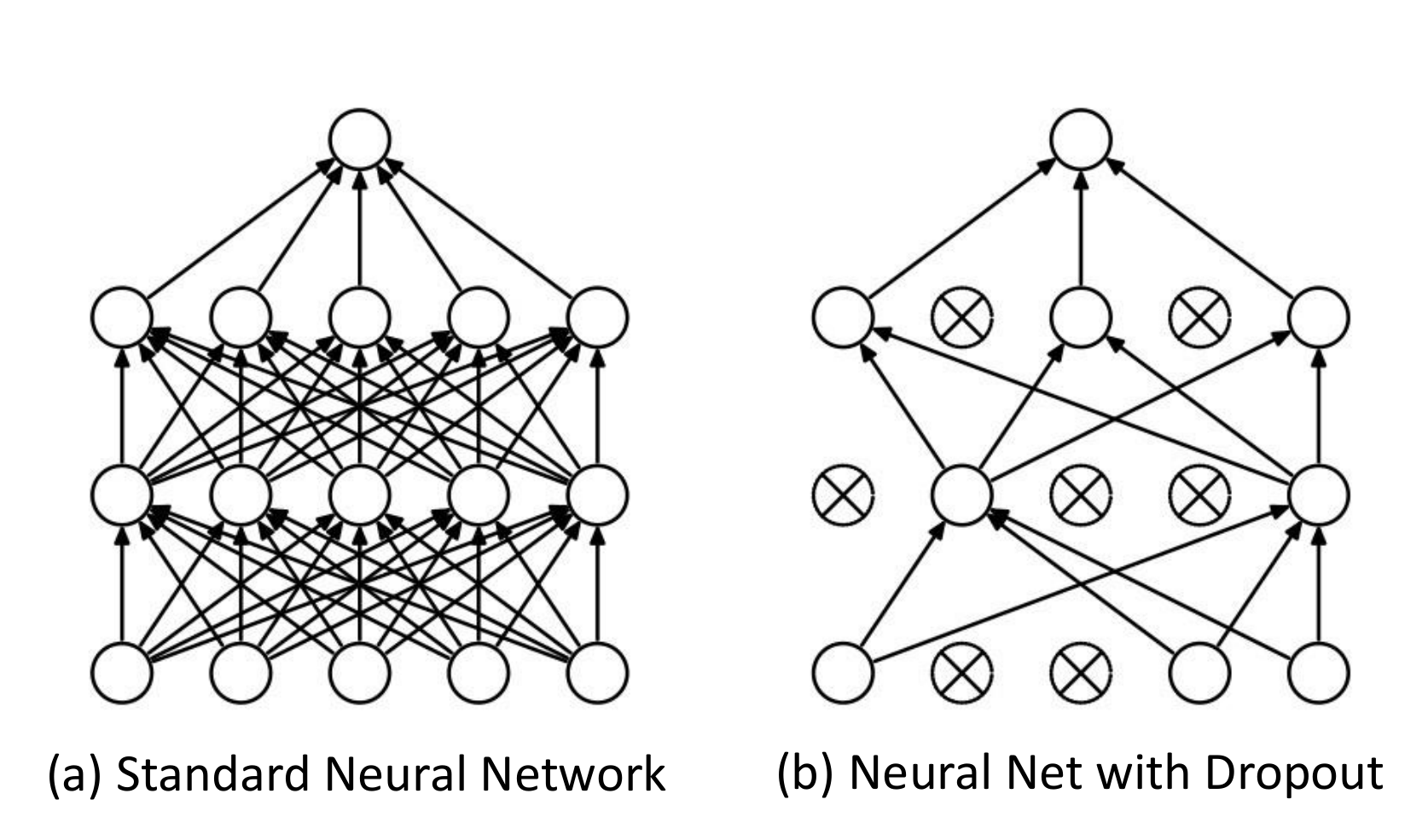}
\caption{An illustration of the dropout mechanism within the proposed CNN. (a) Shows a standard neural network with $2$ hidden layers. (b) Shows an example of a thinned network produced by applying dropout, where crossed units have been dropped.}\label{fig7:DropoutOur}
\end{figure*}
Another strategy adopted during the training phase is \textit{dropout}. Dropout prevents overfitting and provides a way of approximately combining many different neural network architectures exponentially and efficiently. Indeed, dropout can be viewed as a form of \textit{ensemble learning} (see Figure \ref{fig7:DropoutOur}). By dropping a unit out, we mean temporarily removing it from the network, along with all its incoming and outgoing connections. In particular, given a training input $x$ (a mini-batch of gallery examples) and corresponding desired output $y$ (e.g., the class labels of our gallery samples). In our network, dropout, randomly and temporarily deletes the $30\%$ (dropout ratio set to 0.3) of the hidden units in the network, while leaving the input and output units untouched. After this step, the training procedure starts by forward-propagating $x$ through the network, and then back-propagating to determine the contribution to the gradient, therefore updating the appropriate weights and biases.

The learning rate was initially set to $10^{-2}$, and then decreased by a factor of $10$ when the validation set accuracy stopped improving. In total, the learning rate was decreased 3 times, and the learning was stopped after only $20$ epochs. We conjecture that in spite of the smaller number of parameters and the lesser depth of our nets compared to \cite{KrizhevskyNIPS2012}, the nets required less epochs to converge to two (a) implicit regularization imposed by lesser depth and smaller convolutional sizes filter (1 dimension); (b) a good pre-initialisation of filters. The initialisation of the network weights is important, since bad initialisation can stall learning due to the instability of gradient in deep nets. The reason is that the logistic function is close to flat for large positive or negative inputs. In fact, if we consider the derivative at an input of $2$, it is about $\frac{1}{10}$, but at $10$ the derivative is about $\frac{1}{22,000}$. This means that if the input of a logistic neuron is $10$ then, for a given training signal, the neuron will learn about $2,200$ times slower that if the input was $2$.

To circumvent this problem, and allow neurons to learn quickly, we either need to produce a huge training signal or a large derivative. To make the derivative large, the inputs are normalized to have mean $0$ and standard deviation $1$, then we initialized the biases to be $0$ and the weights $w$ at
each layer with the following commonly used heuristic:
\[
    w \in \Big[ \frac{-1}{\sqrt{n}},\frac{1}{\sqrt{n}} \Big]
\]
where $[−a, a]$ is the uniform distribution in the interval $(−a, a)$ and $n$ is the size of the previous layer. The probability that we get a sum outside of our range is small. That means as we increase $n$, we are not causing the neurons to start out saturated \cite{Glorot10understandingthe}.

\subsection{Deep Verification and Recognition }\label{sec7:ExpRes2}

We performed identity recognition in order to investigate the ability of the system in recognizing a particular probe user among the gallery subjects. To this end, we consider conversations which are 55 turns long. After this, we analyze the user verification: the verification performance is defined as the ability of the system in verifying if the person that the probe user claims to be is truly him/herself, or if he/she is an impostor. As a performance measure for the identity recognition, we used the Cumulative Matching Characteristic (CMC) curve like in the previous studies. As comparative approaches, we consider the strategies proposed in \cite{Roffo:icmi2014,Roffo2013}. Given a test sample, we want to discover its identity among a set of N subjects (i.e., N=$94$ in $C$-$Skype$ and N=$50$ in TBKD). In particular, the value of the CMC curve at position $1$ is the probability (also called $rank_1$ probability), that the probe ID signature of a subject is closer to the gallery ID signature of the same subject than to any other gallery ID signature; the value of the CMC curve at position $n$ is the probability of finding the correct match in the first $n$ ranked positions.  As a single measure to summarize a CMC curve, we use the normalized Area Under the Curve (nAUC),which is the approximation of the integral of the CMC curve. For the identity verification task, we report the standard ROC and precision-recall curves. Results show that the deep learning approach is more efficient.

\subsubsection{Identity Recognition}

In the identity recognition task, we performed the same experiments in \cite{Roffo:icmi2014,Roffo2013}. We built a training set, which for each person has a particular number of turns, that is used by the 1D-CNN algorithm to train the system. After training, we applied our approach on the testing set, which was composed of the same number of turns as the training data. The 1D-CNN is used to classify the set of unknown user templates $x_i$ by minimizing the objective cross-entropy cost function of positive pairs $<template, label>_i$. 
The recognition task is performed, then the CMC curve and the related nAUC value calculated. We did the same with the comparative approaches (i.e., \cite{Roffo:icmi2014,Roffo2013} simply calculate distances among features, and computes the average distance among the probe and training conversations). All the experiments were repeated 50 times, by shuffling the training/testing partitions. The results are better with our proposal both in case on nAUC and rank 1 score.

\begin{figure*}[!]
\centering
\includegraphics[width=1.0\textwidth]{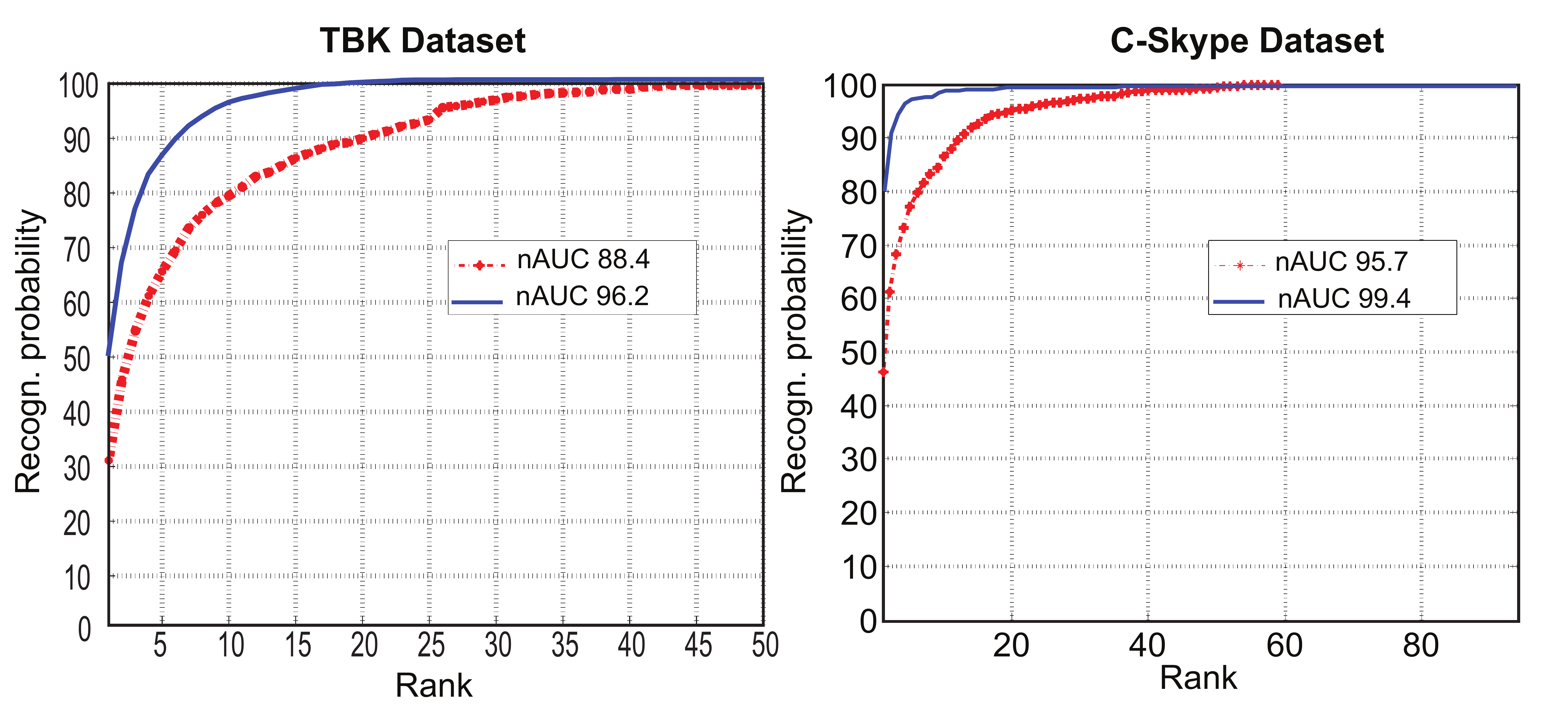}
\caption{Comparison with the previous work. (left) CMC on the typing biometrics keystrokes dataset; (right) Global CMC Curve on C-Skype dataset. nAUC and rank1 probability are reported in the legend.}\label{fig7:DeepRankCMC}
\end{figure*}

In Fig.~\ref{fig7:DeepRankCMC} is reported the CMC curves obtained from the Deep Ranking approach, which gives the identity recognition performance.
This curve is strongly superior than all the others of our previous work, realizing an nAUC of $96.2\%$ and $99.4\%$ on $TBK$ and $C$-$Skype$ respectively. It is worth noting that, the probability of guessing the correct user at $rank$ $1$ is slightly below $50\%$ which is really encouraging. Rank 1 of CMC curves tells us what is the capability of the re-identification system in correctly recognizing the right identity from the gallery database. Interestingly, in the case of the $C$-$Skype$ we derive a $80\%$ of accuracy performed by deep learning on the corpus of $94$ different users. This result strongly overcomes the previous system performance where learning was not used.

\subsubsection{Identity Verification}

Considering the verification task, we adopted the following strategy: given the signature of user $i$, if it matches with the right gallery signature with a matching distance which is ranked below the rank $K$, it is verified. Intuitively, there is a trade-off in choosing $K$. A high $K$ (for example, 94, all the subjects of the $C$-$Skype$ dataset) gives a 100\% of true positive rate (this is obvious by construction), but it brings in a lot of potential false positives. Therefore, taking into account the number $K$ as varying threshold, we can build ROC and precision/recall curves, using the value $K$ as varying parameter to build the curves.
\begin{figure*}[!t]
\centering
\includegraphics[width=0.85\textwidth]{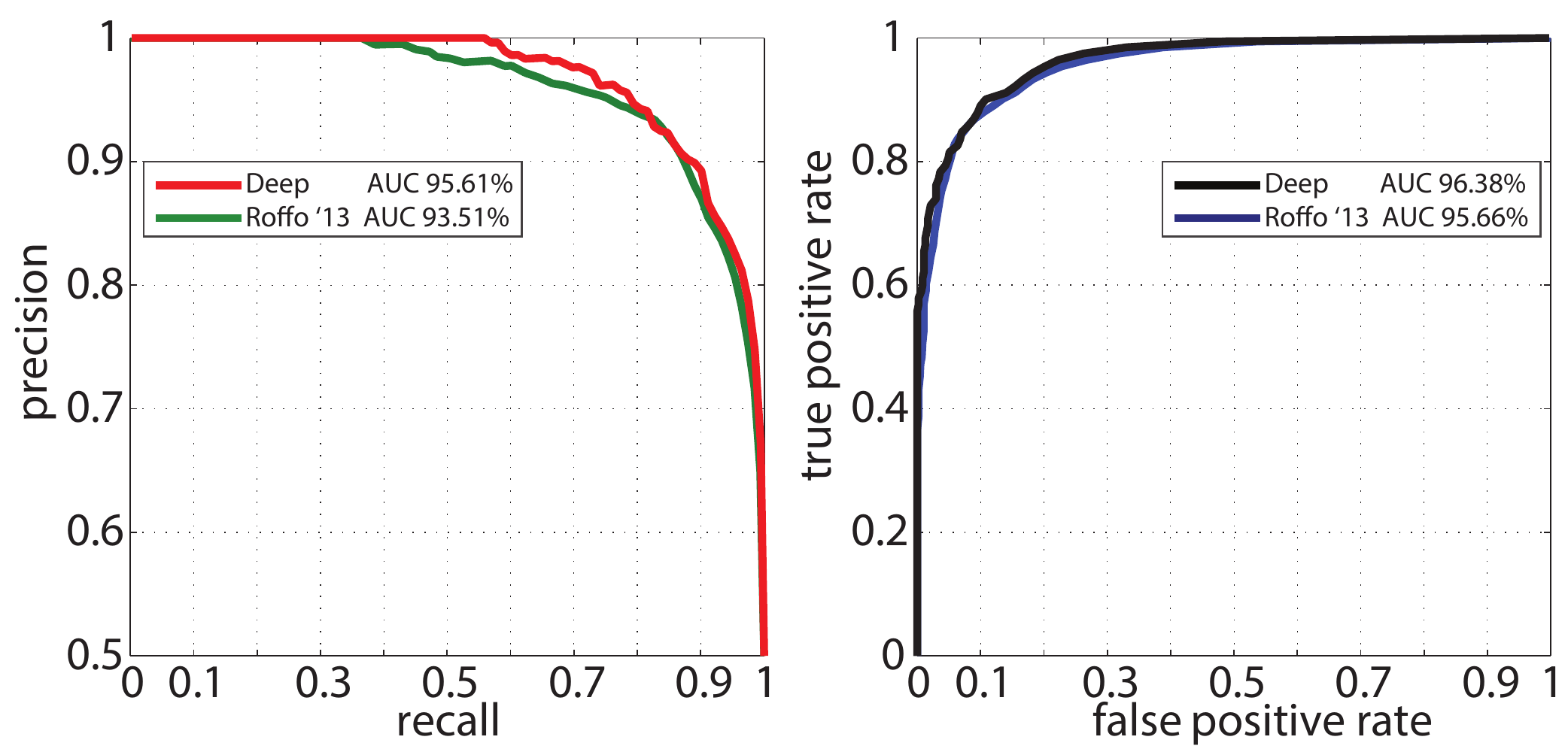}
\caption{Precision-Recall and ROC Curves on the $C$-$Skype$ dataset, N = 94 subjects. Comparison of the deep ranking approach against the Roffo '13 \cite{Roffo2013}.}\label{fig7:PR_ROC1}
\end{figure*}
\begin{figure*}[!t]
\centering
\includegraphics[width=0.85\textwidth]{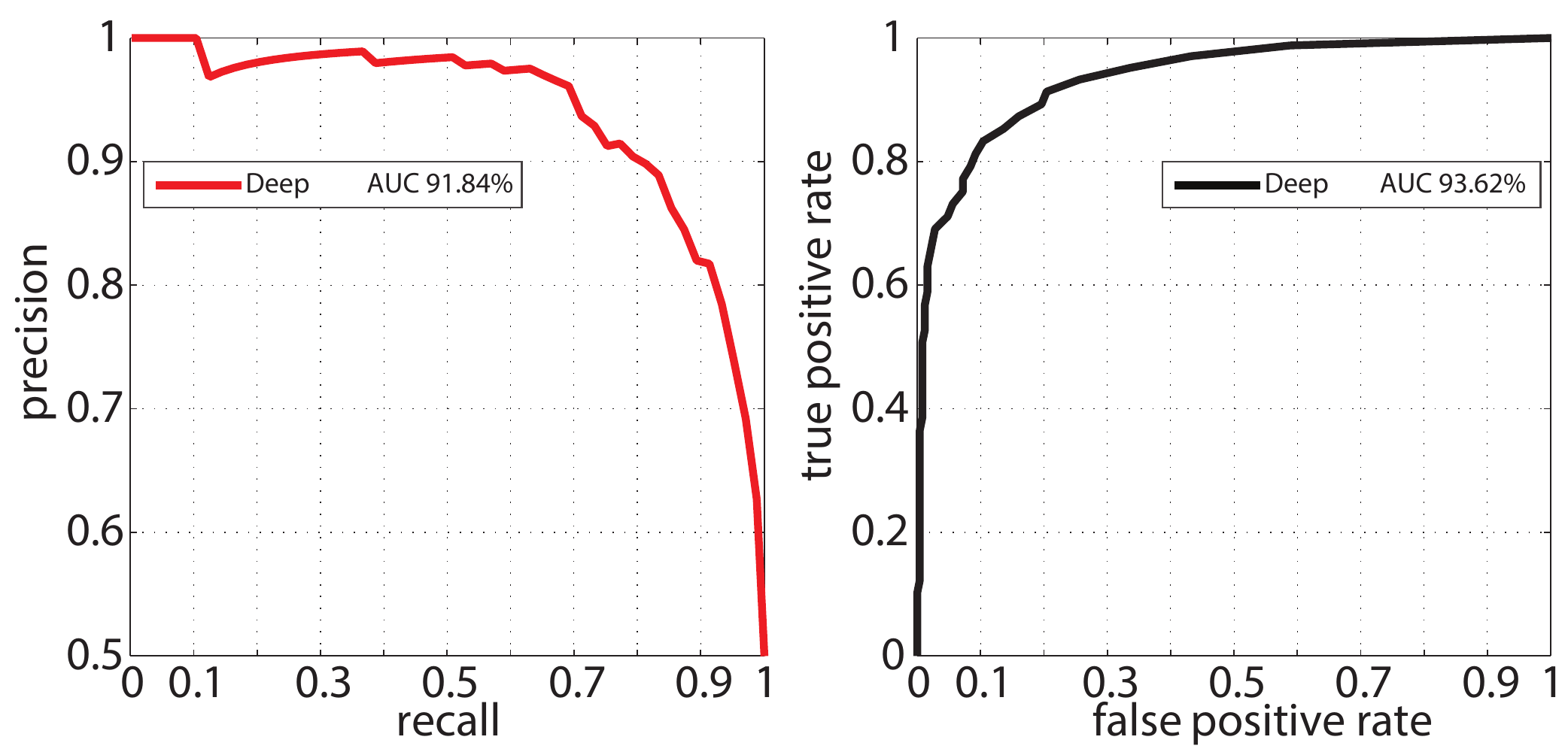}
\caption{Precision-Recall and ROC Curves on the TBK dataset \cite{Roffo:icmi2014}, N=50 subjects.}\label{fig7:PR_ROC2}
\end{figure*}
Considering the nAUC of both the Precision-Recall curves in Figure \ref{fig7:PR_ROC1}, ranking identities by means of deep learning increases the performance from 93.51\% to 95.61\%. On the other hand, the area under the ROC curve increases by 0.72\%. The best compromise between precision and recall is obtained calculating the F1 value, which gives, for our previous work \cite{Roffo2013}, 0.88 for precision 0.90 and recall 0.87, corresponding to the value of K= 45. F1 score of deep ranking is up to 0.89 corresponding to the value of k = 48. The precision of the system at this threshold is 0.89 and recall 0.90.

The same verification experiment is repeated on the TBK dataset, where the number of users is up to 50. Differently from the $C$-$Skype$, the TBK allows the use of different features based on timestamps and accounting for typing rhythms. Therefore, it is of interest to test this features in a deep learning framework. Figure \ref{fig7:PR_ROC2} shows both precision-recall and ROC curves. The F1 score is up to 0.86 corresponding to the value of k = 45. The precision of the system at this threshold is 0.82 and recall 0.91.

\subsection{Summary}
Our hypothesis is that deep neural networks with shallow architectures can perform even better that very deep networks when they are fed with pre-processed features rather than raw data. The pros of this strategy is threefold. Firstly, shorter training times. A shallow architecture has much less parameters and learning can be performed also without expensive hardware and GPUs. Secondly, simplification of models to make them easier to interpret by researchers/users. We know that high-level features come from specific cues designed a priori. Finally, enhanced generalization by reducing overfitting. Since the amount of parameters is much less than a deep architecture the risk to incur in overfitting on limited amount of samples is reduced. A comparative evaluation is given, demonstrating that the proposed approach significantly outperforms all the approaches proposed in this chapter.

The main contribution of this pilot experiment is the investigation of a possible application of DNNs on limited amount of data samples. A shallow architecture has much less parameters and learning can be performed also without expensive hardware and GPUs. It results in shorter training times. We feed the deep network with pre-processed features (represented as histograms). We know that high-level features will be automatically learnt by the network starting from specific cues designed a priori. This fact makes results easier to interpret by researchers. Moreover, since the amount of parameters is much less than a deep architecture the risk to incur in overfitting is reduced. 

The choice of using 1D convolutions is motivated by the nature of the input data, since histograms when normalised can be interpreted as an estimate of the probability distribution of a continuous variable, the deep network will learn specific filters (and hierarchical representations in increasing levels of abstraction) capturing complex patterns among input handcrafted representations. This means that they automatically generate other features from inputs with a higher discriminative power. 

As a result, we can interpret our results, and saying something about the conversational nature of the texts typed during chat exchanges.

\section{Social Signals Through Online Textual Chats}

After the advent of the Internet, the term ``\emph{chat}'' refers not only to informal conversations,
but also to any form of synchronous textual communication. In the most general case, chats 
involve multiple users that, typically, do not interact directly with one another, but post messages 
that can be read - and possibly  reacted upon - by a potentially large number of people using the 
same technological platform~\cite{Uthus2013}. In the particular case of dyadic chats, the participants 
are only two and the interaction is direct. In other words, dyadic chats can be thought of as dyadic
conversations that take place through a textual interaction platform rather than face-to-face. Thus, 
dyadic chats should involve the same social and psychological phenomena that can be observed in 
face-to-face conversations. In particular, dyadic chats should involve the use of nonverbal
behavioural cues capable to convey socially and psychologically relevant information about
the interactants~\cite{Pentland2010}. Still, to the best of our knowledge, no major efforts have been 
done in the computing community to verify whether this is actually the case or not.

The goal of this last part is to fill, at least partially, the gap above and to investigate whether the
typing behaviour of people - meaning the \emph{way} people type and not \emph{what} they 
type - provides information about two of the most important characteristics of an individual, 
namely gender and type of role (\emph{proactive} or \emph{reactive}). In particular, 
the experiments show that both characteristics can be predicted with $79$\% accuracy
without taking into account the content of the data. This is important under at least two
main respects. The first is that it is possible to detect fraudulent attempts to conceal
one's identity or to steal the identity of others, something that might be easier to do through 
a textual interface~\cite{Roffo2013}. The second is that chats play an increasingly more important role
in business-to-customer interactions: ``\emph{Nearly one in five online US 
consumers has used chat - reactive or proactive - for customer service in the past 12 
months}''~\cite{Clarkson2010}. The inference of socially and/or psychologically relevant information 
can be beneficial in both cases.

The proposed approach is based on features that have been shown to be effective in biometrics 
and authorship attribution applications~\cite{Roffo2013,Cristani:2012}. However, compared to these works and previous approaches in the literature, this article takes into account the temporal aspects of
typing behaviour. In particular, the automatic recognition is performed with Conditional Random Fields 
that take as  input the temporal sequence of the key-strokes. This is possible thanks to the key-logging platform adopted to collect the data and, to the best of our knowledge, it has not done previously. 

The experiments have been performed over the $SSP$-$Chat$ $Corpus$ introduced in section \ref{ch4:SSPChat}. $SSP$-$Chat$ is a collection of $30$ dyadic
chat conversations between fully unacquainted individuals, for a total  of $60$ subjects 
($35$ females and $25$ males). The conversations revolve around the Winter 
Survival Task, a scenario in which the
subjects are asked to identify items that increase the chances of survival after a plane 
crash in Northern Canada~\cite{Joshi2005}. In particular, the subjects are given a list
of $12$ items and they have to make a consensual decision for each of them (``\emph{yes}'' 
if it increases the chances of survival and ``\emph{no}'' if it does not). The main
advantage of the scenario is that people, on average, do not hold any expertise
relevant to the topic. Thus, the outcome of the conversations depends on social
dynamics rather than on actual knowledge about the problem.

A chat can be thought of as a stream of keys that are typed sequentially by the
subjects. The sequence can be split into \emph{tokens}, i.e., strings of non-blank
characters delimited by blank spaces. The \emph{rationale} behind this choice
is that such strings should correspond to semantically meaningful units.
Overall, the data includes a total of 33,085 tokens, 21,019 typed by the female 
subjects and 12,066 typed by the male ones. In every dyadic 
conversation, one of the subjects is \emph{proactive} (the person that starts
the chat) and the other is \emph{reactive} (that person that joins the chat). 
Every subject has been randomly assigned one of these two roles.

During the chats, the subjects must press the ``\emph{Enter}'' key to send a sequence
of tokens to their interlocutor. Some subjects press the Enter key every few words
while others do it only after having written long and articulated messages. In
both cases, the chat can be segmented into \emph{chunks}, i.e., sequences of tokens 
delimited by Enter keys. The chunks are the analysis units of the experiments.
\subsection{Approach}\label{sec7:approach}
The approach proposed in this work is applied individually to each of the chunks extracted 
from the dataset and it includes three main steps: the first
is the conversion of each chunk into a vector of features (the \emph{feature extraction}),
the second is the classification of the chunks into socially relevant categories (male/female
or reactive/proactive in the experiments of this work), the third is the aggregation of
the results obtained at the chunk level to obtain gender and role of
a given subject. The rest of this section describes the three steps in detail.
\subsubsection{Feature Extraction}\label{sec7:fsextr}
Every chunk of the dataset is mapped into a feature vector of dimension $D=19$.
The features are non-verbal, i.e., they do not take into account what the subjects type,
but how they type it. The main reasons behind this choice are two. The first is that
applications running in real-world scenarios are likely to be more accepted if they will
respect the privacy of the users. The second is that research on face-to-face interactions
shows that nonverbal behaviour is more likely to leak \emph{honest} information
about social and psychological phenomena. 
\begin{table*}[!ht]
{\small
\centering
  \resizebox{1\textwidth}{!}{%
\begin{tabular}{|c|m{14.0cm}|c|}
\hline
\textbf{Category}  & \textbf{Description} & \textbf{Ref.}\\\hline
\textbf{Token} 
& Total number of tokens (F1); Mean token length (F2); Normalized number of tokens (F3);
Total number of characters (F4); Normalized number of characters (F5). & \cite{Cristani:2012,Roffo2013}
\\\hline
\textbf{Syntactic} &  Number of punctuation marks (F6); Number of ``?'' (F7);  number of ``!'' (F8);  
Number of Back-Spaces (F9);
Normalized number of punctuation marks (F10); 
Normalized number of ``?'' (F11); Normalized number of ``!'' (F12);  Normalized number of 
Back-Spaces (F13); Number of emoticons (F14). & \cite{Roffo2013,Roffo:HBU2014}
\\\hline
\textbf{Chunk} & Chunk duration (F15); Silence time (F16); Typing ratio (F17); Deletion time (F18); Mean typing speed (F19) &\cite{Roffo:HBU2014,Roffo:icmi2014} 
\\\hline	
\end{tabular}
}
  \caption{Synopsis of the features.}
  \label{table:feats}
  }
\end{table*}
The features can be grouped into three categories (see Table~\ref{table:feats}). The first, 
called ``\emph{Chunk}'',  aims at capturing how frequently a subject splits its messages into
chunks and how long these latter are. This is important because it accounts for how the
subjects manage their social presence~\cite{Weinel2011}. When the chunks are short and the 
Enter key is pressed frequently, the users are actually trying to project their social presence
to their interlocutor (the text becomes visible to the other subject of the dyad only after
pressing the Enter key). 

The second category, called ``\emph{Syntactic}'' accounts for how formal (similar
to written text) or informal (similar to spoken conversation) the text of a user is. The number of 
punctuation marks shows how much a user respects the grammar and expresses
intonation according to the way it is done in written texts. The number of deletions
measures how disfluent the typing is, accounting for planning difficulties in elaborating
the text. Finally, the number of emoticons measures how much
the subjects favour the use of informal expressive means over formal expressions
of the affective content, based on text and punctuation. 

The third and last category aims at capturing how quickly a subject types. The duration
of a chunk is the length of the interval between two consecutive Enter keys. The silence
length is the amount of time during which a subject does not type. The typing / silence
ratio is the ratio between the time spent in typing and the time spent in not typing.
The deletion time is the amount of time spent in pressing the \emph{Back Space} key and, 
finally, the mean typing speed is the average number of keys per minute during a chunk.
\subsubsection{Chunk Classification} 
After the feature extraction, a dyadic chat can be represented with a sequence of feature 
vectors $X=\{ \vec x_1, ..., \vec x_N \}$, where $\vec x_i$ has been extracted from the 
$i^{th}$ chunk and $N$ is the total number of chunks. Consider the sequence $T= \{t_1,\ldots,t_N\}$,
where $t_i$ represents a socially relevant characteristic of the person that has authored
chunk $i$. The problem of inferring socially relevant information can then be thought of
as finding the sequence $T^* = \{t^*_1,\ldots,t^*_N\}$ that maximizes $p(T|X)$. In this work, this 
task is performed with linear-chain Conditional Random Fields (CRF)~\cite{Lafferty:2001}. The main 
advantage of the CRFs with respect to other probabilistic sequential models (e.g., the Hidden 
Markov Models) is that no assumption is made about the statistical independence of the observation 
vectors. In the experiments of this work, the elements $t_i$ correspond to gender (\emph{male}
or \emph{female}) and role (\emph{proactive} or \emph{reactive}). 

In order to improve the performance of the approach, the sequence $X$ is fed to a
feature selection approach allowing one to identify a subset of features that can guarantee
a performance as high as possible while limiting as much as possible the number of model
parameters. The feature selection approach is the \emph{Infinite Feature Selection} 
(inf-FS)~\cite{Roffo_2015_ICCV}. The Inf-FS is a graph-based method which exploits the convergence 
properties of the power series of matrices to evaluate the importance of a feature with respect to all 
the other ones taken together. Indeed, in the Inf-FS formulation, each feature is mapped on an affinity 
graph, where nodes represent features and weighted edges relationships between them. In particular, 
the graph is weighted according to a function which takes into account both correlations and standard 
deviations between feature distributions. Each path of a certain length $l$ over the graph is seen as 
a possible selection of features. Therefore, varying these paths and letting them tend to an infinite 
number permits the investigation of the importance of each possible subset of features. The Inf-FS 
assigns a final score to each feature of the initial set; where the score is related to how much the 
given feature is a good candidate regarding the classification task. Therefore, ranking in descendant 
order the outcome of the Inf-FS allows us to perform the subset feature selection throughout a 
\textit{model selection stage} to determine the number of features to be selected. In the
experiments of this work, the Inf-FS has also the advantage of providing insights about the
features that drive the outcome of the classification and, indirectly, about the features that
better account for gender and interaction condition.
\subsubsection{Subject Classification}
Finding $T^*=\arg\max_T p(T|X)$ corresponds to classifying individually each chunk of
a chat. However, both characteristics targeted in this work (gender and role) do not change across the chunks authored by the same subject.
Thus, the subjects are assigned the class that has been assigned to the majority of their chunks. In other words, if most of the chunks typed by a subject have been classed as female, then the subject will be classed as female (the same approach is applied for the classification into proactive and reactive). 
For this reason, the performances are not reported in terms of chunk classification
accuracy, but in terms of subject classification accuracy.

\subsection{Experiments and Results}\label{expres}
The experiments aim at two main goals. The first is to identify difference makers
in typing behaviours when it comes to gender and role. The second
is the automatic classification of the subjects in terms of gender and role. 
The rest of this section shows the results that have been obtained.

\subsubsection{Identification of Difference Makers}
For every feature, it is possible to test whether there are statistically significant differences
between subjects belonging to different classes. This shows whether the feature is
a difference maker with respect to the class, i.e., whether the feature plays a role
in making a difference between the subjects that belong to the different classes. When
the feature is a count, then the statistical test is the $\chi^2$. When the feature is continue, 
the difference has been tested with a $t$-test. In both cases, a difference is considered 
statistically significant when the $p$-value is lower than $0.05$.

For what concerns gender, the main difference makers are the characters ``\emph{?}''
and ``\emph{!}''. Female subjects appear to use them more frequently than male ones
to a statistically significant extent. In particular, all the features that involve the exclamation
mark tend to have higher value in the case of female subjects than in the case of
male ones (the difference is statistically significant in all cases). One possible
explanation is the systemising-empathising theory which shows that women tend
to be more capable than men to establish empathic relationships~\cite{Lawson2004}. In this 
respect, the use of the exclamation mark can be interpreted as an attempt to better
communicate and share emotions. Another important difference maker is the use
of the Back-Space key. Female subjects tend to use it more than male ones to a
statistically significant extent and, as a consequence, female subjects tend to take
more time than men to complete a chunk. This seems to suggest that female
subjects tend to make more corrections and changes than male ones before they
send their messages to the interlocutors. One possible explanation is that,
according to the literature, females tend to be less assertive and more
hesitant than males during discussions~\cite{Thompson2010}. The same observation
can explain why the typing speed and typing to silence ratio tend to be higher 
for male subjects than for female ones.

In the case of the role, the main difference makers for the proactive
subjects are related to the length of the chunks. In particular, proactive participants 
tend to write chunks with a larger number of tokens and a larger number of characters. 
Furthermore, proactive subjects tend to be faster in typing. When it comes to reactive
subjects, the difference maker is the use of the Back Space. In particular, reactive
participants tend to use the Back Space more frequently, spend more time in
pressing the Back Space key and have a larger number of Back Space key presses
per chunk. One possible explanation is that proactive participants spontaneously
drive the chat and, therefore, tend to type more while reactive participants tend
to follow and, hence, tend to be more hesitant and less assertive. However, the
literature does not provide indications about these observations.
\begin{table}[!h]
\centering
\begin{tabular}{|l|c|c|c|c|c|c|}\hline
Task & FS & mAP & B.Prec & B.Rec &  AUC &  CCR \\\hline  
Gender & 17 & 77.9\% & 79.4\% & 77.1\% & 70.1\% & 71.7\% \\\hline
Role & 11 & 75.6\% & 69.2\% & 86.2\% & 75.6\% &  74.5\% \\\hline 
\end{tabular}
\caption{The table reports the performances for gender and role recognition. The 
performance measures are as follows (from left to right): mean Average Precision (mAP), F-measure Best Precision (B.Prec), F-measure Best Recall (B.Rec), Correct Classification Rate (CCR) }
  \label{results}
\end{table}
Overall, the most important indication of the corpus analysis is that several features
are difference makers for gender and role. Thus, gender and interaction 
condition tend to leave physical, machine detectable traces in typing behaviour. 
\subsubsection{Classification}
\begin{table*}[!h]
\begin{center}
\begin{tabular}{|c|c|c|c||c|c|c|}\hline
Feat. & Female & Male & G FS & Proac. & Reac. & R FS \\\hline
F1 & - & - & y & - & - & n \\\hline
F2 & - & - & y & $\uparrow$ & $\downarrow$ & y \\\hline
F3 & - & - & n & - & - & n \\\hline
F4 & - & - & y & - & - & n \\\hline
F5 & - & - & n & $\uparrow$ & $\downarrow$ & y \\\hline
F6 & $\uparrow$ & $\downarrow$ & y & - & - & y \\\hline
F7 & - & - & y & - & - & y \\\hline
F8 & $\uparrow$ & $\downarrow$ & y & - & - & n \\\hline
F9 & $\uparrow$ & $\downarrow$ & y & $\downarrow$ & $\uparrow$ & y \\\hline
F10 & - & - & y & - & - & y \\\hline
F11 & - & - & y & $\downarrow$ & $\uparrow$ & y \\\hline
F12 & $\uparrow$ & $\downarrow$ & y & - & - & n \\\hline
F13 & $\downarrow$ & $\uparrow$ & y & $\uparrow$ & $\downarrow$ & y \\\hline
F14 & - & - & n & - & - & n \\\hline
F15 & $\uparrow$ & $\downarrow$ & y & - & - & n \\\hline
F16 & - & - & y & - & - & y \\\hline
F17 & $\downarrow$ & $\uparrow$ & y & $\downarrow$ & $\uparrow$ & y \\\hline
F18 & $\uparrow$ & $\downarrow$ & y & $\downarrow$ & $\uparrow$ & n \\\hline
F19 & $\downarrow$ & $\uparrow$ & y & $\uparrow$ & $\downarrow$ & y \\\hline
\end{tabular}
\caption{The table shows, for every feature, whether it has been retained by Inf-FS
for Gender (G FS) and Role (R FS). Furthermore, it shows whether a feature is
larger ($\uparrow$) or smaller ($\downarrow$) than the other class to a statistically
significant extent.}
\label{selection}
\end{center}
\end{table*}
The classification experiments have been performed using a leave-one-out
approach. Both CRFs and Inf-FS have been trained 
over all subjects except one, then they have been tested over this latter. The
process has been iterated and, at each iteration, a different subject has been left
out. In this way, it has been possible to keep a rigorous separation between
training and test set. Table~\ref{results} shows the classification results. The
performances are better than chance to a statistically significant extent for both
gender and interaction condition.

Table~\ref{selection} shows the results of the feature selection. In the case of
the gender, all the features that were identified as difference makers in the corpus
analysis have been retained by the Inf-FS as well. However, the feature selection
step retains many other features as well. One possible explanation is that these
features carry information useful for the classification, but the gender effects,
if any, are too weak to be observed with the statistical tests. Collecting more
data will show whether this is actually the explanation. In the case of the interaction
condition, all features identified as difference makers have been retained by the
Inf-FS except one (the Deletion time). However, the feature selection step retains
more features like in the case of the gender. The explanation is likely to be the
same, i.e., that the differences related to these features are not statistically
significant because the data collected so far is not sufficient. Still, these
features provide useful information to perform the classification.

\subsection{Summary}\label{concl}
This last section has presented experiments aimed at inferring socially relevant information
(gender and role) from dyadic textual chats. The experiments have been 
performed over a collection of  $30$ chats ($60$ subjects in total) revolving
around the Winter Survival Task, a scenario commonly applied in social psychology.
The features extracted from the data do not take into account what the subjects type, 
but how they type it. In this respect, the features account for the nonverbal aspects
of the interaction. The results show that it is possible to predict the gender and the 
interaction condition of a subject with accuracy up to $79$\%. However, the most important 
result  is that nonverbal features extracted from dyadic chats appear to account for social 
and psychological phenomena like nonverbal behaviour does in face-to-face conversations. 
This suggests that the methodologies of Affective Computing and Social Signal Processing,
so far applied only to face-to-face conversations, can probably be extended to dyadic
chats. Future work will focus on the collection of more data and on the classification in terms 
of other socially relevant characteristics such as, e.g., the personality traits.
\chapter{Ad Recommendation: Matching, Ranking and Personalization}\label{ch8:ADS}

Nowadays, online shopping plays an increasingly significant role in our daily lives \cite{Forrester}. Most consumers shop online with the majority of these shoppers preferring to shop online for reasons like saving time and avoiding crowds. Marketing campaigns can create awareness that drive consumers all the way through the process to actually making a purchase online \cite{Kim:2011}. Accordingly, a challenging problem is to provide the user with a selection of advertisements they might prefer, or predict how much they might prefer the content of each advert.

Past studies on recommender systems take into account information like user preferences (e.g., user's past behavior, ratings, etc.), or demographic information (e.g., gender, age, etc.), or item characteristics (e.g., price, category, etc.).  For example, collaborative filtering approaches first build a model from a user's past behavior (e.g., items previously purchased and/or ratings given to those items), then use that model to predict items that the user may have an interest in by considering the opinions of other like-minded users.  Other information (e.g., contexts, tags and social information) have also taken into account in the design of recommender systems~\cite{Choi2012309,Lee20102142,NzVald20121186}. 

Another interesting cue taken into account is \textit{personality}. The impact of personality factors on advertisements has been studied at the level of social sciences and microeconomics~\cite{Bosnjak2007,soBob,Turkyilmaz201598}. Recently, personality-based recommender systems are increasingly attracting the attention of researchers and industry practitioners~\cite{Cosley:2003,Rong:7926302,Tka2010}. Personality is the latent construct that accounts for ``\textit{individuals characteristic patterns of thought, emotion, and behavior together with the psychological mechanisms - hidden or not - behind those patterns}"~\cite{Funder}. Hence, personality is a critical factor which influences people's behavior and interests. Attitudes, perceptions and motivations are not directly apparent from clicks on advertisements or online purchases, but they are an important part of the success or failure of online marketing strategies. A person's buying choices are further influenced by psychological factors like impulsiveness (e.g., leads to impulse buying behaviors), openness (e.g., which reflects the degree of intellectual curiosity, creativity and a preference for novelty and variety a person has), neuroticism (i.e., sensitive/nervous vs. secure/confident), or extroversion (i.e., outgoing/energetic vs. solitary/reserved) which affect their motivations and attitudes~\cite{Turkyilmaz201598}. 

Personality has shown to play an important role also in other aspects of recommender systems, such as implicit feedback, contextual information~\cite{Odi}, affective content labeling~\cite{TkalcicLabeling}. With the development of novel techniques for the unobtrusive acquisition of personality (e.g. from social media~\cite{Roffo:HBU2014,Roffo:icmi2014}) this study is meant to contribute to this emerging domain proposing a new corpus which includes questionnaires of the Big-Five (BFI-10) personality model~\cite{rammstedt:2007}, as well as, users' liked/disliked pictures that convey much information about the users' attitudes. There is a high potential that incorporating users' characteristics into recommender systems could enhance recommendation quality and user experience. For example, given a user's preference for some items, it is possible to compute the probability that they are of the same personality type as other users, and, in turn, the probability that they will like new items~\cite{Pennock:2000}. 

\begin{figure}[!t]
    \centering
    \begin{subfigure}[b]{0.49\textwidth}
        \includegraphics[width=\textwidth]{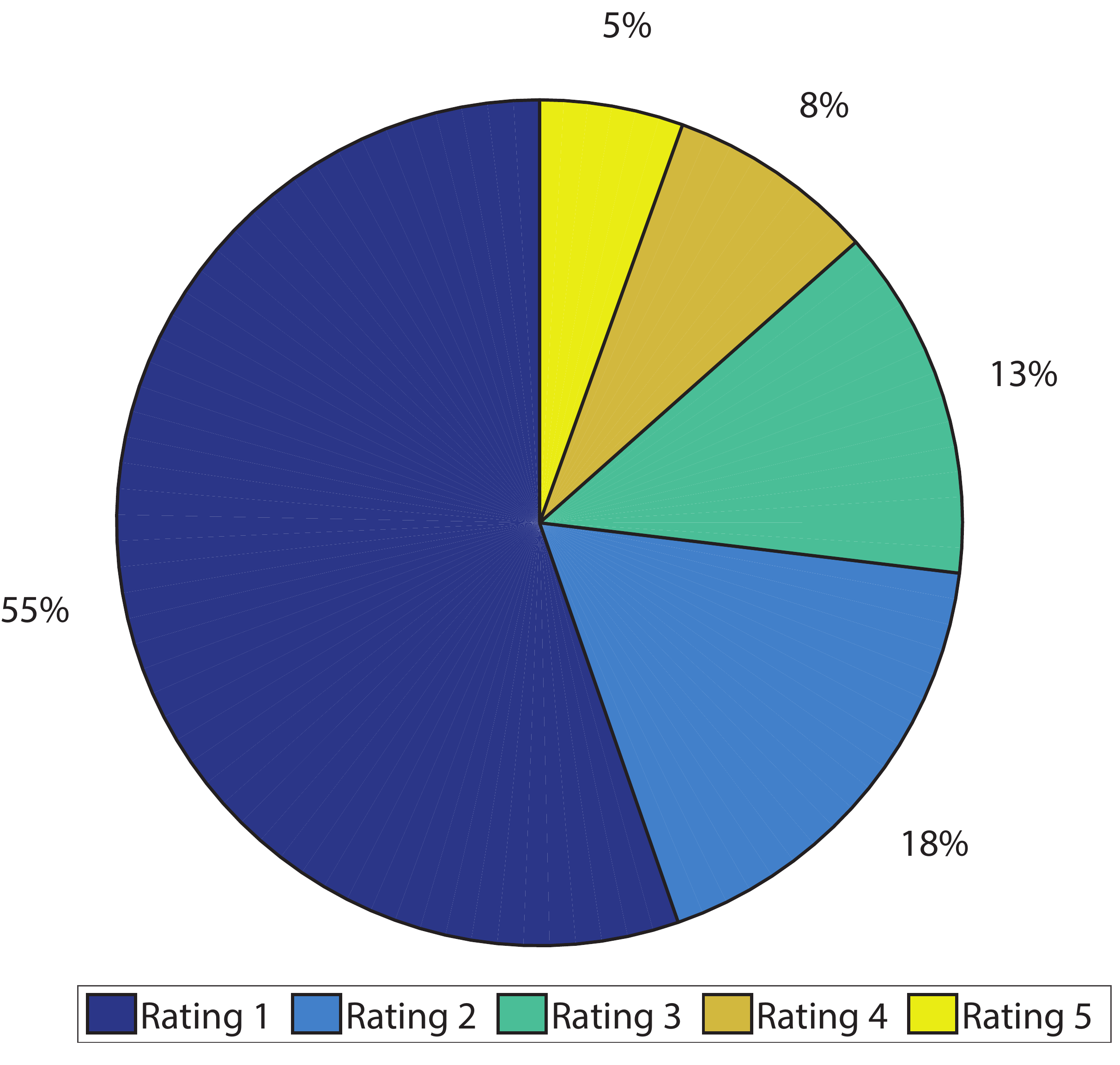}
 \caption{Rating distribution.  }
\label{Figure8:DistrRatings}
    \end{subfigure}
    ~ 
    \begin{subfigure}[b]{0.45\textwidth}
       \includegraphics[width=\textwidth]{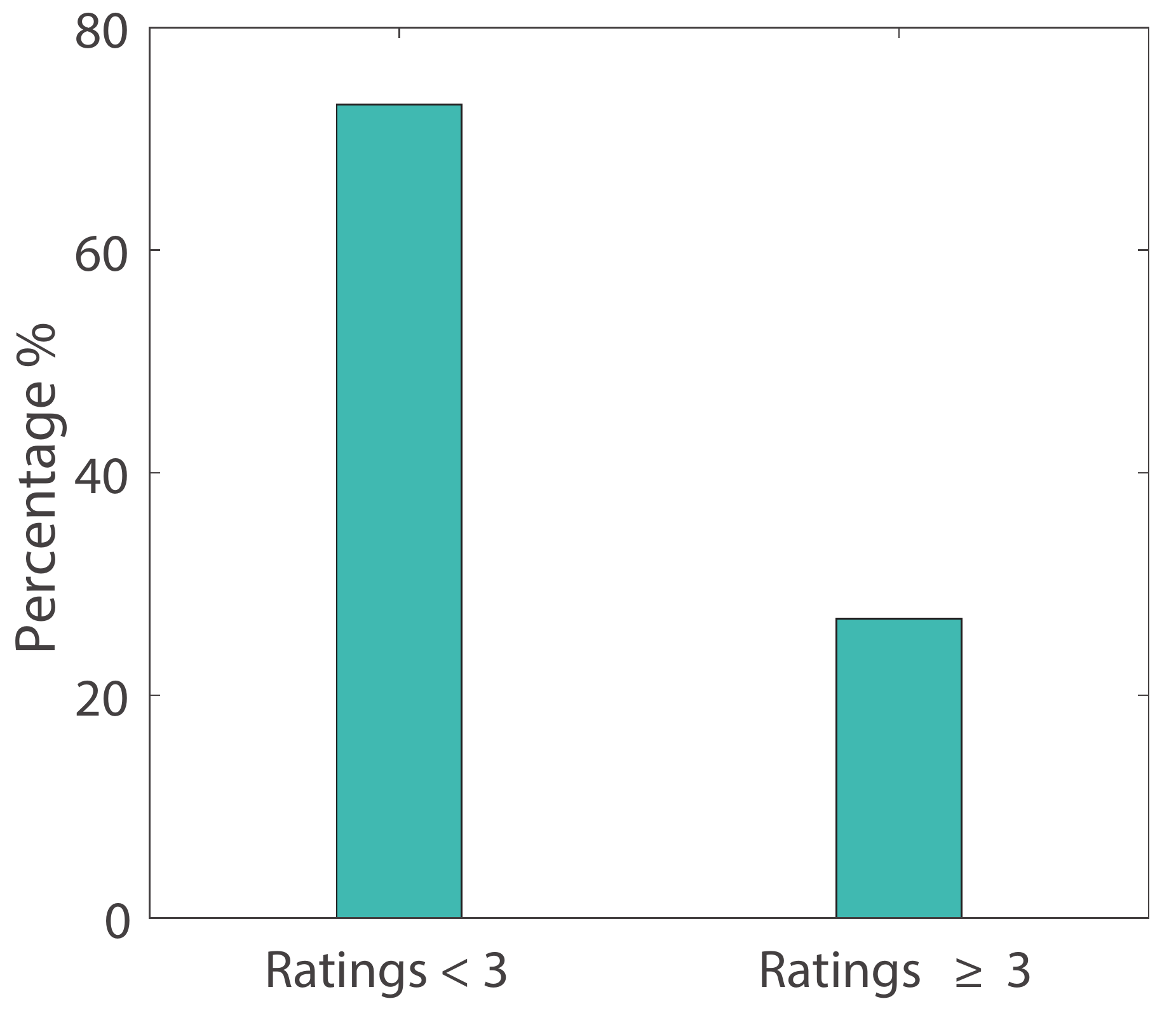}
 \caption{Click distribution.}
\label{Figure8:ADS16_RelevantItems}
    \end{subfigure}
    \caption{(a) A pie chart over 36,000 users' votes of ADS-16. (b) Expressed user's preference on ads (click - No Click).}\label{Figure8:ADS16info}
\end{figure}

In this chapter we carry out prediction experiments performing two different tasks: \emph{ad rating prediction} or \emph{ad click prediction} for ad recommendation. Experiments have been done over the \href{https://www.kaggle.com/groffo/ads16-dataset}{ADS-16} dataset that is a highly innovative collection of 300 real advertisements rated by 120 participants and enriched with the users' five broad personality dimensions. The ADS-16 is described in chapter \ref{ch4:DATASETS} section \ref{ch4:ADS16}. Figure \ref{Figure8:ADS16info} reports some statistics on the distribution of ratings and clicks.  
According to Figure \ref{Figure8:ADS16_RelevantItems}, we labeled adverts as ``clicked''  (rating
greater or equal to three), otherwise ``not clicked'' (rating less than three). The distribution of the ratings across the adverts that were scored by the users turns out to be unbalanced (9,685 clicked vs 26,315 unclicked). 

\begin{figure*}[!t]
  \centering
    \includegraphics[width=1\textwidth]{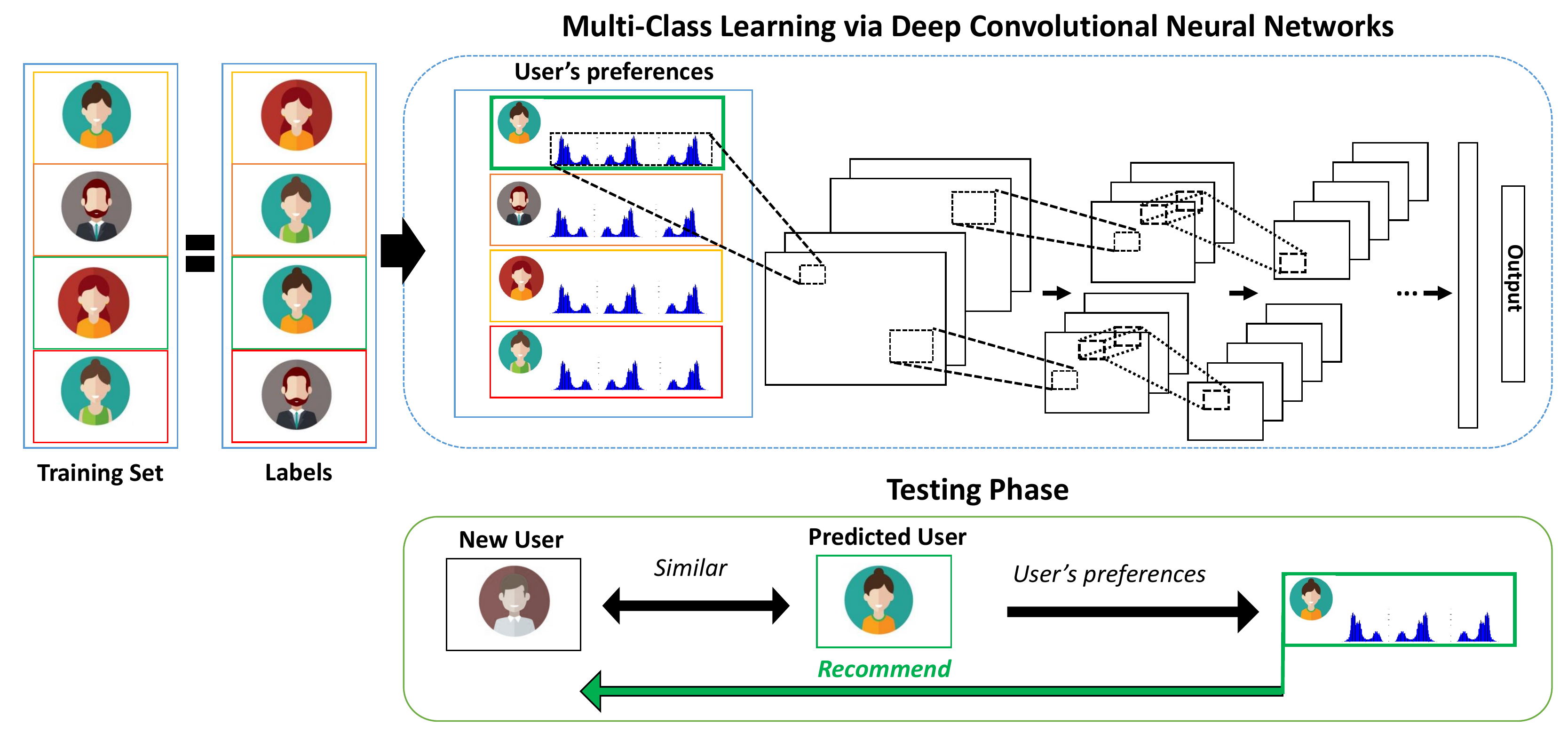}
 \caption{Illustration of the proposed deep learning framework. We aim to learn a deep CNN that assigns a higher similarity score to the positive pair than any negative pairs in each mini batch. Best viewed in color.}
\label{Figure8:Arkads}
\end{figure*}

We propose the use of the convolutional neural network presented in section \ref{sec7:GoingDeep} to solve the ranking problem. Although we inherit the architecture, the pipeline, the number of filters in each layer, their size, and other parameters differ from the network used in re-identification. Figure \ref{Figure8:Arkads}  shows an illustration of the proposed deep learning framework. 
The network is trained in a way to learn the transformation $S$ from a pair of user templates $<x,z>$ to its similarity score. Learning a similarity function between pairs of objects is at the core of learning-to-rank approaches. Since the correct match should be positioned at the top of the recommendation list, we perform ranking as a classification problem, by minimizing the objective cross-entropy cost function of positive pairs. A positive pair is defined for each user $x$ from the training set by linking it to its most similar user $z$ from the same set, where $x \neq z$. The similarity between two templates is defined by an information retrieval measure: the area under the ROC curve. Alternatively, in literature other information retrieval measurement can be found for such a kind of problem like the normalized discounted cumulative gain (NDCG) that measures the performance of a recommendation system based on the graded relevance of the recommended entities \cite{jarvelin2002cumulated}. This metric is commonly used in information retrieval and to evaluate the performance of web search engines. Experimentally, we did not observed a main effect of using one metric or another, since they are only used as a mean to assign class labels.

Therefore, we propose Logistic Regression (LR)~\cite{LIBLINEAR}, Support Vector Regression with radial basis function (SVR-rbf)~\cite{Chang:2011}, and L2-regularized L2-loss Support Vector Regression (L2-SVR)~\cite{LIBLINEAR} as baseline systems for recommendation. These engines may predict user opinions to adverts (e.g., a user's positive or negative feedback to an ad) or the probability that a user clicks or performs a conversion (e.g., an in-store purchase) when they see an ad \cite{roffo2016towards}.

\section{Deep learning with Shallow Architectures}\label{sec8:Framework}

Figure \ref{Figure8:Arkads} gives an illustration of the proposed framework. At the training stage, each training example is associated to the most similar example within the same set by mutual exclusive labelling and then fed into the deep CNN. The network will model similarities among training examples  by minimizing the objective cross-entropy cost function accounting for positive pairs $<user, label>$. The learnt CNN conducts similarity computing in one shot at the test time.  

Before continuing into the formulation, we describe some of the terminologies associated with our problem that will be used later. Without loss of generality, let us consider solving the following item recommendation problem in a single-shot case. Suppose we are given a training set $X = \{ (x_i,y_i), i = 1,2,...,N \}$, where  $(x_i,y_i)$ is a pair of the preferences of the $i^{th}$ user in the training set and their label accounting for the most similar user in terms of interests and preferences, $N$ is the number of users. For a new test user $x$ to be matched against training set, a ranking list should be generated according to the similarity between $x$ and each user in the database. The list of recommended ads for $x$ with respect to the database can be expressed as the list choices of the predicted user (see Figure \ref{Figure8:Arkads} testing phase).

\begin{figure*}[!]
\centering
\includegraphics[width=1.0\textwidth]{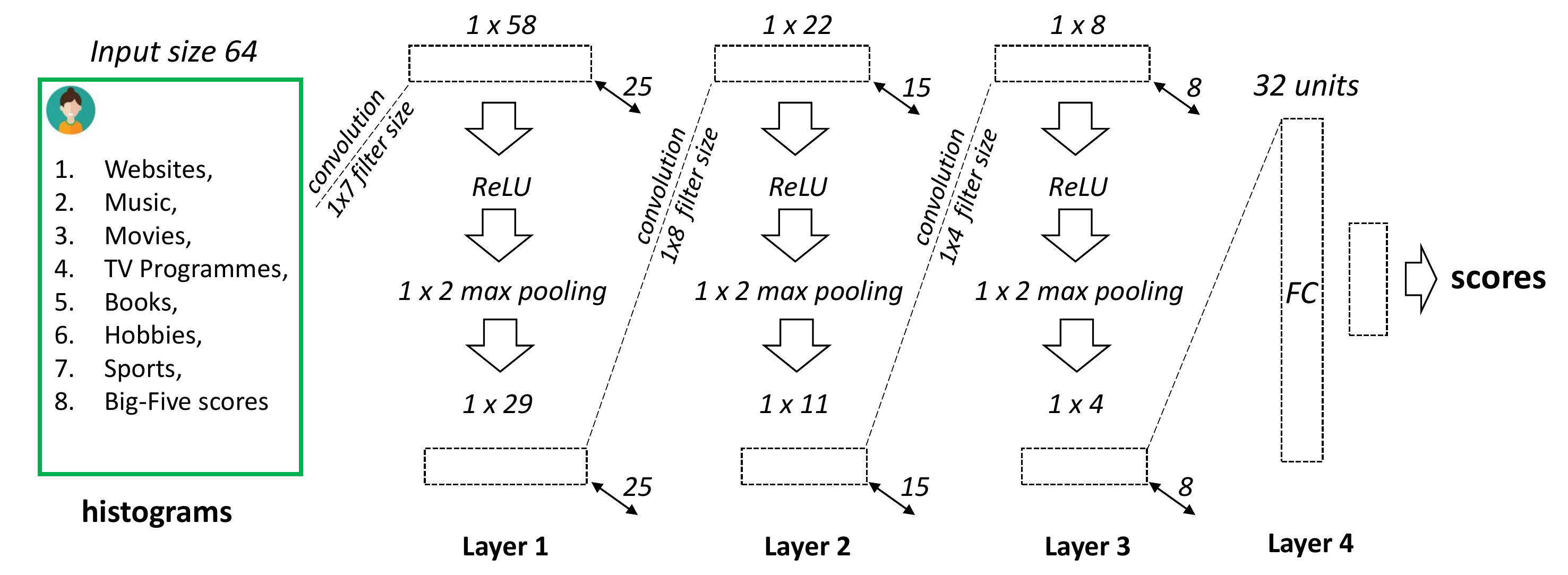}
\caption{Architecture of our deep network. Each conversation in the gallery set is first pre-processed, features are extracted and represented by 64-bins histograms. Then, the $ 1 \times 64$ feature vector is presented as the input, which is convolved with 25 different first layer filters, each of size $1 \times 7$. The resulting feature maps are then passed through a rectified linear unit, max-pooled ($1 \times 2$ regions) to give 25 different $1 \times 29$ feature maps. Similar operations are repeated in the second to third layer. The last layer is fully connected, taking features from the top convolutional layer as the input in vector form. Finally, a classification score for each sample is returned.}\label{fig8:CNNADS}
\end{figure*}

Our deep network comprises three convolutional layers to extract features hierarchically, followed by a fully connected layer. The CNN layer configurations are designed using the principles inspired by \cite{KrizhevskyNIPS2012}. Figure\ref{fig8:CNNADS} shows the detailed structure of our network. A notable difference between classic deep learning and the proposed solution is that we propose to learn via 1D convolutions instead of the 2D ones. 1D convolutions naturally account for the 1D structure of input data. 
Convolutional layers are followed by a non-linear operation. Non-linearities between layers ensure that the model is more expressive than a linear model. Among the many possible activation functions we use the Rectified Linear Unit (ReLU) which is defined in section \ref{sec3:relu}. ReLU avoids zig-zagging dynamics in the gradient descent optimization algorithms, greatly accelerate the convergence of stochastic gradient descent compared to the sigmoid/tanh functions, and it can be easily implemented by thresholding a matrix of activations at zero. In addition to the convolutional layers and non-linearities just described, this network also contains pooling layers. The pooling layers function is to simplify the output information from the convolutional layer. We propose a $1 \times 2$ max pooling, which partitions the input into a set of non-overlapping sub-vectors and, for each such sub-vector, outputs the maximum value. These pooled features can then be used for the next steps.  

Neural networks have a very strong representational power. A neural network with a single layer can represent any bounded continuous function to an arbitrary degree of accuracy (i.e., the \textit{Universal Approximation} theorem \cite{Hornik:1991}). Although, very deep convolutional networks with hundreds of layers have led to significant reductions in errors on competitive benchmarks, research on the possibility of using shallow architectures has been conducted over the last few years. This is the case of \cite{BaCaruanaNIPS2014} where the authors provided evidence that shallow feed-forward nets could learn the complex functions, previously learned by deeper networks, and achieved accuracies previously only achievable with deeper models. They empirically showed that single-layer fully connected feedforward nets trained to mimic deep models can perform similarly to well-engineered complex deep architectures. 

One of the goals of this work is to fill, at least partially, the gap above and to investigate whether well balanced deep networks  - meaning the amount of layers relative to the number of observations - can perform better of complex very-deep convolutional architectures when the number of observations is not enough to support their training. Indeed, overfitting occurs when a model is excessively complex, for example, when having too many parameters relative to the number of observations. A model that has been overfit has poor predictive performance, as it overreacts to minor fluctuations in the training data. 

\subsubsection{Input Data}

Vectorized data representations frequently arise in many data mining applications. They are easier to handle since each data can be viewed as a point residing in an Euclidean space. Thus, similarities between different data points can be directly measured by an appropriate metric to solve traditional tasks such as classification, clustering and retrieval. Unfortunately, many data sources cannot be naturally represented as vectorized inputs. As a result, there is a need of combining these heterogeneous data sources (user demographic information, preferences, etc.) to come up with meaningful and stronger results. The basic assumption is that, once the vectorized representation is obtained, the mining tasks can be readily solved by deep learning algorithms. To address the aforementioned challenge, we present an interesting idea to representation learning which learn from heterogeneous data. At the training stage, data a firstly represented by \textit{histograms} (e.g. in our previous work in section \ref{sec7:GoingDeep}).
\begin{table*}[!t]
\centering
  \resizebox{1\textwidth}{!}{%
\begin{tabular}{|m{2.8cm}|m{3.9cm}|m{8cm}|m{2.5cm}|}
\hline
\textbf{Group}  &\textbf{Type} & \textbf{Description} & \textbf{References}\\\hline
\textbf{Users' \newline Preferences} 
& Websites, Movies, Music, TV Programmes, Books, Hobbies &  Categories of: websites users most often visit (WB), watched films (MV), listened music (MS), watched T.V. Programmes (TV), books users like to read (BK), favourite past times, kinds of sport, travel destinations.& \cite{He:2014,Lee20102142,NzVald20121186}\\
 
\hline
\textbf{Demographic} & Basic information  & Age, nationality, gender, home town, CAP/zip-code, type of job, weekly working hours, monetary well-being of the participant &\cite{NzVald20121186}\\

\hline
 \textbf{Social Signals} &  Personality Traits&  BFI-10: Openness to experience, Conscientiousness, Extraversion, Agreeableness, and Neuroticism (OCEAN) &~\cite{Chen201657,rammstedt:2007}\\
 &  Images/Aesthetics  & Visual features from a gallery of 1.200 \textit{positive / negative} pictures and related meta-tags &~\cite{lovato2012tell}\\ 
\hline
 \textbf{Users' Ratings} &  Clicks & 300 ads annotated with Click / No Click by 120 subjects &~\cite{He:2014,Avila16,wang2010click}\\
 &  Feedback  & From 1-star (Negative) to 5-stars (Positive) users' feedback on 300 ads &~\cite{He:2014,Avila16,wang2010click}\\ 

\hline	
\end{tabular}}
  \caption{The table reports the type of raw data provided by the ADS Dataset. Data of the first and last group can be considered as historical information about the users. }
  \label{tab8:features}
\end{table*}

We extracted the set of features reported in Table \ref{tab8:features}. From the raw data we decided to represent features as histograms. Histograms are then used to feed the deep CNN (see Figure\ref{fig8:CNNADS}). Since some of the features (e.g., musics, websites, movies, etc.) share the same range of values and each value represents a category of interest, these histograms count how many different user's dimensions belong to a particular category. 

\section{Training the CNN}\label{sec8:learning}

In the previous section we presented the details of our network configuration. In this section, we
describe the details of classification CNN training and evaluation. The CNN training procedure generally follows Krizhevsky et al. \cite{KrizhevskyNIPS2012}. Namely, the training is carried
out by optimising the multinomial logistic regression objective using mini-batch gradient descent with cross entropy cost function. When training neural networks cross entropy is a better choice of cost function than a quadratic cost function. A detailed justification is given in section \ref{sec3:crossEntropyCost}. 
The use of SGD in the deep neural network setting is motivated by the high cost of running back propagation over the full training set, since SGD can overcome this cost and lead to fast convergence.
We use $L2$ regularization to train our network. The idea of L2 regularization is to add an extra term to the cost function, a term known as the \textit{regularization term}.
\begin{equation}\label{eq:RegCrossEntropy}
    C = -\frac{1}{n} \sum_{x_j}  \Big[ y_{j} ln(a^L_{j}) + ( 1 - y_{j}) ln(1 - a^L_{j})  \Big] + \frac{\delta}{2n} \sum_w w^2,
\end{equation}
where $L$ denotes the number of layers in the network, and $j$ stands for the $j^{th}$ neuron in the last $L^{th}$ layer. The first term is the common expression for the cross-entropy in $L$-layer multi-neuron networks, the regularization term is a sum of the squares of all the weights in the network. The regularization term is scaled by a factor $\frac{\delta}{2n}$, where $\delta > 0$ is known as the \textit{regularization parameter}, and $n$ is the size of the training set (see Table~\ref{tab:notation11} for the used notation). Intuitively, regularization can be viewed as a way of compromising between finding small weights and minimizing the original cost function. The relative importance of the two elements of the compromise depends on the value of $\delta$. This kind of compromise helps reduce overfitting. In our case, the training was regularised by weight decay (the L2 penalty multiplier set to $5 \cdot 10^{-4}$).
\begin{figure*}[!]
\centering
\includegraphics[width=0.65\textwidth]{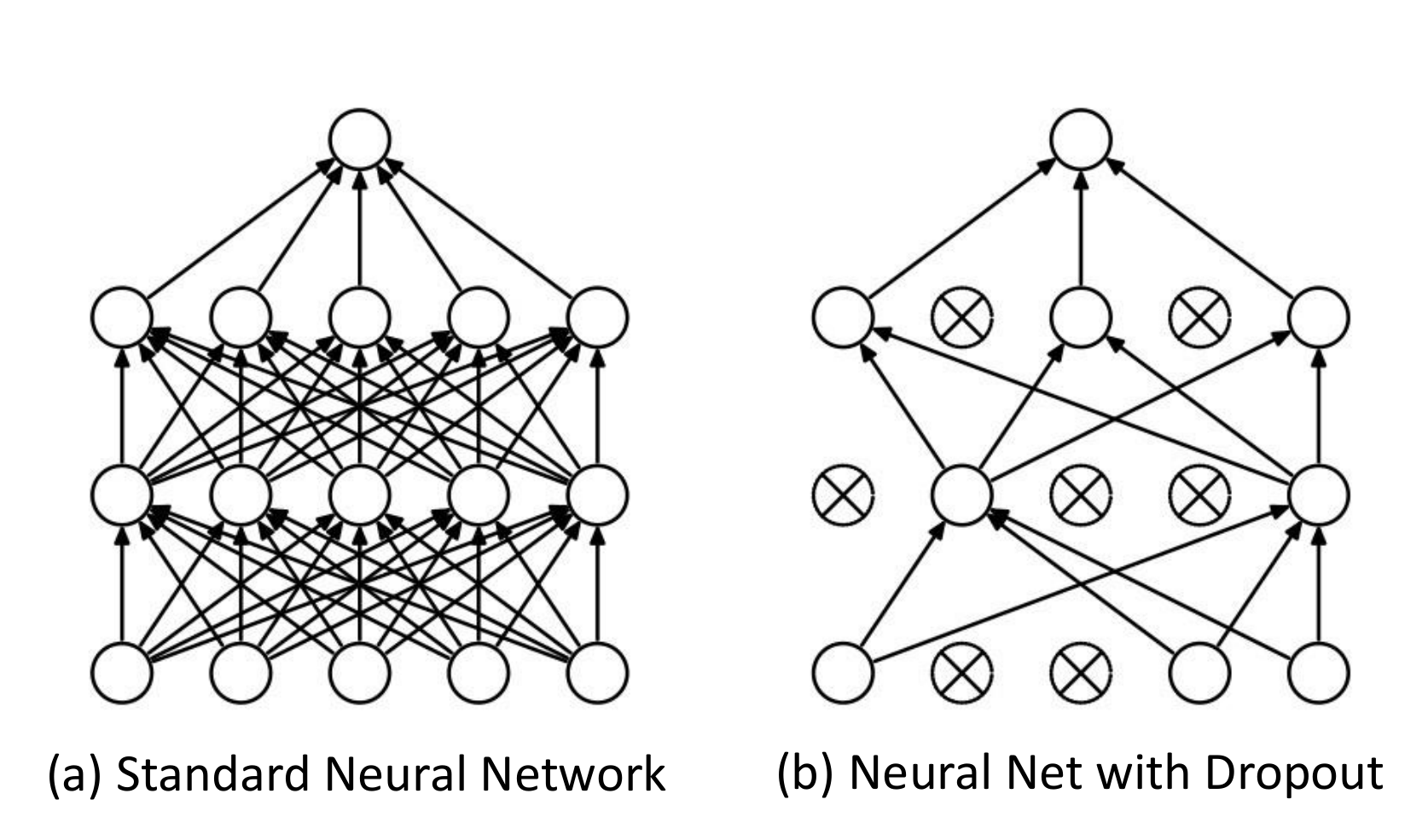}
\caption{An illustration of the dropout mechanism within the proposed CNN. (a) Shows a standard neural network with $2$ hidden layers. (b) Shows an example of a thinned network produced by applying dropout, where crossed units have been dropped.}\label{fig8:DropoutOur}
\end{figure*}
Another strategy adopted during the training phase is \textit{dropout}. Dropout prevents overfitting and provides a way of approximately combining many different neural network architectures exponentially and efficiently. Indeed, dropout can be viewed as a form of \textit{ensemble learning} (see Figure \ref{fig8:DropoutOur}). By dropping a unit out, we mean temporarily removing it from the network, along with all its incoming and outgoing connections. In particular, given a training input $x$ (a mini-batch of training examples) and corresponding desired output $y$ (e.g., the class labels of our training samples). In our network, dropout, randomly and temporarily deletes the $30\%$ (dropout ratio set to 0.3) of the hidden units in the network, while leaving the input and output units untouched. After this step, the training procedure starts by forward-propagating $x$ through the network, and then back-propagating to determine the contribution to the gradient, therefore updating the appropriate weights and biases.

The learning rate was initially set to $10^{-2}$, and then decreased by a factor of $10$ when the validation set accuracy stopped improving. In total, the learning rate was decreased 3 times, and the learning was stopped after only $20$ epochs. We conjecture that in spite of the smaller number of parameters and the lesser depth of our nets compared to \cite{KrizhevskyNIPS2012}, the nets required less epochs to converge to two (a) implicit regularization imposed by lesser depth and smaller convolutional sizes filter (1 dimension); (b) a good pre-initialisation of filters. The initialisation of the network weights is important, since bad initialisation can stall learning due to the instability of gradient in deep nets. The reason is that the logistic function is close to flat for large positive or negative inputs. In fact, if we consider the derivative at an input of $2$, it is about $\frac{1}{10}$, but at $10$ the derivative is about $\frac{1}{22,000}$. This means that if the input of a logistic neuron is $10$ then, for a given training signal, the neuron will learn about $2,200$ times slower that if the input was $2$.

To circumvent this problem, and allow neurons to learn quickly, we either need to produce a huge training signal or a large derivative. To make the derivative large, the inputs are normalized to have mean $0$ and standard deviation $1$, then we initialized the biases to be $0$ and the weights $w$ at
each layer with the following commonly used heuristic:
\[
    w \in \Big[ \frac{-1}{\sqrt{n}},\frac{1}{\sqrt{n}} \Big]
\]
where $[−a, a]$ is the uniform distribution in the interval $(−a, a)$ and $n$ is the size of the previous layer. The probability that we get a sum outside of our range is small. That means as we increase $n$, we are not causing the neurons to start out saturated \cite{Glorot10understandingthe}.

\section{Experiments and Results}

In this section we show results obtained for two types of scenarios: \emph{ad rating} and \emph{ad click prediction}. We conduct prediction experiments to explore the strengths and weakness of using our approach for recommendation. Since a prediction engine lies at the basis of the most recommender systems, we selected some of the most widely used techniques for recommendations and predictions~\cite{He:2014}, such as Logistic Regression (LR)~\cite{LIBLINEAR}, Support Vector Regression with radial basis function (SVR-rbf)~\cite{Chang:2011}, and L2-regularized L2-loss Support Vector Regression (L2-SVR)~\cite{LIBLINEAR}. 
These methods have often been  based on a set of sparse binary features converted from the original categorical features via one-hot encoding~\cite{Lee:2012,Richardson07predictingclicks:}. 
These engines may predict user opinions to adverts (e.g., a user's positive or negative feedback to an ad) or the probability that a user clicks or performs a conversion (e.g., an in-store purchase) when they see an ad. 

Let us say $X = \{ \bar x_1, ..., \bar x_N \}$ is the set of observations, where the vectors $\bar x_i$ correspond to features coming only from the group ``users' preferences'' as described in Table~\ref{tab8:features} and $N = 120$ stands for the number of users involved in the experiment.

Regression is performed over the 300 advertisements. The prediction problem is solved using LR, L2-SVR, and SVR-rbf, while feeding them with the features coming from Table~\ref{tab8:features} and represented as 64-bin histograms. All experiments were performed using an exhaustive cross-validation: Leave-One-Out (LOO) approach (k-fold, k = $1$). 
In k-fold cross-validation, $X$ is randomly partitioned into k's equal sized subsamples (the folds are the maintained the same for each algorithm in comparison). Of the k subsamples, a single subsample is retained as the validation data for testing the model, and the remaining k - 1 subsamples are used as training data. The cross-validation process is then repeated k times, with each of the k subsamples used only once as the validation data. The k results from the folds can then be averaged to produce a single estimation. Note that k=1 in the case of LOO.

\subsection{Ad Rating Prediction}

\begin{figure*}[!t]
  \centering
    \includegraphics[width=1\textwidth]{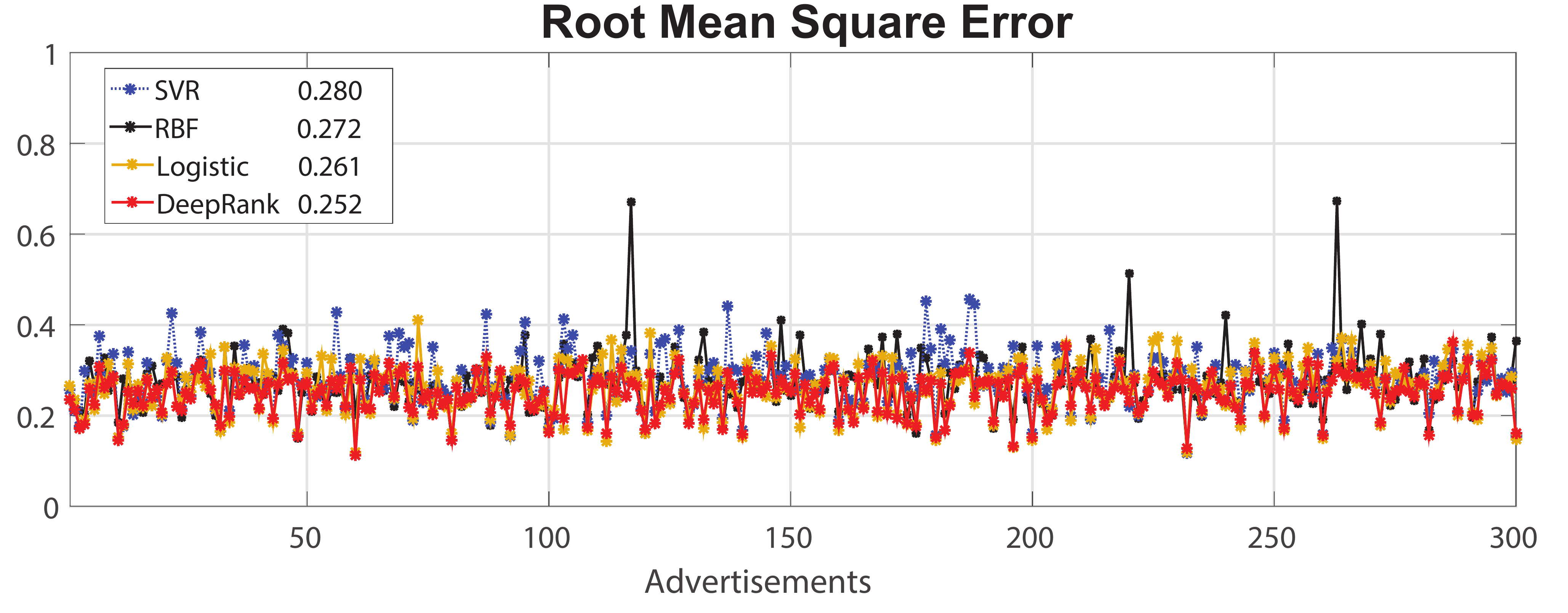}
 \caption{A comparison of rating prediction performance on ADS-16 in terms of RMSEs.}
\label{Figure8:RMSE}
\end{figure*}

\begin{figure*}[!t]
  \centering
    \includegraphics[width=1\textwidth]{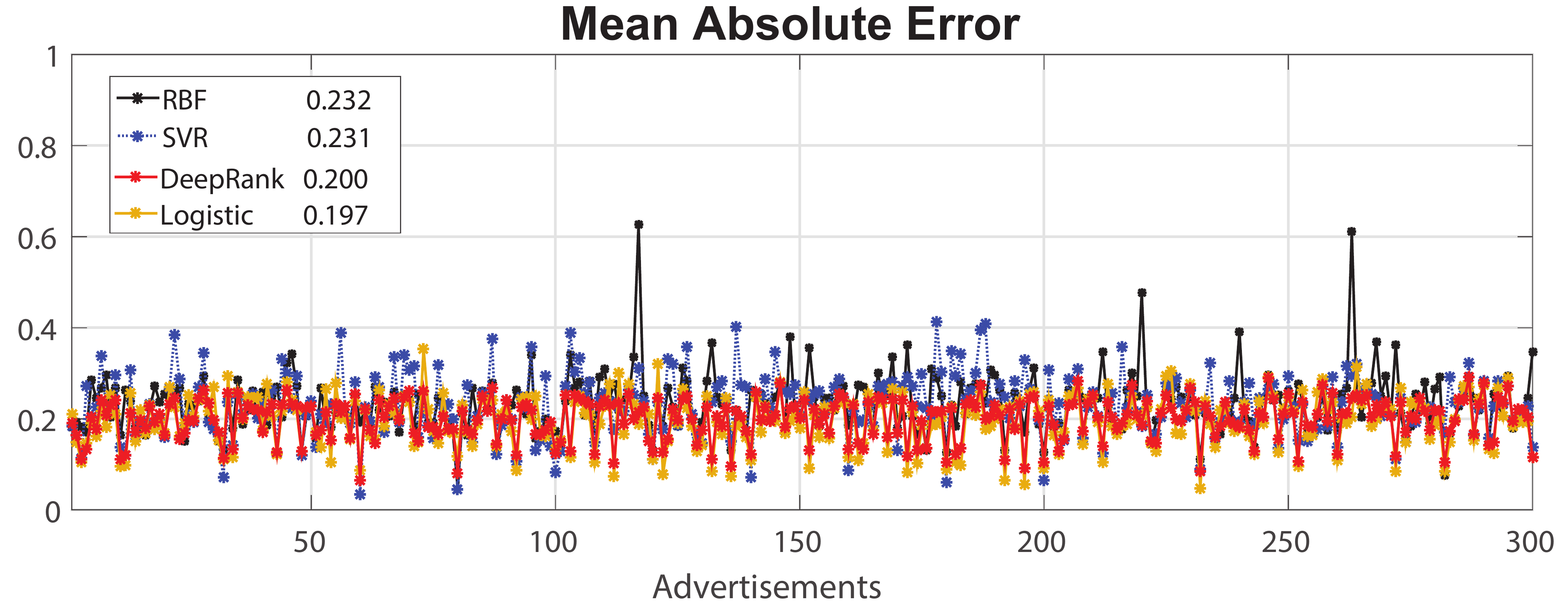}
 \caption{A comparison of rating prediction performance on ADS-16 in terms of MAEs.}
\label{Figure8:MAE}
\end{figure*}

In this section we report results for rating prediction showing that our balanced deep network can improve the prediction performance of the evaluated methods significantly. Figure \ref{Figure8:RMSE} illustrates rating prediction results in term of RMSE for each advert. Results show that our approach followed by Logistic Regression are the most effective methods in providing good predictions. 
Figure \ref{Figure8:MAE} shows the same results in terms of Mean Absolute Errors, as the name suggests, the MAE is an average of the absolute errors $err_{u,a}=|{\hat r_{u,a} - r_{u,a}}|$, where $\hat r_{u,a}$ is the prediction and $r_{u,a}$ the true value. The MAE is on same scale of data being measured.
Statistical evaluation of experimental results has been considered an essential part of validation of machine learning methods. In order to assess if the difference in performance is statistically significant, t-tests have been used for comparing the accuracies. This statistical test is used to determine if the accuracies obtained with our approach are significantly different from the others (whereas both the distribution of values were normal). The test for assessing whether the data come from normal distributions with unknown, but equal, variances is the \emph{Lilliefors} test. 

Results show a statistical significant effect in performance while using DeepRank (p-value < 0.05, Lilliefors Test H=0) and Logistic regression respect to SVR and RBF.

\subsection{Ad Click Prediction}

\begin{figure*}[!t]
  \centering
    \includegraphics[width=1\textwidth]{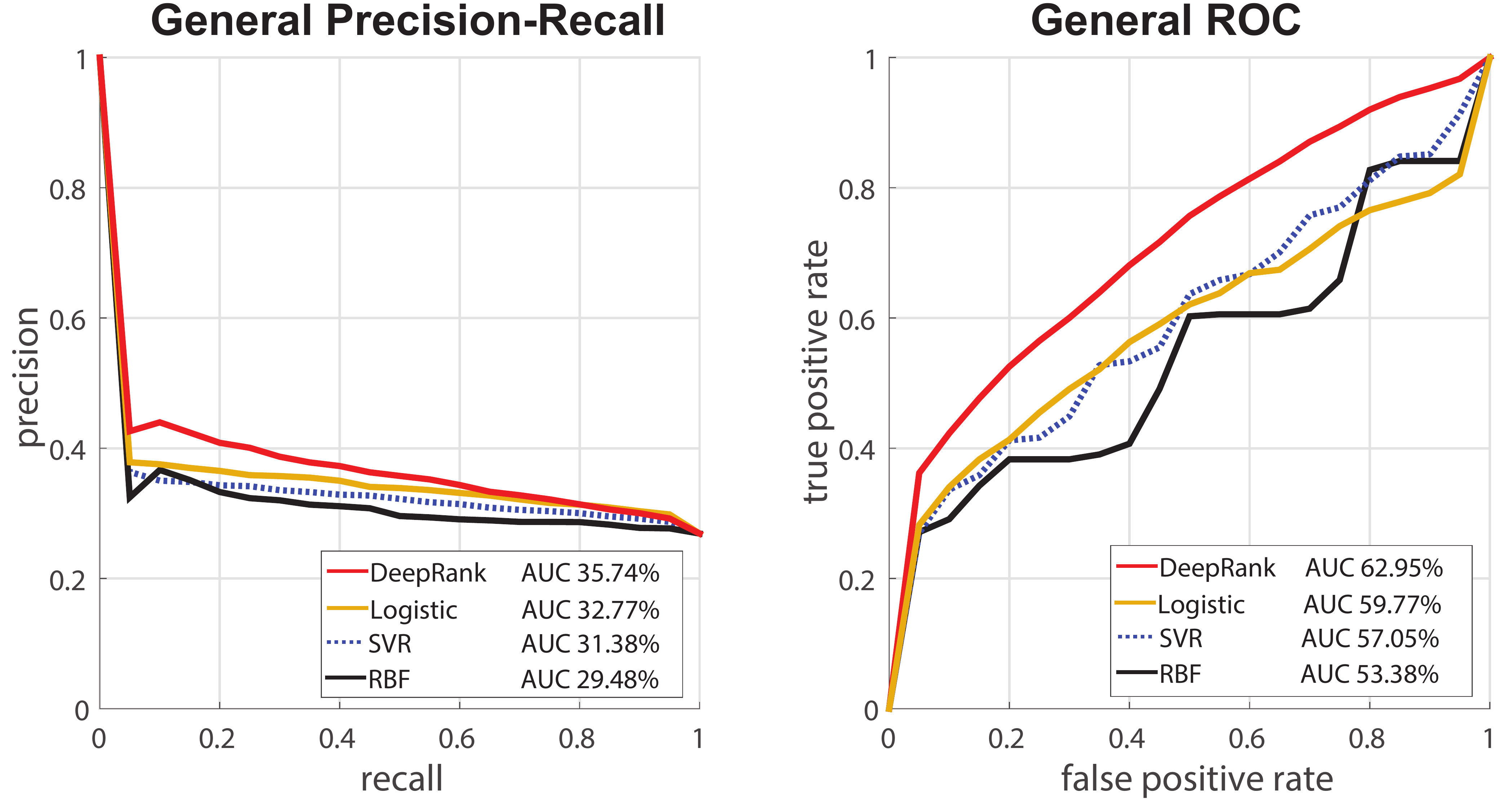}
 \caption{Precision-Recall and ROC Curves.}
\label{Figure8:ROC}
\end{figure*}

This section shows an offline evaluation of click prediction. Along the lines of the previous experiment, a LOO cross-validation is used. The experiment is performed at the advert level, whenever a user showed their interest in a given advert (i.e. rating greater or equal to 3) we labeled the ad as ``clicked'', otherwise ``not clicked'' (rating less than 3). As a result, for each user we obtained a list of 300 labels representing their preference to each ad. We computed precision-recall and ROC curves for each user, and then averaged the resulting curves over users (see the plot in Figure \ref{Figure8:ROC}). This is the usual manner in which precision-recall (or ROC) curves are computed in the information retrieval community \cite{Harman02overviewof,Sarwar:2000,Schein:2002}. Such a curve can be used to understand the trade-off between precision and recall and ROC a typical user would face.
Figure~\ref{Figure8:ROC}.(left) reports the precision-recall curves which emphasize the proportion of recommended items that are preferred and recommended. Figure~\ref{Figure8:ROC}.(right) shows the global ROC curves, which emphasize the proportion of adverts that are not clicked but end up being recommended. The DeepRank (red) curve completely dominates the other curves, the decision about the superior setting for DeepRank is clear in this case.
The Area Under the ROC Curve is calculated as a measure of accuracy, which summarizes the precision recall of ROC curves, we report the AUCs for all the methods in comparison in the legend of Figure \ref{Figure8:ROC}.

\begin{figure*}[!t]
  \centering
    \includegraphics[width=1\textwidth]{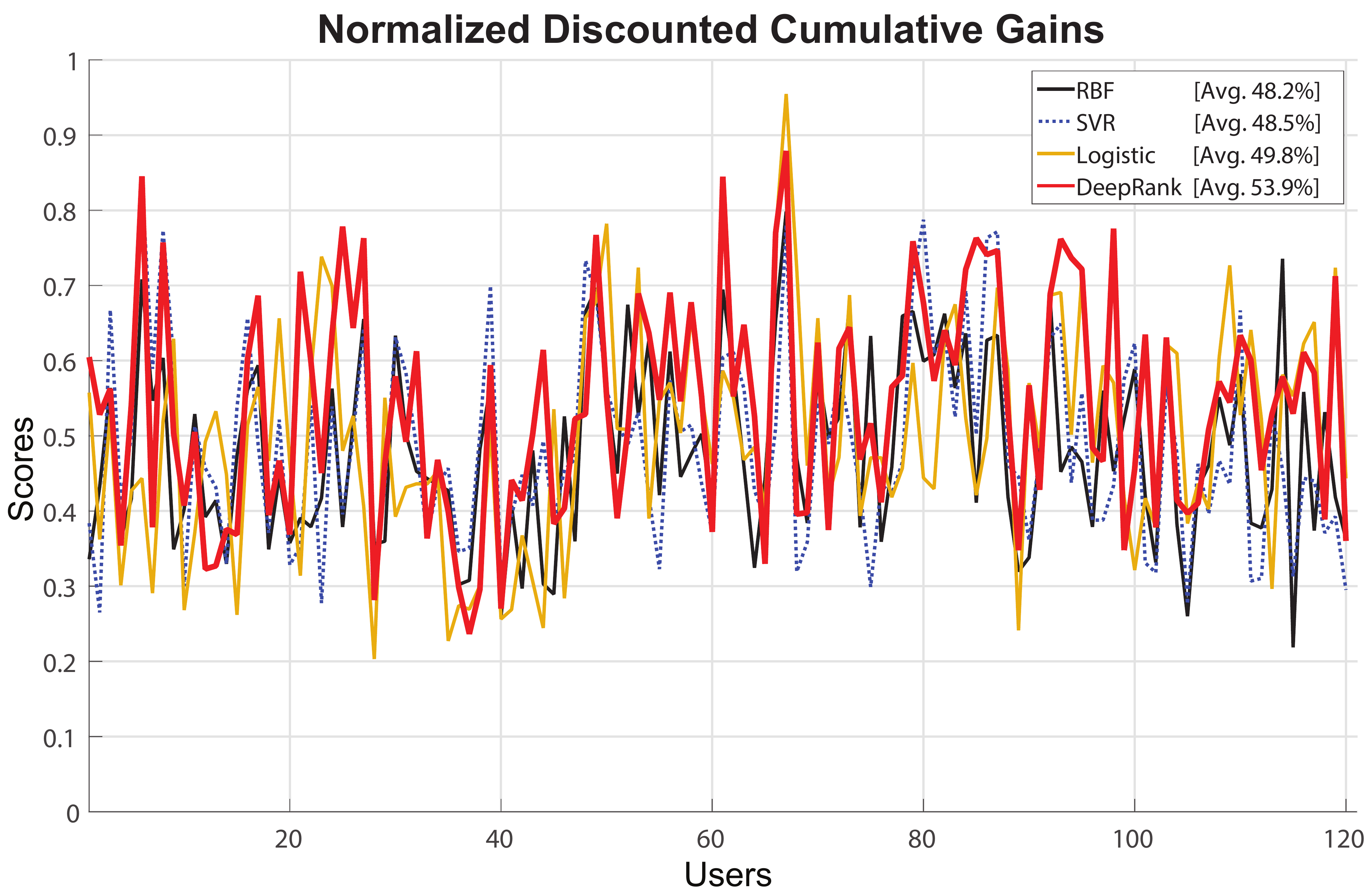}
 \caption{Ranking quality of RBF, SVR, Logistic and DeepRank on ADS-16.}
\label{Figure8:NDCG}
\end{figure*}

Finally, we measures the performance of the recommendation system based on the graded relevance of the recommended entities through the normalized discounted cumulative gain (NDCG). It varies from 0.0 to 1.0, with 1.0 representing the ideal ranking of the entities. 
\[
    DCG_k = \sum_{i=1}^k \frac{2^{rel_i}-1}{log(i+1)}
\]
where $k$ is the maximum number of entities that can be recommended, we set to 15 out of 300. The normalized DCG is given by the ratio between the $DCG_k$ and the maximum possible (ideal) DCG for all the set of users. Using a graded relevance scale of adverts in a recommender system result list, DCG measures the usefulness, or gain, of an ad based on its position in the result list. The gain is accumulated from the top of the result list to the bottom with the gain of each result discounted at lower ranks. 
Figure \ref{Figure8:NDCG}, reports the NDCG for each subject. In a LOO sense, the NDCG is intended as the ability of the system in producing a good recommendation for the user $u_i$ when she is not present in the training set, that is to say, the user is unknown. 
The average NDCG values, in the legend of Figure \ref{Figure8:NDCG}, indicate the superior ability of our approach in providing highly relevant adverts earlier in the result list.

\section{Summary}

Learning a similarity function between pairs of objects is at the core of learning-to-rank approaches. However, before learning can take place, such pairs need to be mapped from the original space into some feature space encoding various aspects of their relatedness. Feature engineering is often a laborious task and may require external knowledge sources that are not always available or difficult to obtain. Recently, deep learning approaches have gained a lot of attention from the research community and industry for their ability to automatically learn optimal feature representation for a given task, while claiming state-of-the-art performance in many tasks in computer vision, speech recognition and natural language processing. In this work, we proposed the use of deep learning architectures with fewer parameters in the non-visual task of ad recommendation. Experiments have been performed on two types of scenarios: \textit{Ad-Rating} and \textit{Ad-Click} prediction. In this work, we provided a new CNN architecture in a learning-to-rank sense. The designed architecture (4-layers) consists of 1-dimensional convolutional filters that results to be useful when learning from very few data. This pilot work represents one of the very first works on deep ranking for advert recommendation. A comparative evaluation is given, demonstrating that the proposed approach significantly outperforms other standard approaches to rank such as logistic regression, support vector regression and its variants. 

\chapter{Conclusions and Final Remarks}\label{ch9:conclusions}

This thesis investigated the role of ranking in the wide field of pattern recognition and its applications. In particular, in this work we analysed the ranking problem from two different methodological perspectives: \textit{ranking to learn}, which aims at studying methods and techniques to sort objects with the aim to enhance models generalization by reducing overfitting, and \textit{learning to rank}, which makes use of machine learning to produce ranked lists of objects to solve problems in those domains that require objects to be sorted with some particular criteria.  

In the first part of this work we addressed the merits of ranking features to improve classification models. The central premise when using a feature ranking technique is that a designed feature set often contains many features that are either redundant or irrelevant, and can thus be removed without incurring much loss of information. We have seen that redundant or irrelevant features are two distinct notions, since one relevant feature may be redundant in the presence of another relevant feature with which it is strongly correlated. Actually, the exact definitions of relevancy vary across problem settings. For example, in classification, where the aim is to find useful features that separate well two or more classes, \textit{relevant} is a feature which is highly representative of a class \cite{Guyon:2003:IVF:944919.944968}. We proposed novel solutions in feature selection that efficiently remove redundant or unwanted cues from the information stream. Two novel graph-based ranking algorithms have been proposed: the unsupervised \textit{Infinite Feature Selection} (Inf-FS) and the supervised approach \textit{Eigenvector Centrality} (EC-FS).  
The Inf-FS (introduced in \ref{ch5:infFS11}) ranks features based on path integrals \cite{Feynman:1948aa} and the centrality concept on a feature adjacency matrix. The path integral formalism, is a tool for calculating quantum mechanical probabilities. The Feynman's recipe applied to a particle travelling on a manifold, considers all the possibilities for the particle travelling between two positions in space-time. Not only the straight-line approach, but also the possibility of the particle turning loopings and making diverse detours. Each path has its own amplitude, which is a complex number, in order to calculate the total amplitude between two space-time events the Feynman's receipt states that all the complex amplitudes have to be added up. The standard way to interpret these amplitudes is the probability to measure the particle at position $B$ at time $t_B$, knowing that it was at position $A$ at time $t_A < t_B$, which is given by the the square absolute value of the amplitude associated to those two events. Therefore, we derived a discrete form for path integral. Then, we mapped the space-time to a simplified discrete form without time, that is: \textit{a graph}. Finally, the framework used to estimate the most probable position where to find a particle has been switched to the novel problem of finding the most likely relevant feature (see Sec. \ref{sec5:Quantum} for details). 
From a machine learning perspective, the Inf-FS approaches feature selection as a regularization problem, where features are nodes in a graph, and a selection is a path through them. The application of the Inf-FS to 13 different datasets and against 8 competitors, led to top performance, notably setting the absolute state of the art on 8 benchmarks. We showed that this ranking method is also robust with a few sets of training data, it performs effectively in ranking high the most relevant features, and has a very competitive complexity. 
We provided many insights into the method from a probabilistic perspective (in Sec. \ref{sec:markovProcesses}) and from a graph theory point of view (in Sec.\ref{sec5:graphtheory}).
The former shows that the solution of the Inf-FS can be seen as the estimation of the expected number of periods that a Markov chain spends in the $j^{th}$ non-absorbing state given that the chain began in the $i^{th}$ non-absorbing state. Perhaps this interpretation comes from the specification of the matrix $S$ as the infinite sum. $A^l(i, j)$, where $A$ denotes the adjacency matrix, is the probability that the process which began in the $i^{th}$ non-absorbing state will occupy the $j^{th}$ non-absorbing state in period $l$. The latter shows how identifying the most important nodes corresponds to individuate some indicators of centrality within a graph. This fact motivated the exploration of some centrality measurements such as the Eigenvector Centrality. As a result, a second feature selection method called EC-FS, exploits the convergence properties of power series of matrices thereby individuating candidate features, which turn out to be effective from a classification point of view. Like the Inf-FS, also the EC-FS is a graph-based method - where features are the nodes of the graph. The EC-FS is a supervised method (by construction) weighted by a kernelized adjacency matrix, which draws upon best-practice in feature selection while assigning scores according to how well features discriminate between classes. We discussed how the method estimates some indicators of centrality to identify the most important features within the graph. The results are remarkable: EC-FS has been extensively tested on 7 different datasets selected from different scenarios (i.e., object recognition, handwritten recognition,biological data, and synthetic testing datasets), in all the cases it achieves top performance against 7 competitors selected from recent literature in feature selection. EC-FS approach is also robust and stable on different splits of the training data as proved by the kuncheva's stability index. Given that, we investigated the interrelations of the two algorithms for the classification task. The purpose of this analysis was to obtain more insights into the Inf-FS formulation, with the aim of gaining a better understanding of its strength and weakness. An interesting result came from this analysis. The EC-FS does not account for all the contributions given by the power series of matrices (or sub-solutions). On the other hand, the Inf-FS integrates each partial solution (i.e., $S = A^1+A^2+...+A^l, l\to \infty$), that together help the method to react in a different way in presence of noise in data or many redundant cues. 

Ranking to learn has been explored in the real-time application of visual tracking. In this part we evaluated a collection of modern feature selection approaches, used in off-line settings so far.  We investigated the strengths and weaknesses of these algorithms in a classification setting to identify the right candidates for a real-time task. We selected four candidates who meet the requirements of speed and accuracy for visual tracking. Finally, we showed how these feature selection mechanisms can be successfully used for ranking features combined with the ACT system \cite{Danelljan_2014_CVPR},  and,  at the same time,  maintaining high frame rates (e.g., ACT with Inf-FS operates at over 110 FPS). Results show that our solutions improve by 3\% up to 7\% the ACT tracker where no feature selection is done. Moreover, ACT with mutual information resulted in a very impressive performance in precision, providing superior results compared to 29 state-of-the-art tracking methods. In this setting, we presented our contribution (i.e., Dynamic Feature Selection Tracker - DFST) accepted to the $A^*$ visual object tracking challenge 2016. The DFST uses the Inf-FS as ranking engine to sort visual features according to what happens in the image stream, with the objective of selecting the first $4$ most discriminative features, the ones that better separate the foreground from the background. The ACT tracker is then used to predict the next position of the target.

The second part of the thesis focused on the problem of learning to rank. Firstly, we investigated the different problematics related to biometric verification and identification in text chat conversations. We disclosed a new facet for biometrics, considering the chat content as personal blueprint. Therefore, among the many types of biometric authentication technologies we focused on \textit{keystroke biometrics}. As a result, we explored novel solutions of feature designing from the handcrafting of soft-biometric cues to automatic feature learning solutions.  In particular, we proposed a pool of novel turn-taking based features, imported from the analysis of the spoken dialogs, which characterize the non-verbal behavior of a IM participant while she/he is conversing. We introduced the concept of turns as key entity where the features have to be extracted in text chat conversations. On a test set of 94 people, we demonstrate that identification and verification can be performed definitely above the chance. 
We moved a step forward and showed how putting these features into a learning approach, which is capable of understanding the peculiar characteristics of a person, enables effective recognition and verification. In particular, we offered a first analysis of what a learning approach can do, when it comes to reduce the information needed to identify a particular user. The results are surprisingly promising: with just 2 turns of conversation, we are able to recognize and verify a person strongly above chance. This demonstrates that a form of behavioral blueprint of a person can be extracted even on a very small portion of chats. Another different framework of learning to rank has been used based on deep convolutional neural networks designed to the problem of recognizing automatically the identity of chat participants while respecting their privacy. We proposed a CNN architecture with a reduced number of parameters which allows learning from few training examples. As for the re-identification task, we used the areas under the CMC curves to evaluate our re-identification system. Results show that ranking identities by means of deep learning increases the accuracy from 88.4\% to 96.2\% on the TBK dataset, and from 95.7\% to 99.4\% on the C-Skype. Finally, we provided evidence that the methods and architectures developed for the biometric authentication system above were also suitable for advert recommendation tasks, where learning is used to rank ads according to users' preferences. We introduced a CNN architecture consisting of only four layers (i.e., shallow architecture) that results to be useful when learning from very few examples. This pilot work represents one of the very first prototypes on deep ranking for advert recommendation. A comparative evaluation was given, demonstrating that the proposed approach significantly outperforms other standard learning approaches to rank such as logistic regression, support vector regression and its variants. The main contribution of this pilot work was the investigation of possible applications of CNNs on limited amount of training samples. Indeed, a shallow architecture has much less parameters and learning can be performed also without expensive hardware and GPUs. It results in shorter training times. We fed the deep network with pre-processed features (represented as histograms). During training, we know that high-level features will be automatically learnt by the network starting from specific cues designed a priori. This fact makes results easier to interpret by researchers. Moreover, since the amount of parameters is much less than a deep architecture the probability to incur in overfitting is reduced. \\
\noindent
The choice of using 1D convolutions is motivated by the nature of the input data, since histograms when normalised can be interpreted as an estimate of the probability distribution of a continuous variable, the deep network will learn specific filters (and hierarchical representations in increasing levels of abstraction) capturing complex patterns among input handcrafted representations. This means that they automatically generate other features from inputs with a higher discriminative power while preserving the interpretability of research results (at least partially).
\\
\noindent
In addition, we collected and introduced representative datasets and code libraries in different research fields. For example, we made publicly available different datasets, such as the \href{https://www.kaggle.com/groffo/ads16-dataset}{ADS-16} dataset for computational advertising released on kaggle repository, or the C-Skype, TBKD, among others. This work also produced useful tools in terms of source code and libraries, such as the Feature Selection Library \href{https://uk.mathworks.com/matlabcentral/fileexchange/56937-feature-selection-library}{FSLib} that has been recognized and awarded by Mathworks in 2017 for its impact on the Matlab community (5+ ratings, more than 300 downloads pcm), and the \href{http://www.votchallenge.net/vot2016/trackers.html}{DFST tracker} released to the international Visual Object Tracking (VOT) challenge 2016.\\

\section{Future Perspectives}\label{ch9:results}

This study points to many future directions. Future work concerns deeper analysis of particular mechanisms of the proposed ranking algorithms, new proposals to try different methods, or simply curiosity. There are some ideas that we would have liked to try during the description and the development of the methods for feature ranking and selection, authentication, biometric verification, re-identification, and recommendation. Based on the results presented in this thesis, we are planning to continue our research on several topics.

\subsubsection{Feature Ranking and Selection}

Traditionally there have been two schools of thought in machine learning: the generative and the discriminative learning. Generative methods learn models which explain the data and are more suitable for introducing conditional independence assumptions, priors, and other parameters. Contrarily, discriminative learning methods only learn from data to make accurate predictions. The combination of both is a widely explored topic in machine learning. The feature selection method presented in this thesis is a purely discriminative method. We plan to work on feature selection methods which combine the generative and discriminative approaches. 

Recent advances in causal inference have opened up new possibilities in dealing with the problems of feature selection. We want to perform a paradigmatic shift from traditional statistical analysis to causal analysis of multivariate data \cite{pearl2009causal}. We are planning to continue on the general theory of causation based on the Structural Causal Model (SCM) described in \cite{pearl2003causality}, which
subsumes and unifies different approaches to causation, and provides a coherent mathematical foundation for the analysis of causes and counterfactuals. In particular, we wish to answers to two types of causal queries: 1) queries about the causal effects of certain kind of features on the nature of the problem, and (2) queries about direct and indirect effects. These are causal questions because they require some knowledge of the data-generating process; they cannot be computed from the data alone, nor from the distributions that govern the data. Remarkably, although much of the conceptual framework and algorithmic tools needed for tackling such problems are now well established, they are hardly known to researchers who could put them into practical use. This research direction rest on contemporary advances in four areas:
\begin{itemize}
    \item Counterfactual analysis
    \item Nonparametric structural equations
    \item Graphical models
    \item Symbiosis between counterfactual and graphical methods.
\end{itemize}
 

\subsubsection{Computer Vision Applications}

Object tracking is one of the most important tasks in many applications of computer vision. Many tracking methods use a fixed set of features ignoring that appearance of a target object may change drastically due to intrinsic and extrinsic factors. The ability to identify discriminative features at the right time would help in handling the appearance variability by improving tracking performance. We analysed different ranking methods in terms of accuracy, stability and speed, thereby indicating the significance of employing them for selecting candidate features for tracking systems, while maintaining high frame rates. We improved the performance of the standard methods, while keeping their fast performance during runtime. In particular, the \textit{Infinite Feature Selection} mounted on the Adaptive Color Tracking \cite{Danelljan_2014_CVPR} system operates at over 110 FPS resulting in what is clearly a very impressive performance. This combination was proposed at the international challenge on visual object tracking \cite{KristanLMFPCVHL16}. 

This analysis points towards future work to increase the robustness of the tracker. We are developing novel strategies which make the Inf-FS a supervised method. Supervision can help in selecting those features which better distinguish the foreground from the background. Intrinsic and extrinsic factors affect the target appearance by introducing a strong variability. Therefore, it is necessary to have some criteria to automatically select the right amount of features with regard to the context of the target object. 

Another future direction consists into make a strong connection between the weighted adjacency matrix of the Inf-FS and the temporal aspects that a tracking problem involves, so as to appropriately manage the presence of spatially-correlated background clutter or distractors.

\subsubsection{Biometric Verification and Identification}

This study demonstrates that a form of behavioral blueprint of a person can be extracted even on a very small portion of chats. We believe therefore that this work has the potential to open up the possibility of a large range of new applications beyond surveillance and monitoring. From the interesting results obtained over the proposed datasets we derive that the methodologies of Affective Computing and Social Signal Processing, so far applied only to face-to-face conversations, can probably be extended to dyadic chats. Future work will focus on the collection of more data and on the classification in terms of other socially relevant characteristics such as, e.g., the personality traits, negotiation outcomes, interaction modes, and so forth.

Moreover, we want to explore treating text as a kind of raw signal at character level, and apply temporal 1D Convolutional Neural Networks on it. In this thesis we only used CNNs in a ranking task starting from a set of designed features to exemplify CNNs ability to understand text chats. We know that CNNs usually require large-scale datasets to work, therefore we used a shallow architecture which consists of very few parameters. CNNs can be applied to text classification or natural language processing without any knowledge on the syntactic or semantic structures of a language. Character-level is a simplification of engineering, and it could be crucial for a single system that can work for different languages, since characters always constitute a necessary construct regardless of whether segmentation into words is possible. Working on character has the advantage that abnormal character combinations such as misspellings and emoticons may be naturally learnt. These models can serve as a basis for the next generation of ranking systems for biometric verification, re-identification and beyond. 

\subsubsection{Ranking for Recommender Systems}

We believe that deep learning is one of the next big things in recommendation systems technology. The past few years have seen the tremendous success of deep neural networks in a number of complex tasks such as computer vision, natural language processing and speech recognition. Despite this, only little work has been published on deep learning methods for the recommender systems. Notable exceptions are deep learning methods for music recommendation \cite{van2013deep,hamel2011temporal}, and session-based recommendation. With the help of the advantage of deep learning in modeling different types of data, deep recommender systems can
better understand what customers need and further improve recommendation quality. 
 
\appendix
\chapter{The FSLib: A Feature Selection Library for Matlab}\label{app:A}
 
Feature Selection (FS) is an essential component of machine learning and data mining which has been studied for many years under many different conditions and in diverse scenarios. FS algorithms aim at ranking and selecting a subset of relevant features according to their degrees of importance (i.e., minimum redundancy and maximum relevance). FS can reduce the amount of features used for training classification models, alleviate the effect of the curse of dimensionality, speed up the learning process, improve model’s performance, and enhances data understanding.

The Feature Selection Library \href{https://uk.mathworks.com/matlabcentral/fileexchange/56937-feature-selection-library}{(FSLib)} is a widely applicable MATLAB library for Feature Selection (FS) \cite{roffo2016FSLib}. It is publicly available on the \href{https://uk.mathworks.com/matlabcentral/fileexchange/56937-feature-selection-library}{MATLAB File Exchange} repository where it received more than $3,000$ unique downloads only in 2016 (avg. ~300 downloads pcm, 5-starts ratings), becoming one of the most popular and used toolbox for feature selection in Matlab. It received the \textbf{Mathworks official recognition for the outstanding contribution} (in February 2017). The library provides the user with a set of 14 feature selection methods collected from recent literature on the topic. Many of them have been translated from other programming languages to matlab scripts. All the methods proposed with the toolbox share the same function prototype to facilitate performance evaluation. The toolbox is provided with a demo script and for each method its source code. 
\begin{table}[!ht]
\small
\centering
\begin{tabular}{| c |p{2.9cm} |C{0.7cm}| C{0.7cm} |C{2.5cm}| C{2.5cm}|}
\hline
ID &Acronym &   \small{Type} & \small{Cl.} & Compl.  \\\hline
1 & Inf-FS~\cite{Roffo_2015_ICCV}  &f&u& $\mathcal{O}(n^{2.37}(1+T) )$\\
2 & EC-FS~\cite{RoffoECML16} &f&s& $\mathcal{O}(Tn +n^2)$   \\
3 & mRMR~\cite{Peng05featureselection} &f&s& $\mathcal{O}(n^3 T^2)$\\
4 & Relief-F~\cite{liu2008} &f&s& $\mathcal{O}(iTnC)$   \\
5 & MutInf~\cite{Hutter:02feature} &f& s&$\sim\mathcal{O}( n^2 T^2)$\\
6 & FSV~\cite{Bradley98featureselection,Grinblat:2010} & w& s& N/A  \\
7 & Laplacian \cite{HCN05a} &f&u& $N/A$   \\
8 & MCFS \cite{Cai:2010}&f&u& $ N/A$   \\
9 & SVM-RFE \cite{Guyon:2002} & e& s& \small{$\mathcal{O}(T^2 n log_2n )$}  \\
10 & L0\cite{Guyon:2002} &w&s& $ N/A$   \\
11 & Fisher~\cite{Quanquanjournals}   &f&s& $\mathcal{O}(Tn)$   \\
12 & UDFS \cite{Yang:2011}  &f&u& N/A   \\
13 & LLCFS \cite{zeng2011feature}  &f&u& N/A   \\
14 & CFS \cite{Guyon:2002} &f&u& N/A   \\\hline
\end{tabular}
\caption{List of the feature selection approaches provided with the \emph{Feature Selection Library (FSLib)}. The table reports their \emph{Type}, class (\emph{Cl.}), and complexity (\emph{Compl.}). As for the complexity, $T$ is the number of samples, $n$ is the number of initial features, $i$ is the number of iterations in the case of iterative algorithms, and $C$ is the number of classes.}
\label{table:compmethods33}
\end{table}
This appendix provides an overview of the feature selection algorithms included in the toolbox among: filter, embedded, and wrappers methods. For each algorithm, Table \ref{table:compmethods33} reports its \emph{type}, that is, \emph{f} = filters, \emph{w} = wrappers, \emph{e} = embedded methods, and its \emph{class}, that is, \emph{s} = supervised or \emph{u} = unsupervised (using or not using the labels associated with the training samples in the ranking operation).  Additionally, we report computational complexity (if it is documented in the literature). The FSLib is publicly available on File Exchange - MATLAB Central at: \url{https://goo.gl/bvg1ha}.

From Table \ref{table:compmethods33}, the toolbox contains the following methods: Inf-FS  that stands for Infinite Feature Selection  ~\cite{Roffo_2015_ICCV}, EC-FS is the Feature Selection via Eigenvector Centrality proposed in  ~\cite{RoffoECML16,roffomelzi2016}, mRMR is the supervised minimum redundancy maximum relevance algorithm \cite{Peng05featureselection}, Relief-F \cite{liu2008}, MutInf stands for Mutual Information \cite{Hutter:02feature}, FSV stands for Feature Selection via concaVe minimization and support vector \cite{Bradley98featureselection,Grinblat:2010}. Laplacian for the method in \cite{HCN05a} which uses the Laplacian scores in an unsupervised manner. The unsupervised feature selection for multi-cluster data is denoted MCFS in \cite{Cai:2010}, the widely known SVM-RFE with recursive feature elimination discussed in \cite{Guyon:2002} and variants like L0 norm based selection methods, Fisher stands for the classic fisher score \cite{Quanquanjournals}, L2,1-norm  regularized  discriminative  feature selection  for  unsupervised  learning is called UDFS \cite{Yang:2011}, then LLCFS stands for feature selection and kernel learning for local learning-based clustering \cite{zeng2011feature} , finally the correlation-based feature selection is called CFS.

\clearpage
\newpage

\chapter{The Winter Survival Task scenario}\label{app:B}
 
\large \textbf{The Scenario}

You are member of a rescue team. Your duty is to provide assistance to any person facing
dangerous situations in a large area of Northern Canada. You have just received an SOS
call from a group of people that survived a plane crash and report on their situation as
follows: Both the pilot and co-pilot were killed in the crash. The temperature is -25C, and
the nighttime temperature is expected to be -40C. There is snow on the ground, and the
countryside is wooded with several rivers crisscrossing the area. The nearest town is 32.2
km ($\approx$ 20 miles) away. We are all dressed in city clothes appropriate for a business meeting.
The survivors have managed to extract 12 objects from the plane. But they have to leave
the site of the accident, carrying only a few objects the less the better - in order to increase
their chances of survival.\\

\noindent
\large \textbf{The Mission}

Your mission is to identify the objects most likely to maximize the chances of survival of
the plane passengers. The protocol includes two steps:

\begin{itemize}
    \item \textbf{Step 1 - Individual Step}  \\
    On the web page 1, you access a form showing the 12 items (Fig.\ref{fig:items}) and you have to decide for each one of them whether it is worth carrying or not. You must select your choice, YES or NO (YES: they have to carry it, NO: must not carry it).
    \item \textbf{Step 2 - Conversation}  \\
    You will have a text chat conversation with another member of the rescue team in order to decide together
which objects must be carried and what objects must be left in the plane.
\end{itemize}
As the text chat is a matter of life and death for the survivors, you will follow an emergency discussion protocol:
\begin{figure*}[ht]
\centering
\includegraphics[width=0.8\textwidth]{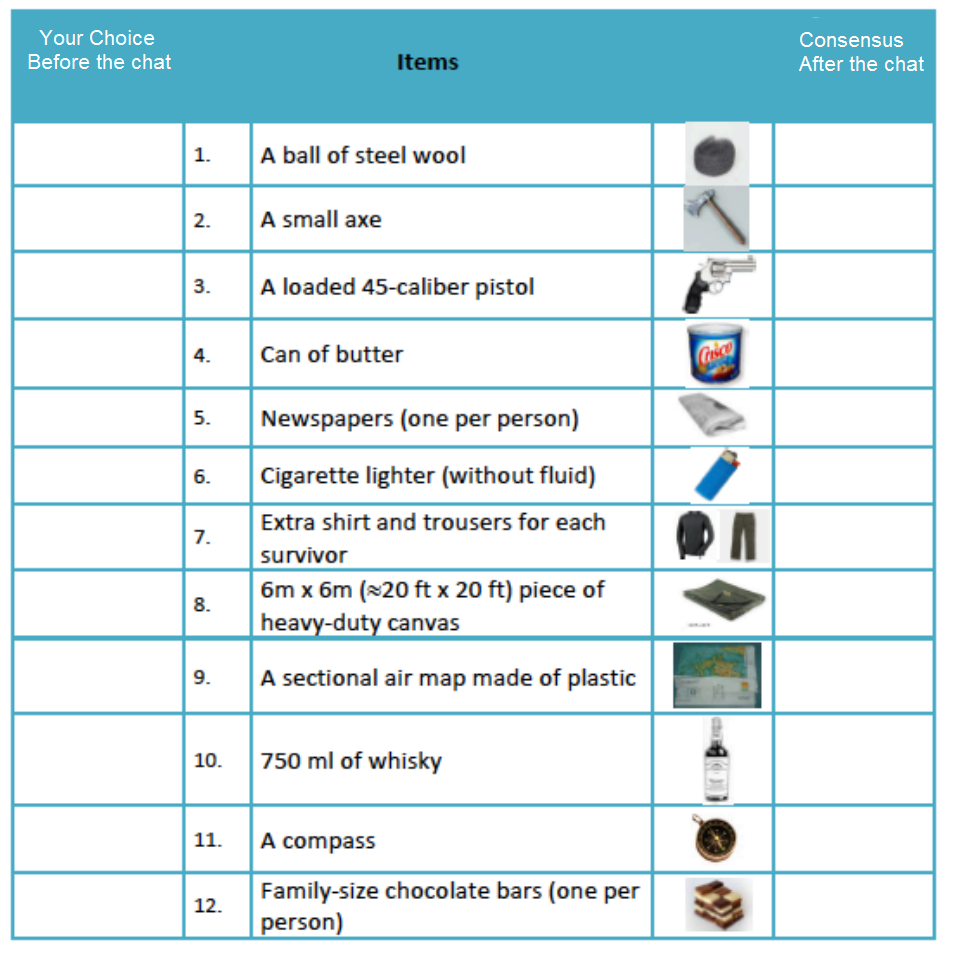}
\caption{List of 12 items.}\label{fig:items}
\end{figure*}

Please consider the following:
\begin{itemize}
    \item Discuss one object at a time and move onto the next only after a consensual decision has been made.
    \item Once a decision has been made, do not go back and change the decision about previous objects. Discuss the objects in the order shown on the attached list.
    \item Do not interrupt the chat until all objects have been discussed and all decisions have been made.
\end{itemize}

At the end of the conversation you have to submit the form with the items, completed with YES or NO decisions for each item. The results must be the same for both you and your colleague. \textbf{The text chat conversion will be recorded}.\\

\noindent
\large \textbf{Rewarding Scheme}

You will receive $\pounds$6 for your participation, but you can significantly increase your reward if you make the right decisions. Some objects are actually necessary and must be carried while others should be left on the crash site:
\begin{itemize}
    \item You receive $\pounds$3 extra, each time you decide to carry an item that must actually be carried (a right item)
    \item You lose $\pounds$3, each time you decide to carry an item that must not actually be carried.
    \item You lose $\pounds$3 for each decision marked on your list that is different from the one of your colleague.
\end{itemize}
\noindent
\textbf{In any case, a payment of $\pounds$6 is guaranteed for your participation.}
\center
\textit{Thanks for your participation!}

\clearpage
\newpage

\begin{singlespace}
\bibliography{mainbib}
\bibliographystyle{plain}
\end{singlespace}

\end{document}